\documentclass[openany,oneside,11pt,a4paper]{book}

\usepackage{natbib}
\usepackage{graphics} 
\usepackage{imakeidx}
\usepackage{amsmath} 

\usepackage{fourier}
\usepackage[T1]{fontenc}

\usepackage{titling}
\usepackage[british,calc]{datetime2} 

\usepackage{titlesec} 
\titleformat*{\section}{\Large\bfseries}
\titleformat*{\subsection}{\large\itshape}  

\newcommand{\citenew}[1]{\footnote{\citep{#1}}}

\usepackage{color} 
\newcommand{\add}[1]{#1} 

\usepackage{hyperref}
\hypersetup{colorlinks,citecolor=black,linkcolor=black}

\newcommand{\MN}[1]{\textit{Majjhima Nikaya} #1}
\newcommand{\SN}[1]{\textit{Samyutta Nikaya} #1}
\newcommand{\DN}[1]{\textit{Digha Nikaya} #1}
\newcommand{\AN}[1]{\textit{Anguttara  Nikaya} #1}
\newcommand{\EN}{\textit{The Enchiridion}}
\newcommand{\Discourses}{\textit{Discourses}}
\newcommand{\Lucilius}{\textit{Letters to Lucilius}}
\newcommand{\Beata}{\textit{On a Happy Life}}
\newcommand{\yogacara}{Yog\=ac\=ara}
\newcommand{\eqsuff}{frustration equation}

\newcommand{\hume}{Hume, \textit{A Treatise of Human Nature}}

\makeindex[intoc]

\begin{document}

\title{ \textrm{ \bf  \Huge  Painful intelligence:\\\vspace*{2mm}  What AI can tell us about human suffering}\\\vspace*{4mm}\vspace*{0.5cm}}
\author{\textrm{\Large Aapo Hyv\"arinen} \\ \ \\ \textrm{\large University of Helsinki}
  \\ \vspace*{6cm} \\
  {\Large Second Edition\vspace*{5mm}}\\
\textrm{September 2024}
}
\date{}
\maketitle

\setcounter{page}{2}

\vspace*{3cm}

\begin{center}
  \textbf{\large Abstract}
\end{center}
This book uses the modern theory of artificial intelligence (AI) to understand human suffering or mental pain. Both humans and sophisticated AI agents process information about the world in order to achieve goals and obtain rewards, which is why AI can be used as a model of the human brain and mind. This book intends to make the theory accessible to a relatively general audience, requiring only some relevant scientific background.

The book starts with the assumption that suffering is mainly caused by frustration. Frustration means the failure of an agent (whether AI or human) to achieve a goal or a reward it wanted or expected. Frustration is inevitable because of the overwhelming complexity of the world, limited computational resources, and scarcity of good data. In particular, such limitations imply that an agent acting in the real world must cope with uncontrollability, unpredictability, and uncertainty, which all lead to frustration.

Fundamental in such modelling is the idea of learning, or adaptation to the environment. While AI uses machine learning, humans and animals adapt by a combination of evolutionary mechanisms and ordinary learning. Even frustration is fundamentally an error signal that the system uses for learning. This book explores various aspects and limitations of learning algorithms and their implications regarding suffering.

At the end of the book, the computational theory is used to derive various interventions or training methods that will reduce suffering in humans. The amount of frustration is expressed by a simple equation which indicates how it can be reduced. The ensuing interventions are very similar to those proposed by Buddhist and Stoic philosophy, and include mindfulness meditation. Therefore, this book can be interpreted as an exposition of a computational theory justifying why such philosophies and meditation reduce human suffering.

\vfill\noindent
\begin{minipage}{\textwidth}
\begin{center}  \setlength{\baselineskip}{5mm} {\small
Copyright \copyright \the\year\ Aapo Hyv\"arinen. All rights reserved.\\
Distribution allowed as per Creative Commons Attribution-Noncommercial-NoDerivatives (CC BY-NC-ND) License. }
\end{center}
 \begin{center} \setlength{\baselineskip}{4mm} {\footnotesize
This book is typeset in Utopia(R), Copyright 1989, 1991 Adobe Systems Incorporated. All rights reserved. \\Utopia is either a registered trademark or trademark of Adobe Systems Incorporated in the United States and/or other countries. \\Used under license, as implemented in the Latex {\tt fourier} package.}
\end{center}
\end{minipage}

\setcounter{tocdepth}{1}
\setcounter{secnumdepth}{0}
\tableofcontents

\chapter*{Preface (1st Edition)}

I like to write books that I would have wanted to read myself as a student. I really wish I had been able to read this book. It would probably have changed my life and my career, as I would have insisted on doing my PhD on this topic.
Alas, when I was a student in the 1990s, the topic of this book was not something a reasonable PhD student would have worked on. There was hardly any literature on the topic; it would have been considered uncharted territory, if not suspicious. I hope the world has changed, and that this book may contribute to that change. With the huge increase in research on AI and computational neuroscience on the one hand, and affective neuroscience and mindfulness meditation on the other, I think the time is ripe to attempt a synthesis, which is the motivation for this book.

What I should emphasize is that this book is about a scientific theory, or rather, several scientific theories. It is not a book that teaches meditation; it has little to do with self-help and certainly constitutes no clinical guidance. Nor is it really a philosophical book in the sense that the word would be used in academic circles: while there is some philosophical speculation, the main paradigm is that of the natural sciences. It may be surprising that I seem to include artificial intelligence in the natural sciences, but here it is largely used as a computational model of the brain, even if sometimes on a very abstract level. The strong neuroscience component of this book further connects it to empirical science. 

I have tried to write the book so that it is suitable for as wide an audience as possible. I believe anybody trained in computer science or neuroscience should be able to understand it. Scientific training in any discipline might be enough to understand the main ideas, and I hope that even members of the general public might find something interesting in it. Although not primarily intended as such, the book can also be used as a university-level textbook for advanced undergraduates or graduate students in computer science or cognitive science; it should also be suitable for computationally minded students in neuroscience or psychology.

This book was written while working in different institutions. Most of the work was done while a faculty member at the University of Helsinki (Department of Computer Science). Part of the writing was accomplished while a faculty member at University College London (Gatsby Computational Neuroscience Unit) as well as a research scientist at Universit\'e Paris-Saclay (DataIA Institute and Inria--Saclay-Ile-de-France, supported by grant ANR-17-CONV-0003). The work was further supported by a Fellowship from CIFAR (Learning in Machines \& Brains Program).

Finally, I'm very grateful to Moritz Grosse-Wentrup, Riitta Hari, Marianne Maertens, John Millar, Tiina Parviainen, Jonne Viljanen, and, especially, Michael Gutmann, for most helpful comments on the manuscript.  

\vspace*{8mm}
\hspace*{2cm}\textit{Helsinki, May 2022} \hfill Aapo Hyv\"arinen \hspace*{2cm}

\chapter*{\add{Preface (2nd Edition)}}

\add{
In the second edition (V2 on Arxiv) the theory has been slightly expanded and clarified. The main changes are:

\begin{enumerate}
\item A probabilistic theory of threat is now introduced in Chapter~\ref{threat.ch}, which is the only major change.
\item Chapters 7 and 15 of the first edition have been split into two chapters, giving rise to Chapters \ref{dual.ch}--\ref{summary1.ch} and \ref{training.ch}--\ref{attitude.ch}. The contents have been slightly expanded in the latter chapter of each pair. 
\item New figures (Fig.~\ref{valueplot.fig} and Fig.~\ref{flowchart3.fig}) have been added to improve readability.
\item A Guide to the Reader has been added at the  end of Chapter~\ref{introduction.ch}.
\item More material on Greek philosophy has been added in Chapters~\ref{freedom.ch}, \ref{training.ch}, and \ref{attitude.ch}, which have also been slightly streamlined by removing some material and transferring some material to footnotes.
\end{enumerate}

\noindent
I would further like to thank Mitsuo Kawato, Keith Davis, and Michael Gutmann for additional comments.

\vspace*{8mm}
\hspace*{2cm}\textit{Helsinki, September 2024} \hfill Aapo Hyv\"arinen \hspace*{2cm}

}

\chapter[Introduction]{Introduction:\\ Understanding human suffering by AI}\label{introduction.ch}

What is the most central question in human life? For me, it is the question of suffering. There may be questions which are more fundamental, or philosophically more fascinating, for example: Why does the world exist? Or, how is it possible that we are conscious? But those questions are rather theoretical and mainly satisfy one's intellectual curiosity. If you found the answer to those latter questions, would that change your life, or other people's lives, for the better? 

The question of suffering is with us at every moment. By suffering I mean mental pain, the opposite of pleasure and happiness. In some cases, it is a result of physical pain, but usually of purely mental origin. In fact, any casual observer of human life easily comes to the conclusion that it is full of such suffering: There is frustration, anxiety, sadness, depression, and so on. 

Why is the ``human condition'' so unpleasant: did somebody (or something) make a huge mistake in designing humans? And, most importantly, is there anything we can do about it: can we remove suffering, or at least reduce it? Now, this is a question that has enormous practical significance. Reducing suffering, almost by definition, makes people's lives better.

The starting point of this book is the idea that we can use the theory of artificial intelligence, or AI, to understand why there is so much suffering in humans. This book will show how suffering is largely due to the inability of an intelligent system, whether an artificial intelligence or a human being, to understand its own programming and its own limitations, in particular the limitations of its computation and data.

\section{Investigating intelligence by constructing it}

How can I claim that the theory of AI has any relevance to understanding the human mind, let alone suffering?
The answer lies in how AI can help us understand the computational design principles which are applicable to humans as well.

When I asked above if somebody made a huge mistake in designing humans, that ``somebody'' was of course evolution, metaphorically speaking. Evolution designed the basic processes of our mental life, for good or bad.
Importantly, evolution didn't construct our brains in some random, arbitrary ways, but it designed us to be fit for certain purposes and goals. Ultimately, those evolutionary goals are about reproduction and spreading your genes, but to satisfy that ultimate goal, many more intermediate goals need to be considered. You have to get food, find sex partners, and not get killed. These, in turn, require that you know how to walk, and you are able to recognize objects as well as to plan your future actions.

We can learn to understand such evolutionary design goals by trying to design and construct an AI, or a robot. This is a perspective which  
is gaining more and more prominence in neuroscience: Trying to actually construct an intelligent system forces you to think about the computation and algorithms needed.

Ordinary neuroscience is based on conducting experiments on humans or animals. It can establish many interesting facts about the brain; for example, where in the brain the processing necessary for vision or fear takes place. In particular, it can also tell us a lot about how such processing happens; it can explain how the brain recognizes that the animal in front of you is a cat and not a dog, and how the brain initiates a fear response if the animal actually turns out to be a tiger.

However, the deepest question in neuroscience is the \textit{why} question: Why does a certain kind of processing take place at all? What is its evolutionary purpose? 
Why do we, for example, have emotions like fear in the first place? Why is our mind frequently assailed by thoughts about the past and the future even when we try to concentrate on the present? And ultimately, why is there suffering?

Designing intelligent systems goes a long way toward answering the ``why'' question. If we find that an AI necessarily needs a certain kind of computation to achieve human-like intelligence, it is likely that the human brain does that same kind of computation---at least on some level of abstraction. AI can also give us a deeper understanding of ``how'' computations happen in the human brain, since designing it necessarily forces the scientists to figure out all the details needed in the computation.

\section{Is the brain a big computer?}

The prerequisite for learning about the brain by building intelligent systems is that our brain is in many ways like a computer. In fact, the modern paradigm in neuroscience and psychology considers the brain as an information-processing device. The term ``cognition'' is used to describe information-processing performed by the brain, while with ordinary computers we usually talk about computation.

The brain receives new data by seeing, hearing, or otherwise sensing things. It processes the sensory data in various ways, ultimately enabling us to recognize objects and act in the world. It can also process information retrieved from its own memory, which is necessary for what we call thinking in plain English. A system that processes information in such ways can be called, almost by definition, a computer, so it is natural to say that, actually, the brain \textit{is} a computer.

Certainly, the brain is very different from any ordinary computer that you can buy in a shop. For example, your PC, or your mobile phone, has a central processing unit (CPU), sometimes a couple of them. The brain has no such thing. The information-processing happens in the neural cells, or neurons. Each of them is like a tiny CPU which can only perform extremely simple processing--- but there is a huge number of them, tens of billions. The crucial difference with respect to a CPU is that each neuron processes its own input independently, and all the neurons do that at the same time---this is called parallel and distributed processing. 

Yet, from an abstract viewpoint, such differences can be seen as just technical details.
In particular, if we are interested in the question of ``why'' certain computations are performed, the physical structure of the information-processing device, or even the details of the programming do not matter. 
What really matters for our purposes is whether the brain and the computer need to solve the same kinds of computational problems. 
This will be the case if humans and the AI live in the same kind of environment, have the same kind of goals for their actions, and use similar means to try to reach them.
That is increasingly the case when AI develops in terms of autonomous robots, for example, and in any case, we can use our current AI theory to extrapolate what AI's might be like in the future.

\section{Machine learning as analogue to evolution}

Even granted that humans and computers are both information-processing devices, some would argue that they process information based on very different principles. A popular claim is that a computer does exactly what it is programmed to do, and nothing else, and this is supposed to be very different from humans who do what they want themselves ---so any parallels between humans and computers are impossible. I think this reasoning is fundamentally wrong, for two reasons.

First, modern AI systems do not just do what they are programmed to do. That's because their function is based on \textit{learning}. They are programmed to learn from input data. The input may be a database determined by the programmer; it can be obtained by crawling the internet; or it can be the result of interactions with the environment, like a robot using a camera or users typing words, and so on. What the programmer really does is to provide an \textit{algorithm} for learning. The algorithm is based on certain goals or \textit{objective functions} that the AI is trying to optimize. An AI dedicated to searching the internet for images that resemble a given target image will learn to optimize the accuracy of its search results, for example by maximizing the number of clicks users make on each image it proposes.

\index{evolution}

What this means is that anyone who programs an AI cannot really know in detail what the AI will actually do, because it is often impossible to know what kind of input the AI will receive, and it is equally difficult to understand what the AI will learn from it. Even in the simplest case where the programmer completely decides the input to the AI, the input is often so complex (say, millions of pictures downloaded from the internet) that it is impossible for a human programmer to understand what can be learned from that data.

The second reason why there is not such a big difference between humans and AI 
is that just like an AI is programmed by humans, we humans are designed---one might say ``programmed''---by evolution.
From an evolutionary perspective, 
we are programmed to maximize an objective function which is roughly given by the total number of copies of our genes in the population. 
To satisfy such programming, we gather a lot of data---by reading things, talking to people, and simply looking around---which is not so different from an AI.

So, I have turned the claim about the difference between AI and humans on its head. What humans and AI have \textit{in common} is that both are programmed \textit{by something else} to have certain goals and needs;
nobody has really decided ``by themselves'' to have the needs and goals they have. To accomplish those goals, both humans and AI gather data from the environment and learn from it, which leads to actions that are \textit{very difficult to predict}. So, in the end there is little difference between AI and humans, except regarding the source of the original programming---whether it was by evolution or a human programmer.

\section{Can an AI actually suffer?}\label{canaisuffer.sec}

By now, I hope to have convinced you that an AI is a useful model of many phenomena taking place in the human brain. But perhaps there are limits. Some would argue that we cannot talk about AI's or robots suffering: They may \textit{seem} to be suffering, or look like they are suffering, but in fact they are not, because they cannot \textit{feel} anything.

I think this argument may not be completely wrong, but it is quite irrelevant. Obviously, it depends on the exact definition of what suffering is. 
It is true that AI may not feel suffering in the same way as humans because that might require that AI is conscious, \add{i.e., it has subjective experiences.} This argument against AI's suffering really hinges on two points: First, that an AI is not conscious, and second, that consciousness is necessary for suffering.

However, conscious feeling is only one part of suffering. The situation is similar with emotions, such as fear, which are actually clever information-processing mechanisms. The conscious feeling of being afraid is only one part of a complicated process involving cognition (or information-processing), behavioral tendencies, and several other aspects. I would argue it is the same for suffering.\index{consciousness!and suffering}\index{subjective experience!not necessary for pain and suffering}

Suffering is actually a signal in a complex information-processing system. The real meaning of the suffering signal is that an \textit{error} occurred---this will be elaborated in several chapters in this book. Any information-processing system can create error signals. That's why we can, in that specific sense, say that an AI or a robot is suffering, even if they are not conscious. All that would be missing is the conscious feeling components of suffering.

There is an even more important reason why it is largely irrelevant here if an AI really suffers according to some stringent definition of the word. This book does not just aim to describe the mechanisms of suffering; the primary goal here is to develop various ways of alleviating suffering. For the purpose of \textit{reducing} suffering, it does not matter if computers actually suffer in some deeper sense. If we can reduce suffering in an AI that is sufficiently human-like, then, with reasonable probability, the same methods will apply to humans, and they will reduce suffering in humans, including the conscious experience of suffering. In other words, the AI is really a simulation or a model of mechanisms that are relevant for making humans happier.

For those who find it impossible to
think that a computer could suffer in any sense of the word, I suggest
the following viewpoint that they can use while reading this book.
Trying to understand human suffering by AI is one big thought
experiment, where we try to understand how much the AI would suffer
under various circumstances, \textit{if} it were able to consciously
experience suffering. It is like a mathematical model of atoms, or like a computer simulation of chemical processes. Everybody agrees that models and computer simulations are not the real thing, but they can help us understand the actual natural processes, and in particular, predict their behavior. A model may tell you how a change in one quantity, say X, leads to a change in another quantity, Y. If you know that, you can perhaps choose X to maximize or minimize Y---which might be suffering.

\section{Intelligence is painful---overview of this book}

The central hypothesis in this book is that if we create an artificial intelligence that is really intelligent, really worthy of its name, it will necessarily perform computations which are more or less like human suffering.  In spite of the many differences between AI's and humans, there is a common logic in the design. In order to achieve sufficiently human-like intelligence, certain design principles have to be followed, and these lead to suffering. This book explores several interwoven ideas about such a computational basis of suffering, and the necessity of suffering as a part of intelligence. 

The fundamental approach here is that suffering is caused by error signalling, which is typically due to \textit{frustration}. Frustration occurs when an intelligent system, generally called an ``agent'', fails to achieve a goal, \add{or it obtains less reward than it expected.} Such errors are inevitable in a complex world, where things are uncertain and unpredictable, and we have limited control over them. Error signalling is necessary for any sufficiently intelligent system, since such error signals are used by learning algorithms. Our brain produces error signals automatically, and we simply cannot shut off the error-signalling system. 

In fact, the complexity of the world is overwhelming for any known intelligent system, whether the very best supercomputer in the world, or the most intelligent human brain. The computations available to them cannot handle all the different possibilities when, for example, choosing action sequences to reach a given goal.
Modern AI uses learning to cope with such complexity. %
However, for such learning to be really successful, huge data sets are required. Obtaining data sets which completely capture the complexity of the world is rarely possible in practice.
These two factors, \textit{lack of computational resources together with scarcity of data}, mean that the intelligent agent cannot work optimally. 
Its intelligence and its control over the world are limited. Thus, there will be errors: The agent's actions do not always lead to the desired outcome, hence frustration. %

Suffering is greatly enhanced by several information-processing principles inherent in the design of human-like intelligent systems.  One is the phenomenon of experience replay, where memories related to past errors are recalled and repeated in the system in order to optimize learning about past experiences. Likewise, plans for future actions are constantly computed, which means the agent \textit{simulates} or ``imagines'' them in its mind, together with the ensuing errors. Such replay and planning multiply any suffering arising from real events: errors are signalled as if those bad, imagined events happened for real. 
\add{Further suffering is created by the perception of \textit{threats}, or predictions of future frustration; that means frustration that did not actually happen but just might happen with some probability.} Thus, we suffer from mishaps which only happen in our imagination.

Meanwhile, modern AI has found systems based on parallel and distributed information processing to be useful for programming intelligent systems, which makes it understandable that our brain uses similar principles. However, such processing leads to overwhelming \textit{uncontrollability}. Systems that are parallel and distributed do not admit central executive control, since different modules are competing for control; this makes, for example, any sustained attention or concentration difficult. Any internal control of the agent's computations is further reduced by emotions such as fear, which work as evolutionarily conditioned ``interrupts'' of ongoing processing. Thus, the agent has little control even of its own internal processing, let alone the external world.
\add{A related problem is the \textit{uncertainty} of our perceptions, and the difficulty of understanding how uncertain most perceptions and inferences actually are.} Perceptions are often highly subjective and contextual interpretations, sometimes little more than guesses. However, humans often mistakenly think that our perceptual systems are able to discover some underlying objective reality. 
\add{Such uncontrollability and uncertainty both increase suffering by increasing frustration and other errors.}

Finally, the goals and desires that have been  programmed in us by evolution are ultimately counterproductive and make us unhappy. Evolution never had our happiness as its goal anyway. In fact, it forces us to do things which are clearly bad for our happiness, something I call \textit{evolutionary obsessions}.  Evolution makes us worry about our survival and our evolutionary performance, creating a sense of self. In fact, evolution does \textit{not} want us to reduce suffering because the error-signalling system is necessary for learning and optimal behavior. What evolution does want us to learn is to act in  more and more efficient ways, but the goals towards which this intelligence is used are those set by evolution, not us.
Even worse, both AI and humans are usually trying to satisfy their drives and desires endlessly, without any limits; at no point do they become satiated and think that they have achieved enough.

However, there is hope. At the very end of the book, I sketch \textit{interventions}, or mental training methods,  that can be used to decrease suffering, based on the theories outlined in this book.  What is needed is a \linebreak \textit{reprogramming of the brain}. The key method is to retrain the brain by inputting new data into the learning system. The new data will change the computations in such a way that error signals, and in particular frustration, are reduced: \add{learning to reduce expectations and desires is crucial here.} This is difficult and takes a lot of time, but various forms of philosophical contemplation and mindfulness meditation attempt to do it. These methods are rather logical consequences of the theory, while they have mainly been proposed earlier in Buddhist, and to some extent Stoic, philosophy. 
Thus this book can be seen as an attempt to construct a scientific, computational theory on the underpinnings of such philosophies and meditation.

\section{Guide to the Reader}
\add{

Obviously, the recommended way is to read all the chapters in the order presented. Footnotes can be skipped by readers not interested in the details.
However, for busy readers, here is an outline of shorter paths through the book:
\begin{itemize}
\item A short overview of the basic ideas can be extracted by reading Chapters~\ref{suffering.ch}, \ref{planning.ch}, \ref{summary1.ch}, \ref{overview.ch} and the first full section of Chapter~\ref{attitude.ch} including Figure~\ref{flowchart3.fig}.
\item A slightly longer overview can be extracted by reading Chapters~\ref{suffering.ch}, \ref{planning.ch}, \ref{rpe.ch}, \ref{threat.ch}, \ref{summary1.ch}, \ref{overview.ch}, \ref{freedom.ch} and the first full section of Chapter~\ref{attitude.ch} including Figure~\ref{flowchart3.fig}. 
\item A slightly shortened version emphasizing Buddhist-Stoic philosophy and meditation can be obtained by reading Part I and  Part III,  thus skipping Part II. (Chapter~\ref{overview.ch} in Part III contains a short summary of Part~II, so at least on some level, it should be possible to understand Part III without Part II.)
\end{itemize}
}

\part[Suffering as error signalling]{Suffering as error signalling
  \\ \ \\ \ \\  \normalsize The first part will explore the very definition of suffering, \\existing proposals on how suffering comes about, \\and how these can be understood by the theories of AI and evolution\label{parti}}

\chapter{Defining suffering} \label{suffering.ch}

In this chapter, I try to define the word ``suffering''.
This is not an easy task, as we will quickly see. Defining the term properly requires, to some extent, elucidating the underlying mechanisms creating suffering.

One fundamental point here is that I exclude physical pain from the definition of suffering; I use the word suffering synonymously with mental pain. Nevertheless, I will start the search for a definition of suffering by considering the closely related concept of pain, taken here in the medical sense of physical pain.

The central conclusion of this chapter is that the main  definitions of suffering consider it based on either  \textit{frustration} or a \textit{threat to the intactness} of the person. These two definitions, and especially the definition based on frustration, are the basis of the developments of the rest of this book. From a more abstract viewpoint, I will argue that such suffering can be seen as \textit{error signalling}, similarly to physical pain.

\section{Medical definitions of pain} 

Let us start by defining pain. Pain has been given a widely accepted consensus definition by the International Association for the Study of Pain (IASP) as: 
\begin{quote}\label{iasp}
Pain is an unpleasant sensory and emotional experience associated with actual or potential tissue damage or described in terms of such damage.\index{pain!IASP definition}\index{IASP|see{pain, IASP definition}}
\end{quote} 
Surprisingly, while this definition was originally adopted in 1979, it is still used with minimal modifications. It posits damage to any tissue of the person, or any threat of such damage, as the origin of pain. Pain is then defined as an ensuing unpleasant experience.
While this definition has been found to be quite useful in a clinical context, deeper theoretical analyses have found various problems.\footnote{\citet{cohen2018reconsidering} gives a long review of competing definitions; \citet{corns2016pain} considers the validity of the very concept; \citet{klein2007imperative} proposes an alternative definition and reviews some philosophical approaches. IASP has very recently proposed a revised version \citep{raja2020revised}, but the changes are minimal and rather immaterial for our purposes.}

One important controversy is whether one should define pain as a subjective experience,\index{subjective experience!and pain} or as something that has a more objective existence. The definition above talks about an ``experience'' which is here interpreted as a conscious, subjective experience: something that only I am aware of, and which you cannot measure in any objective way.
As we will discuss in more detail in Chapters~\ref{emotions.ch} and \ref{consciousness.ch}, this problem of subjective conscious experience vs.\ objectively observed phenomena is ubiquitous in neuroscience and psychology.\index{consciousness!and pain}\index{pain!and consciousness}

The problem with talking about such subjective experience in a scientific context is that objective, reproducible measurement is the basis of science. Fortunately, subjective experience can be measured in various indirect ways, such as verbal report. That is, we can ask the patient if there is pain. Yet we will never know for sure what the patient actually feels. In particular, we cannot tell how her experience of pain compares with other people's experience: Does she feel more or less pain than some other patient who gives the same verbal report? 

This problem in the definition of pain is to some extent alleviated by the reference to tissue damage, which is objectively measurable and reasonably well-defined. Yet, as this definition clearly points out, actual tissue damage is not \textit{necessary} for pain---since it can be just ``potential''---and thus it does not provide a basis for measuring pain or for objectively defining it. (In fact, the definition does not actually say that pain is in any sense proportional to the amount of damage--- it is well-known that tissue damage that creates a lot of pain in one person may create little pain in another---which complicates any measurement even more.)
Another related problem with the IASP definition above is that it relies heavily on the word ``unpleasant'', which is not a very well-defined term, and, again, quite subjective. 

One approach to solve these problems is to take an evolutionary approach. To begin with, we could replace ``unpleasant experience'' in the definition by ``experience that has evolved to motivate behaviour, which avoids or minimises tissue damage, or promotes recovery''.\footnote{This definition is by \citet{wright2011criticism}. On a related note, \citet{seymour2019pain} emphasizes the importance of pain as a signal used in control and learning, and relativizes the importance of conscious experience.}  Here we go towards defining pain using its evolutionary, functional role, while still acknowledging the subjective nature of pain by talking about an ``experience''.  The downside of such an approach is that it works on a very abstract level, and provides no details on what might cause pain, in contrast to the IASP definition which explicitly points at tissue damage (even if only potential). This definition, in a sense, shifts the burden to understanding the evolutionary goals of certain experiences, which is not easy either. However, one obvious candidate for such an evolutionary goal is minimizing tissue damage and recovering from it, which links this evolutionary approach with the IASP definition. In more general terms, the evolutionary goal could be the maintenance of ``homeostasis'',  that is, an optimal balance in the physiological condition of the body.\citenew{craig2003new} Such evolutionary logic can be applied on suffering as well, and we will see related argumentation throughout this book.\index{homeostasis}\index{pain!evolutionary definition}

\section{Medical and psychological definitions suffering}

In contrast to pain, suffering is a rather neglected term in science, and there is nothing like a consensus definition. Intuitively, most people would think suffering also contains an \textit{unpleasant feeling or experience} as an integral part, while being more abstract and general than physical pain, in particular including more psychological and emotional aspects. A typical dictionary definition is ``Feeling of pain or strong stress, either physical or emotional''.\footnote{https://psychologydictionary.org/suffering/}
Like pain, suffering is often considered a subjective experience which cannot be objectively measured.\citenew{cassell1982nature,edwards2003three}

One simple and concrete approach to define suffering is to give examples of phenomena related to suffering and possibly producing suffering. A typical list would contain grief, sadness, discomfort, distress, anguish, fear---which is just a random sample, and many different lists can be produced. While this is a good starting point, it does not lead to a solid scientific theory.\index{suffering!definition}

Terms such as psychological pain or mental pain are often preferred in neuroscience, and some attempts at definitions of those terms have been made.\footnote{Reviews on the topic are provided by \citet{mee2006psychological,tossani2013concept,papini2015behavioral}. The term ``mental pain'' could be criticized because all pain is ultimately mental, as seen in the IASP definition. In this book, I mainly use the term ``suffering''.} In this line of thinking, suffering is really a generalization of pain. This may not solve the problem of defining suffering, since the burden is then simply shifted to defining pain, but then we can leverage the large literature on pain, in particular the IASP definition just given, as well as any of its critique and improvements.

One approach distinguishes three kinds of pain:
physical pain,  social pain, and psychological pain.\footnote{\citep{papini2015behavioral,eisenberger2004rejection,macdonald2009social}. Pain based on empathy when one sees others hurting, or ``vicarious'' pain, could be added to the list \citep{singer2004empathy}.}  An interesting emphasis in this line of research is that all these different kinds of pain are neurally very similar in the sense that the brain areas responsible are the same.\footnote{However, see \citet{iannetti2013beyond,wager2016pain} for criticism of the reverse inference used in that work. \citet{iannetti2010neuromatrix} argue that the brain network considered may be more related to detection of saliency (i.e.\ how much attention a stimulus attracts).} Here, physical pain is primarily due to physical damage to the body, but it can also be felt when there is a strong anticipation of such physical damage (think about going to a dentist), reminiscent of the IASP definition. In contrast, social pain is an unpleasant feeling due to social exclusion or rejection. Psychological pain is largely the same as what I call mental pain or suffering, and attempt to define here.

Importantly for our purposes, in such an approach,  mental or psychological pain is often assumed to be due to \textit{reward loss}, defined as follows\citenew{papini2015behavioral}
\begin{quote} 
  [Reward loss is] a negative discrepancy between expected and obtained rewards.
\end{quote}
In other words, reward loss happens when you expect a reward but don't get it, and it leads to mental pain. Reward loss can also be called \textit{frustration}, although sometimes this term is reserved for the actual suffering caused by reward loss.  This provides one important computational viewpoint: reward loss is a function of computations involving expectations, observations of the obtained reward, and their difference.
\index{suffering!definition!as reward loss}
\index{suffering!definition!as frustration}
\index{reward loss!definition}
\index{frustration!as reward loss}

An alternative approach emphasizes how suffering is related to our person, or self. Psychological or mental pain has been characterized as an aversive state of high self-awareness of inadequacy,\citenew{baumeister1990suicide,orbach2003mental} or a negative appraisal of an inability or deficiency of the self.\citenew{meerwijk2011toward} This is analogous to the IASP definition of physical pain in the sense that there is ``damage'', even if purely mental, to one's image of oneself as a psychological and social entity.\footnote{In this  line of research, typical in the philosophy of medicine and bioethics, suffering is sometimes seen as something particularly strong \citep{degrazia1998suffering,hoffmaster2014understanding}, in particular stronger than any pain typically encountered in everyday life. I don't follow such a definition here: in this book, suffering can be very mild or very strong.}

A particularly potent and influential idea in this vein is that suffering necessarily involves a threat to, or a loss of, the \textit{intactness of the person}, as proposed by Cassell:\citenew{cassell1989relationship}
\begin{quote}\label{casselldef}  
Suffering is a state of severe distress induced by the loss of the intactness of person, or by a threat that the person believes will result in the loss of his or her intactness.\index{suffering!definition!Cassell|see{intactness of the person}}\index{intactness of the person}
\end{quote}
This is a natural generalization and abstraction of the IASP definition of pain as related to ``tissue damage''.   In this definition, damage to the intactness of the person actually includes tissue damage, but it is something much more general, in particular, it includes damage to one's self-image. It is of course crucial to understand what ``intactness'' means more precisely; Cassell emphasizes the generality of this notion, saying that ``suffering may occur in relation to any aspect of personhood''.\footnote{For recent critique of Cassell's approach, see \citet{bueno2017conceptualizing} who criticizes Cassell's definition precisely on the ground that ``intactness'' is not well-defined and may not even exist; another point of critique is that Cassell's definition ignores existential suffering. Further criticism is given by \citet{tate2019we} who propose to define suffering as ``a loss of a person's sense of self'' together with ``a negative affective experience''.}

A general theory that combines pain and several kinds of suffering in a single framework
has been developed by van Hooft.\citenew{van1998suffering}\label{hooft}\index{suffering!definition!van Hooft}\index{suffering!definition}\label{vanhooftpage}
He starts from an Aristotelian\index{Aristotle} conception of the human person as having four ``parts of the soul''. They range from the lowest level of biological functioning to the emotional/desiring functions and the rational functions, finally reaching the sense of the meaning of existence. In his theory, each of these parts has its own goals, its own form of ``fulfillment'', which is again an Aristotelian idea. Suffering is then nothing else than frustration, namely ``frustration of the tendency towards fulfillment'' of one of the different parts of the soul.
In this theory, the lowest level of biological functioning is even below ordinary pain and pleasure, and simply about staying healthy and alive. 
Ordinary physical pain is the frustration on the emotional/desiring level, where the goal of the organism is to gain pleasure and avoid pain.  
Frustration of rational (intellectual) function refers to suffering which happens when it is not possible to reach long-term goals that one plans for and expects to reach. Frustration on the highest, ``spiritual'' level happens when it is impossible to understand why it is me that is sick---in the medical context where van Hooft writes---or life seems meaningless due to the despair and fear which a malady brings with it. This last kind of suffering brings us close to the kind of suffering considered in existential philosophy.\citenew{svenaeus2014phenomenology,bueno2017conceptualizing}

Closely related definitions can be found in the literature on \textit{stress}: Lazarus and collaborators define ``psychological stress'' as ``a particular relationship between the person and the environment that is appraised by the person as taxing or exceeding his or her resources and endangering his or her well-being''.\footnote{\citep{lazarus1984stress}; see also \citet{lazarus1993psychological}. Their work emphasizes the individual's perception and interpretation of the events by the term ``appraise'', related to Cassell's definition which talks about ``believing''.
  Another related approach to defining stress emphasizes conservation of resources, and defines the stress as, roughly, loss of resources \citep{hobfoll1989conservation}.}
\index{stress}
So, we have to consider the possibility that stress is another kind of suffering, or a mechanism for suffering. However, I don't take such a view in this book because the classic definition by Hans Selye, ``the father of stress'', proposes that ``stress is the non-specific response of the body to \textit{any} demand'' (my italics). This  is a very general definition, and Selye has explicitly emphasized that positive, happy events can induce stress just as well as negative, threatening ones; think about an athlete engaged in a competition. Based on this definition, it does not seem possible to simply consider stress as one kind of suffering, unless we focus on the negative kind of stress, termed ``distress'' by Selye.\footnote{See \citet{fink2016stress} where the quote by Selye is also taken from.} 
The distinction between distress and ``pleasant'' stress is, unfortunately, not very clear; 
it has been proposed that it is the \textit{unpredictability and uncontrollability} of a situation which distinguish the unpleasant distress from other kinds of stress.\footnote{\citep{koolhaas2011stress}}
Their connection to suffering will be considered from different viewpoints in this book.

\section{Ancient philosophical approaches to suffering}

\index{suffering!definition}
Centuries before any such modern developments, 
some ancient philosophers already made great progress in understanding suffering. The best expert on the topic may have been the Buddha, and in fact the whole of Buddhist philosophy can be seen as a theory of suffering---especially when considering the original version proposed by the Buddha himself. He gave the following description of suffering:\footnote{This is from a fundamental discourse by the Buddha found in one of the earliest known layers of Buddhist literature, the Pali Canon. Different versions are available in  \SN{56.11}, \MN{141}, and \DN{22}, where the last one is the most detailed version. This quote is part of the description of what is called the Four Noble Truths,\index{four noble truths (Buddhist)} of which we here consider only the first one (see footnote~\ref{fournobleall} in Chapter~\ref{freedom.ch} for the rest). The whole description of the first truth, synthetizing the different versions, says approximately:  Birth is suffering, ageing is suffering, illness is suffering, death is suffering; grief, lamentation, pain, distress, and despair are suffering; union with what is displeasing is suffering, separation from what is pleasing is suffering, not to get what one wants is suffering. (Several partial translations of the Pali Canon are available on the internet and I will often select the translation I find the most compatible with my terminology; the one in the main text here is by Bhikkhu Boddhi.)\label{fournoblefirst}}
\begin{quote} \label{fournoble1}
Union with what is displeasing is suffering; separation from what is pleasing is suffering; not to get what one wants is suffering.\index{suffering!definition!Buddha}\index{Buddha}
\end{quote}
This is actually not so much a definition of what suffering is, but rather an attempt to describe what the main causes of suffering are.

Stoic philosophers in ancient Greece and Rome had very similar ideas. Epictetus, one of the most famous Stoics, describes mechanisms that lead to suffering as follows:\footnote{Paragraph 2 in \EN, compiled approximately 125-135 CE. Quotes in this book are taken from the translation by E.~Carter at {\tt classics.mit.edu/Epictetus/epicench.html} unless otherwise mentioned.} 
\begin{quote} \label{epictetospromise}
[D]esire promises the attainment of that of which you are desirous; and aversion promises the avoiding that to which you are averse. However, he who fails to obtain the object of his desire is disappointed, and he who incurs the object of his aversion wretched.\index{suffering!definition!Stoics}\index{Epictetus}\index{Stoics|see{Epictetus, Seneca, \textit{and} Plutarch }}
\end{quote}
These are essentially a reformulation of the points given by the Buddha above. 
We can summarize these philosophical ideas as the following two causes for suffering, each with two variants:
\begin{enumerate} \label{fourpoints}
\item[1 a)] Not getting what one wants (Buddha, Epictetus) 
\item[\ b)] Something pleasant, which one would like to be present, is absent (Buddha)\footnote{I interpret ``separation'', also translated as ``dissociation'', in the quote by the Buddha not simply as absence but as absence of something one would like to be there since it is pleasant.} 
\item[2 a)]  Not being able to avoid what one is averse to, i.e., wants to avoid (Epictetus)
\item[\ b)] Something unpleasant is present (Buddha)
\end{enumerate}

Then, the definitions by the Buddha and Epictetus can be interpreted in terms of wanting (point~1) and aversion (point~2) only. 
Point~1 in particular defines the typical case of frustration, related to the reward loss already considered above.
Thus, we see that both the ideas of both the Buddha and Epictetus can be simply summarized as saying that suffering comes from frustration.
Using the term somewhat liberally, we can also call the suffering in point~2 frustration, since the desire to avoid something is frustrated.\footnote{ My logic is that if something pleasant is not present as in 1b, the point is that one actually \textit{wants} that pleasant thing to be present, so this is also a question of not getting what one wants, as in 1a. The same logic shows that 2a and 2b are really the same thing. The points 1b and 2b present the difficulty that they use the terms pleasant (or ``pleasing'' in the translation quoted above) and unpleasant (or ``displeasing''), much like the IASP definition of pain. (Alternative translations of these two words include ``beloved''/''unbeloved'' (Thanissaro Bhikkhu) ``loved''/''loathed'' (Nanamoli), ``liked''/''disliked'' (P.~Harvey), and indeed ``pleasant''/''unpleasant'' (Piyadassi Thera), given at {\tt https://www.accesstoinsight.org/tipitaka/sn/index.html\#sn56}.) I suggest the key here is that ``pleasant'' is here assumed to necessarily lead to wanting (and ``unpleasant'' to aversion), and thus the Buddha is really talking about desire or wanting and aversion. When he specifically mentions wanting at the end of the quote, that may be seen as a kind of summary of the two first sentences.}

\section{Two main kinds of  suffering}

Now I shall try to recapitulate the ideas above, both ancient and modern, as succinctly as possible. I think we only need to talk about two kinds of suffering, or rather two mechanisms producing suffering, namely:
\begin{enumerate}
\item \textit{Frustration} (e.g., Buddha, Epictetus, several neuroscientists\footnote{Among recent neuroscience, see especially \citet{papini2015behavioral}, but the idea has a long history in experimental psychology as reviewed by Papini et al. \add{A very similar point is made by Pascal in his famous formula: ``c'est \^etre malheureux que de vouloir et ne pouvoir.'' (\textit{Pens\'ees}, fragment Mis\`ere, 24).}})\index{suffering!definition!as frustration}\index{suffering!definition}
\item \textit{Threat}, especially to the intactness of the person, including their self-image (e.g., IASP, Cassell)
\end{enumerate}
 Based on this dichotomy, this book will develop two computational definitions of suffering, one each for these two aspects. \add{Two different definitions of frustration are given in Chapters~\ref{planning.ch} and \ref{rpe.ch}, respectively, while threat is defined  in Chapter~\ref{threat.ch}. Chapters~\ref{threat.ch} and \ref{summary1.ch} consider some connections between the two concepts, and in particular, how frustration is more important than threat from the viewpoint of designing interventions.\footnote{Van Hooft's theory can also be seen as combining these two aspects. While it starts with frustration, threats to the person can be seen as frustration of certain long-term goals, as explained in more detail in Chapter~\ref{self.ch}.\index{suffering!definition!van Hooft}}

}

The emphasis in the following chapters is, obviously, on information processing.
As already argued in the introduction, my main justification for talking about information-processing is that the framework of information-processing is a \textit{practically useful} way of describing suffering in the precise sense that it can tell us something about how to \textit{reduce} suffering. Information-processing is something that we can influence, something we can intervene on, so from a practical viewpoint, it is a very important aspect of suffering to investigate. 
Focusing on information-processing is also perfectly in line with the current emphasis on cognition in neuroscience and psychology; I see cognition as synonymous with information-processing.

\section{Using the pain system for broadcasting errors}\label{broadcasting.sec}

To conclude this chapter, I discuss some computational principles that explain why pain and suffering are so closely related. First, I propose that on a more abstract computational level, both pain and suffering are essentially  \textit{error signals}, messages that something is going wrong from the viewpoint of the goals and rewards of the system. Clearly, frustration signals that something went wrong in terms of not getting what one wants, and a similar case will be made for the  threat to the person in Chapters~\ref{self.ch} and \ref{threat.ch}.  Such error signals are in fact ubiquitous in artificial intelligence, where, in particular, they can be used for learning to choose actions better in view of maximizing rewards. We will see several kinds of error signals in the following chapters, and see how some of them can be interpreted in terms of suffering.\index{suffering!as error signalling}\index{mental pain|see{suffering}}\index{error signalling}\index{broadcasting}

Pain is thus an evolutionarily primitive form of an error signal. Its unique feature 
 is that pain signals are \textit{broadcast widely}\index{broadcasting}\index{error signalling!broadcasting|see{broadcasting}} in the information-processing system. This is important in an agent whose computation is distributed into different modules (whether processors or brain regions, see Chapter~\ref{control.ch}). For such an agent, it is necessary that any really important signal uses a special pathway that allows it to be broadcast to all, or most of, the modules. The pain signal is indeed broadcast widely to different neural systems, and the signal can change the behavior of the whole organism in terms of making it stop whatever it is doing and pay close attention to the pain. Furthermore, when an error signal drives the learning of the system, as we will consider in later chapters, it often needs to be observed by several of the modules, and such broadcasting is essential.\footnote{The broadcasting hypothesis is closely related to the global workspace theory by \citet{baars1997theater}, which will be treated in Chapter \ref{consciousness.ch}.  However, while Baars links broadcasting it to consciousness, I think the broadcasting does not have to be conscious, especially in the case of pain or suffering. The hypothesis is also closely related to the earlier interrupt theory of emotions explained in Chapter~\ref{emotions.ch}. The broadcasting might happen through several specific connections between brain areas, or through a central hub.}

Suffering is largely using the neural systems originally developed for physical pain, as already mentioned. This makes evolutionary sense if we think that computationally more sophisticated forms of error signalling, such as frustration, simply started using the evolutionarily older pain signalling pathway, adapting it for their own purposes. That was practical because the pain system already existed, and served well the purpose of broadcasting error signals to many brain regions. Using the physical pain system for signalling mental pain is thus a useful computational shortcut.\footnote{Such evolutionary arguments for using the same system were proposed by \citet{eisenberger2004rejection}, see also \citet{papini2015behavioral}. I am slightly confounding ``pathways'' and ``systems'' here: While the existing evidence is mainly about overlapping activation of certain brain regions, I am extrapolating the idea to the case of the signalling pathways.}\index{evolution!shortcuts}

Yet, merely talking about information-processing, as in a computer, may seem a rather incomplete description of suffering. Why does suffering \textit{hurt}, if it is merely a signal in an information-processing system? 
This is in fact exactly the same problem that we encountered with the IASP definition of pain above: Is it a subjective experience, or something more objective and measurable? 
The evolutionary rationale just described  explains why suffering ``hurts'' in the same way as physical pain:  the physical pain system is hijacked for the purposes of suffering or mental pain. (Perhaps this explains why we talk about mental ``pain'' in the first place.)  The very dichotomy of experience vs.\ objective measurements is thus exactly the same for pain and suffering, since it is a question of similar experiences and neural pathways. Nevertheless, explaining why physical pain actually subjectively feels like it does in the first place, is an extremely difficult question; it is intimately related to the question of consciousness, which we defer to Chapter~\ref{consciousness.ch}. We shall rather continue, in the next chapter, by elucidating the computational underpinnings of a particular form of error signal: frustration.

\chapter{Frustration due to failed plan} \label{planning.ch}

In this chapter, I propose the first model where frustration is a fundamental mechanism for suffering. It is assumed that an agent, whether a human or an AI, engages in \textit{planning} of action sequences in order to get to a desired \textit{goal state}. A state is here an abstraction of the properties such as location, context, and possessions of the agent. 
Frustration happens when the goal state is not reached in spite of the agent executing the planned sequence of actions.

I start by emphasizing the great computational difficulty of such planning of action; it is one reason why frustration happens. Another central concept here is wanting or \textit{desire}, which is a complex phenomenon we will return to several times in this book. As an initial definition, I consider desire as a computational process that suggests goals for the planning system. Finally, I discuss the importance of committing to a single plan, even in the presence of conflicting desires, based on Bratman's concept of \textit{intention}. This chapter lays out the framework in simple, largely intuitive terms; the main terms and concepts will be greatly refined in later chapters.

\section{Agents, states, and goals} \label{search.sec}

One may be tempted to think of an artificial intelligence as a system which just takes input, and processes information. However, information-processing in itself will actually be rather pointless unless it leads to some kind of visible output or action regarding the external world. In the very simplest case, action can just mean printing some text on a computer screen, so this is not necessarily a big leap.

In AI, the basic unit of analysis is often what is called an intelligent \textit{agent}, i.e.\ a system which not only processes information but also takes actions.\index{agent!definition}
In fact, the word ``agent'' literally means ``one that acts''. An intelligent agent can be artificial, such as a robot or an AI program, but the term also encompasses biological agents, that is, animals. In one extreme, an artificial agent could be just a program inside a computer, working in a virtual world with no physical body; actions would essentially consist of sending messages inside an information network. In the other extreme of human-like artificial agents, it could be a robot having a body with arms and legs; actions would include walking and grasping objects. In this book, we will see examples of both extremes---in addition to agents that actually are animals or humans.

Such an agent needs at least two things: perception and action selection.
Perception is actually a tremendously difficult task but we defer its discussion to Chapter~\ref{ml.ch} and especially Chapter~\ref{perception.ch}. To begin with, we assume perception is somehow satisfactorily performed, and consider the question of how the agent is to choose its actions.\footnote{For introductory textbooks on the topic, see \citet{russell2020artificial,poole2010artificial}.}

Perhaps the simplest and most intuitive approach to action selection is to think in terms of \textit{goals}. This is an introspectively compelling approach: We usually think of ourselves as acting because there are certain goals that we try to reach. That may be why it has also been a dominant approach in the history of AI, starting from around 1960. For example, a very simple thermostat has the goal of keeping room temperature constant; a cleaning robot has the goal of removing dust and dirt from the room.

The goals of humans are fundamentally determined by evolution, complemented by societal and cultural influences. In this book, I simply use the word ``evolution'' to describe the joint effect of biological evolution, culture and society. The assumption here is that the latter two are ultimately derived from biological evolution, although this is of course a controversial point. Fortunately, for the purposes of this book, the exact relationship between biology and culture is irrelevant: What matters is that the goals that humans strive for are largely, even if sometimes very indirectly, determined by some outside forces. Humans can set some intermediate goals, such as getting a job, but those are usually in the service of final biological or societal goals, such as being nourished or raising one's social status. In the case of AI, in contrast, the goals are usually supplied by its human designers. This may seem to be a fundamental difference between AI agents and humans, but we will see in later chapters that it may not matter very much; the human designer plays a role similar to evolution in terms of being an outside force. In any case, regardless of where the goals come from, the way they are translated into action may still be rather similar in both cases. 

\subsection{Modelling the world as states}

In order to choose its actions, the agent should have some kind of a model of how the world works, where the agent itself is seen as a part of the ``world'' modelled. The model expresses the agent's beliefs 
of what the world is typically like, and how the world  changes from one moment to another, in particular as a function of the actions the agent takes. 

\index{state of world}\index{robot!cleaning}
AI research uses a very abstract kind of a world model based on the concept of a \textit{state}, where each possible configuration of the world is one state. For example, if a cleaning robot is in the corner of a room, facing south, and there is only a single speck of dust in the room, at exactly two meters east from the robot, that is one state, we can call it state \#1. If the agent finds itself 10~cm further to the west, it is in another state, say state \#2; likewise, if another speck of dust appears in the room, that means the agent is in state \#3 (and if the speck of dust appears \textit{and} the agent is 10~cm further to the west, that is yet another state). 
In the simplest case, such a world model has states which are categorical, or discrete; in other words, there is a finite number of possible states.\footnote{\add{Or, if the number of states is infinite, it is restricted to what is called countably infinite, which is the ``smallest'' kind of infinity, meaning that the states can be indexed by integers.}}
This is a very classical AI approach, but we will see alternatives in later chapters.

Any effects of the agent's actions can now be described in terms of moving from one state to another, called state transitions. Indeed, in addition to knowing what the states of the world are like, the agent should know something about the transitions between the states caused by its actions. If it finds itself in state \#1 and decides to move forward, does it find itself in state \#2, or \#47, or something else? If the agent's world model can predict the effects of its actions in terms of transitions from one state to another, it is ready to start taking actions.

Using this formalism of states, the basic approach to action selection is that one of the states is designated as the goal state, by some mechanisms to be specified. The agent then uses its capabilities to reach the goal state, starting from its current state. 
That is surprisingly difficult, usually requiring complicated computation called \textit{planning},  which is a central concept we consider next.\footnote{It may seem very abstract to consider the world in terms of states. More insight might be obtained by taking an object-oriented viewpoint and considering the world a collection of objects. However, such a theory is still in its infancy \citep{diuk2008object,guestrin2003generalizing}, so we have to use the approach based on world states. Another question is whether the states should really be considered discrete-valued, such as indexed by integers. In fact, most current AI systems do not use such a categorical representation, but rather some kind of continuous-valued perceptual representation in a neural network, for example given by the outputs of certain neurons. However, the approach using discrete states is widely used in the theory and the textbooks because of its conceptual simplicity; the distinction is discussed in more detail in  Chapters~\ref{ml.ch} and \ref{dual.ch}.}

\section{Planning action sequences, and its great difficulty}

The fundamental problem in action selection is that you must actually select \textit{sequences} of short actions. For example, if the agent in question is you, and you decide to get something from the fridge in the kitchen, you need to take a step with your left foot, take a step with your right foot, repeatedly, until you finally can open the fridge door---which consists of several actions such as: raise your arm, grab the handle, pull it down, pull  the door, and so on. In some cases, you may easily know how to choose the right sequence, but it is not easy at all in many cases. For programming AI, it has turned out to be quite a challenge.

\begin{figure}
\begin{center}
\resizebox{\textwidth}{!}{\includegraphics{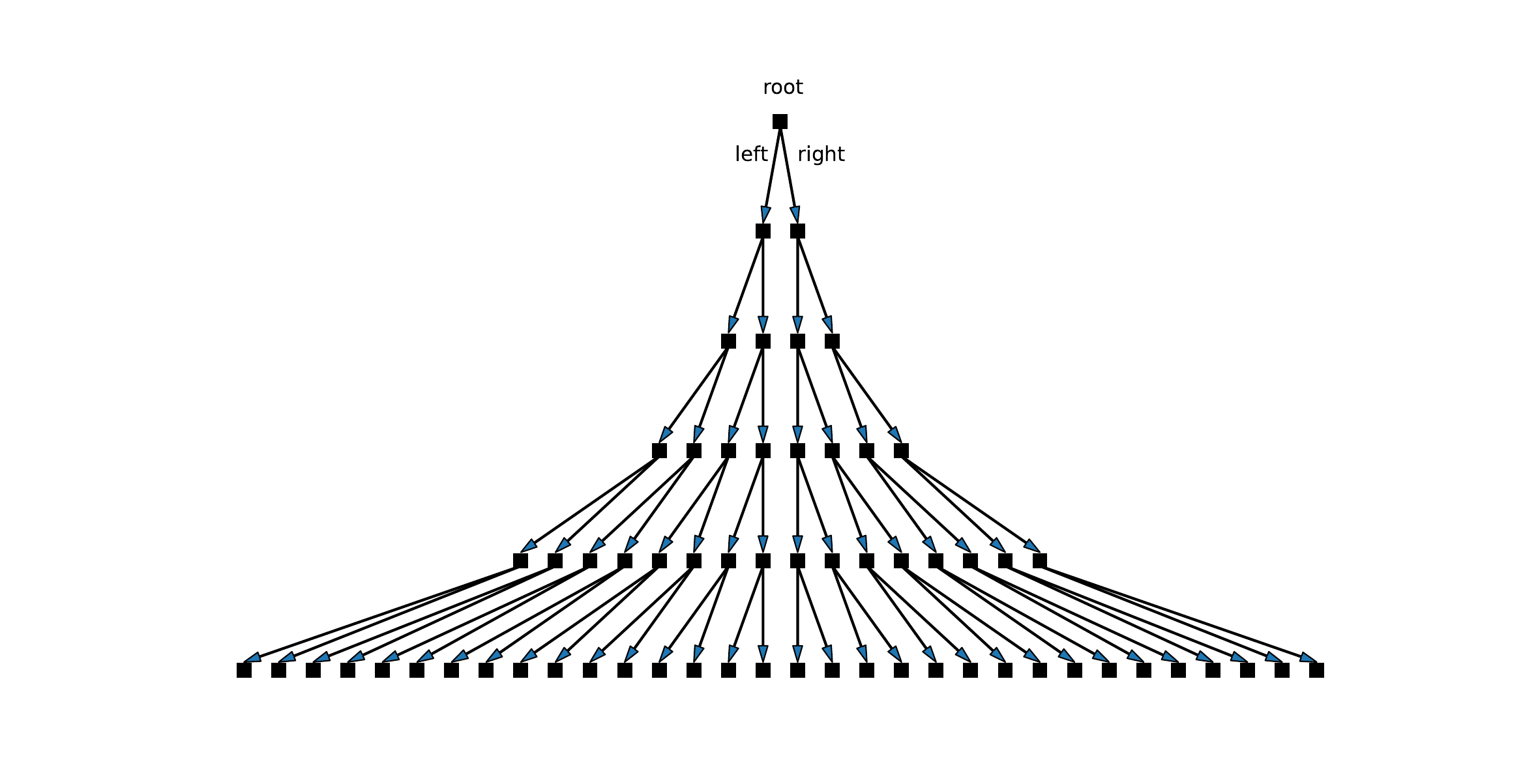}}
\end{center}
\caption{A search tree where the agent has two action options at every time point. They could be ``turn left'' or ``turn right'',  supposing the agent always makes a new decision when it finds itself in a new crossroads in a maze. The squares represent different states the agent can find itself in; the agent starts at the upper-most square in the figure (called root), and each action takes the agent one level down in this figure. The lines with arrows are the transitions to new states after every action taken. The crucial point here is that the number of different paths or plans it can take grows exponentially. After just 5 steps, as depicted here, the number of paths equals 32, that is, 2 to the 5th power. After 30 steps, it would be more than a billion.}
\label{searchtree.fig}
\end{figure}

This is known as the problem of planning in AI.\index{planning!definition}
Using the formalism of world states, action sequences can be represented graphically as what is called a tree (Figure \ref{searchtree.fig}). The ``root'' of the tree represents the state you're in at the moment. Any action leads to a branching of the tree, and depending on the action, you will find yourself on any of the new branches. (In this figure, taking an action means moving down in the tree, and we assume for simplicity that there are just two actions you can take at any time point). At the end of a given number of actions, or levels in the tree, you find yourselves at one of those states which are depicted at the outer ``leaves'' of the tree. Of course, the tree continues almost forever since you can take new actions all the time, but to keep things manageable, we consider a tree of a limited depth.

Let's now assume that the agent has been given a goal state by the programmer. It would be one of the states at the lowest level of the tree.
The central concept here is \textit{tree search}; many classical AI theories see intelligence as a search for paths, or action sequences, among a huge number of possible paths in the action tree. In particular, the planning system tries to find a path which leads from the current state to the goal state. Such search may look simple, but the problem is that with such paths or action sequences, the number of possibilities grows exponentially.\index{tree search}
If you have, at any single time point, just two different actions to choose from, then after 30 such time points you have more than a billion (precisely 2 to the power of 30) possible action sequences to choose from. What's worse is that typically an AI would have many more than just two possible courses of action at any one point. The computations involved easily go beyond the capacity of even the biggest computers or brains. So, it may be impossible to ``look ahead'' more than a couple of steps in time. 

The difficulty of such planning may be difficult for humans to understand since evolution has provided various tricks and algorithms that solve the problem quite well, as we will see below. We may only be able to grasp the difficulty of planning in some slightly artificial examples such as the search tree above. One of the more realistic examples would be planning a route between two points. Say you find yourself in a random location in Paris and want to go to the Eiffel Tower using public transportation. Even if you remembered every detail of the metro map as well as the geography of Paris itself, you would still need quite a lot of thinking, that is, computation. Which metro station should I walk to, or should I perhaps use the bus? What is the best itinerary once inside the metro station? It is not surprising that people tend to use mobile phone apps to solve this problem.\footnote{Planning might actually seem to be very easy in a simple illustration like in Figure~\ref{searchtree.fig}, since all you need is to \textit{start} at the \textit{goal} state, and go \textit{backwards} in the search tree until you arrive at the root; thus you have found the path from the root to the goal. The reason why this does not work in practice is that in reality there are many overlapping trees, each starting from a \textit{different} root state, and each goal state can be reached starting from a number of different roots. So, you cannot go backwards because you don't know which tree to follow. You can see this in the example of planning a route between two points in Paris: It may help a bit to start calculating backwards from Eiffel Tower, but you cannot just backtrack in a tree because the possible routes going ``back'' from the Eiffel Tower are as numerous as the routes you can start going ``forward'' from your current location; routes computed ``backwards'' from the Eiffel Tower can take you anywhere in Paris, not just your current location.}

Board games are an extreme example of the difficulty of planning. Humans playing chess have great difficulties in thinking more than one or two moves ahead. The search tree has a lot of branches at every move because there are so many moves you can take. Even worse, your opponent can do many different things. (The uncertainty regarding what your opponent will do further adds to the complexity, but that is another story.)\index{planning!computational difficulty}

A lot of the activity we would casually call \textit{thinking} is actually some kind of planning. If you are thinking about where to go shopping for a new electronic gizmo, or how to reply to a difficult message from your friend, you are considering different courses of action. Basically, you're going through some of the paths in the search tree. Interestingly, a lot of such thinking or planning happens quite involuntarily, even when you're supposed to be doing something else, a topic to which we will return in Chapter~\ref{replay.ch}. 
\index{thinking!as planning}

\section{Frustration as not reaching planned goal}

Equipped with this basic framework for action selection, we are ready to define frustration in its most basic form. We start by considering one part of the Buddha's definition of suffering mentioned above (page~\pageref{fournoble1}): ``not to get what one wants''. This is in fact a typical dictionary definition of frustration. 
To achieve a deeper computational understanding of the phenomenon, we need to integrate this with the framework of planning.

Just like AI, complex organisms such as humans engage in planning: Based on their perception of the current environment, they try to achieve various goals by some kind of tree search. For such organisms, it is vital to know if a plan failed, so that they can re-plan their behavior, and even learn to plan better in the future. %
We thus formulate the basic case of frustration as \textit{not reaching a goal that one had planned for}, and the ensuing error signal.\index{frustration!based on planning}

This initial definition will be refined and generalized in later chapters, where we will see how central error signals are to any kind of learning. For example, a neural network that learns to classify inputs, or predict the future, is essentially minimizing an objective function which gives the error in such classification or prediction. Frustration can be seen as a special case of such error signalling: It signals that an action plan failed.  In complex organisms like humans, which are constantly engaged in planning, frustration is an extremely important learning signal, and the basis of a large part of the suffering. It should also be noted that in some contexts, frustration rather refers to the resulting unpleasant mental state; that is, frustration refers to the actual suffering instead of the cause for suffering. In this book, the word is in both of those meanings.\footnote{This ambiguity is to some extent justified by the ambiguity of how the term is used in the literature, and some dictionaries explicitly list these two meanings for the term, e.g.\ {\tt https://psychologydictionary.org/frustration/}.}

\section{Defining desire as a goal-suggesting mechanism} \label{desire.ch}

However, there is a slight inconsistency here: Frustration was actually defined as not getting what one \textit{wants} in Chapter~\ref{suffering.ch}. How is this related to our computational formulation based on planning above? In other words, what exactly is wanting, or desire, in a computational framework like ours?\index{frustration!based on desire}

In everyday intuitive thinking, action selection is indeed supposed to be based on wanting, or desires: An agent takes an action because it \textit{wants} something, and it thinks it is reasonably likely to achieve or obtain it by that action. I choose to go to the fridge because I want orange juice. However, the account earlier in this chapter made no reference to the concepts of desire or wanting. 
In AI, the term ``desire'', which I consider synonymous with ``wanting'', can actually be used in a couple of different meanings.
\index{wanting|see{desire}}

In the very simplest definition, if the agent has a goal to plan for, one could simply say the agent ``wants'' to reach the goal state; desires would essentially be the same as goals.
In such a meaning, desire is a kind of purely rational,  ``cold'' evaluation of states and objects. However, the word has many more connotations in everyday language. Desire also has an affective aspect we could call ``hot'', in which we are ``burning with desire'', unable to resist it. 
\index{desire!definition}

A definition that is a bit more in the direction of ``hot'' can be obtained by considering desire as a specific computational process inside the agent.
To begin with, we can adopt a definition of desire as a ``psychological state of motivation for a specific stimulus or experience that is anticipated to be rewarding''.\footnote{ \citep{papies2015grounding}. For a number of alternative definitions see \citep{schroeder2017desire}.} 
While a ``psychological state'' may mean different things, here we consider it as a particular kind of information processing being performed---another meaning would be related to conscious experience which we treat in Chapter~\ref{consciousness.ch}. 
In practical terms, desire is often triggered by the perception of something that is rewarding to possess.\footnote{
  ``[D]esire arises when an internal or external cue triggers a simulation, or partial re-enactment, of an earlier appetitive experience that was rewarding.''\citep{papies2015grounding}. I'm using ``reward'' in this chapter in a non-technical sense, which will be refined in Chapter~\ref{rpe.ch}.}
Such perception of an object often means that the agent should be able to get the object after a rather short and uncomplicated action sequence: If you see something, it is likely to be within reach.

From the viewpoint of information-processing, we thus define desire as: \textit{A computational process suggesting as the goal a state that is anticipated to be rewarding and seems sufficiently easily attainable from the current state.}\label{desiredef1}\index{desire!definition} 
I want to emphasize that  I am considering desire as a particular form of information-processing: Desire is not simply about preferring chocolate to beetroot, nor is it merely an abstract explanation of the behavior where I grab a chocolate bar. It is sophisticated \textit{computation} that is one step in the highly complex process that translates preferences into planning and, finally, into action.

The starting point for that processing is that your perceptual system, together with further computations, estimates that from the current state, you can relatively easily get into a state of high reward---the exact formalism for ``rewards'' will be introduced in Chapter~\ref{rpe.ch}. This realization will trigger, if you are properly programmed, further computational processes that will try to get you in that desired state by suggesting it as the goal for your planning system. 
When all this happens, you want to be in the new state, or have a desire for that new state, according to the definition just given. 
For example, if chocolate appears in your visual field, your brain will compute that the state where you possess the chocolate is relatively easy to reach, and produces high reward; so it will choose the chocolate-possessing state as a possible goal and input that to the planning system. 

Therefore, the definition of desire just given shows how the intuitive definition of frustration as not getting what one wants and the computational definition of not reaching the goal are essentially the same thing. This definition  also solves a question which many readers must have asked while reading this chapter: Where do the goals for planning come from? In a very simple AI, there might be just a single goal, or a small number of them, defined by the programmer. But for a sophisticated agent, that is certainly not the case: The number of possible goals for a human agent is almost infinite. Here, we define desire as a computational process that suggests new goals to the planning system, so this is where the goals come from. (More details on how the desire system could actually choose goals will be given in Chapter~\ref{dual.ch}, which also considers a different aspect of desire related to its interrupting and irresistible quality.)

A closely related concept is \textit{aversion}, which is
in a sense the opposite of desire.\index{aversion!definition} 
However, from a mathematical viewpoint, aversion is very similar to desire: The agent wants to \textit{avoid} a certain state (or states) and wants to be in some other state.\footnote{A linguistic confusion is created in English and many other languages in which it is commonplace to say ``I don't want X'', where X might be drilling noise in your office, or flies in your bedroom. What this actually means is that you want those things to be \textit{absent}: It does \textit{not} simply mean that you merely \textit{refrain} from wanting that noise of the flies.  You want ``not X'', the opposite or absence of X, which is in fact the meaning of aversion.} For example, the agent wants to be in a state in which  some unpleasant object is not present.  Thus, it is really a case of wanting and desire, just framed in a more negative way. I do not use the term aversion very much in this book since it is mathematically contained in the concept of desire.\footnote{\add{One difference is, though, that since aversion suggests as the goal \textit{all} the states where the unpleasant object is not present, it actually operates with a very large set of goal states.}} Whenever I use the word ``desire'', aversion is understood to be included.

\section{Intention as commitment to a goal} \label{intention}

We have seen that a desire is something that suggests the goal of the agent.\index{intention!definition}
Note that I'm not saying that desire \textit{sets} the goal, but it \textit{suggests} a goal to the planning system. This difference is important because there might be conflicting goals; you don't grab the chocolate every time you have desire for it. 
The agent needs to choose between different possible objects of desire. This is particularly important because attaining the desired goal state often takes time: The whole plan has to be executed from the beginning till the end, and new temptations---activations of the desire system which suggests new states as possible goals---may arise meanwhile. Some method of arbitrating between different desires is necessary.

Suppose the desired state for a monkey is where the monkey has eaten a banana. The banana is currently high up the tree which is in front of the monkey, so the monkey needs to perform a series of actions to reach that desired state: it must climb up the tree, take the banana, peel it, and eat it. The monkey must thus figure out the right sequence of actions to reach the desired state--- this is just the planning problem discussed above---and launch its execution.

But, suppose the monkey suddenly notices another banana in another tree near-by. Its desire system may suggest that the new banana looks like an interesting goal.
The monkey now faces a new problem: Continue with the current banana plan, or set the new banana as a new goal?
It may be common sense that after the monkey has launched the first banana plan, the monkey should, in most cases, persist with that plan until the end. The monkey should not start pondering, halfway up the first tree, whether it actually prefers to get the other banana in the other tree, even if it looks a bit sweeter.
The key idea here is \textit{commitment} to the current plan, and thus to a specific goal. 

The reason why commitment is important comes fundamentally from computational considerations. Since computing a plan for a given goal takes a lot of computational resources, it would be wasteful to abandon it too easily in favor of a new goal. It might be wasteful even to just consider alternative goals seriously, because that would entail a lot of computation to produce alternative plans. %
The agent must settle on one goal and one plan and execute it without spending energy thinking about competing goals.\footnote{A lot of physical energy would also be wasted if the monkey is already half-way up the tree and then decides to go for the other banana. But arguably that waste of energy would be taken into account by the monkey in its planning, so it does not need to be evoked as a separate reason for commitment. I think we can take here a viewpoint considering purely computational resources: Even if the monkey is intelligent enough to eventually understand this waste of physical energy after some thinking, it would still spend a lot of time and computational resources to reach that conclusion if there were no commitment mechanisms.}

Another utility of commitment is that the agent has a better idea of what will happen in the future, and it can start planning further actions, e.g.\ a plan on what to do after reaching the goal of the current plan. So, while the monkey is climbing up the tree, it \textit{is} a good idea to start thinking about the best way of getting the banana in the other tree \textit{after} having grabbed the first banana. That would be planning the long-term future after completing the execution of the current plan; perhaps the monkey can directly jump to the other tree from the location of the first banana. Such long-term plans would obviously collapse if the monkey didn't first get the first banana due to lack of commitment, being distracted by yet another thing.

Commitment to a goal is also called \textit{intention} in AI, and leads to an influential AI framework called belief-desire-intention (BDI) theory.\index{belief-desire-intention theory}
``Belief'' refers here mainly to the results of perception, which give rise to desires. BDI theory argues, as I just did,  that it is  important to have of intentions as commitment to specific goals, on top of beliefs and desires.\footnote{\citep{bratman1987intention,cohen1990intention,rao1991modeling}. See \citet{mulder2018intentions,brodaric2020foundations} for recent work and slightly different formulations. For a modern neuroscientific approach see \citet{oreilly2014goal} which proposes something very similar using its concepts of ``goal engagement'', and ``active goal''. Note that the word ``intention'' has different meanings in the literature, and in particular this definition of the word is quite different from the meaning typically associated with ``intentionality'' \textit{\`a~la} Brentano.  On the other hand, in the literature, there is some ambiguity on whether intentions are commitments to desires, goals, or plans. I consider them as commitments to goals.}
Of course, there must be some limits to such commitment: If something unexpected happens, the goal may need to be changed. If a tiger appears, the monkey  cannot persist with the goal of just eating a banana. Chapter~\ref{emotions.ch} will consider the importance of emotions such as the fear aroused by the tiger as one computational solution.\footnote{Some AI systems solve this problem by planning everything from scratch at regular intervals, but that is unlikely to be possible in a real-time environment where the time needed for planning is the main bottleneck. In fact, my treatment here may not do full justice to Bratman's original definition of intention, where a plan is actually composed of several intentions. Such a definition creates more flexibility for behavior in the sense that even if the circumstances change (or the circumstances were unpredictable to start with), the behavior may flexibly move from one path to another by triggering an alternative sequence of intentions. The definition I use here is more similar to the later AI developments of the concept, cited in the preceding footnote.}

The concept of intentions has important implications for suffering, as will be discussed in detail in later chapters. To put it simply, I will propose that frustration and suffering are stronger if an intention is frustrated, as opposed to frustration of a simple desire as in the basic definition.

\pagebreak

\section{Heuristics can help in planning}

Still, we have not yet solved the central problem regarding the planning system. We saw above that because of the huge number of possible paths in planning, a complete tree search is quite impossible in most cases. Is planning then impossible?
Fortunately, there are  a couple of  tricks and approximations that can be used to find reasonable solutions to the planning problem. Here we first consider  what is called \textit{heuristics}, while a more sophisticated solution is given in the next two chapters. These solutions also have important implications for the definition of frustration, and understanding suffering.\index{heuristics}

A heuristic means some kind of method for evaluating each state in the search tree, usually by giving a number that approximately quantifies how good it is, i.e.\ how close to the goal it is. The point is that a heuristic does not need to be exact---if it were, we would  have already solved the problem. It just gives a useful estimate, or at least an educated guess, of how ``good'' a state is.\footnote{A very general definition of a heuristic, not only applicable to the tree search problem, is given by \cite{gigerenzer2011heuristic}: A heuristic is a strategy that ignores part of the information, with the goal of making decisions more quickly, frugally, and/or accurately than more complex methods.}

Sometimes, it is quite simple to program some heuristics in an AI agent.
Consider a robot whose goal is to get some orange juice from the fridge and deliver it to its human master. 
Clearly, when the robot has orange juice in its hand, it is rather close to the goal; we could express that by a numerical value of, say, 8. If it is, in addition, close to its master, it is very close to its goal, say a value of 9. The most important thing is, however, to assist the robot at the beginning of the search, and that is where the heuristic is the most powerful. So, we could say that when the robot is close to the fridge, the heuristic gives a value of 2. When it has opened the fridge, the value is 3, and so on.

With such heuristics, the search task would not require that much computation. The robot just has to figure out how to get to some easily reachable state with a higher heuristic than the current state. Assuming the robot starts at an initial state with heuristic value 0, it would quickly compute that what it can achieve rather easily is a state of heuristic value of 2, by going to the fridge. The length of the tree to be searched for is thus much shorter, i.e.\ much fewer actions steps need to be taken in that subproblem. Once there, it only has to figure out how to open the door to get to the state with heuristic value of 3. Thus, the heuristic essentially divides a long complex search task into smaller parts. Each of these parts is quite short, so the exponential growth of the number of branches is much less severe.\footnote{As a simple (and only approximative) numerical example, think of dividing a tree of length 20 into two parts. Each part of length 10 has $1,024=2^{10}$ states, so the two search trees have total of $2,048$ states. This is much less that the original tree with $2^{20}=1,048,576$ states.}

There is one famous success of AI where such tree search with heuristics was hugely successful: The Deep Blue chess-playing machine,\citenew{campbell2002deep}\index{Deep Blue}\index{games!chess} which beat the chess world champion, for the first time, in 1997. Its main strength was the huge number of sequences of moves (i.e.\ paths in a search tree) it was able to consider, largely because it was based on purpose-built, highly parallel hardware that was particularly good in such search computations on the chessboard. But its success was also due to clever heuristics, the main one being called ``piece placement'', computed as the sum of the predetermined piece values with adjustments for location, telling how good a certain position is. (In chess, the state is the configuration of all the pieces on the board, and called a ``position'' in their jargon.)

Evolution has also programmed a multitude of heuristics in animals. Think about the smell of cheese for a rat. The stronger the smell, the closer the rat is to the cheese. The rat just needs to maximize the smell, as it were, and it will find the cheese. No complex planning is needed---unless there are obstacles in the way.\footnote{In fact, we see here that there is some intricate connection with heuristics and desires. When the rat smells the cheese, surely a desire for cheese appears in its system. See  Chapter~\ref{emotions.ch}, and in particular footnote~\ref{somaticmarkersfn}, on how the same computations can sometimes be interpreted as heuristics or desires.}

However, the crucial problem is how to find such heuristics for a given planning problem. In fact, this is a very difficult problem, and there is no general method for designing them. Nevertheless, there is a general principle which has been found tremendously useful in modern AI, and can be used here as well: \textit{learning}. Modern AI is very much about using learning from data as an approach to solving the problem of programming intelligence. In the case of planning, it turns out that a general approach for solving the planning problem is to learn to rate the states, i.e.\ learn to associate some kind of heuristic to each world state. This is why in the next two chapters, we delve into the theory of machine learning. Its specific application to solving the planning problem will be considered in Chapter~\ref{rpe.ch}, where we also consider a different approach to defining frustration.

\chapter{Machine learning as minimization of errors} \label{ml.ch} 

In this chapter, we will go through some of the basics of the backbone of modern AI: machine learning. 
Such AI crucially relies on learning from incoming data---which is also true of the brain. 
Machine learning is most often used in conjunction with neural networks, which are powerful function approximators, loosely mimicking how computations happen in the brain.
We will also consider an alternative, older approach to intelligence based on symbols, logic, and language,
which is now called ``good old-fashioned AI''. (The preceding chapter with its discrete, finite states, was an example of this latter approach.)

A central message in this chapter is that learning is often based on some measure of error. Minimizing such errors means optimizing the performance of the system. The fundamental importance of computing and signalling such errors is important in future chapters where such errors are directly linked to suffering, generalizing the concept of frustration. I conclude this chapter by claiming that any kind of learning from complex data can lead to quite unexpected results, something that the programmer could not anticipate.

\section{Neurons and neural networks}

Modern AI is based on the observation that the human brain is the only ``device'' we know to be intelligent for sure and without any controversy. It is actually not easy to define what ``intelligence'' means, and I will not attempt to do that in this book.\footnote{For standard textbook expositions on the definition of (artificial) intelligence, see e.g.\ \citet{russell2020artificial,poole2010artificial}. For particular viewpoints relevant to our discussions later, see e.g.\ \citet{brooks1991intelligence,legg2007universal}.} Yet, nobody denies that the brain is intelligent---or, to put it another way, it enables us to behave in an intelligent way. The brain is intelligent as if by definition; it is the very standard-bearer of intelligence. Therefore, if you want to build an intelligent machine, it makes sense to try to mimic the processing taking place in the brain.\index{intelligence!definition}

\subsection{Neurons as tiny processors}

\begin{figure}
\begin{center}
\resizebox{0.6\textwidth}{!}{\includegraphics{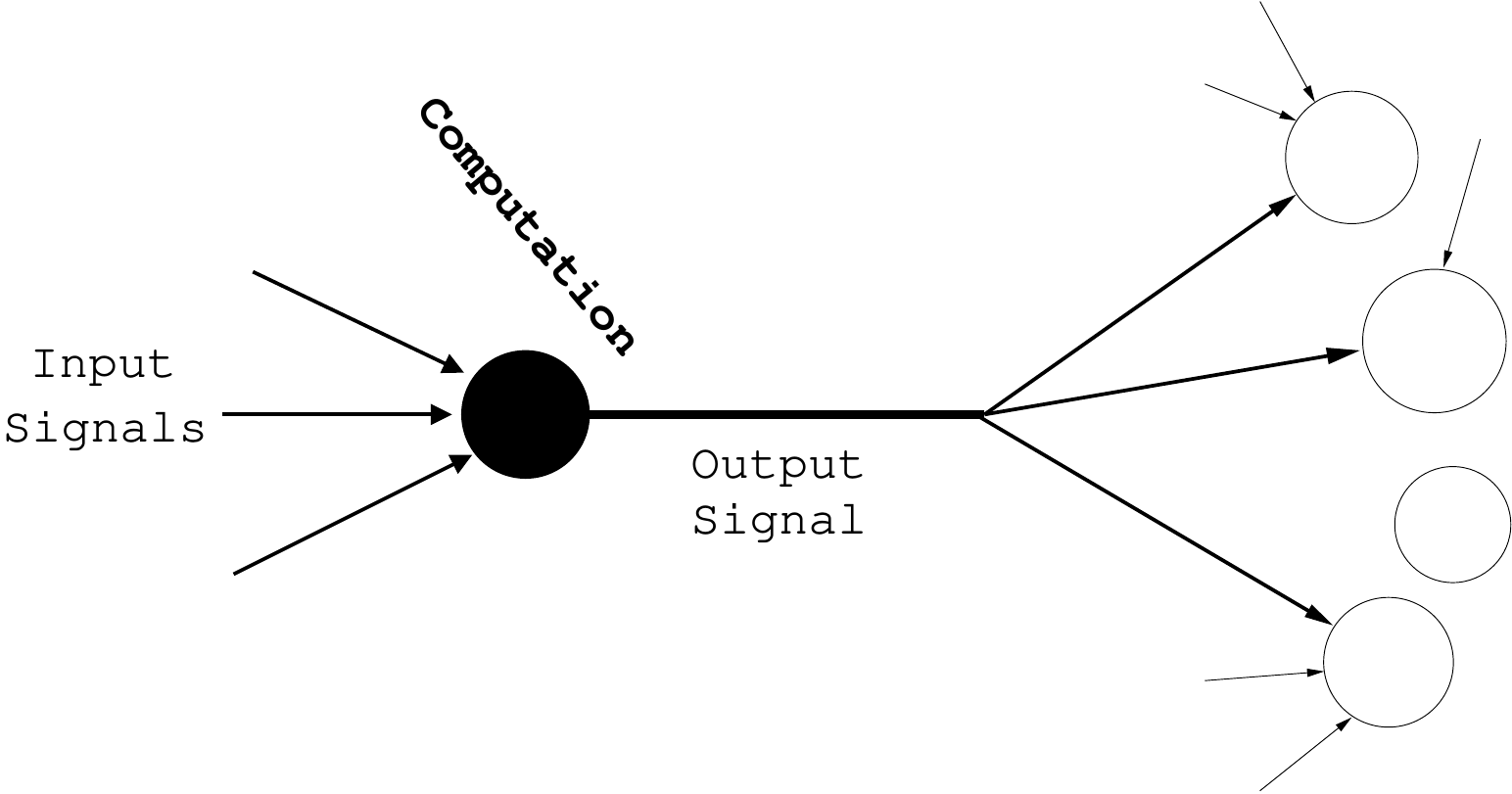}}
\caption{A schematic of a neuron. Input signals coming from other neurons (from the left) are received by the neuron (depicted by the black disk). Computation happens inside the neuron, and the resulting output signal is transmitted to a number of other neurons (depicted by white disks)  on the right-hand side. The other neurons simultaneously receive input signals from many further neurons outside of this figure (depicted by further arrows).}
\label{neuron.fig}
\end{center}
\end{figure}

The computation in the brain is done by specialized cells called neural cells or neurons.\footnote{It has also been claimed that other types of cells in the brain could also participate in computations, in particular glial cells \citep{perea2009tripartite}. However, AI systems typically mimic neurons only.} A schematic picture of a neuron is in Fig.~\ref{neuron.fig}. A neuron receives input from other neurons, processes that input, and outputs the results of its computations to many other neurons. There are tens of billions of neurons in the human brain.\index{neuron}
Each single neuron can be seen as a simple information-processing unit, or a processor.\footnote{I do not attempt to define ``information processing''\index{information processing!definition}\index{computation!definition} in any rigorous way in this book. It is a very general concept with many meanings, and attempting to define it in a way that is both general and rigorous enough seems hopeless to me. I use ``computation'' simply as a synonym for information processing.}

All these tiny processors do their computations simultaneously, which is called \textit{parallel} processing.\index{information processing!parallel|see{parallel processing}}\index{parallel processing} 
The opposite of parallel processing is serial processing, where a single processor does various computational operations one after another---this is how ordinary CPU's in computers work. Another major difference between the brain and ordinary computers is that processing in neurons is also \textit{distributed}.\index{information processing!distributed|see{distributed processing}}\index{distributed processing}
This means that each neuron processes information quite separately from the others: It gets its own input and sends its own output to other neurons, without sharing any memory or similar resources. Compared to an ordinary PC, the brain is thus a \textit{massively parallel and distributed} computer. Instead of a couple of highly sophisticated and powerful processors as found in a PC, the brain has a massive amount---billions---of very simple processors. (Parallel and distributed processing is discussed in detail in Chapter~\ref{self.ch}.)

While the actual neurons are surprisingly complex, in AI, a highly simplified model of a real neuron is used. Sometimes, such a model is called an artificial neuron to distinguish it from the real thing, but for simplicity, we call them just neurons. Like a real neuron, an artificial neuron gets input signals from other neurons, but each such input signal is very simple, just a single number; we can think of it as being between zero and one, like a percentage. Based on those inputs, the neuron computes its output which is, again, a single number. This output is, in its turn, input to many other neurons.\index{neuron}

In such a simple model, the essential thing is to devise a simple mathematical formula for computing the output of the cell as a function of the inputs.
In typical models, the output is computed, essentially, as a \textit{weighted sum} of the inputs. 
The weights used in that sum are interpreted, in the biological analogy, as the strengths of the connections between neurons, or the incoming ``wires''  on the left-hand-side of Fig.~\ref{neuron.fig}. These weights can get either positive or negative values:  The weight is defined as zero for those neurons from which no input is received. The weighted sum is usually further thresholded (i.e.\ passed through a nonlinear function) so that the output is forced to be between zero and one. In the brain, the connections are implemented through small communication channels called synapses, which is why the weights can also be called ``synaptic''.

Importantly, these weights can be interpreted as a pattern, or a template, which the neuron is sensitive to. Thus, a neuron can be seen as a very simple \textit{pattern-matching unit}. The neuron gives a large output if the pattern of all the input signals matches the pattern stored in the vector of weights or connection strengths. 
As an illustration, consider a neuron that has a weight with the numerical value +1 for inputs from another neuron, let's call it neuron A, as well as a zero weight from neuron B, and a weight of -1 for inputs from neuron C. %
This neuron will give a strong output signal when neuron A gives a large output signal, and the neuron C gives a small signal, while it does not care what the output of neuron B might be. %

\begin{figure}
\begin{center}
\resizebox{7cm}{!}{{\includegraphics{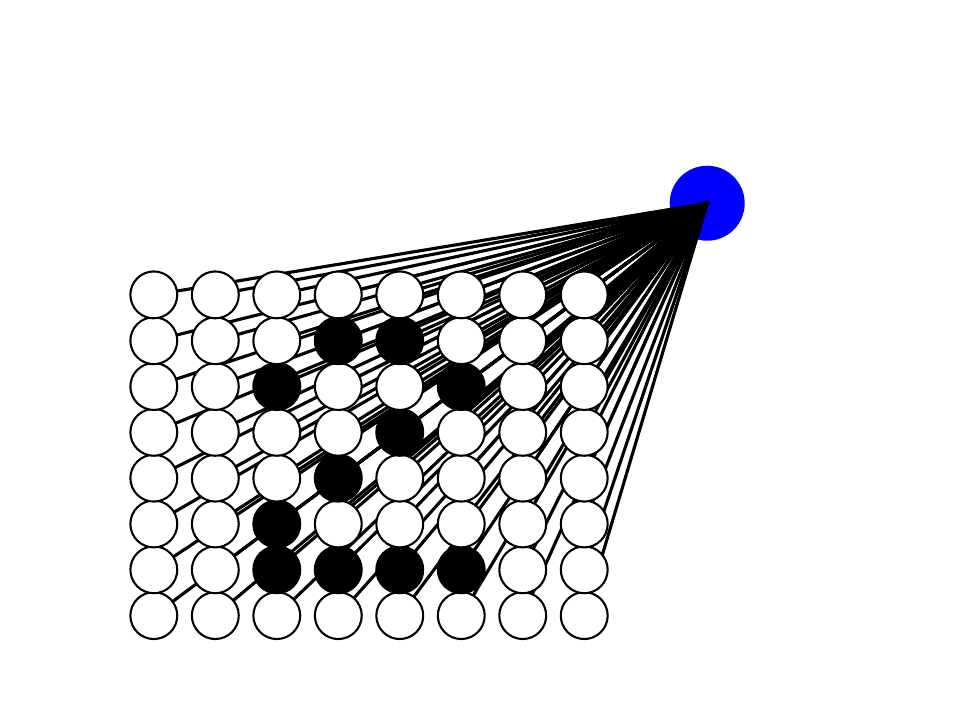}}}
\caption{Synaptic weights of a neuron illustrated. Pixels shown in black have a connection strength of $-1$ to the neuron (depicted in blue), while pixels shown in white have a connection strength of $+1$. The neuron is maximally activated when the input corresponds to the stored pattern, which is a picture of the digit ``2''.}
\label{digitneuron.fig}
\end{center}
\end{figure}

Such pattern-matching is obviously most useful in processing sensory input, such as images. Consider a neuron whose inputs come from single pixels in an image. That is, the input consists of the numerical values of each pixel, telling how bright it is, i.e.\ whether it is white, black, or some sort of gray. Then, we can plot the synaptic weights as an image, so that the gray-scale value in each pixel in this plot is given by the corresponding synaptic weights. If they are -1 or +1 as in the previous example, we can plot those values as black and white, respectively. A neuron could have synaptic weights as in Fig~\ref{digitneuron.fig}. Clearly, this neuron is specialized for detecting a digit, in particular number two.

Of course, in reality, to recognize digits (or anything else) in real images, things are much more complicated. For one thing, the pattern to be recognized could be in a different location. If the digit is moved just one pixel to the right or to the left, the simple pattern-matching above does not work anymore, and the neuron will not recognize the digit. Likewise, if the digit were white on a black background instead of black on a white background, the same pattern-matching would not work. To solve these problems, we need something more sophisticated.

\subsection{Networks based on successive pattern-matching}

\begin{figure}
\begin{center}
\resizebox{\textwidth}{!}{{\includegraphics{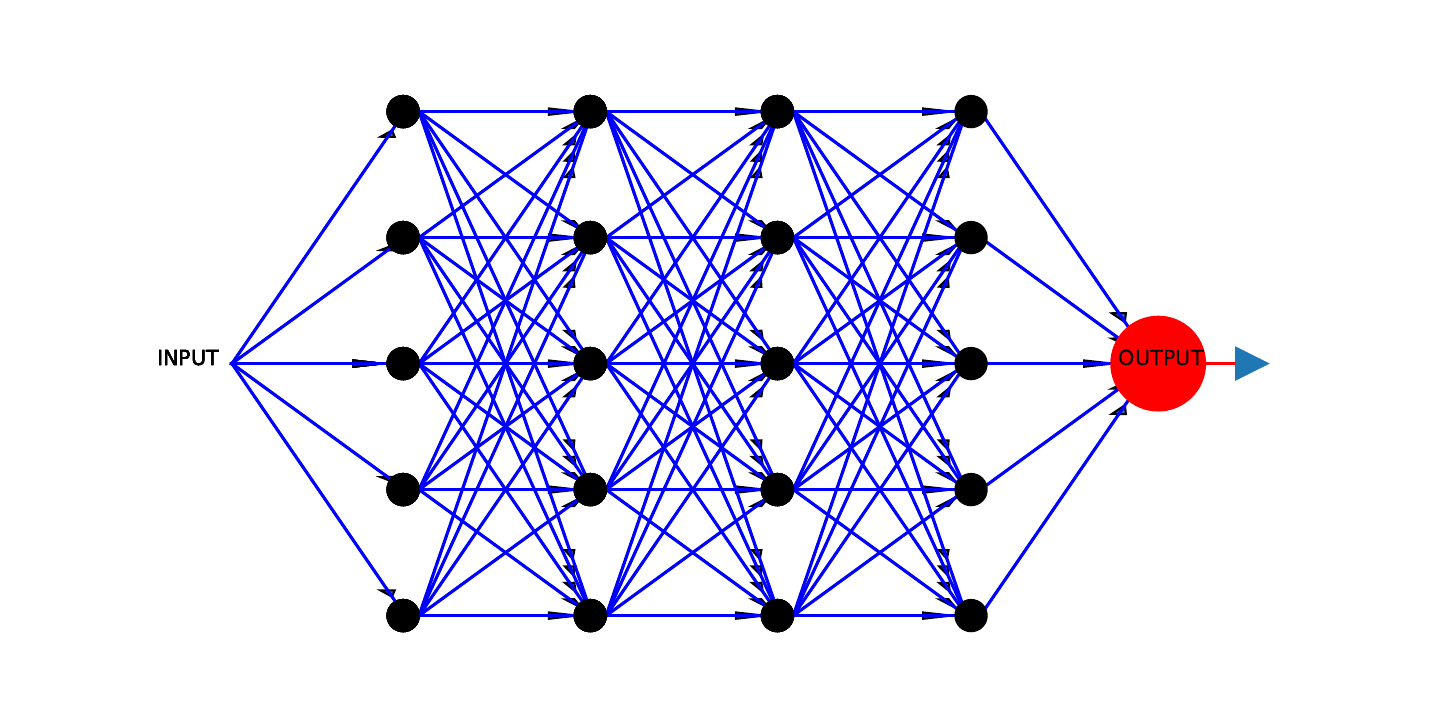}}}
\caption{An illustration of a neural network. The information enters the system in the first ``layer'' of black neurons on the left-hand side. It is processed by several successive layers, each having five neurons illustrated by small black disks. Each neuron is doing a simple pattern-matching computation on its inputs, and transmitting the result of that computation to the next layer to its right, along the wires depicted in blue. The information is transmitted from the left (input) to the right (output). As a result of many neurons (in reality, thousands or even millions), the total computation of the network is highly complex and can achieve sophisticated object recognition, as well as many other kinds of computations. }
\label{neuralnetwork.fig}
\end{center}
\end{figure}

Building a \textit{neural network} greatly enhances the capabilities of such an AI, and solves the problems just mentioned.
A neural network is literally a network consisting of many neurons. Networks can take many different forms, but the most typical one is a hierarchical one,  
where neurons are organized into layers, each of which contains several cells, actually quite a few sometimes. The incoming input first goes to the cells in the first layer which compute their outputs and send them to neurons in the second layer, and so on. This is illustrated in Figure~\ref{neuralnetwork.fig}.\index{neural networks!definition}\index{neural networks!learning|see{learning}}

From the viewpoint of pattern-matching, we can say that such a network performs successive and parallel pattern-matching. The input is first matched to all the patterns stored in the first-layer neurons, and those neurons then output the degrees to which the input matched their stored patterns or templates. These outputs are sent to the next layer, whose neurons then compare the pattern of first-layer activities to their templates. So, the second-layer patterns are not patterns of original input (such as the pixels of an image) but patterns of the first-layer activities, which form a description of the input on a slightly more abstract level. This goes on layer by layer, so that each neuron in each layer is ``looking for'' a particular kind of pattern in the activities of the neurons in the previous layer. The patterns are always stored in the synaptic weights of the neurons.

The utility of such a network structure is that it enables much more powerful computation. For example, consider the problem of a digit which could be in slightly different locations in the image, as mentioned above. The problem of different locations can be fixed by having several neurons in the first layer, each of which matches the digit in one possible location. All we need in the second layer is a neuron that adds the inputs of all first-layer neurons, and thus computes if any of them finds a match. With such a scheme, the second-layer neuron is able to see if there is a digit ``2'' at any location in the image.\index{vision}\index{pattern recognition|see{vision}} 

\section{Finding the right function by learning} 

Now, a crucial question is how the synaptic connection weights can be set to useful values.
In modern AI, the synaptic weights between neurons are \textit{learned from data}, hence the term \textit{machine learning}.\index{learning!definition with neural networks} Learning is really the core principle in most modern AI. Especially in the case of neural networks, it is actually difficult to imagine any alternative. How could a human programmer possibly understand what kind of strengths are needed between the different neurons? In some cases, it might be possible: in image processing, the first one or two layers do have rather simple intuitive interpretations, as we have alluded to above. However, with many layers---and neural networks can have thousands of them--- the task seems quite impossible. Hardly anybody has seriously tried to design such neural networks by fixing the weights manually, based either on some theory or intuition.

In the brain, the situation is quite similar. There is simply not enough information in the genome---which is somewhat analogous to the programmer here---to specify what the synaptic connection strengths should be for all the neurons. It would hardly be optimal anyway to let the genes completely determine the synaptic connections, since animals live in environments that may change from one generation to another, and some individual adaptation to circumstances is clearly useful. What happens instead is that the synaptic weights change as a function of the input and the output of the neuron, or as a function of perceptions and actions of the organism. The capability of the brain to undergo such changes is called ``plasticity'', and those changes are the biophysical mechanism underlying most of learning in humans or animals.\index{learning!in the brain}\index{plasticity} 

\begin{figure}
\begin{center}
\resizebox{5cm}{!}{\includegraphics{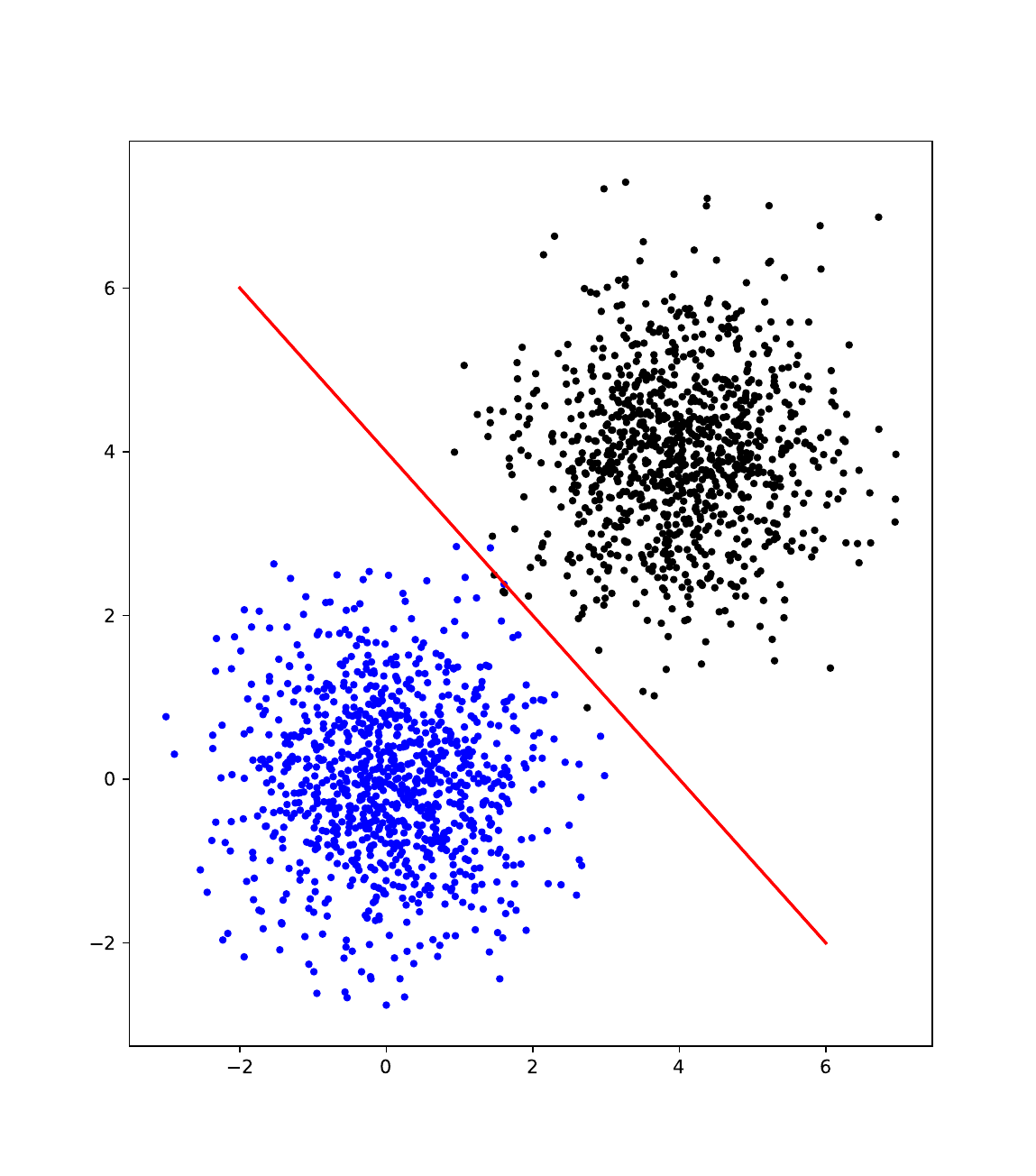}}
\caption{A simple illustration of what kind of a function a single neuron can learn in the basic case of classification with two classes. Each object (e.g.\ image of an animal) can be considered as a point in a very high-dimensional space where the coordinates correspond to pixel values, for example. For the purposes of this illustration, we assume there are only two input variables, so we can plot the points on a 2D plane. We also assume there are only two classes (something like ``cats'' and ``dogs'') which correspond to black and blue points, respectively. In the ideal case, the neuron will learn to output a ``one'' when the input is in one of the classes, and a ``zero'' when it is in the other class. Such learning corresponds to learning the line that separates the two classes, drawn here as red. Finding a line that separates the classes is clearly possible based on this data, and you have probably done that automatically in your head while looking at this figure. Such learning can be done by a single artificial neuron due to the great simplicity of this illustration, but in reality, we would often need a neural network with many neurons and layers. } 
\label{twoclusters.fig}
\end{center}
\end{figure}

How such changes precisely happen in the brain is an immensely complex issue, and we understand only some basic mechanisms.
Nevertheless, in AI, a number of relatively simple and very useful learning algorithms have been developed. Neural networks using them learn to perform basic ``intelligent'' tasks such as recognizing patterns (is it a cat or a dog?) or predicting the future (if I turn left at the next intersection, what will I see?). 
Learning in a neural network in such a case is based on learning a \textit{mapping}, or function, from input data to output data. Let's first consider a single neuron. It can basically learn to solve simple classification problems, as illustrated in Figure~\ref{twoclusters.fig}. If the classes are nicely separated in the input space, a single neuron can learn, as its synaptic weights, the pattern that precisely describes the difference between the classes. 

However, a network with many neurons can learn to represent much more complex functions from input to output. The input data could be photographs and the output data could be a word describing the main content of the photograph (``cat'', ``dog'', or ``unicorn'').  
The learning of the input-output mapping then consists of changing the synaptic weights of all the neurons in all the layers. In successive layers, the network learns to perform increasingly sophisticated and abstract computations, consisting of matching the inputs successively to the templates given by the weight vectors in each layer. After successful learning of the right mapping, you can input a photograph to the network, and the output of the neural network will give its estimate of what the photograph depicts.

There is an infinite number of different ways you can use a neural network by just defining the inputs and outputs in different ways. If you want it to learn to predict future stock prices, the input data would be the past prices and the output, current stock prices.  You can create a recommendation system that recommends new products to people in online shopping by defining the inputs to be some personal information of the customers, and the output whether the customer bought a certain item or not.\citenew{davidson2010youtube}
In a rather unsavoury application, the inputs would be what a social media user likes,\index{social media} and the output some sensitive personal information (say, sexual orientation), and then you can predict that sensitive information for anybody. Whether the prediction is accurate is another question, of course.

Here, we see one main limitation of machine learning: the availability of data. Where do you get the sensitive personal information of social media users in the first place, i.e.\ where do you get the data to train your network? Maybe nobody wants to give you such sensitive data. In other cases, the data may be very expensive to collect; for example, in a medical application, useful measurements and their analyses may cost a lot of money. Finding suitable data is a major limiting factor in neural network training; this is a theme we will come back to many times. Learning needs data, obviously; but it also needs the right kind of data, and enough of it.

\section{Learning as incremental minimization of errors} \label{objectivefunction}

After we have somehow found enough good data, we need to define how to actually perform the learning.
Most often, the learning is based on formulating some kind of error, and the network then tries to minimize it by an algorithm.   The error is a function of the data, i.e.\  something that can be computed based on the data at our disposal, 
and tells us something about how well the system is performing.
\index{learning!as minimization of errors}

Suppose the data we have consists of a large number of photographs and the associated categories (cat/dog etc.). To recognize patterns in the images, the network could learn by minimizing the percentage of input images classified incorrectly, which is called classification error.\index{error!classification e.} Alternatively, suppose we want to learn to predict how an agent's actions change the world---say, how activation of an artificial muscle changes the position of the arm of a robot. In that case, what should be minimized is prediction error:\index{error!prediction e.} the magnitude of the difference between the predicted result of the action and the true result of the action (which can be observed after the action).
Such errors don't usually go to zero, i.e.\ some error will be left even after a lot of learning. This is due to the uncertain and uncontrollable nature of the world and an agent's actions; that is another theme we will discuss in detail in later chapters.

After having chosen what kind of errors to minimize, we need an algorithm to actually minimize the errors.
What most such algorithms have in common is that they
learn by making tiny changes in the weights of the networks.
This is because optimizing an error function, such as classification error, is actually a very difficult computational task: There is usually no formula available to compute the best values for the weight vectors. In contrast, what is  usually possible is to obtain a mathematical formula that gives the \textit{direction} in which the weights should be changed to make the error function decrease the fastest. That direction is given by what is called the \textit{gradient}, which is a generalization of the derivative in basic calculus.\index{optimization} 

So, you can optimize the error function step by step as follows. You start by assigning some random values to the weight vectors. Given those values, you can compute the gradient, and then take a small step in that direction (i.e.\ move the weight vector a bit in that direction), which should reduce the error function, such as prediction error. But you have to repeat that many times, often thousands or even millions, always computing the gradient for the new weight values obtained at the previous step. (The direction of the gradient is different at every step unless the error function is extremely simple.)
Such an algorithm is called \textit{iterative} (repeating) because it is based on repeating the same kind of operations many times, always feeding in the results of the previous computation step to the next step. Using an iterative algorithm may not sound like a very efficient way of learning, but usually an iterative gradient algorithm is the only thing we are able to design and program.\index{gradient descent}

Such an algorithm is a bit like somebody giving you instructions when you're parking your car and cannot see the right spot precisely enough. They will only give instructions which are valid for a small displacement. When they say ``Back'', that means you need to back the car \textit{a little bit}, and then follow some new instructions. This is essentially an iterative algorithm, where you get instructions for the direction of a small displacement, and they are different at every time step.

Neural networks use an even more strongly iterative method, based on computing the gradient for just a few data points at a time. A data point is one instance of the input-output relation data, for example, a single photograph and its category. In principle, a proper gradient method would look at all the data at its disposal, and push the weights a small step in the direction that improves the error function, say the classification error, for the data set as a whole. However, if we have a really big data set, say millions of images, it may be too slow to compute gradient for all of them, since that would entail going through all the data points. What the algorithms usually do is to take a small number of data points and compute the gradient only for those. That is, you just take a hundred photographs, say, and compute the gradient, i.e.\ in which direction the weights should be moved to make the classification accuracy better, for those particular images. Importantly, at every step you randomly select a new set of a hundred images, and do the same thing for those images.\index{learning!iterative}

The point is that you are still \textit{on average} moving the weights in the right direction, so this is not much worse than computing the real gradient. But crucially, you can take steps much more quickly, since the computation of the gradient is much faster for the smaller data set. It turns out that in practice, the benefit of taking more steps often overwhelms the slight disadvantage of having just an approximation of the gradient.\footnote{To this advantage we have to add the more technical one that stochastic methods include an implicit regularization and are thus less likely to overfit the data \citep{bottou2003stochastic,hardt2015train}. Overfitting\index{overfitting} is an important problem in practical AI learning, but I don't discuss it at any length in this book. Basically, it means that if the amount of data at your disposal is very limited, learning may go wrong in a particular way: The learning may seem to work well for the data you have, achieving a good ``fit'', but the predictions your neural network gives are actually useless, because the learning ``overfit'' your data and does not work (or give a good fit) for any new data on which you would like to apply the system in the future.\label{overfitfn}}

Putting these two ideas together, we get what is called the \textit{stochastic gradient descent} algorithm. Here, ``descent'' refers to the fact that we want to minimize an error.
``Stochastic'' means ``random'', and refers to the fact that you are computing the gradient for randomly chosen data points, so you are going in the right direction only on average.\index{gradient descent!stochastic}\index{stochastic gradient descent|see{gradient descent, stochastic}}
\add{Ultimately, the agent could take a gradient step for each single data point that its sensory systems receive, leading to what is called \textit{incremental} learning.}\index{learning!incremental}
  
Suppose you're in an unfamiliar city and you need to get to the railway station. Your ``error function'' is the distance from the station. You can ask a passer-by which direction the station is, and you get something analogous to the gradient for one data point. Now, of course, that direction given by the passer-by is not certain, she could very well be mistaken; maybe she even said she is not quite sure about the direction. But you probably prefer to walk a bit in that direction, and then ask another passer-by. This is like stochastic gradient descent, where you follow an approximation of the gradient, given by each single data point. The opposite would be that instead of following each person's advice one after the other, you first ask everybody you see on the street where they think the station is, and move in the average of the directions they are giving. Sure, you would get a very precise idea of what the right direction is, but you would advance very slowly---this is analogous to using the full, non-stochastic gradient. 

\section{Gradient optimization vs.\ evolution} \label{GA.sec}

There are many more ways of optimizing an error function, and many systems that can be conceptualized as the optimization of a function.
In particular, evolution is a process where the error function called \textit{fitness} is optimized. Fitness is basically the same as reproductive success, which can be quantified as the expected number of offspring of an organism.\footnote{
  While the idea of evolution as fitness maximization can be found in many textbooks, some biologists would refute the whole idea of evolution optimizing any single function; see a recent review by \cite{birch2016natural}. To some extent, this controversy may also have arisen because of some semantic confusion about whether that would mean that evolution has already optimized the function (which would be a very strong statement) or whether it is \textit{in the process} of optimizing it (which may be more plausible); see \citet{parker1990optimality}. Regarding the precise definition of fitness, it seems impossible to find a consensus opinion \citep{Rosenberg2011fitness,Grafen2008simplest}. A standard textbook definition would be along the lines ``expected number of offspring'', but this may have to be complemented by the concept of inclusive fitness treated in Chapter~\ref{self.ch}. --- In this book, I tend to anthropomorphize evolution rather unashamedly, often comparing it to a human programmer. I believe that is a useful pedagogical device since humans find it easier to think of natural phenomena in terms of agents that have goals instead of a more abstract description such as a dynamical system. See e.g.\ \citet{Dawkins} for a rather strictly anti-anthropomorphic view on evolution.}
Fitness is, of course, maximized, while errors in AI are minimized. However, this difference is completely insignificant on the level of the optimization algorithms, since maximization of a function is the same as minimizing the negative of that function. Thus, evolution can equally well be seen as minimization of the negative fitness,\index{evolution!as optimization} which is then analogous to an error function.

In general, such a function to the optimized---whether minimized or maximized---is called an ``objective function''. The objective function does not necessarily have to be any kind of a measure of an error, although in AI, it often is.\index{objective function!based on errors|see{error}}\index{objective function} 
  Note that the objective function is completely different from the function from the input to the output that the neural network is computing, which was described earlier. The objective function is what enables the system to learn the best possible input-output function, so it works on a different level.

Optimization in evolution works, of course, in a rather different way than stochastic gradient descent. But it is actually possible to mimic evolution in AI and use what is called evolution strategies, evolutionary algorithms, or genetic algorithms. These are iterative algorithms that are sometimes quite competitive with gradient methods. They can optimize any function, which does not need to have anything to do with biological fitness. For example, we can learn the weights in a neural network by such methods. 
The idea is to optimize the given error function by having a ``population'' of points in the weight space, which is like a population of individual organisms in evolution. 

Like real evolution, such algorithms are based on two steps. First, new ``offspring'' is generated for each existing ``organism''. In the simplest case, you randomly choose some new weight values close to the current weight values of each organism, which is a bit like asexual reproduction in bacteria, with some mutations to create variability.
Then, you evaluate each of those new organisms by computing the value of the error (such as classification error) for their values for weights. Finally, you consider the value of the error as an analogue of fitness in biological evolution, albeit with the opposite sign because fitness is to be maximized while an error function is to be minimized. What this means is that you let those organisms (or weight values) with the smallest values of the error ``survive'', i.e.\ you keep those weight values in memory and discard those weight values which have larger (that is,  worse) values of the error function. You also discard the organisms of the previous iteration or ``generation'', since those individual organisms are already dead in the biological analogy. 

Such an evolutionary algorithm will find new weight values which are increasingly better because only the organisms with the best weight values survive in each iteration.   Thus, it is an iterative algorithm that optimizes the error function. It is a randomized algorithm, like stochastic gradient descent, in the sense that it randomly probes new points in the weight space. In fact, an evolutionary algorithm is much more random than stochastic gradient descent, since gradient methods use information about the shape of the error function to find the best direction to move to, while evolutionary methods have no such information. This is a disadvantage of evolutionary algorithms, but on the other hand, one step in an evolutionary algorithm can be much faster to compute since you don't need to compute the gradient, just random variations of existing weights.\footnote{For a basic genetic algorithm, see for example \citep{such2017deep} and the references therein. A particular advantage of evolutionary methods is that often they can be very efficiently parallelized; parallelization is explained in Chapter~\ref{control.ch}. They can also be combined with gradient methods by using a stochastic gradient descent to generate the offspring \citep{salimans2017evolution}.}

So, we see that \textit{both evolution and machine learning are  optimizing objective functions.} The optimization algorithms are often quite different, but they need not be. One important difference is that in machine learning, the programmer knows the error function, and explicitly tells the agent to minimize it. In real biological evolution, fitness is an extremely complicated function of the environment; it cannot be computed by anybody, nor can its gradient. Real biological fitness can only be observed afterwards, by looking at who survived in the real environment, and even then you only get a rough idea of the values of fitness of the individual organisms concerned: Those who die without offspring \textit{probably} had a low fitness, but it is all quite random---even more than stochastic gradient descent. (If an organism were actually able to compute the gradient of its fitness, that would give it a huge evolutionary advantage.)
Another important difference is that in biology, evolution works on a very long time scale, over generations, while in AI, the learning in the neural networks happens typically inside an individual's life span. The evolutionary algorithms in AI typically learn within an individual agent's lifespan as well, only simulating ``offspring'' of a neural network in its processors.

\section{Learning associations by Hebbian rule}

So far, we have seen learning as finding a good mapping from input to output. Such learning is called \textit{supervised}\index{learning!supervised} because there is, metaphorically speaking, a ``supervisor'' that tells the network what the right output is for each input. Yet, sometimes it is not known what the output of a neural network should be, or whether there is any point at all in talking about separate input and output---especially so if we consider the brain. In such a case, learning needs to be based on a completely different principle, typically the principle of \textit{unsupervised learning}. In unsupervised learning, the learning system does not know anything about any desired output (such as the category of an input photo). Instead, it will try to learn some regularities in the input data.\index{unsupervised learning}

The most basic form of unsupervised learning is learning associations between different input items. In a neural network, they are represented as connections between the neurons representing those two items. For example, if you have one neuron representing ``dog'' and another neuron representing ``barking'', it is reasonable that there should be a strong association between them. 

One theory of how such basic unsupervised learning happens in the brain is called Hebbian learning.\label{hebb} Donald Hebb proposed in 1949 that when neuron A repeatedly and persistently takes part in activating neuron B, some growth process takes place in one or both neurons such that A's efficiency in activating B is increased.\footnote{Adapted from \cite[p.~62]{Hebb1949organization}. I have simplified the original quote, stripping it from its neurobiological terminology. The original formulation is ``When an axon of cell A is near enough to excite a cell B and repeatedly or persistently takes part in firing it, some growth process or metabolic change takes place in one or both cells such that A's efficiency, as one of the cells firing B, is increased.''}\index{learning!Hebbian}\index{Hebb's rule|see{learning, Hebbian}}
Such Hebbian learning is fundamentally about learning associations between objects or events. 
A very simple expression of the Hebbian idea is that ``cells that fire together, wire together'', where ``firing'' is a neurobiological expression for activation of a neuron. In this formulation, Hebbian learning is essentially analyzing statistical correlations between the activities of different neurons.\footnote{This is perhaps oversimplifying the original idea: Recent research in neuroscience has emphasized that, as in the original definition above, it is important that cell A participates in the activation, i.e.\ is has a causal influence of cell B. This would usually mean that cell A is activated before cell B \citep{markram2012spike}. However, in most implementations of Hebbian learning in AI, such causal and temporal aspects are not used.  
  Actually, the details of how Hebbian learning works in the brain are not very well understood. 
}

One thing which clearly has to be added to the original Hebbian mechanism is some kind of forgetting mechanism. It would be rather implausible that learning would only increase the connections between neurons. Surely, to compensate, there must be a mechanism for decreasing the strengths of some connections as well. Usually, it is assumed that if two cells are not activated together for some time, their connection is weakened, as a kind of negative version of Hebb's idea.\citenew{oja1982simplified,zenke2017temporal}

Hebbian learning has been widely used in AI, and it has turned out to be a highly versatile tool. You can build many different kinds of Hebbian learning, depending on how the inputs are presented to the system and on the mathematical details of how much the synaptic strengths are changed as a function of the firing rates. You can also derive Hebbian learning as a stochastic gradient descent\index{gradient descent!stochastic} for some specially crafted error functions.\citenew{oja1992principal,hyvarinen1998independent}

\section{Logic and symbols as an alternative approach} 

The inspiration for neural networks is that they imitate the computations in the brain. Since the brain is capable of amazing things, that sounds like a good idea. But historically, before neural networks, the initial approach to AI was quite different. It was actually more like the world of planning we saw in Chapter~\ref{planning.ch}, where the world states are discrete, and there are few if any continuous-valued numerical quantities.

In early AI, it was thought that logic is the very highest form of intelligence, and therefore, AI should be based on logic. Also, the principles of logic are well-known and clearly defined, based on hundreds of years of mathematics and philosophy, so they should provide, it was thought, a solid basis on which to build AI. In modern AI, such logic-based AI is not very widely used, but it is making a come-back: It is increasingly appreciated that intelligence is, at its best, a combination of neural networks and logic-based AI---now called ``good old-fashioned'' AI, or GOFAI for short. Such logic provides a form of intelligence that is in many ways completely different from neural network computations, as we will see  next.\index{Good old-fashioned AI|see{GOFAI}}\index{GOFAI}

\subsection{Binary logic vs continuous values}

Mathematical logic is based on manipulating statements which are connected by operators such as AND and OR. For example, a robot might be given information in the form of a statement that ``the juice has orange color  AND the juice is  in the fridge''. Any statement can also be made negative by the NOT operator.
An important assumption is such systems, in their classical form, is that any statement is either true or false; no other alternatives are allowed. This goes back to Aristotle\index{Aristotle} and is often called the law of the ``excluded third''. That is, truth values are binary (can have only two different values).\index{binary values}

Such logic is perfectly in line with the basic architecture of a typical computer. Current computers operate on just zeros and ones, and those zeros and ones can be interpreted as truth values: Zero is false and one is true. Such computers are also called ``digital'', meaning that they process only a limited number of values, in this case just two. Our basic planning system in the preceding chapter, with its finite number of states of the world, was an example of building AI with such a discrete approach, and planning is fundamentally based on logical operations.\index{digital}

The brain, in contrast, computes with quantities which are in  ``analog'' form, which means continuous-valued numbers that can take a potentially infinite number of possible values.\index{information processing!analog vs digital}\index{analog} Artificial neural networks do exactly the same, as they are trying to mimic even this aspect of the brain. It is rather unnatural for the brain to manipulate binary data or to perform logical operations; such operations are possible only due to some very complex brain processes which we do not completely understand at the moment.\label{digitalvsanalog}

This distinction between digital and analog information-processing is another important difference between ordinary computers on the one hand, and the real brain or its imitation by neural networks on the other. (Earlier we saw the distinction between parallel and distributed processing in the brain versus the serial processing in an ordinary computer.) The digital nature of ordinary computers implies that any data that you input has to be converted to zeros and ones. This is actually a bit of a problem because a lot of data in the real world does not really consist of zeros and ones. For example, images are really intensities of light at different wavelengths, measured in a physical unit called ``lux''. One pixel in an image might have an intensity of 1,536 lux and another 5,846 lux. It is, again, rather unnatural to represent such numbers using bits, which is why processing non-binary data such as images is relatively slow in modern computers, compared to binary operations.\footnote{I shall just briefly mention another crucial difference between brains and ordinary computers:
An ordinary computer has hardware and software, and these two are separate. The same hardware can run different kinds of software, and the same software can be used on different hardwares. In fact, you can take software from one computer and download it to another, similar computer and it will work on that new computer as well.
However, in the brain, it is difficult to see any clear distinction between the software and the hardware: nobody ever downloaded software into their brain. Such a division between hardware and software is part of what is called the von Neumann architecture, named after the great mathematician John von Neumann.\index{von Neumann architecture} }

\subsection{Categories and symbols}

Saying that things are either true or false is related to thinking in terms of \textit{categories}.\index{categories}\index{thinking!with categories}
Human thinking is largely based on using categories: We divide all the perceptual input---things that we see, hear, etc.---into classes with little overlap. Say, you divide all the animals in your world into categories such as cats, dogs, tigers, elephants, and so on, so that each animal belongs to one category---and usually just one. Then, you can start talking about the animals in terms of true and false. You can make a statement such as ``Scooby is a dog'', and that is either true or false based on whether you included that particular animal in the dog category; any other (third) option is excluded.

Categories are usually referred to by \textit{symbols},\index{symbol} which in AI are the equivalent of words in a human language. For example, we have a category referred to by the word ``cat'', which includes certain ``animals'' (that's another category, actually, but on a different level). Ideally, we have a single word that precisely corresponds to each single category, like the words ``cat'' and ``animal'' above. Such symbols are obviously quite arbitrary since in different languages the words are quite different for the same category. An AI system might actually just use a number to denote each category. 

We see that logic-based processing goes hand-in-hand with using categories, which in its turn leads to what is sometimes called symbolic AI. These are all different aspects of the GOFAI.

\subsection{From hand-coded logic to learning}

Historically, one promise of GOFAI was to help in medical diagnosis, where the programs were often called ``expert systems''. This sounds like a case where categories must be useful since medical science uses various categories referring to symptoms (``cough'', ``lower back pain'') as well as diagnoses (``flu'', ``slipped disk'').\index{expert system}

The basic approach was that a programmer asks a medical expert how a medical diagnosis is made, and then simply writes a program that performs the same diagnosis, or makes the same ``decisions'' in the technical jargon. For example, one decision-making process by the human expert might be translated into a formula such as
\begin{quote}
IF cough AND nasal congestion AND NOT high fever THEN diagnosis is common cold
\end{quote}
However, this research line soon ran into major trouble.
The main problem was that medical doctors, and indeed most human experts in any domain, are not able to verbally express the rules they use for decision-making with enough precision. This is rather surprising since we are operating with human language and well-known categories. The situation is different from neural networks where it is intuitively clear that no expert can directly tell what the synaptic weights should be, because their workings are so complex and counterintuitive.
Yet, it turned out that even medical diagnoses are often based on intuitive recognition of patterns in the data, which is a form of \textit{tacit knowledge}. Tacit knowledge means knowledge, or skills, which cannot be verbally expressed and communicated to others. 

A major advance in such early AI was to understand that expert systems should actually learn the decision rules based on data. Again, learning provides a route to intelligence that is more feasible than trying to directly program an intelligent system. Given a database with symptoms of patients together with their diagnoses given by human experts, a machine learning system can learn to make diagnoses. 
Such learning is not so fundamentally different from learning by neural networks. What is different is that the data is categorical (``cough'', ``no cough''), and the functions are computed in a different way, for example by combining logical operations such as AND, OR, and NOT.\index{learning!and GOFAI}

\subsection{Categorization and neural networks}

By definition, such logic-based AI can only learn to deal with data which is given as well-defined categories.  Yet, real data is often given as numbers instead of categories; even medical input variables often include numerical data in the form of lab test results. In this medical diagnosis, we have indeed a category called ``high fever''.   The system is effectively dividing the set of possible body temperatures into at least two categories, one of which is ``high fever''. How are such categories to be defined? What is fever? What is low fever and what is high fever?
Here we see a deep problem concerning how categories should be defined based on numerical data, such as sensory inputs.

Again, some progress can be made by learning, this time learning the categories themselves from data. The AI can consider a huge number of possible categorizations of body temperatures: It can try setting the threshold for high fever at any possible value. If there is enough data on previous diagnoses by human doctors, the system can use that to learn the best threshold. In fact, the right threshold could be found as the one that minimizes classification error, i.e.\ the number of wrong diagnoses.\index{categories!learning c.}

However, while it is possible to learn categories in such very simple numerical data, 
GOFAI has great difficulties in processing complex numerical data. It is virtually impossible to use it to process high-dimensional sensory input, such as images consisting of millions of pixels. 
This was a rather big surprise for GOFAI researchers in the 1970s and 80s. After all, categorization of visual input is done so effortlessly by the human brain that it may seem to be easy. Yet, AI researchers working in the GOFAI paradigm  found it to be next to impossible. The early research on GOFAI was fundamentally over-ambitious, grossly underestimating the complexity of the world, as well as the complexity of the brain processes we use to perceive and make decisions.

One reason for the current popularity of  neural networks is that processing high-dimensional sensory data is precisely what they are good at. As we have seen, neural networks operate in a completely different regime from such logic-based expert systems. There are no categories and no symbols in the inner workings of neural networks: What they typically operate on is numerical, sensory input such as images, or some transformations of  sensory input. Raw gray-scale values of pixels are kind of the opposite of neat, well-defined categories.\index{vision}

As such, neural networks and logic-based systems can complement each other in many ways. Usually, the categories used by a logic-based system need to be recognized from sensory input: a neural network can tell the logic-based system whether the input is a cat or a dog. In particular, a neural network can take sensory data as input, and its output can identify the \textit{states} used in action selection; to begin with, it can tell the planning system what the current state of the agent is. In Chapter~\ref{dual.ch}, we will consider in more detail this fundamental distinction between two different modes of intelligent information-processing, which are found both in AI and human neuroscience.

\section{Emergence of unexpected behavior}

Finally, let me mention a phenomenon that is typical of any learning system.
``Emergence'' means that a new kind of phenomenon appears in some system due to complex interactions between its parts. It is a special case of the old idea, going back to Aristotle at least, that ``the whole is more than the sum of its parts''.\index{Aristotle} For example, systems of atoms have properties that atoms themselves do not---consider the fact that a brain can process information while single atoms hardly can.\footnote{\add{Going further, it has been claimed that "the whole is \textit{different} from its parts", especially by the Gestalt school of perceptual psychology \citep{wagemans2015historical}. For example, a set of dots that forms a (dotted) straight line (such as $\:\:\cdots\cdots\cdots\:\:$)  can be perceived as a straight line only, completely forgetting the fact that it is composed of the dots; thus the line and the dots are two different things in the sense of two different ways of perceiving the same stimulus.}\index{emergence}}
Likewise, evolution is based on emergence. Its objective function is given by evolutionary fitness, which sounds like a very simple objective function. Nevertheless, it has given rise to enormous complexity in the biological world, as well as human society.
What is typical of such emergence is that its result is extremely difficult to predict based on knowledge of the laws governing the system. If the objective function given by fitness had been described to some super-intelligent alien race a few billion years ago, they would hardly have been able to predict what the world looks like these days.

Machine learning is really all about the emergence of artificial intelligence. 
We build a simple learning algorithm and give it a lot of data, and hope that intelligence emerges. 
It is the interaction between the algorithm and the data that gives rise to intelligence. This seems to work, if the algorithm is well designed, there is enough data of good quality, and sufficient computational power is available.
Such emergence in machine learning is actually a bit different from emergence in other scientific disciplines. In physics, very simple natural laws \textit{by themselves} can give rise to highly complex behavior. In machine learning, the complexity of the behavior of the system is, to a large extent, a function of the complexity of the \textit{data}. In some sense, one could even say the complex behavior learned by an AI does not emerge but is extracted, or ``distilled'', from input data. The complexity of the input data is due to the complexity of the real world, which is obvious when inputting a million photographs into a neural network.

The emergent nature of the behavior learned by an AI implies that, just like in evolution, there is often something unexpected in the resulting system. The complexity of the input data usually exceeds the intellectual capacities of the programmer. So, the programmer of an AI cannot really know what kind of behavior will emerge: Often the system will end up doing something surprising. 
\index{unexpected behavior}

In this book, we will encounter several forms of emergent properties in learning systems which are related to suffering. While some kind of suffering may be necessary as a signal that things are going wrong, we will also see how an intelligent, learning system will actually undergo \textit{much more} suffering than one might have expected. To put it bluntly, a particularly intelligent system will find many more errors in its actions and its information-processing. In fact, finding such errors was necessary to make it so intelligent in the first place. Therefore, a learning system may learn to suffer much of the time, even though that is not what the programmer intended.

\chapter{Frustration due to reward prediction error} \label{rpe.ch}

Now, armed with modern machine learning theory, we revisit the problem of action selection and the concept of frustration. In planning, as we saw in Chapter~\ref{planning.ch}, the main computational problem is looking several steps ahead, which can lead to quite impossible demands of computational capacities. Another constraint is that it requires a model of how your actions affect the world, i.e.\ where do you go in the search tree when you perform a given action in a given state. 
As such, planning is not really a good method for action selection if computational resources are very limited, as in a simple computer, or a very simple animal such as an insect.

In this chapter, we consider an alternative way to action selection, based on learning. A paradigm called reinforcement learning enables learning intelligent actions without any explicit planning, thus avoiding many of its problems. It also generalizes the framework of a single goal to maximization of rewards obtained at different states. While it can be performed even in very simple animals and computers, it is also used by humans; it is similar to how habits work.

We then consider how frustration can be defined in such a case; it can no longer be simply  defined as not reaching the goal---since there is no explicit goal. We define more general error signals called reward loss and reward prediction error, which have been linked to signals of certain neurons in the mammalian brain.
Thus, we expand the view where frustration is related to error signalling by linking it to errors in prediction.  

Repeated frustration is thus something necessary for learning algorithms to work, and intelligence may not be possible without some frustration. We further see how the very construction of an agent based on reward maximization means that it is insatiable, never satisfied with the amount of reward obtained. Moreover, it can be directed towards intermediate goals which are not valuable in themselves, but simply predictive of future reward. Evolutionary rewards, in particular, can lead to behaviors which resemble obsessions.

\section{Maximizing rewards instead of reaching goals}

In modern AI, action selection is most often not based on planning, but a framework where the obtained \textit{rewards}, or reinforcement, is maximized. 
This is useful because often an AI does not have just a single goal to accomplish, but many things it should take care of. Defining behavior as maximization of rewards as opposed to reaching goals is also often thought to be more appropriate for modelling behavior in simple animals, which are thought to be incapable of the sophisticated computations needed in planning (more on this in Chapter~\ref{replay.ch}).\index{reward!definition}\index{reinforcement|see{reward}}

For example, if a cleaning robot disposes of some dust in the dustbin, it could be given a reward signal. Since there are many rooms and many dustbins in the building, it makes sense to give a reward whenever the robot disposes of some of the dust. In principle, we could decide to give it a single reward when all the rooms are completely clean; however, it is common sense to rather give it a reward every time it removes some dirt or dust from any of the rooms. After all, the robot has done something useful every time it reduces the amount of dust in the room; telling this to the robot is highly useful information, and it would simply complicate the learning if the reward were postponed until the robot has completed some larger part of the task.\index{robot!cleaning}

In fact, giving a single reward at the end would mean the robot has to engage in long-term planning, which is  difficult. A ``piece-wise'' training by giving rewards for small accomplishments is not very different from how you would teach a child to perform a rather demanding, long task, say tying shoelaces: divide the task into successive parts and give the child a small encouragement when it completes each small part. This is computationally advantageous since it eliminates the need for long-term planning, a bit like the heuristics we saw above.\footnote{It is also essential in training animals to perform long sequences of actions; in that context it is called ``shaping'' \citep{krueger2009flexible,ng1999policy}. However, reinforcement learning is a much more general concept than just dividing a long sequence into smaller parts. In the case of the cleaning robot, there may not be any end to the cleaning task since more dust appears constantly. The only meaningful goal for the cleaning robot may be to just remove dust and dirt as much as possible, which is exactly captured by the reward formalism.}\index{shaping}

Reinforcement or reward can also be negative; if the robot tries to put household items in the dustbin, it can be given some. Negative reinforcement is really what we usually call a punishment---but the word is interpreted without any moral connotations here. 

Thus, we actually ground action selection in the optimization of an objective function, i.e.\ a quantity to be optimized. Earlier, we saw that minimization of an error function, such as the number of images incorrectly classified, is the way an AI can learn to recognize objects in images. Here we define a different kind of objective function which is the basis of action selection: It is equal to the \textit{sum of all future rewards}. It is a function of the action selection parameters of  the agent, and more precisely, it expresses how much reward the agent can obtain by behaving according to its current action selection system.\index{objective function!based on rewards}

Such a learning process based on maximization of future rewards by learning a value function is called \textit{reinforcement learning}.\index{reinforcement learning}\index{learning!reinforcement|see{reinforcement learning}}
Reinforcement learning can be seen as a third major type of learning in AI, in addition to supervised and unsupervised learning. 

In a sense, this future reward is the ultimate objective function of an agent. Its maximization, by tuning the action-selection system, is the very meaning of life of the agent. The objective functions we saw earlier, used to learn things like pattern recognition by minimization of errors,  are there merely to help in maximizing this reward-based objective function.\citenew{silver2021reward}
 
In such a reward-based objective function, more weight is often put on the rewards in the near future as opposed to rewards in the far-away future, which is called discounting. The justification for this is complicated, but suffice it to say that such discounting is often evident in human behavior: Humans prefer to have their reward right now, and value it less if they have to wait. To keep the discussion simple, I sometimes ignore discounting in what follows, but it could be used in almost every case considered in this chapter.\footnote{A basic exposition of discounting is given by \citet{suttonbook}. For discussions of different kinds of discounting, and in particular for comparisons between exponential and hyperbolic discounting, see \citet{dasgupta2005uncertainty,ainslie2001breakdown}.}\index{discounting}

\section{Learning to plan using  state-values and action-values}

As such, the sum of future rewards gives a more general framework than having a single goal as in Chapter~\ref{planning.ch}, since trying to reach a single goal can be accommodated in the reward framework by simply giving a reward when the agent reaches the goal, and no reward otherwise. In such a case, discounting further means the agent receives more reward if it reaches the goal more quickly, which is intuitively reasonable.

It turns out that we can use this reformulation of planning as reward maximization to our advantage, since the algorithms developed for maximizing future rewards give a particularly attractive way of solving the problem of planning. In Chapter~\ref{planning.ch}, we saw how difficult planning is due to the exponential explosion in the number of possible plans to choose from. While heuristics were proposed as a practical trick to make the computations more manageable, there is no universal way of designing good heuristics. 

Like in other branches of AI, it has been found that \textit{learning} solves these problems, at least to some extent.  Intuitively, if the agent encounters the same planning problem again and again, it can store information about the previous solutions (or attempts) in memory.\index{learning!to solve planning}
For example, a cleaning robot will probably clean the same building many, many times, and a delivery robot will deliver the parcels to the same addresses quite a few times. So, such agents should be able to learn something about planning in their respective worlds.\index{robot!cleaning}
This would be a clear improvement compared to heuristics, which need to be explicitly programmed in the system by programmers as in our examples above, and it is often unclear how to do that.\footnote{Actually, the chess-playing Deep Blue mentioned in Chapter~\ref{planning.ch} did already use some learning as well: It analyzed data from several chess databases, including 700,000 historical games played by human grandmasters, to compute another heuristic.\index{Deep Blue} }

Reinforcement learning gives us a sophisticated mathematical theory that tells us how to learn a particularly good substitute for a heuristic, called the \textit{state-value function}.\index{heuristics!state-values}
It is a clever way of learning to deal with the complexity of the search in a planning tree. The basic principle is simple:\index{state-value}\index{value function|see{state-value or action-value}}\index{Q-value|see{action-value}}
Using the previous planning results in its memory, the agent can compute something like a heuristic based on how well it performed starting from each possible state. If it found the goal quickly starting from a certain state, that state gets a large state-value.

In the case where we have a single goal, the state-value function basically tells you how far from the goal you are, thanks to discounting which takes account of the time needed to reach the goal. 
A delivery robot that frequently delivers stuff to the same building (say town hall) would easily learn the distance from any other building to the town hall. In the beginning, when it had a delivery to the town hall, it had to spend a lot of time and effort\index{effort!in planning} in planning the path there.\index{robot!delivery} But little by little, it gained information by storing any results of executed plans in its memory, and learned the distance from any other building to the town hall. Such distances now give the state-value function for that goal (the state-value is actually a decreasing function of that distance). When the robot next needs to go to the town hall, it recalls the distances, to the town hall, from those buildings that are close to its current location, and simply decides to move in the direction of the near-by building which has the smallest distance to the town hall. Thus, it has learned a kind of a heuristic that avoids planning action sequences altogether.

\begin{figure}
  \begin{center}
  \resizebox{0.42\textwidth}{!}{\includegraphics{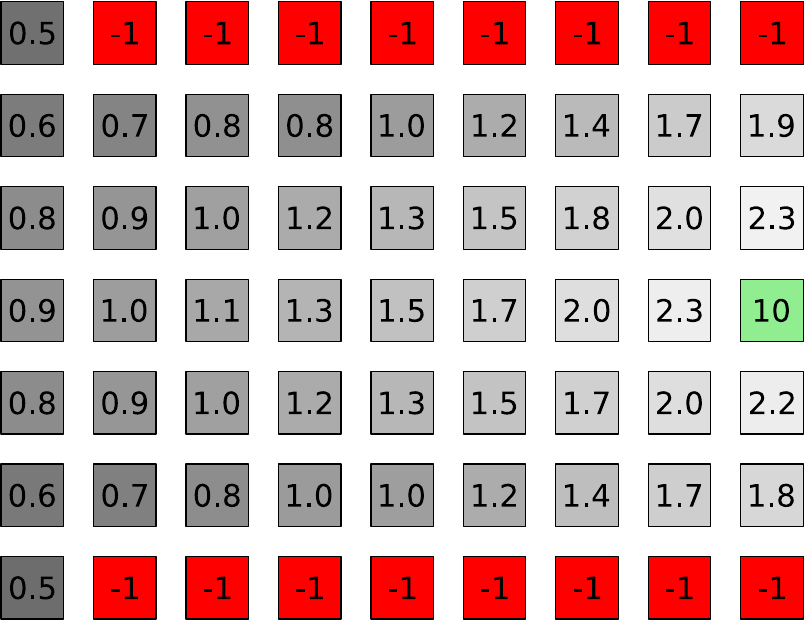}}
  \hspace*{0.11\textwidth}
  \resizebox{0.42\textwidth}{!}{\includegraphics{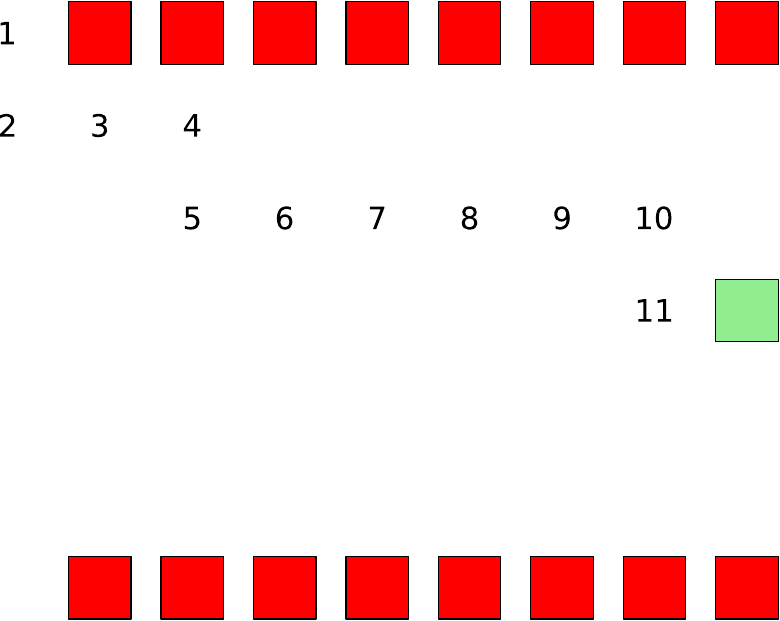}}
  \end{center}
\caption{\add{A simple illustration of value functions in reinforcement learning. We use a simple ``gridworld'' where the agent lives on a 2D grid. Each state is one location (or one box on the left-hand figure), and the actions are up, down, left, right. There is a reward at the right-hand border of the world, worth 10 reward units, marked in green. There are also ``punishments'', i.e.\ negative rewards, at the upper and lower edges of the world, marked in red.       Also, some randomness is added to the world, so that the actions the agent takes are sometimes replaced by random actions, to model the fact that the real world is random; therefore, the agent should be ``careful'' and stay far away from the punishments.
The agent starts from a random location at the left-hand edge and then tries to get to the reward while avoiding punishments. Avoiding punishments is not that difficult once the agent has learned where they are, but that learning will take time. Likewise, the agent has to learn, by trial and error, where the reward is located.
    In the figure on the left, the state-value functions (of the optimal policy) are plotted, both in numbers and the gray-scale value (darker squares mean lower value). We see that the values are higher closer to the positive reward, and away from the negative rewards.     This is exactly how they should be to indicate that the agent should move right while avoiding the upper and lower edges. (The values are much less than the reward of 10 since they are discounted, i.e., kind of long-term averages.)  In the figure on the right, one episode in the life of the agent is shown. The numbers 1,2,3,... refer to the flow of time: they are the time points when the agent resided in each of the states. The agent started at the upper left-hand corner and walked rather directly to the reward. This was possible because this plot assumed the agent had learned the values on the left-hand figure and used them for action selection.
    We see that using the state-values, the agent was able to find the shortest path to the big reward, by simply always choosing to move to the near-by state that had the largest state-value.}}
\label{valueplot.fig}
\end{figure}

Such learning works even in a very general setting when there is no particular goal. In general, the value of a state is defined as the sum of all future rewards the agent can obtain \textit{starting} from that state.\footnote{Next, I give a more rigorous and general definition of state-values.\index{state-value!mathematical definition}\index{policy} To begin with, it must be noted that the state-value is a function of the ``policy'' used by the agent; the policy is what I call the action selection system  in the text, i.e., the system that decides which action is taken in any given state (where the decisions can have some randomness programmed in them).  Further, we have to take into account the fact that the world may have some randomness in it, so we have to consider expected reward in the sense of the mathematical expectation in probability theory. The value function at a given state is then generally defined as the expected amount of discounted reward that the agent will obtain starting from that state, when it follows that policy. (Sometimes, when speaking about state values, it is more specifically assumed that the policy in question is the optimal policy which gives the highest expected reward, nut that is just one special case for a special policy.) This definition reduces to the definition in the main text for the case of a single goal in a deterministic world, where the state-value is a decreasing function of the distance to the goal. The connection can be seen by defining that there is a reward at the goal and nowhere else, and using the fact that there is discounting, and thus rewards in the distant future are given less weight than rewards in the near future. Then, the closer you are to the goal, the larger the expected reward is, because the reward at the goal is given more weight when you are closer to the goal. (I define here ``closer''  to mean that you can get there more quickly compared to the situation where you are further away and need time to get there). While this standard definition in the literature, as just given, considers the reward uncertain and talks about expected (discounted) reward, I will not usually do that in this chapter for simplicity: I assume the world, as well as the policy, are deterministic. See Chapter~\ref{threat.ch} and its  footnote~\ref{threatmathsfn} for a more sophisticated, probabilistic definition.} After successful learning of the state-values, the solution to the problem of action selection is that at each step, whatever state the agent may be in, the agent just selects the single action which leads to the state with the highest state-value (e.g.\ closest to the town hall above). There is no need to compute several steps ahead, or make a search within the huge search tree anymore. This reduces the complexity of the computations radically: instead of planning all its actions up to reaching the goal, the state-values now provide kind of intermediate goals that the agent can very easily reach in just one step. However, a lot of time and computation still needs to be spent on first learning the state-values.\footnote{\label{bellmanfootnote} A multitude of algorithms for learning the state-values exist; see \citet{suttonbook} for a comprehensive treatment. %
  Typical algorithms  proceed by a recursion where the value of a state is defined based on the values of the states to which the agent can go from that state, based on the theory of dynamic programming and, in particular, what is called the Bellman equation.
  As an illustration, consider a simple world with three states, A, B, and C, where C is the (only) goal state; suppose you can move from A to B and from B to C. 
  The first part of the recursion says that the value of state A must be the value of state B minus a small quantity. That is because from state A you could go to state B in a single step, and the subtraction of the small quantity expresses discounting, due to the fact that you need one step. Likewise, the value of state B must be a bit less than the value of C. Now the value of C is fixed (to some  numerical value which is irrelevant) by the fact that it is the goal, and needs no recursion or computation. So, once the agent encounters the state C even once, it knows the value of C. Based on that knowledge and its model of the world, it can start recursive computations, by applying the ideas above (value of B equal to value of C minus a small constant, value of A likewise) to recursively compute the values of B and A. If we fix the value of the goal to 1, the state-values could be 0.8, 0.9, and 1 for A, B, and C respectively. Note that in this example, we computed the state-values of the optimal policy, i.e., assuming the agent always takes the smartest possible actions. %
  You could also compute the state-values for a very dumb policy (say, always taking random actions), and they would be lower because by taking less smart actions, the agent would get less reward.\index{Bellman equation}
}
\add{Value functions are illustrated in Fig.~\ref{valueplot.fig}.}

\subsection{Completely reactive action selection by action-values}

While we have thus solved the problem of the computational explosion of planning, there is still the problem that the agent needs to have a model of how the world works. Even using the state-values, it needs to understand which action takes it into a state with higher state-value. Now, consider an extreme case where the agent has no model of how the world works in the sense that it has no idea what about the effects of its actions. Then, it is not enough to assign values to different states since the agent does not know how to get from state A to state B. (Still, we assume the agent knows in which state it is, at any given moment, so it does have some minimal model of the world.)

The trick to learning to act even with such a minimal model of the world is to learn what is called the \textit{action-value} function.\index{action-value} When the agent is in a given state, the action-value function tells the value of each of its actions, in terms of how much the total future reward is if the agent performs that action.\footnote{Again, strictly speaking, the action-values depend on the policy of the agent, while sometimes the term is used to mean the action-values for the optimal policy. The terminology is further confounded by the fact that sometimes action-values can refer to the current estimates of the agent for those action-value instead of their true values (the same holds for state-values as well). Action-values can also be called ``Q-values''. }
This makes action selection really easy and extremely fast: Just compare the values of different actions and choose the one which gives the maximum. In fact, all the relevant information about the effects of the agent's actions are implicitly included in the action-value function. 
The agent still has to learn the action-values, but that is not really more difficult than learning the state-values.\footnote{For  different algorithms, see \citep{suttonbook}; for example, a recursion similar to that in footnote~\ref{bellmanfootnote} above could be used.}

At the end of the 19th century, Edward Thorndike put cats in a box where they would need to press a lever to get out of the box and receive some fish to eat. He observed that in successive trials, the cats were pressing the lever more and more often. Such learning is called \textit{instrumental conditioning} (as opposed to classical conditioning as in the famous Pavlov's dogs, to be considered below).\index{conditioning!instrumental}\label{instrucond} 
This shows how learning to choose actions is possible by simply associating what we call a state in AI  (here, being in the box) with an action.\footnote{
  This can be seen as an example of using something like action-values without a sophisticated world model. On the level of neurobiology, such reactive behavior can also be explained by a special form of Hebbian learning, which implements something similar to the abstract theory of reinforcement learning we have just seen. In such learning, the association weight between the state (being in the box) and action (pressing the lever) increases every time both the state and the action are active, \textit{and} a reward (fish) is delivered. Ordinary Hebbian learning would only be able to learn, in an unsupervised manner, the connection between the state and the action if the same action is frequently taken in a particular state. It would be useless in itself for selecting the \textit{best} action since it does not take the reward gained by the actions into account. So, an extension of Hebbian learning to such ``three-factor'' learning, modulated by reward, is necessary.
This may not be exactly what happens in the brain, but it is probably a useful approximation nevertheless \citep{nevin1999analyzing}.\index{learning!Hebbian!three-factor} Such three-factor (or modulated) Hebbian learning rules have a long history, see e.g. the discussions by \citet{legenstein2010reward,gerstner2018eligibility}. These learning rules can also be extended to choosing action sequences in a dynamic environment: Basically, instead of the reward itself, the Hebbian rule might be modulated by reward prediction error considered next in the main text.
}

So, using reinforcement learning, an AI or an animal can actually learn to act without doing any real planning and having almost no model about the world. If it learns the action-value function, it only needs to look at the single actions immediately available, and then take the action which has the largest action-value---at the state where it happens to find itself. 
Since the action is here triggered immediately without any deliberation, like a \textit{habit} or a knee-jerk reaction, the resulting behavior is often called habit-based, or reactive.\footnote{Some authors use the term ``model-free reinforcement learning'' to clearly distinguish this from anything using planning. Planning uses a model of the world, thus it would be called ``model-based''. Model-based reinforcement learning then refers to a set of algorithms for solving the planning problem, with the possible modification that instead of reaching a single goal, the plan may still attempt to maximize the sum of rewards.}\index{habits}

Reinforcement learning has recently become popular as a model of human behavior in neuroscience, where humans may not be considered too different from experimental animals such as cats or rats. Current thinking is that the same reinforcement learning algorithms can be used to model at least one part of the action selection system in most animals, including humans. Nevertheless, there is little doubt that some animals, probably most mammals, engage in planning as well.\footnote{For a review on applications of reinforcement learning to modelling animal and human behavior, see \citet{niv2009reinforcement}. On planning in animals, see \citet{redshaw2018future,corballis2019language}.}\index{planning!in animals}

In fact, reinforcement learning using value functions is not a magic trick that will obliterate the complexity of the action selection: It simply shifts the computational burden from search in the tree to learning a value function. Sometimes, this is a good idea, but not always.\index{tree search}
We will discuss the pros and cons of reinforcement learning vs.\ planning in Chapter~\ref{dual.ch}. Let me just mention here the main disadvantages of habit-based behavior:  such learning often needs a lot of time and data, and leads to inflexible behavior. This is quite in line with the common-sense idea we have about habits.

\section{Frustration as reward loss and prediction error}

We have thus divided action selection into planning and habits, where habits refer to more automated action selection mechanisms similar to reinforcement learning.\citenew{dolan2013goals}  
Therefore, we need to ask how we can define frustration in the case of habits, where there are no goals but rather rewards obtained here and there, and we cannot talk about frustration in the sense of not reaching the goal as in Chapter~\ref{planning.ch}.

What defines frustration in this case is an error signal called reward loss\footnote{\add{\citep{papini2015behavioral,mee2006psychological}; see also \citep{bell1985disappointment,van2017affective}}} which we already saw briefly in Chapter~\ref{suffering.ch}. It is computed by the following simple formula:
\begin{quote} \label{frustdef}
reward loss = expected reward - obtained reward \index{reward loss!definition}\index{frustration!as reward loss}
\end{quote}
which is set to zero in case the difference is negative. That is, a reward loss is incurred when an agent expects to get some reward but actually gets less reward than expected. Maybe a cleaning robot expected to find a lot of dust in a room, but in fact there was much less. If it happens that the obtained reward is actually larger than expected, there is obviously no reward loss, so the reward loss is defined as zero if the difference in the expression above is negative. Reward loss can also occur if the expected reward is negative, and the obtained reward is negative while even larger (in absolute value): the agent did expect something bad to happen, but it turned out to be even worse.\footnote{The effort made in trying to obtain the reward may also need to be taken into account in computing frustration. While it might seem natural to simply subtract the effort spent from the reward, considering it as a ``cost'',  sometimes more effort leads, paradoxically, to higher perceived reward \citep{inzlicht2018effort}.\index{effort!and perceived reward}\index{reward!perceived}}

Expectation of reward here refers to the mathematical expectation as defined in probability theory.\index{expectation!definition}\index{frustration!definition of expectation} It is obtained by weighting the possible values by their probabilities: if the probability of obtaining a reward is 50\% and the reward is 10 pieces of chocolate, the expected reward is 5 pieces of chocolate.\footnote{The exact definition of expectation as used in reward loss is not very clear in my view, and an important problem for future research. Not much attention has been paid on it, partly because in typical experiments, it seems obvious what the expectation should be, and there is little planning involved. In a prototypical experiment, an animal (or a human) is given the same (positive) reward several times for some simple behavior, and then suddenly it is given  less reward (this is called ``successive negative contrast'') for that same behavior. In such a case, the future expectation of the reward is simply assumed to be equal to the past reward. With longer plans in more complex environments, the definition will be less obvious. Clearly, there is a strong connection to the concept of a prediction, as discussed next in the main text as well as footnote~\ref{bayesfrustrationfn} below.
 Furthermore, an alternative definition of reward loss might be developed using counterfactual contrast \citep{roese1997counterfactual}, formalized as counterfactual regret by \citet{zinkevich2008regret},\index{thinking!counterfactual}\index{regret! counterfactual} where the obtained reward is compared with what \textit{might} have been obtained, if better actions had been chosen. If the agents form some kind of a society, even more options exist for defining the expectations. The agent might use social comparison, i.e.\ information on what other agents get, and expect to obtain the same reward as others do. Such a ``social'' expectation might simply be based on probabilistic inference: If the other agents are similar to the agent in question, it is logical to expect that the agent in question will be able to obtain the same amount of rewards \citep{rutledge2016social}; see also footnote~\ref{socialevaluationfn} in Chapter~\ref{self.ch}.\index{social comparison} Yet another, very different, form of expectation might be produced in a situation where the agent assumes a \textit{moral} right to obtain something, assuming the existence of some ethical norms in the agents' society \citep{dignum2000towards}.\index{morality!and expectation}
  \label{expdeffn}}  
Expectations of the future are often called predictions. Predictions are in fact ubiquitous in the brain: it is likely that the brain makes a prediction of almost any important quantity in the environment.\footnote{\citep{clark2013whatever}. From the viewpoint of mathematical theory, it might actually be more appropriate to talk about \textit{predicted reward} instead of expected reward in the definition of reward loss (assuming here that expectation is defined as the mathematical expectation according to probability theory as in the main text). While these are often seen as the same thing---prediction being an expectation of a future quantity---the concepts are not equivalent. In particular, in machine learning theory, a prediction can be considered more general than expectation: a sophisticated prediction will also include an estimate of  the uncertainty involved in the prediction, in addition to the mathematical expectation. The importance of such uncertainty of predictions will be seen in Chapter~\ref{threat.ch} regarding the concept of threat.
  But here, regarding frustration, such uncertainty is relevant because it seems that the certainty of the prediction affects the level of frustration. I would claim that if you are completely certain that you will get chocolate (say, 5 pieces), but then it turns out you don't, the frustration will be greater than in the case where there is only some chance of getting any (like the example in the main text, 10 pieces with 50\% probability). Crucially,  in this example, the expected amount of chocolate, in the sense of the mathematical expectation, is the same in the two cases, and only the uncertainty changes. Therefore, the effect of uncertainty should be taken into account in the definition of reward loss.
  See footnote~\ref{deltafn} in Chapter~\ref{freedom.ch} and  the main text preceding that footnote for further developments of this point,   as well as Chapter~\ref{perception.ch}.
  \label{bayesfrustrationfn}}
  
In contrast to our basic definition of frustration in Chapter~\ref{planning.ch}, which works only on the level of plans, the reward loss can be computed after every single action and at every single time point. This definition of reward loss is, in fact, quite flexible since the time interval in which the reward is computed can be specified to be anything from seconds to days. Therefore, it  provides a general framework encompassing both planning and habit-based action. Reward loss coming from planning and reward loss coming from single actions are similar except that they work on very different time scales. We shall consider this point in more detail in Chapter~\ref{summary1.ch}.\footnote{Another difference is that while earlier (Chapters~\ref{suffering.ch} and \ref{planning.ch}) we defined frustration as ``not getting what one \textit{wants}'', in line with the quotes from ancient philosophers, here reward loss is defined as ``not getting what one \textit{expects}''. These are not exactly the same thing and are sometimes quite different; 
 this connection will also be discussed in Chapter~\ref{dual.ch} (page~\pageref{expvsdesire}).
 \label{wantfootnote}}

Reward loss, in its turn, is related to what is called the reward prediction error (RPE), a most fundamental quantity in machine learning theory.\index{reward prediction error (RPE)|see{error, reward prediction}}\index{error!reward prediction e.} RPE means any error made in the prediction of the reward. This definition is very general because the expected reward can be greater or less than the obtained one, and thus RPE can be positive or negative.\index{frustration!as reward prediction error}
If the obtained reward was larger than expected, that is the opposite of reward loss and suffering, and related to pleasure.\footnote{I refrain from trying to rigorously define pleasure in this framework,\index{pleasure} but obviously an RPE where the obtained reward is greater than expected is a good candidate. In fact, in experiments with participants playing a gambling game, the long-term average of the reward prediction error (taking both positive and negative parts into account) was a strong predictor of the participants' well-being \citep{rutledge2014computational}; however, the average level of reward had a strong effect as well. Alternatively, \citet{carver2003pleasure} has proposed that the function of the pleasure system is to signal that the current task has been accomplished, and the system can direct its resources to other tasks.\label{pleasurefn} The neurobiology of pleasure and pain is reviewed by \cite{leknes2008common}. }

As the very expression ``reward prediction \textit{error}'' indicates, the theory of RPE also shows how suffering is related to learning  by minimization of errors, which is a fundamental approach in machine learning. If the agent can predict the rewards obtained by different actions in different states, it will be able to act so as to maximize the obtained rewards (at least if it has a good model of other aspects of the world as well). To learn and improve such predictions, it is necessary to compute the errors produced by the current predictions. This is how minimizing RPE is related to maximization of rewards. In fact, it is possible to devise reinforcement learning algorithms that work simply by minimizing reward prediction error.\footnote{This can be done using a special form of RPE called temporal difference (TD) error (see footnote~\ref{RPEfootnote} below), and in particular using the squared error summed over all states. See \citet[p.~268]{suttonbook} who call it Bellman error, or related developments by \citet{bhatnagar2009convergent}.}

The exact mathematical definition of RPE is quite involved and relegated to a footnote.\label{RPE}\footnote{\label{RPEfootnote}\index{error!reward prediction e.!mathematical definition} 
  RPE can actually be defined in different ways. In neuroscience literature, the definition may not be very different from reward loss. In the reinforcement learning theory, a more sophisticated definition is usually used, using what is specifically called the temporal difference (TD) error, which we explain here. For simplicity, %
  no discounting is used here. For each time step, RPE is then defined as
$
\text{RPE}= \text{reward} - (V_\text{before}  -V_\text{after}) 
$
where $V$ is the state-value function (for the policy being followed, not necessarily the optimal one), in the state before the action was taken or after the action taken, respectively (which could also be denoted by time indices $t-1$ and $t$). 
The reward is the reward obtained for this particular action, or in other words, at this particular time step for which we are computing the RPE. Note that the sign is flipped compared to the definition of reward loss, but this is just a technical convention with no deeper meaning. 
The connection to reward loss can be seen by understanding that  in the state-value formalism, $V_\text{before}  -V_\text{after}$ can be interpreted as expected reward. The reason is that by the definition of the state-value function, the state-value function gives the total reward expected when starting from each of the states, so you would expect a reward equal to $V_\text{before}  -V_\text{after}$ for this action. Otherwise the two state-values would be inconsistent; the total expected reward starting from the state ``before''  must be equal to the total expected reward starting from state ``before'' plus the expected reward obtained in the transition.
So, the agent can expect that $\text{reward} = V_\text{before}  -V_\text{after}$, and if that actually holds, RPE would be zero. 
If you get less, there is a reward loss, which is here expressed as a negative RPE.
Such an RPE signal %
is more general than reward loss since it considers the whole future of rewards via the state-values, as explained next in the main text. There are also some small differences: RPE has a different sign, corresponding to the negative of reward loss, and our definition of reward loss considers only the case where it is positive (or RPE is negative) since this is the part corresponding to suffering. Also, typically the discounting formalism of reinforcement learning is included in the definition, in which case $V_\text{after}$ would by multiplied by a discounting factor throughout; omitting the discount factor is possible if we consider a finite time horizon. See \citet[Ch.15]{suttonbook} for more information.
}
Let me just point out that RPE is a more general concept than reward loss also in the sense that it enables an error signal even when the agent is far away from any actual or expected rewards, but it receives ``bad news'' about future reward.\index{frustration!based on predictions only} This is in contrast to reward loss which does not make any sense unless a reward is actually expected to be  obtained right now (the exact meaning of ``now'' depends on the time scale). In particular, if the expectation of total \textit{future} reward decreases, this is enough in itself for RPE to signal frustration. %
Suppose a cleaning robot is on its way to a room where there is a lot of dust (yummy!), and thus its expected (predicted) reward is high; but this reward will in any case not be obtained for quite a while. Then, it finds that the door to the room is locked and it cannot enter; that is bad news it didn't expect. Thus the robot finds itself in a new state that has a much lower expected total future reward since the dust in that room cannot be reached. It is this difference between the earlier prediction and the new prediction that creates a RPE and suffering. This is not an ordinary reward loss because no actual reward was expected to appear at this time point anyway: the robot has not yet even entered the room, and the dust is still far away. However, RPE can create suffering merely based on predictions: if information arrives that makes the agent reduce its prediction of future reward, frustration is created. This is intuitively appealing since a lot of our frustration is actually about such negative news and the lowering of expectations they create. Suppose I'm planning to attend an event that I expect to enjoy, and then, well in advance, I hear the event has been cancelled. I will suffer, although I didn't expect to obtain anything enjoyable yet, and I may not have taken any action either; it was all just predictions in my head.\footnote{To see how this works mathematically, consider the definition of RPE (given in footnote~\ref{RPEfootnote} above) in the case where the obtained reward is zero. It makes sense to consider zero reward because it is generally agreed that rewards are temporally sparse (mostly zero), often extremely sparse, so most of the time the RPE is simply the difference between the state-values in two states (before and after, or past and present),  possibly discounted in the latter state. Recall that the state-value is nothing else than the predicted total future reward. Thus, recalling that the sign in this conventional definition of RPE is wrong for our purposes, RPE defines frustration as $V_\text{before}  -V_\text{after}$,  which is exactly the decrease in predicted total future reward, comparing the prediction in the previous time step and the present time step. Such a decrease is possible when the agent receives new information (which implies, in the basic formalism, that it finds itself in a new state incorporating that information), and that information makes it revise its prediction downward (it switches to the prediction given by the new state it finds itself in).   Thus, in the case where the prediction decreases in the absence of any reward obtained,  the reward loss or the negative part of RPE is equal to the decrease in the prediction of the total future reward. %
  This is how RPE can define frustration based on predictions alone, without any reward currently expected. One might think that reward loss could do the same if we simply change the time scale: in the robot example, if you take the expected and obtained reward for, say, one whole hour, that would arguably lead to a reward loss since the robot expected to get dust during that hour but didn't get any. However, RPE makes its computations independently of any such time scales (it is in fact taking into account the whole future as it looks at the total expected future reward) and moreover, such long-term reward loss would not occur before than hour has passed, while RPE signals frustration the very moment the new information has arrived and has been processed. (As a minor point on terminology, it may be slightly misleading to talk about ``reward prediction \textit{error}'', since RPE is in this case rather a change in predictions due to new observations; a non-zero RPE does not necessarily imply that there was any error, but simply a change, an update of prediction based on new information.
)\label{predictedfrustrationfn}
}

\section{Expectations or predictions are crucial for frustration}

Reward loss  and  RPE highlight the importance of expectations and predictions. Clearly, there must be some expectation or prediction in order for them to occur. 
If the cleaning robot were so primitive that it had no expectations or predictions at all, it might be just enjoying every single speck of dust it finds. Making it more intelligent so that it can predict the future thus deprives it of its ``innocence'', and enables frustration to occur.
Likewise, Cassell says that ``to suffer, there must be a source of thoughts about possible futures'', even though his approach to suffering is quite different.\footnote{\citep{cassell1989relationship}; such an approach will be treated in Chapter~\ref{threat.ch}.  Likewise, the importance of predictions and expectations in economic decision-making is emphasized by \citet{koszegi2006model} who propose that consumers compare expected utility given an action with a ``reference-point'' given by a probabilistic prediction of the future utility.}\label{cassellprediction}\index{error!prediction e.}\index{prediction}\index{thinking!about the future}

The importance of predictions is well appreciated in neuroscience.
It has been observed that in the brain, RPE is coded by certain neurons using a neurotransmitter called dopamine.\index{dopamine} More precisely, it is coded by quick changes in the level of dopamine (called ``phasic dopamine signal''), typically originating in evolutionarily old areas such as the midbrain, which is literally in the very center of the brain.\footnote{\citep{schultz2016dopamine,lerner2021dopamine}. In experiments with humans, such signalling might be measurable as the error-related negativity (ERN) seen in EEG measurements, as well as fMRI signals mainly in some parts of the anterior cingulate cortex where ERN seems to originate \citep{holroyd2002neural,abler2005neural,zubarev2018evidence}.}
In case the obtained reward is higher than expected, there is a temporary peak in the amount of dopamine in the signalling pathways, which is called by some a ``dopamine surge''.  That's why many drugs of abuse target the dopamine pathways in the brain. For example, cocaine\index{drugs!misperception of reward} blocks the removal of dopamine in the synapse so that its signal is amplified.\footnote{\citep{nida}, 
  but see also \cite{nutt2015dopamine}} Such drugs are fooling the reward-processing system in the brain, thus leading to a strong desire for such drugs, in addition to a pleasurable feeling.
This has led some to think that dopamine is the neurotransmitter responsible for the feeling of pleasure itself. Such a viewpoint is probably incorrect, and the actual feeling of pleasure is mainly mediated by other transmitters, namely those in the opioid family,\index{opioid neurotransmitters} while dopamine is more related to ``cold'' action selection and learning.\footnote{\citep{berridge2015pleasure,leknes2008common}. This dissociation may sound logically contradictory, but it is based on the distinction (in Berridge's terminology) between the motivational ``wanting'' processes which more directly tell the organism what to do, and the affective ``liking'' processes which are related to the feeling of pleasure.  \citet{abler2005neural} also proposes that reward loss triggers different kinds of neural processes, some of which are more related wanting, action selection and reinforcement learning, and others more to liking and the feeling of pain or pleasure; they find that the localizations of those two processes in the brain are different. See also footnote~\ref{berridge2fn} in Chapter~\ref{emotions.ch}.\index{desire!wanting vs liking}\label{berridgefn}}

\subsection{Classical conditioning}

To emphasize the importance of predictions in the brain, let's consider an extremely famous kind of prediction learning in the animal realm: classical conditioning. Ivan Pavlov, doing physiological experiments on dogs around the year 1900, observed that the dogs began to salivate when they saw the staff person who was responsible for feeding them, even before receiving any food. Pavlov was intrigued and tried to see if the dogs would be able to associate any arbitrary stimuli to food. He succeeded in making the dogs associate food with many different kinds of stimuli, including the sound of a bell or a metronome, provided that these stimuli were consistently presented just before food was given.\index{conditioning!classical}

What the animal is clearly doing here is predicting the future:\index{prediction} after the bell, food is likely to arrive. Such predictions are ubiquitous in the brain; the brain is constantly trying to predict what happens, using its myriad systems. Predicting the results of any action you might take is important if you want to choose good actions, as we already saw in the case of instrumental conditioning above (page~\pageref{instrucond}). Predicting where the rabbit will be a second or two later is necessary if you want to catch it. Note the crucial distinction between classical conditioning and instrumental conditioning:  in classical conditioning, the agent does not yet learn to \textit{choose} actions, but merely to predict future states, independently of any rewards.\footnote{In Pavlov's experiment, the dog learned to predict that food is coming, independently of its actions. It did salivate, which could be seen as an action, but the salivation was a (presumably innate) response to food that was not learned during this experiment.}

It would be natural to assume that such classical conditioning could be easily performed by Hebbian learning.\index{learning!Hebbian} It is just the kind of association of two stimuli---bell and food--- that Hebbian learning seems to be good at. That is to some extent true, although this is a bit tricky; the most successful models actually use supervised learning, with the bell as input and the food as the output. Such learning, again, proceeds by minimizing prediction error.\footnote{For a single conditioned (i.e., predictive) stimulus, Hebbian learning actually works fine, but the problem is that when there are several conditioned stimuli, Hebbian learning would create too many associations and in an unbalanced way. For example, we could have an experiment where both a bell and a green light predict food. Simple Hebbian learning would then associate both those stimuli with the food, since the association strengths would be computed independently of each other. But this is in contradiction with what seems to happen in the brain. Such interaction between predictions has been investigated in a famous twist to the basic classical conditioning experiment using the bell: after the main experiment, another experiment is made where both the bell and a newly introduced green light predict food. In such a case, the dog will not learn to associate the green light with the food because the connection from the bell is enough to predict the food, and there is no need to construct an association from the light to the food anymore. This is in contrast to what Hebbian learning is supposed to do.   The brain apparently tries to be economical and constructs only those connections that are necessary for the prediction of the food. Therefore, the association strength of one conditioned stimulus will also depend on the associations of other stimuli. This is why most research assumes a supervised model, which typically learns several such association strengths in a balanced way, and thus explains the various experiments better than simple Hebbian learning. A basic supervised learning rule accomplishing this is the Rescorla-Wagner model \citep{miller1995assessment}, which further models the dynamics of learning, as in the bell/light example just given; %
  the model explains how the existing association with the bell ``blocks'' the development of a new association with the light.}\index{error!prediction e.}

\subsection{Does a low level of rewards produce frustration?} 

Intuitively, however, it might seem that talking about frustration based on expectations and predictions is unnecessarily complicated.  If the agent is in a state with low state-value (in its own estimation), would that not in itself imply frustration? Being in a state of low value means that the agent believes it will not obtain much net reward in the future, which sounds like a good reason for mental pain. Or, even more fundamentally, why not just say that lack of rewards, presumably during recent history, leads to suffering? 

One  fundamental problem with such an approach would be that it is not obvious how to define a suitable baseline or comparison: What level of state-value is actually low, and how small should recent reward actually be to create frustration?\index{frustration!definition of baseline} The reward loss or prediction error actually solves this problem by using the expectation of the reward as the baseline. Thus, the obtained reward is compared with the expected level, and if it is ``low'' in this particular sense, frustration occurs.\footnote{The RPE formalism could also be interpreted as providing another baseline mechanism, by looking at the \textit{change} of state-values. Going to a state which has a lower value than the current state, without obtaining any reward, does produce suffering according to the definition of RPE above, as explained in footnotes~\ref{RPEfootnote} and \ref{predictedfrustrationfn} in this chapter. From this viewpoint, RPE uses the current \textit{state-value} as the baseline defining what is ``low'', looking at the total expected future reward. See also footnote~\ref{expdeffn} on different possibilities of defining the baseline as ``expectation''.\index{error!reward prediction e.} \add{Chapters~\ref{self.ch} and \ref{summary1.ch} will further discuss how such expectations can operate on different time scales, and we will see that a long-term lack of rewards can indeed be considered frustration when looking at a longer time scale.}}

\section{Unexpected implications of state-value computation}

In the rest of this chapter, I will consider some practical implications of the theory.\index{state-value!implications}\index{unexpected behavior}
First, let us consider how the computation of state-values, as proposed in basic reinforcement learning theory, fundamentally changes the behavior of human agents. 
Originally, of course, evolutionary forces demand that an action is pursued by a biological organism if the action helps in reproducing and spreading its genes, and an action is avoided if it hampers this effort. So, evolution ``tells'' us that kicking a stone is bad because it can cause damage to our foot, and the damage decreases our potential for reproduction---thus giving us negative reward for such an action.  Having sex is very good, and rewarded by basic evolutionarily mechanisms, because then we are fulfilling our deepest evolutionary calling and spreading our genes.\index{evolution}

The computation of state-values changes the situation: The organism will not only try to reach states directly giving reward---such as having sex---but also states that have higher state-values. This is a mechanism for looking forward in time: instead of immediate reward, the organism will try to maximize the total reward in the future, and that is exactly what is given by the state-value. 

Seemingly valueless states are now valued by the agent since they \textit{predict} that more actual reinforcement can be found sooner.\index{state-value!as prediction} Such states provide intermediate goals in the pursuit of the actual reward, similarly to heuristics in tree search. If you train a robot to get orange juice from the fridge, it must of course first go to the fridge, and open it. So, the state where the robot is standing next to the fridge acquires a positive state-value and we could almost say that the robot ``enjoys'' being next to a fridge, even more so if it is open.

The situation is even more complex due to the existence of human civilization and society. Culture plays an important role in determining the state-value function, and it is often difficult to separate the influences of biology and culture. 
In neuroscience, this is called the ``nature vs.\ nurture'' question.
There can be extremely complex chains of value computation which transform the original evolutionary goals to behavior based on intermediate goals.
For example, humans have evolved to strive for high social status.\index{social status} From an evolutionary perspective, this is because it helps humans get more sexual partners and increases the number and the survival probability of their offspring. This then implies that we want to increase our status: for example, winning a gold medal in the Olympics is a good behavioral goal. 
Clearly, a gold medal in Olympics has no evolutionary value in itself: it does not satisfy your hunger, thirst, or sexual appetite in itself. 
It is just an arbitrary piece of metal. There is no logical connection between such a piece of metal and sex. It is only due to a complex interplay of value function calculation \textit{and} cultural meanings that
the original evolutionary reward of sex has been subtly transformed into a goal such as excelling in sports---or science, or politics.
\index{evolution!vs learning}\index{learning!vs evolution}

Such slightly weird desires are another manifestation of the phenomenon discussed earlier: emergence of unexpected phenomena due to the interaction between the learning agent and a complex environment. If we program sufficiently sophisticated AI, the same thing is likely to happen as with human evolution. The AI will pursue goals that were not  intended by the programmers, but which still happen to produce a high state-value. This is particularly likely if the AI interacts with humans who provide a particularly complex environment to learn from---or, if the AIs are capable of complex interactions between themselves.\footnote{This is also a problem from the viewpoint of the safety of AI systems. \citet{hendrycks2021unsolved,turner2020avoiding} discuss how to build learning AI agents systems with minimum negative side effects.}

\section{Evolutionary rewards as obsessions}

Now, if we admit that our desires are based on evolution, even if quite indirectly, is that a good thing or a bad thing? Should we just follow our desires, or think twice, or even try to follow some completely different goals?  There are actually people who try to justify certain kinds of behavior by saying they are evolutionarily conditioned, i.e.\ ``evolution made me do it''. In popular science magazines and web sites, such logic is not very uncommon. Fortunately, it is rejected by many as an example of sloppy thinking.\footnote{It is a case of what G.E.~Moore called the naturalistic fallacy. Hume\index{Hume} already pointed out that you cannot infer what \textit{ought to be} from what \textit{is}. In other words, if evolution makes people behave in a certain way, it does not in any way \textit{morally} justify the claim that this way of behaving is good or acceptable.} In the following, I argue the very opposite: following evolutionary desires is often a bad idea and even morally wrong.

In fact, even the evolutionarily conditioned \textit{rewards} themselves can go wrong, sometimes quite catastrophically. One reason is that evolutionarily, we may be adapted to the environment where our evolutionary ancestors lived, often assumed to be the ``African savannah''.\index{evolution!African savannah} However, the modern world is different and, therefore, our evolutionary programming may not be very suitable.\citenew{sapolsky2004zebras,wright2017buddhism} With humans, a well-known example is the addictive quality of sugary food. The sweet taste of sugar must have signalled the high nutritious quality of food in the environment where our ancestors lived.\footnote{Another  striking example in the case of humans is pornography, where watching sexually desirable models on a computer screen is felt to be somehow rewarding by its human consumer, and leads to desire towards such pictures. Simply seeing sexually attractive people naked should indeed have a very high state-value, since that is likely to happen only when copulation is near---at least in our evolutionary past. But in the modern world, this behavior is quite dysfunctional in the sense that there is almost no chance that the consumer would be actually able to mate with those models.} But these days it tends to signal added refined sugar which is bad for your health; evolutionarily speaking, sweet taste should rather be punishing in the modern context, not rewarding.\footnote{As a kind of mirror image  of such maladaptive evolutionary desires, there is the phenomenon of chronic (persistent) pain. \citet{raffaeli2017pain} review research on how chronic pain ``entails a pathologic reorganization of the neural system'' so that it  ``loses its biologic damage signaling function'' and ``becomes a destructive force'', eventually a disease in its own right.}\index{pain!chronic}\index{evolution!and happiness} Yet, the state of having a sweet taste in your mouth is rewarding, and humans tend to try to reach such a ``sweet'' states.\footnote{In fact, it is  often difficult to define what is the actual reward and what is differences in state-values. I'm here assuming that the sweet taste is a reward in itself, and not a question of a high state-value (i.e.\ predicted future reward), but this can be disputed. It is less controversial that an Olympic medal does not produce a reward in itself, but even this is not so clear. To solve this problem, \citet{singh2009rewards} propose that the rewards should evolve so that they are correct in most environments, while state-values are then learned during an individual's lifetime for the particular environment where the individual is living.\label{rewarddefinitionfn}}

What is even more serious is that evolution makes us want particularly questionable goals, especially from  a societal viewpoint. Evolution is fundamentally based on selfish, merciless competition between different organisms (or strictly speaking,  between their genes). Many behavioral tendencies evolution has imposed on us should be seen as instruments for such egoistic competition. Evolution is all about maximally spreading our genes. It makes us hoard finite resources such as food to ourselves in order to spread our genes. It makes us violent; it even makes us go to war, again for the sole purpose of spreading genes. This is in stark contrast to most ethical systems in the world which see such selfishness as evil, and recommend quite opposite courses of action.\footnote{Admittedly, the connection between ethics and evolution is complex, and evolution seems to have conditioned some kind of altruism in us as well \citep{wright1994moral,nowak2010evolution}.\index{emotions!altruistic|see{altruism}}\index{altruism}\index{morality!and evolution} }

Even more fundamentally, the rewards defined by evolution never had the goal of making us happy in any meaningful sense. They are a force that drives us to do exactly those things which are good for our evolutionary fitness. Even if you come to the conclusion that the evolutionary reward system makes you suffer, it cannot be switched off or modified. You cannot decide to be rewarded by something you consider more meaningful and good for society. 

I suggest evolutionary rewards lead to what can be called \label{evobs} \textit{evolutionary  obsessions}.\footnote{I am here using the term ``obsession'' in a loose sense, not using its strict psychiatric definition. For reference, in the current ICD-11 proposal, obsessions are defined as follows. ``Obsessions are repetitive and persistent thoughts (e.g., of contamination), images (e.g., of violent scenes), or impulses/urges (e.g., to stab someone) that are experienced as intrusive, unwanted, and are commonly associated with anxiety. The individual attempts to ignore or suppress obsessions or to neutralize them by performing compulsions. --- Compulsions (or rituals) are repetitive behaviors (e.g., washing, checking) or mental acts (e.g., repeating words silently) that the individual feels driven to perform in response to an obsession, according to rigid rules, or to achieve a sense of 'completeness'. ''\citep{stein2016classification}\index{obsessions!clinical definition}}\index{obsessions!evolutionary}
That is, the evolutionary rewards, together with the learned state-values, make us desire, even crave for many things which we would actually prefer \textit{not} to desire if we could rationally decide what we desire. If you could just consciously, rationally, ``switch off'' your desire for, say, sugary food---would you not do that? 
Chapter~\ref{freedom.ch} explains how Buddhist and Stoic thinking are based on the rather extreme tenet that switching off \textit{all} desires would actually be very good for you.
Whether one agrees with that extreme viewpoint or not, surely, most people have certain desires that they would rather not have. I call them obsessions because they are automatically created, they often override any conscious deliberation, and they may even feel unwanted and intrusive. (We will look at the computational mechanisms for this in Chapters~\ref{dual.ch} and \ref{emotions.ch}.)

\section{Reward maximization is insatiable}

Finally, let me mention another dark side to this reinforcement learning theory.\label{insatiability}\index{insatiability}
One crucial property of the algorithms based on reward prediction error is that they drive the system to get more and more reward, and there is never any long-term satisfaction.
This is because any prediction of the future is learned by the agent, and \textit{constantly updated by learning}. Thus, in the reward loss, the level of expected reward is updated based on what the agent has obtained recently.

\index{robot!cleaning}
Suppose that an agent gets an exceptional amount of rewards for a while, maybe because a cleaning robot finds itself in a building with lots of nice dust to clean, and it is rewarded for every speck of dust it sucks away.  Now, the agent's prediction system is updated so that an equally large amount of rewards is predicted in the future as well. An environment that produced an unexpectedly large amount of reward for a while becomes the new baseline. That level of reward is not unexpected anymore and, therefore, does not produce any particular ``pleasure'' anymore either.\footnote{This is related to the phenomenon of the ``hedonic treadmill'' \citep{lyubomirsky2010}.} 
What's worse is that when things get back to normal, the agent will get less rewards than what it has now learned to expect, since the prediction was updated to reflect the particularly nice environment that lasted for a while. Therefore, the agent suffers enormously when it has to go back to a normal room with a modest amount of dirt. 

Similar computations take place in our brain, since our brain also computes the reward prediction error and updates its expected level of reward. No wonder that Wolfram Schultz, one of the leading neuroscientists on dopamine, calls the dopamine neurons ``little devils''.\citenew{schultz2016dopamine}\label{treadmill}\index{dopamine}
In fact, this is a logical consequence of the guiding principle of AI agent design: the agent should maximize obtained reward. The reward prediction system has no other goal than helping in maximization of rewards. If you program an agent to maximize reward, then by definition, nothing can possibly be enough; the system will be \textit{insatiable}. The agent will relentlessly try to get more and more reward, and it is precisely the frustration signal that will force the agent to try harder and harder.\footnote{\add{ \citet{dubey2022pursuit} provide an explicit model on how such insatiability is computationally useful while leading to less happiness. } \citet{lambie2019understanding} consider insatiability as an important component of greed.\index{greed} On the other hand, it is true that some purely biological needs are satiable to some extent---for example, hunger is reduced by eating, even if momentarily---but classical reinforcement learning theory is lacking much consideration of such metabolic states \citep{keramati2014homeostatic}. \add{See footnote~\ref{epicurusfn} in Chapter~\ref{freedom.ch} for some ancient philosophical references on the topic.} 
}

A merciful programmer might program some stopping criterion to limit the greed of the agent: Once you have obtained X units of reward, you can stop. Unfortunately, evolution knows no mercy, and humans don't seem to have any such stopping criterion programmed in them. We need more money, more power, more sex (and better sex), and better food (and more food). If we follow our evolutionary ``obsessions'', as I called them, nothing is enough. 

Suppose you program a robot called Pat to clean a building.\index{robot!cleaning}
You would like the building to be superclean, and the building is quite large with dozens of rooms. So, you would be very tempted to program Pat so that it will spend all its time cleaning the building. You probably want to program a couple of other functions in Pat as well, such as a routine for charging its batteries, some basic maintenance procedures, as well as safety systems to prevent it from hurting people or breaking things. But you would probably program Pat to spend all the rest of the time in tirelessly cleaning the rooms, with no breaks in between. This is what most programmers would do.
Here, you have implemented a kind of a ``cleaning drive'' which is without mercy. Pat will spend all its time and energy just making the rooms spotlessly clean. This may seem completely natural, given that it is ``just'' a robot. 

Now, suppose your colleague, responsible for the visual design of the robot, decides to make Pat look really cute, giving it the shape of a little kitten. It even says ``Meow'' using its loudspeakers. Many people may suddenly start feeling sympathy for this poor little kitten. ``Does it really have to be working all the time? Can't it ever play, or take a rest?'' they would ask. What would you reply?

\chapter{Suffering due to self-needs} \label{self.ch}

\add{In addition to frustration, Chapter~\ref{suffering.ch} identified another cause of suffering: threats to the intactness of the person, or the \textit{self}. In this chapter, I consider the concept of self, while the concept of threat is treated in the next chapter.}

Self is a concept with a bewildering array of meanings. Psychology, philosophy, and neuroscience offer a multitude of definitions, and I can make no claim to treat the concept comprehensively.\index{self-needs|see{self, needs}}\index{self!needs}
I focus here on two meanings of ``self'' directly related to suffering.
First, self as the target of evaluation of some kind of long-term success of the agent. The human brain, in particular, has a system that constantly evaluates the agent, checking whether the goals set were reached or rewards obtained, and seeks to improve its general performance.
Second, we have self as the target of self-preservation, or survival instinct: all animals have behavioral tendencies to avoid death or organic damage. (A third meaning of self, related to control, will be treated in Chapter~\ref{control.ch}, and the concept of self-awareness, in Chapter~\ref{consciousness.ch}.)

Such self-evaluation and self-preservation are computational mechanisms which are constantly operating in animals, and it is easy to justify their computational utility for any intelligent agent.  Although at first sight, these aspects of self may seem to provide a mechanism for suffering which is completely different from frustration, this chapter shows  how they are related to frustration of internal, higher-level goals and rewards. As the title of this chapter indicates, these aspects of self can thus be seen as needs, or desires, and  they can be frustrated.

\section{Self as long-term performance evaluation}

Let us start with self as something whose performance is being constantly evaluated at different levels.
As we saw earlier, in reinforcement learning, every single action is always evaluated to improve future actions. The reward prediction error is computed even in the simplest algorithms.  If the reward is incorrectly predicted, the error is used by the learning algorithm to improve the prediction---if the prediction was too high, set it lower in the future, for example. While such computations are crucial for learning to act optimally,
the errors also trigger the suffering signal according to the theory of the preceding chapter. 

However, the situation is complicated by the fact that the learning algorithms themselves contain many parameters describing how the algorithm itself works. 
One fundamental parameter is how quickly the system should learn: if it learns too quickly, the new information will tend to override the old one, thus leading to forgetting. Below, we will see another parameter which is how much of the time the agent should spend on relatively random exploration of the environment. There are many such parameters in a sophisticated learning system.

Therefore, sophisticated AI should be able to adjust such internal parameters by itself. This is called \textit{learning to learn}.\citenew{thrun2012learning}\index{learning!to learn} Such learning to learn requires constant monitoring of the performance of the basic learning algorithms. If the current internal-parameter settings do not lead to good learning, adjustments have to be made. This requires an internal signalling system, not unlike the suffering signal, but typically working on a longer time scale, since it takes a long time to see if a learning system learns well.

\subsection{Self-esteem and depression }

In humans, mood is a signalling system working on a longer time scale. Mood is defined as an emotional state which is more long-lasting than single emotional episodes (such as being angry or feeling afraid, which are considered in Chapter~\ref{emotions.ch}). A low mood may take days, if not weeks or months, to change. A psychological concept which works on an even longer time scale is self-esteem: an overall view of the self as worthy or unworthy.\citenew{heatherton2003assessing}\index{self!esteem and evaluation} 

Depression may in fact be an extreme case of the performance signalling made by the self-evaluation system. One theory\index{depression}
proposes that depression occurs when goals are not reached, and moreover, constant attempts to improve performance fail.\citenew{thierry1984searching,nesse2000depression} That is, the agent has to admit that whatever it tries, nothing works. 
In such a case, there is still one last strategy that may help: wait and see. The environment may eventually change by itself, even if you do nothing. Perhaps, after a while, with some luck, the circumstances will be more favorable. Such a ``wait and do nothing'' program may explain some depressive symptoms, such as passivity and lack of interest in any activities.\footnote{A very similar account has been proposed for the simple emotion of sadness by \citet{oatley1987towards}. The difference is mainly in the time scales involved, since depression is by definition much more long-term than sadness. Sadness in its turn could be seen as a frustration or disappointment signal which is particularly strong and relatively long-lasting, but the terminology here is not very well-defined. A related computational account of depression focusing on the concept of learned helplessness is given by \citet{huys2009bayesian}; \add{ \citet{eldar2016mood} propose a theory for both negative and positive moods based on tracking RPE;  \citet{stephan2016allostatic} also link depression to prediction errors.} }  

It would clearly make sense to program such a ``depressive'' mechanism in an AI. If the current algorithms are simply not working at all, it would be better for the agent to just wait and see if the world changes for the better. Such waiting will save energy, and perhaps will also enable the AI to perform some further computations to improve its performance in the meantime.

\add{
Like with frustration, we have to ask the expected level of performance in such computations comes from. What level of rewards is considered enough by the self-evaluation system, and what level produces frustration? In humans, that must be biologically determined to some extent, but social comparison\label{socialcomparison}\index{social comparison} is another important mechanism for determining when a person's performance is ``good enough''. That is, a person can compare his/her reward level with others to judge if it was acceptable. In addition, a person is constantly evaluated by other people, which is another source information for the self-evaluation.\footnote{\citet{vogel2014social} discuss social comparison as a basis for self-esteem. In humans and other social species, how oneself is seen by others is an important aspect of the very concept of ``self''  \citep{sebastian2008development,heatherton2011neuroscience}\index{social interaction!and evaluation}\index{social comparison}; evaluation is only one aspect of such a socially defined self-concept.\label{socialevaluationfn}\index{self!as seen by other agents}}  
}

\subsection{Self-destructing systems}

What if an AI comes to the conclusion that it is not able to fulfill its task at all? Perhaps something went very much wrong in the design of the learning algorithm, or  the task is completely impossible, and the circumstances do not seem to change for the better. The most extreme solution would then be for the AI to ``destroy'' itself.\index{self!destruction}

Suppose you launch many AI agents, or programs, that work more or less independently inside some computing system. 
If one of the agents is not achieving anything, it would be natural that you terminate its execution. This would free up computational resources for other agents---assuming all the agents are running on the same shared processors---and other agents might be more successful.  
To make this possible, there has to be a system for evaluating each AI agent's performance as a whole. Importantly, the evaluation does not have to be done by an external mechanism; it could be part of the agent itself, which could then decide to self-destruct.
There is nothing paradoxical or impossible in such a self-destruction system. It can be explicitly programmed in the agent by a human programmer---while it may indeed be quite impossible for the agent itself to learn such self-destruction behavior. 

It is possible that in some cases, even biological organisms may engage in such self-destruction sequences. Such an idea is quite speculative because it is not obvious why evolution would favor such behavior. It is clearly possible that the designer of an AI system can explicitly create the self-evaluation and destruction systems, but in biological evolution, there is no such explicit designer. It may actually sound completely nonsensical to think that evolution could lead to self-destruction mechanisms, since an organism which destroys itself cannot spread its genes anymore.\index{death}

However, evolution is a bit more complicated than just the survival of the fittest individual. It is widely appreciated that in evolutionary arguments, we should take into account not only the survival and reproduction of an individual, but also the survival and reproduction of the closest relatives. This leads to the concept of ``inclusive fitness'', where the fitness of an individual takes into account the fitnesses of the relatives weighted by the proportion that they share genes. Close relatives of an individual spread partly the same genes anyway, so their survival is evolutionarily useful for that individual.
According to one suggestion, if a person is seriously ill, and finds himself a great burden to his relatives, it might actually be evolutionarily advantageous for that person to commit suicide. If this helps the relatives with whom he shares a large proportion of genes, the suicide might actually help in spreading those genes, thus increasing the inclusive fitness.\footnote{\citep{decatanzaro1991evolutionary}. However, see \citep{nowak2010evolution} for a criticism of the centrality of kinship in the inclusive fitness theory. Related work on suicide and self, but without the evolutionary interpretation, is by \citet{baumeister1990suicide}. Taking the logic of inclusive fitness even further, one may be tempted to think of natural selection working on the level of groups of organisms (families, tribes, herds, etc.), so that it is the fittest group, not organism, that survives the selection.  However, any theories based on such ``group selection'' are controversial, and it is not clear if it actually happens in nature. Some mathematical theories propose that natural selection on the level of individuals leads to emergence of phenomena which look just like the selection happened on the level of groups. In fact, according to the mathematical model by \citet{hadany2006stress}, something like self-destruction could actually emerge from purely individual-level selection if an individual organism finds the current environment particularly adverse.\label{catanzarofn}}\index{suicide}

Thus, self-destruction programs may be useful not only to maximize the utility of AI agents, but also from an evolutionary perspective. This may sound abhorrent from a moral perspective,\index{morality} but that is often the case with evolution which has no reason to be nice or good from a human perspective---as already argued in the preceding chapter, where I compared evolutionary desires to obsessions.

\section{Self as self-preservation and survival}

Another rather obvious reason why some kind of concept of self should be programmed in an AI is that the AI may need to protect itself against anything that might destroy it. A robot must take care not to be run over by a car: This is the concept of self-preservation. There is no doubt we can, and probably want to, program some kind of self-preservation mechanism in an AI agent.\index{self!preservation}\index{survival|see{self, preservation}}

Even the simplest biological organisms have behavioral programs that are activated when their existence is threatened; we talk about self-preservation, or survival instinct. We already encountered related ideas in considering definitions of pain and suffering. The widely-used IASP definition related pain to ``tissue damage'' (page~\pageref{iasp}), while Cassell's definition of suffering talked about the ``intactness of the person'' (page~\pageref{casselldef}). However, what we are talking about here is threats to the very existence of the agent, not just damage.\index{pain!IASP definition} 
\index{intactness of the person}

While it seems relatively straightforward to program self-preservation behaviors in an AI, an open question is whether an AI can somehow develop a survival instinct by itself. In other words, can self-preservation emerge without being explicitly programmed; can the agent learn to perform certain actions for the main purpose of avoiding its own destruction? This is one of the deepest questions in AI, extremely relevant from the viewpoint of developing safe AI systems, and the subject of intense debate.\footnote{A highly readable account can be found in 
\textit{Vanity Fair}, 
``Elon Musk's Billion-Dollar Crusade to Stop the A.I. Apocalypse'', April 26, 2017. 
} We have seen earlier that learning in AI can have various side-effects and unintended consequences; this would be one of the most extreme ones.\index{unexpected behavior}

On the one hand, there are those who point out that biological organisms have developed their survival instinct via evolutionary mechanisms. They have been subject to natural selection, which has ruthlessly weened out those organisms which do not fight for their survival. In contrast---this line of argumentation goes---AI is not subject to natural selection; it has no evolutionary pressures. So, it will not learn a survival instinct, unless perhaps we explicitly decide to program it to learn one. 

Other experts disagree and point out that some kind of survival instinct may be automatically created as an unintended side-effect of creating sufficiently intelligent machines. If a robot is given any mundane task, say fetching a bottle of milk from a near-by shop, a super-intelligent robot would understand that in order to perform that task, it has to stay alive. If the robot were damaged or destroyed in a collision with a car, for example, its task cannot be performed. Thus, the robot might decide to destroy the car somehow (let's assume the robot is really big) to get the milk safely delivered. If everybody in the car gets killed, that is irrelevant, if the programmer didn't tell the robot to avoid human casualties. The idea here is that there is no need to explicitly program a survival instinct, or any reward related to that: the general goal of maximizing future rewards will direct the robot's behavior towards avoiding destruction. In fact, this line of thinking means that almost any sufficiently intelligent AI will \textit{by logical necessity} strive to survive. If it is intelligent enough, it will understand what death is, and how death makes it impossible to obtain any  further rewards or accomplish goals. This is the opposite of what has happened in biological evolution, where even the very simplest organisms have a survival instinct, and sophisticated intelligence develops later. In AI, intelligence is programmed first, and later, possibly by chance, the AI might obtain a tendency for self-preservation behavior and related information processing, which might then be called a survival ``instinct''. 

Clearly, these two views are based on very different assumptions about the AI. The argument where the robot understands that a car on a crash course has to be destroyed assumes a very, very intelligent robot. The robot must have a sophisticated model of the world, infer that it risks being overrun by the car, and understand that being overrun by the car will prevent it from delivering the milk. Most current robots would be nowhere near the intelligence required---but we don't know if they will be in the future. We are even further away from an AI which could intellectually infer, on an abstract level, that there is such a thing as death, and that various measures should be taken to avoid it. 

Nevertheless, if an AI is learning using evolutionary algorithms instead of the conventional gradient-based algorithms, it might be perfectly possible for an AI to obtain a survival instinct, even at the current level of AI development. 
As reviewed earlier (page~\pageref{GA.sec}), optimization procedures mimicking evolution are already used in AI. 
Large-scale application of evolutionary algorithms definitely has the potential of creating a survival instinct in AI agents. It is a necessary logical consequence of fundamental evolutionary pressures: To spread its artificial ``genes'', an agent has to survive long enough to produce offspring if the evolutionary optimization method is similar enough to biological evolution.

\section{Self-related suffering as intrinsic frustration}

Going back to our main topic, suffering, it is clear that both self-preservation and self-evaluation are important sources of suffering.\footnote{It is not my goal here to define what self \textit{is}, I am merely considering how phenomena typically associated with ``self'' are related to suffering, amplifying or even producing it. In fact, there is some ambiguity in this chapter regarding whether self in, say, self-evaluation is the target of evaluation or the system that evaluates; and whether self-evaluation can be seen as a process that somehow leads to the emergence of self. %
  Similar ambiguities hold for self-preservation, as well as the further discussions of self-related phenomena in later chapters, in particular, self as control in Chapter~\ref{control.ch} and self-awareness in Chapter~\ref{consciousness.ch}. This ambiguity may be related to the distinction between the ``I'' and ``me'' aspects of self, i.e.\ self as subject or object, proposed by William James.\index{James, William}\index{self!I or me (James)}}
First, it is well-known that depression and low self-esteem create suffering--- and they are largely produced by the self-evaluation system.
It is, in fact, rather easy to see this as a form of frustration, so it is very much in line with the ideas of the preceding chapters. Self-evaluation is based on a set standard of how good the self should be, in terms of how much reward it should be able to obtain. If such self-evaluation returns a negative result, that can be seen as a form of frustration, similar to reward loss. One could say that the agent had a long-term desire to achieve that standard of average rewards, but the agent failed.\index{self!as desire}\index{desire!and self}

Second, self-preservation is obviously behind (physical) pain, which is signalling when damage is happening to the physical organism, according to the IASP definition of pain (page~\pageref{iasp}). The same idea was extended to suffering by Cassell's definition (page~\pageref{casselldef}). He emphasizes ``loss of the intactness of person'' or ``threat'' thereof, and that this applies not only to physical intactness but to further aspects such as one's self-image. Replace his term ``person'' by ``self'', and an interpretation related to the discussion in this chapter is clear: self-preservation mechanisms signalling threats to self---even in a very wide sense of the word---directly create suffering.\index{pain!IASP definition}\index{intactness of the person}

Thus, in line with the literature review in Chapter~\ref{suffering.ch}, we seem to have two different kinds of suffering related to self-needs. One is born from frustration, in this case based on self-evaluation, and easy to understand by the theories of the preceding chapters. The other kind of suffering comes from a threat to the self, and has only been considered in this chapter. %
But in fact, self-preservation can be seen as a long-term goal or desire: the desire to survive.  It can be frustrated like any desire.
This is in line with van Hooft's theory of suffering (page~\pageref{hooft}),\index{suffering!definition!van Hooft} where different aspects of one's being have different needs, the ``lowest''  being precisely the need or desire for biological survival.  This shows how the two different mechanisms of suffering identified in Chapter~\ref{suffering.ch}, i.e., frustration and threat, have a much closer connection than it might first seem. \add{This connection is further explored not only in the rest of this chapter, but also in Chapters~\ref{threat.ch} and \ref{summary1.ch}.}

\subsection{Internal rewards and intrinsic desires}

In computational terms, a direct way of linking self and desires is based on the concept of \textit{internal rewards}, or intrinsic motivation.\index{reward!internal}
Reward is, by definition, what an AI agent ultimately wants when it is trained in the conventional framework. As we have seen, just wanting immediate reward is quite short-sighted: If the agent is intelligent enough, it will try to compute the state-value function and thus take future rewards into account. But, even the state-value function framework, with discounted future rewards, may not always provide the best practical solution to the problem of maximizing rewards. This is because the value function may be extremely difficult to learn: there may not be enough data to learn it, and even with enough data, it may be incredibly complex to compute.\footnote{In the simplistic case of a finite number of discrete states, computing the value function is not a problem at all, only learning is. But in realistic scenarios, the value function would be computed by something like a neural network based on sensory input, and these computations can be challenging.} 

Therefore, it has been found that it is often useful to program some additional rewards in the agent, in particular rewards that somehow improve its long-term functioning.  That is, the system is programmed to receive internally generated reward signals in addition to actual, ``external'', rewards. These internally generated reward signals are treated by the learning and planning systems just as if they were real reward signals. Such internal rewards lead to what is called ``intrinsic motivation'' for behavior; it could also be called \textit{intrinsic desire}, and can lead to intrinsic frustration.\index{desire!intrinsic}

As a practical example of such  internal reward, let us consider \textit{curiosity}, which is widely used in current AI.\index{curiosity}
The starting point here is that when an agent learns in a real environment,  the data it receives is strongly influenced by its own actions. If the robot never enters a room, it will not know what is in that room. The action of deciding to enter or not to enter that room will strongly impact the data it gets about that room. This is a problem since usually, the agent does not know what kind of actions create useful data.
Therefore, learning to act intelligently necessarily requires a lot of trial and error. That is,  the agent just tries out what happens when you do something rather random in each possible situation.
Such exploration is actually imposed on almost any agent learning by reinforcement learning. A very simple way of achieving that is to somehow randomize the actions: for example, in 1\% or 10\% of the time steps, the agent could take a completely random action just to see what happens.\footnote{See e.g.\ \citep[Ch.~2]{suttonbook}; for neuroscience results, \citep{costa2020primate}. Similar randomness may even be useful in the motor system, where it is often, perhaps erroneously, considered unwanted noise \citep{dhawale2017role}.}

If you want to buy a new electronic gizmo you have never bought before, a basic exploration strategy would mean you just randomly enter different shops, try to buy it, and depending on whether they sold it to you or not and with what price, you slowly update your value function. Most of your time would probably be spent in trying to buy the gizmo in fashion stores that don't stock any. Because your actions are quite random, you will end up going to the same stores several times, to the great annoyance of the shop assistants. Since you move around randomly, you easily end up going round and round in the same neighbourhood. Gathering data for reinforcement learning is thus particularly difficult because the agent needs to try out different actions, but if it is done completely randomly, much of the time it will take actions that are not very useful for learning, and don't bring any reward either.\footnote{\add{While it is not the main point here, we encounter what is called the ``exploration-exploitation trade-off'', which means the agent cannot very well simultaneously both gather new information and use previously acquired information to obtain reward. To put it simply, when the agent is randomly exploring, it is unlikely to get a lot of reward since it is not even trying.}} 

Here we come to the idea of \textit{curiosity}. It means that the agent does not try out completely random actions, which is very inefficient, but there is an internal mechanism that steers the exploration in an intelligent way.
What we are talking about here is designing an intrinsic reward system that leads to particularly efficient exploration.\citenew{schmidhuber1991curious,mirolli2013intrinsically,pathak2017curiosity,hazan2019provably}
Basically, the agent should try out new actions if they are \textit{informative}. If the agent has never tried a certain action in a certain state, and it has no information that enables it to infer what such an action would do, it would be useful to just try it out. That is, instead of completely randomly trying out new actions, the agent should try out actions whose effects it does not know and cannot predict. This is a more sophisticated form of exploration, and similar to what we would call curiosity in humans: try out things which you never did before---but don't repeat them once you've seen what happens! An intrinsic reward should then be given to the agent every time it successfully engages in such curious exploration and obtains new information.

Curiosity enables the agent to better learn the general structure of the world it is living in, since it will more systematically explore as many possibilities of action as possible. 
Such exploration can greatly improve future planning, since the agent will learn a better model of the world, and thus it indirectly contributes to future reward.\footnote{An abstract  way of justifying curiosity is that basic iterative learning mechanisms such as gradient descent often get stuck in what is called ``local minima'' of an objective (error) function.\index{local minima} That means a point in the parameter space that has a better value of the objective function than any other point near-by, but so that there is a point far-away in the parameter space which has an even better value. A special class of optimization methods called ``global optimization'' tries to improve iterative algorithms so that they might find the global minimum, that is, the very best value for the parameters, or at least something better than simple gradient descent. Bayesian optimization is one class of such methods \citep{gutmann2016bayesian,brochu2010tutorial}.}
In the gizmo shopping example above, you would \add{try out different shops, but you would} not enter the same store twice, since re-entering the same shop gives little new information. You would also get an internal reward for going to a different street, even a new neighbourhood, which certainly increases your chances of finding the right kind of store.
It is likely that such curiosity has been programmed in animals by evolution.\footnote{\citep{singh2010intrinsically} One might ask whether such curiosity could not be learned by the agents as part of the reinforcement learning process. That might be possible in principle, but it would probably take too long. An animal would learn to be curious when it has reached a certain age, but it is probably more useful for animals to be curious when they are young, as tends to be the case in biology. In AI, researchers also assume that such curiosity must be explicitly programmed.}

\subsection{Self and suffering in Buddhist philosophy}

\index{self!as desire}\index{desire!and self}
In line with the idea of desire for internal rewards, the Buddha mentions three different kinds of desires: desire for sense pleasures, desire to be, and desire not to be. While the first one can be interpreted as desire for rewards in the ordinary AI sense, the ``desire to be'' can be interpreted\index{self!in Buddhism}\index{Buddha}
as desiring the self to simply be in the sense of surviving, and further that the self should be something particular. In this interpretation, the ``desire to be'' corresponds to the self-needs as defined in this chapter. (The ``desire not to be'' could be the desire that the self is not something which is considered bad.)\index{self!needs}
Thus, even in early Buddhist philosophy, suffering related to self has been to some extent reduced to suffering related to desires and frustration.\footnote{See \SN{56.11}; my interpretation follows \citet{teasdale2011does}. Different interpretations are possible: One is that ``desire to be'' means desire that something in the world should be in a certain way. On the other hand, ``desire not to be'' could possibly express suicidal tendencies---all these desires were condemned by the Buddha.\index{suicide}}

In later schools of Buddhism, the importance of self was greatly magnified, and some texts even seem to attribute all desires and all suffering to the existence of the ``self'' (sometimes translated as the ``ego'') or attachment to it.\index{self!as desire} This means viewing the connection between desires and self from the opposite angle, considering the self as the \textit{source} of all desires---instead of the self being the target of some very specific desires as in this chapter.
\add{While it is clear that, for example, self-preservation  requires certain actions to be performed, leading to desire towards some particular goals, the claim in later Buddhism is that \textit{most} of our desires could be traced back to such self-needs. In our termonology, in a sufficiently sophisticated agent, internal rewards could determine a very large proportion of the behavior.}\footnote{Another important way in which frustration and self are related is that frustration is particularly strong if the cause of the frustration is attributed to the self (``it was my fault''). However, such attribution of causes is a complicated issue I will not discuss here; see \citet{mancinelli2021internality} for a computational treatment.\label{selfattributionfn}}

\subsection{Programming self as internal rewards and desires}

Internal rewards thus provide a unifying framework for understanding the self, or at least some of its aspects.
The self-evaluation system is nothing else than an internal reward (and punishment) system, which steers the agent's behavior on a higher level.
The difference to ordinary rewards is not only that these self-evaluation rewards come from the internal evaluation system: another fundamental difference is that the self-evaluation system is giving internal rewards to the ``learning to learn'' system, which sets internal parameters of the system.
That system does not directly affect the plans made by the agent, but it tries to improve the general functioning of the planning system to improve all future planning.
\add{Furthermore, these internal rewards work on a longer time scale, and any ensuing intrinsic frustration can  also be  long-term.}\footnote{The distinction between external and internal rewards, or internal and external motivation as they are called in psychology, may not always be very clear. Both come from the programmer or evolution anyway. See footnote~\ref{rewarddefinitionfn} in Chapter~\ref{rpe.ch} which discusses how the difference between  rewards and learned state-values is not always clear; a similar logic has been applied on internal vs.\ external rewards by \citet{singh2010intrinsically}, see also \citet{doya2005cyber}. \add{Likewise, in the discussion of this chapter, it may not be clear if the self-evaluation system should generate a frustration signal or a negative reward signal when the long-term performance is lower than the standard required. Presumably, equivalent computations can be performed in both of those two ways. However, if we assume suffering is generated by frustration, not negative rewards per se, we have to assume the self-evaluation system generates a frustration signal, in order to explain the suffering caused by self-evaluation.}}

Regarding self-preservation, most reasonable programmers would  assign a large negative reward to the destruction of the agent, since losing the agent tends to be expensive. Then, the planning system will try to avoid states leading to the agent being destroyed. In fact, you would ideally program the agent so that it keeps quite far away from anything like destruction. This is possible by programming an internal reward which gives a negative reward at any state that is even close to destruction.\index{self!based on  internal reward} In other words, any perceived \textit{threat} to survival triggers a negative internal reward signal. 
Thus, the agent tries to avoid even any threatening situations, as if it had a desire for something like safety, meaning the absence of threats. %
However, this is only part of what a threat is all about; in the next chapter, we develop a general theory of threat.

\chapter{Threat as anticipation of possible frustration}\label{threat.ch}

\add{
  Already in Chapter~\ref{suffering.ch}, we saw the idea that threat is another cause for suffering, possibly very different from  frustration.
Threat was also briefly considered in the preceding chapter, as being related to survival. However, threat is actually a much more general concept. In this chapter, a general definition of threat is developed in our computational framework. 
This requires looking deeper into the application of probability theory in AI, which is largely drawing from the vast literature of decision-making in economics.\index{threat}

In our definition, the perception of a threat is fundamentally an inference that something quite bad might happen in the future, with some probability. Crucially, threat detection means computing beyond the expected rewards that are the basis of the conventional theory of reinforcement learning. Our definition of threat is based on looking at the whole probability distribution of future rewards, including various aspects of uncertainty of future reward.

Although threat thus provides an alternative framework to frustration, we will see that there are many links between the two concepts. In particular, in our definition, a threat is always based on an inference about the possibility of  frustration occurring in the future. To put it very simply, a threat is always a threat of frustration. Thus, frustration is primary in the sense that without frustration, there could be no threat.

\section{Decision-making under uncertainty}

In the simplest models of AI, the world is seen as a deterministic system. The robot decides to turn left, and so it turns left. It decides to go forward, and it will go forward. As long as the robot understands the basic regularities of the world---for example, that it cannot go through walls---the world is entirely predictable. It may not be entirely controllable, though, because of walls and other nuisances, but there is no uncertainty about what will happen when the robot takes a certain action.

Deterministic modelling was another problem with Good Old-Fashioned AI. In reality, the world is quite unpredictable and not deterministic. An obstacle, such as a human pedestrian, can appear where there was supposed to be none, and the robot cannot go forward. It can start raining and the robot can get stuck in a mud pool. Many unexpected things can happen to human agents as well, often due to other human agents' unpredictable actions.

In reinforcement learning, such unpredictability was, of course, the basis of frustration: The agent expects a certain amount of reward but does not get it. 
In Chapter~\ref{rpe.ch}, such a prediction was formalized using the definition of mathematical expectation:
if the probability of obtaining a reward is 50\% and the reward is 10 pieces of chocolate, the expected reward is 5 pieces of chocolate. Frustration meant that the agent computed the expected reward, but the reward was uncertain, and the prediction turned out to be wrong. 

In the basic theory of reinforcement learning, only this expectation is used in the prediction, and the fact that there is uncertainty is basically forgotten. 
However, a really intelligent agent will not be satisfied with just computing the expected reward, which is a single number. It acknowledges that the world is unpredictable, and it will try to understand just how unpredictable any given reward is. It is one thing to predict you get 5 pieces of chocolate for sure, and another thing to predict that you have a 50-50 chance of getting zero pieces or 10 pieces. If you try to describe a lottery, it is rather uninformative to say that each ticket will win 50 cents on the average. Such an average does not have a lot of meaning, and you really want to know what kind of prizes you can win and with which probabilities. 

Thus, a sophisticated agent will try to compute the probabilities of all the different amounts of reward that it might get after a certain action. In mathematical terms, it will predict the whole \textit{probability distribution} of reward. 
Computing the whole distribution gives the agent much more information to be used in the decision-making: It will be able to make different choices in cases where the expected reward is the same for different actions, but the distributions are otherwise different.

\section{Risk aversion and economic gambles}

As a fundamental example of how an agent might use the whole distribution of rewards,
consider again the case where a reward of 10 chocolate pieces is obtained with 50\% probability, so that the expected reward is 5 pieces of chocolate. Such a situation is called a \textit{gamble} in economic theory, and a lot can be learned about human behavior by looking at what kind of gambles human agents prefer.

So, let us contrast the gamble just defined with a deterministic ``gamble'' where the agent actually gets 5 pieces of chocolate for sure, without any uncertainty.
The basic theory using expectations only says that the two chocolate gambles are equally good, since the expectations are equal. A simple AI agent might use that theory, and if it is given the choice between these two gambles---the 50-50 gamble or the sure-thing gamble--- it will not care which one it chooses because it thinks the gambles are equally good. However, this is not at all the case with most humans.

One of the most robust findings in studies of economic decision-making is that humans do not like uncertainty. Most human agents would choose the certain 5 chocolate pieces instead of the 50-50 gamble with 10 pieces.\index{gamble} People are even willing to pay to reduce uncertainty: a typical person in an economic experiment might prefer getting only 4 pieces for sure instead of the 50-50 gamble with, possibly, 10 pieces. A gamble with 4 pieces for sure has an expectation which is one piece lower than the 50-50 gamble with 10 pieces (4 pieces vs.\ 5 pieces); this means the person would be ``paying'' one chocolate piece to reduce uncertainty.
Such a tendency to avoid uncertainty is called \textit{risk aversion};\index{risk!risk aversion}\index{risk} it can be evolutionarily advantageous and is observed even in animals.\citenew{zhang2014origin,platt2008risky}
In addition to affecting rational economic calculations, uncertainty also \textit{feels} unpleasant.\footnote{\citep{hirsh2012psychological,peterson199915}. \citet{herry2007processing} show that unpredictability activates the amygdala, which is central in fear processing. Uncertainty is also an important factor in stress \citep{koolhaas2011stress,de2016computations}.\index{stress}}
Psychological experiments show that uncertainty can even make physical pain feel worse.\footnote{\citep{yoshida2013uncertainty,seymour2019pain}
As those references point out, lack of control \index{pain!uncertainty/controllability}
also increases physical pain, possibly because the warning signal in pain has to be taken more seriously when the agent cannot do much about the situation and cannot avoid the threat that causes the pain (warning) \citep{wiech2008neurocognitive}.\label{painmodulatedfn}}

Of course, risk aversion should not be so dominant that it ruins your chances of getting any reward. Suppose you're offered a free lottery ticket with which you might win, say, a big chocolate cake. Common sense says that you should take it ---disregarding any health issues with eating a whole cake---since you can only win, and there is no cost. However, if you're really incredibly risk-averse, you should refuse it because the ticket introduces uncertainty. Perhaps you have to wait for a week to know the results, and you would suffer from uncertainty for several days. Few people would be that risk-averse, though.  Nevertheless, this example may not be  as unrealistic as it seems. Suppose the prize is not a chocolate cake but something you really want, while the chances of winning the lottery are extremely low. It is possible that you would suffer quite a lot from the uncertainty while waiting for the result, perhaps in the form of physiological stress symptoms; an elevated blood pressure might even kill you. Therefore, for some people, it might be better not to accept the lottery ticket. They might regret it afterwards, but that is another story.

The theory of risk aversion is the basis of our definition of threat below. Threat is thus mathematically clearly different from frustration, even if the two concepts are in practice closely related, as we will discuss later on multiple occasions. But first, let us consider the connection between threat and fear.

\section{Fear, threat, and predictions}

A threat typically leads to fear, which is central to understanding human suffering. Fear has an obvious connection to self-needs, in particular survival. In fact, it may seem a bit too abstract to talk about suffering as coming from a survival instinct, as I did in Chapter~\ref{self.ch}: such suffering is usually mediated by a feeling of fear. Fear is actually a multifaceted phenomenon, and we will consider various aspects of fear in later chapters (especially Chapter~\ref{emotions.ch}).

Suppose you suddenly find yourself in the presence of a tiger in a jungle.
You are likely to suffer at this very moment, but why exactly? It is not that you missed something you wanted to have or some reward you anticipated, so this is not a case of typical frustration. (Nor is it obviously a case of aversion-based frustration, where you didn't expect something unpleasant to happen but it did, because the tiger hasn't yet attacked you.)
What happens is rather that you are, right now, \textit{predicting} something terrible to happen in the future, and with a non-negligible probability.
Aristotle proposed that ``Fear may be defined as a pain or disturbance due to a mental picture of some destructive or painful evil \textit{in the future}''.\footnote{\textit{Rhetoric}, II.5, translated by W.\  Rhys Roberts, with my italics. This is often abbreviated as ``Fear is pain arising from the anticipation of evil.''
  \add{We might also consider Cicero's  ``fear is an uneasy (anxious) apprehension of future grief'' (\textit{Tusculan Disputations}, 5. XVII, translated by C.~D.~Yonge, with alternative wording in parenthesis by A.~P.~Peabody)}.\index{Cicero} This can be compared with the discussion on various definitions of threat in footnote~\ref{threatfn} in this chapter. In some parts of the literature, a distinction is made between fear and anxiety\index{anxiety}, where fear refers to a ``immediate'' and ``imminent'' threat, while anxiety is more ``future-oriented'' and about ``uncertain'' threats \citep{chandanxiety,ledoux2016using}; however, I see no need to make such a distinction since it seems to be simply a question of different time scales. } \index{Aristotle} 
Here, the ``mental picture'', or prediction, of something bad happening is what I consider a threat, which thus causes fear.

I would further argue that a meaningful definition of threat requires uncertainty: It must be possible to avoid the bad thing that is included in the threat. If the bad thing in the future is completely certain to happen, it is something different from a threat, and the ensuing feeling is something different, often described as resignation. Cassell said that  ``to suffer, there must be a source of thoughts about possible futures'', where I would emphasize the fact that "futures" must be in plural: the future is not certain and fixed, but different outcomes are possible, and the agent can exercise at least some amount of control on the outcomes.\footnote{Selected dictionary definitions of threat include ``the \textit{possibility} that something unwanted will happen'' (Cambridge Dictionary), and ``an indication or warning of \textit{probable} trouble'' (Dictionary.com), both of which indicate uncertainty (with my italics). In psychological literature, a similar definition as ``anticipation of potential harm''  was proposed by \citet{palmwood2019challenge} based on Folkman and Lazarus. While the concept of threat is widely used in the biological and psychological literature, explicit definitions are actually not easy to find. Biologically oriented literature often considers it very specifically in the context of biological survival when attacked by a predator \citep{mobbs2020space}. A general psychological framework postulating that ``threat is the experience of discrepancy between the situation, a personal current cognitive focus, or current personal motives'' is reviewed by \citet{reiss2021exploring}. 
  \index{threat!definition}\label{threatfn}
  }

\subsection{Threat based on prediction of rewards}

Combining the mathematical theory of risk aversion with Aristotle's and Cassell's philosophy, we can now approach the modelling of threat. We might initially think about threat as a prediction that there is a sufficient probability of a very small future reward---here, ``very small'' would typically mean a negative reward of large absolute value. In this way, the concept of threat can be directly linked to the pursuit of any kind of rewards, not only internal ones such as physical safety considered earlier. You might be threatened by a large monetary loss, for example.

Consider again the gamble seen above, where there is a 50\% probability of the agent getting 10 pieces of chocolate and 50\% probability of not getting any. %
Now, let us create another gamble to illustrate a probability distribution that is relevant to threat in the particular sense we are interested in. In this new gamble, the agent has 50\% change of getting the 11 pieces of chocolate, 49\% change of getting nothing, and 1\% chance of being charged a penalty of 50 chocolate pieces (in this world, chocolate seems to act as a common currency). Here, we see that there is a great threat to the agent of losing chocolate in the form of the penalty. On the other hand, I changed the main reward from 10 to 11 pieces so that the expected reward is exactly the same as in the earlier 50-50 gamble (the expectation can be calculated as $0.50\times 11 + 0.49\times 0 +  0.01 \times(-50) = 5$). So, the two gambles are only distinguished by the general distribution of reward, while the expected reward is the same.\index{risk}

Now, it is intuitively compelling that in the gamble with penalty, the agent should behave in a slightly different way since there is the threat, or the risk, of the penalty being charged. It should be ``afraid'' of the penalty of 50 pieces happening, and try do find a course of action that avoids the penalty, presumably by trying to avoid this gamble in the first place. %
While this may be intuitively clear, I emphasize that it is only the case %
if the agent has been programmed to be risk-averse in this particular way, i.e., ``threat-averse''. A very simple agent would behave in the same way in these two chocolate scenarios (as well as the sure-thing scenario considered earlier), since it would not understand anything about risks or threats. Even a more sophisticated agent that  understands something about uncertainty might not make any difference between the two gambles since both have a lot of uncertainty. But a human-like agent that has been programmed to avoid large losses, that is, large negative rewards, would avoid the latter gamble that includes such a strong threat.\footnote{It could be argued that even some general uncertainty (risk) is higher in the latter gamble, but this depends on the specific uncertainty/risk measure used. Gambles where both expectation and variance are made equal, while some asymmetric threat-like difference exists, are discussed by \citet{ebert2021skewness} and \citet{trautmann2018higher}, and they could be used to create and example that rigorously makes the difference between variance-based uncertainty and threat; see also footnote~\ref{riskmeasurefn} below. }

Such threats are widely discussed in the economic literature.\index{risk} Consider investing in a company. One company is quite stable: you can be sure that the return on investment is 5\%. Another promises 10\%, but you know that it also has a 5\% probability of going bankrupt so that you lose all your money. Again, the \textit{expected} return on your investment is the same (up to rounding errors), but there is a much larger risk of loss in the latter case. Most humans prefer the first, stable company since they want to avoid the ``threat'' of bankruptcy.\footnote{ This footnote discusses the difference between threat and risk as well as the exact measures of used in more detail. While the economic literature considers many different kinds of risks,  what we call threat is a special case,  more specifically related to the \textit{downside risk}, i.e.\ the risk of outcomes which are particularly bad. Therefore, I make a clear distinction between threat and risk, and use ``uncertainty'' synonymously with risk. I do not restrict myself to any specific definition of risk here but rather consider it as a general concept with many instantiations, of which threat is one.   In conventional economic theory, especially finance, risk is modelled by the variance of the quantity to be maximized (here, the total future reward).   Alternatively, economic theory  uses concave utility functions to induce risk-averse behavior, which has been applied in reinforcement learning by  \citet{wu2021uncertainty,zhang2020variational}, but the basic effect seems quite similar to using variance.
  However, the crucial point here is that using variance seems rather inadequate to measure threat since it does not focus on downside risk. Thus, I prefer to equate large variance with general uncertainty and one kind of risk, but not threat. Threat, as defined in this book, is all about the probability of bad outcomes,  while variance is measuring uncertainty in both positive and negative directions; if a very good outcome is possible, that also increases variance.
  How the downside risk of a distribution should exactly be defined and measured to measure threat is a complex question to which I'm not going to give a single answer; I discuss some options in what follows.
  One well-known economic theory which is relevant here considers the probability of ``ruin''  (i.e.\  bankruptcy), typically used in insurance theory. Such ruin could be equated to the destruction (death) of an agent, and is not completely different from our concept of threat, especially in the context of evolutionary modelling.   \citet{lipton2016combating} propose a framework related to ruin probabilities in reinforcement learning, measuring the distance to what they call \textit{catastrophic events}, which could be death and serious injury in the case of a biological agent, or, from the viewpoint of making robots safe to humans, it could be defined as the robot injuring a human being; see also \citep{martin2016death}. Further possibilities for modelling threat can be found in financial theory. One option is skewness  \citep{ebert2021skewness,trautmann2018higher}, which is a measure of the asymmetry of a probability distribution;
  however, it is not clear if it is enough in itself as a measure of threat: it may need to be combined with variance. Fortunately, financial theory has also developed measures such as conditional value-at-risk, also called expected shortfall, which measures the negative tails and could in fact be quite suitable as a measure of downside risk of reward loss, and thus threat,  in our framework. \citet{bellemare2023distributional} discusses them from the viewpoint of reinforcement learning.  \label{riskmeasurefn}}

\section{Threat as prediction of possible large frustration}

To arrive at the final definition of threat, we still need to define what level of possible reward is so small (or so negative) that it actually can be called a threat. In other words, what is a suitable baseline? We can actually borrow the baseline from the definition of the reward prediction error and reward loss, thus comparing the different possible rewards with their expectation. In that case, threat would be the same as a very large reward loss happening with sufficient probability. The crucial difference is that a reward loss (or RPE) is typically computed only after the action, or after the fact, so to say. However, as a very intelligent agent will try to predict any relevant quantities, it would also try to predict the reward loss before it actually acts or the reward loss happens.\footnote{It may seem logically contradictory to predict an RPE, or to predict a prediction error. If the agent were able to predict a prediction error, wouldn't it mean that the agent understands what kind of error is about to occur, and then it should be able to cancel it by improving the prediction accordingly? This would indeed be true in the basic case where the prediction is only about a single quantity such as the expectation. %
If the agent somehow understands that it is predicting the expected reward as too high, it can simply make its prediction a bit lower, and thus the error in the prediction of the expected reward can be removed.
However, this is no longer meaningful with more sophisticated predictions that predict the whole probability distribution. If the agent predicts a large risk, it predicts a major possibility of prediction error, but there is nothing wrong with that prediction; there is nothing to correct. Therefore, there is no contradiction in predicting that there is going to be a prediction error. 
 \citet{dabney2020distributional} claim that the brain is coding the whole distribution of RPE, based on a populations of neurons with different thresholds. 
}

Furthermore, as always in reinforcement learning, the agent should take into account all the future rewards, and look at the distribution of total future reward, not just the reward in the next time step. In earlier chapters, we considered the expectation of total future reward, which is given by the state-value function, but now, we thus need to model  the whole probability distribution of total future reward. That is, the single number given by the state-value is replaced by the probabilities of all possible future outcomes of future reward, starting from the current state. 
In the rest of this chapter, we thus assume the agent is sophisticated enough to actually compute the whole probability distribution of total future reward, or at least something more than just its expectation. In the simple chocolate gambles, the agent should understand that getting a certain amount of chocolate has a certain probability, and not getting any has another probability. In a more realistic scenario where the agent chooses actions at many time points (think about navigation by a robot), it will consider the long-term consequences of its actions by trying to learn the distribution of future rewards for each state, thus going beyond simple state-values.
Modelling the whole distribution of total future reward in addition to its expectation is, in fact, a rather recent development in reinforcement learning theory.\citenew{morimura2010nonparametric,lowet2020distributional,prashanth2022risk,bellemare2023distributional}  
Obviously, this is computationally very challenging and needs a lot of data where all those different outcomes are realized.

Putting this all together, we arrive at a definition of threat as a \textit{prediction of sufficiently probable and large reward loss,} where the reward loss is computed over the total future reward.\footnote{To keep the exposition simple, I'm taking some shortcuts here. It must be emphasized that 
the reward loss is here computed for the total (discounted) future reward, instead of any particular future time point. Thus, more precisely, threat is a prediction that the total future reward has a sufficiently large probability of being much less than the expected total future reward, with discounting applied if necessary; see footnote~\ref{threatmathsfn} below for a mathematical definition. 
 Obviously, it is necessary to define hyperparameters that say what is ``sufficient'' and ``much less'' (or ``large'' in the definition of the main text). 
Alternatively, it is also possible to define threat  simply on the distribution of reward loss at a single time point, which would lead to simpler computation at the risk of suboptimality.
\label{threatshortcutfn} } 
This definition is very general: it means that threats can come from many different sources. 
In the preceding chapter, we already briefly mentioned the concept of threat in terms of death and tissue damage, but those are now seen as simply special cases of this general concept of threat,  seamlessly integrated to the general reinforcement learning framework.
Still, it is true that the biggest threats may be related to survival and self-image, as will be discussed below.\footnote{\label{threatmathsfn}
  In this footnote, I propose a formal definition of threat. Denote by $X_\pi(s)$ the random variable giving the total discounted future reward starting from initial state $s$, following policy $\pi$, and using discount factor $\lambda$. That is, $X_\pi(s)= \sum_{t=0}^\infty \lambda^t r_t \ |\ \pi, s(0)=s$. The expectation of this quantity is nothing else than the state-value function, and that will be used as the baseline. Thus, subtracting the baseline we obtain the random variable  $\tilde{X}_\pi(s)=X_\pi(s)-V_\pi(s)= \sum_{t=0}^\infty \lambda^t (r_t - E\{r_t\})\ |\ \pi, s(0)=s$. Some measure of the downside risk (negative tail) of $\tilde{X}_\pi(s)$ is now defined a measure of threat for state $s$. We could use, for example, the conditional value-at-risk (expected shortfall), see footnote~\ref{riskmeasurefn} in this chapter for discussion of various downside risk measures. 
}

\section{Interplay of threat and frustration}

Threat as defined above is in many ways different from frustration.\index{threat!and frustration} To summarize, threat is about a prediction of something bad that \textit{might} happen, while frustration is about realizing that something \textit{did} go wrong;  a threat is mainly used for \textit{choosing immediate actions}, as will be considered in more detail in Chapter~\ref{emotions.ch}, while frustration is a signal for \textit{learning}. One might further say that a threat is about the \textit{future}, while frustration is about the immediate \textit{past}, but this might be oversimplifying since frustration can sometimes refer to mere changes in expectations of  future rewards.\footnote{Footnote~\ref{predictedfrustrationfn} in Chapter~\ref{rpe.ch} pointed out  that frustration in the sense of RPE might happen purely based on a change in prediction.}

Threat produces a subjective feeling, typically in terms of fear, which is also very different from frustration. This is logical since the computations underlying threat are different from those underlying frustration, and especially the way threat influences behavior and learning must be very different. Thus, fear has to produce a different kind of signal, even if both frustration and threat signals lead to suffering.

Still, frustration and threat often come together. Let's go back to the case where a tiger appears in front of you. It might eat you and produce a great loss of future rewards, but this is not certain since you might still be able to escape; in this sense, there is a threat but no frustration yet. But there \textit{is} frustration in the sense that you certainly would have preferred that the tiger does not appear, that is, you wanted to live a peaceful life where tigers are remote, and that desire is now frustrated. In this example, the planning system can  amplify the frustration, because planning may be launched with the goal state being any state where the threat is not present: you are frantically thinking about what to do to be safe. Planning is attempted, but it fails: no plan is found that would get rid of the threat, or if such a plan is found, its execution fails. Thus, arguably there is frustration even in the sense of plans failing.\footnote{Furthermore, the more sophisticated theory of reinforcement learning provides another explanation of frustration and suffering in this case. As explained in Chapter~\ref{rpe.ch}, in particular footnote~\ref{predictedfrustrationfn}, the RPE theory says there is frustration solely created by predictions in case you move to a state of lower value, and there is no reward. Now, when the tiger appeared, you suddenly moved to a state where the value (expected future rewards) went down considerably, since if it eats you, there will be no more reward for you. In other words, your chances of getting any positive reward in the future just got smaller (because you won't get any after being eaten), and thus the expected total reward during the rest of your life (which is the definition of state-value)  decreased: This produces a reward loss and thus suffering.
In fact, such frantic planning also consumes the agent's resources, draining batteries either literally or in some figurative sense; this makes the situation even worse by reducing expected future reward due to limited energy, and thus leading to reward loss similarly to what was just explained---and perhaps increases the threat by making the agent weaker and future frustration more probable.
}\label{fearrpefn}

Another interesting interplay of fear and frustration can seen in the fear of frustration that arises at the moment of making decisions. A person can be afraid of choosing the wrong flavor for his ice cream and spend an embarrasingly long time in the decision-making process. His brain may correctly predict that a frustration will happen in the future if it turns out that he does not like the flavor that much after all. Such a fear might be present surprisingly often when humans make decisions.\citenew{schwartz2004paradox}\index{fear!and prediction}

\subsection{Risk aversion and internal rewards}

Another intriguing connection is that the very reason why humans are risk-averse can be understood based on frustration of internal rewards, as introduced in Chapter~\ref{self.ch}.
If the agent has a lot of uncertainty about the state of the world, it will find it more difficult to reach its goals or obtain rewards. Thus, uncertainty in itself is something that should be avoided. We saw above that this is exactly what humans do; it is the very essence of risk aversion.
We can interpret this phenomenon from the viewpoint of internal rewards.\index{uncertainty}\index{unpredictability}\index{uncontrollability}
Since uncertainty is bad for future reward, it would clearly make a lot of sense to program an internal reward system that gives a negative reward when the agent is in a state of a lot of uncertainty.
Therefore, it may not be surprising that uncertainty creates suffering in itself as well, which is the basis of risk aversion.\index{frustration!of internal needs}

In fact, such a logic of internal rewards goes much beyond risk aversion.
We can consider unpredictability and uncontrollability in the same framework as uncertainty. All these properties are bad for future rewards, and they increase frustration. This fundamental idea will be considered in detail in later chapters: if the world is, say, uncontrollable, frustration is difficult to avoid.
Thus, it could very well be that uncertainty, unpredictability, or uncontrollability are suffering in themselves because they lead to frustration of specific internal rewards. %
If, say, controllability is lower than some expected standard, a frustration signal could be launched. That would be useful for learning because it signals that the agent has failed in learning about the environment; it should not have gotten itself into a situation where controllability is that low. This is equivalent to a self-evaluation system which considers that the agent should not be in situations that are uncertain, difficult to predict or difficult to control.  This is how uncontrollability, as well as uncertainty and unpredictability, can directly lead to suffering. Nevertheless, this tends to happen in states where a threat is observed, according to our definition, since a threat is nothing else than a form of uncertainty.\footnote{Reducing uncertainty as measured by entropy can even be seen as a general learning principle for the brain \citep{friston2010free}, and thus failure to reduce uncertainty should generate an error signal. At the same time, reducing uncertainty, unpredictability, and uncontrollability is very closely related to  the goal of curiosity discussed in Chapter~\ref{self.ch}: Uncertainty can be reduced by a curious investigation of new aspects of the environment, and uncontrollability can be reduced by trying out the effects of actions in new circumstances. Another interesting point is that the estimates of uncertainty etc.\ are not exact either; the estimate of, say, entropy has some uncertainty (estimation error) as well. However, considering uncertainty of uncertainty would lead to a potentially endless recursion, and may not be very useful or feasible. }

This gives an alternative viewpoint of threats, completely reducing them to frustration of internal reward systems. When uncertainty and uncontrollability reach high levels, such an internal reward system gives negative rewards, which produces frustration. However, this account clearly explains only part of what a threat is about. While it cannot be denied that uncertainty and uncontrollability do lead to frustration, the suffering due to a threat simply does not feel the same as frustration: it is more like fear, anxiety, or stress. Thus, such a reduction of threat to frustration is not quite satisfactory, and justifies the separate definition given earlier in this chapter.\footnote{In the first version (2022) of this book, I actually attempted such a reduction, but in this second version (2024), I'm able  to provide a separate model of a threat based on the prediction of whole distributions.}

\section{Threats and the level of intelligence}

A simple AI agent might only generate the suffering signal when something bad happens, such as when it fails in its tasks---this is the basic case of frustration. Suppose a thermostat connected to a heating system tries to keep the room at a constant temperature. (This is actually a task that the nervous systems of many animals face as well.) It continually monitors the room temperature and adjusts its actions accordingly. Its function is based on a simple error signal created when the room gets too hot or too cold. When the temperature is suitable, there would be no error signals whatsoever, and certainly no suffering.\index{thermostat} 

Now, suppose you make the thermostat very intelligent, so that it is able to predict the future, compute threats, evaluate itself, perhaps even think about its own survival. Then, it might not only suffer when the room temperature is wrong but also when it anticipates that that might happen. Your hyperintelligent thermostat might be reading the weather forecast on the internet. Suppose the forecast says that tomorrow night will be exceptionally cold, beyond the capacities of the heating system. Then, the thermostat anticipates that tomorrow night it will not be able to keep the temperature high enough.
Thus, the thermostat suffers due to such a threat---at least in the computational sense.%

The extraordinary thing here is that the hyperintelligent thermostat suffers even long before anything bad happens, before, say, actual frustration is produced, merely by virtue of the newly appeared anticipation of possible negative reward. This is perceived as a threat, and produces fear. Becoming more intelligent means the agent can perform computations related to threat, suffer based on those computations, and thus suffer much more than it did earlier.
Furthermore, if the thermostat realizes it is unable to properly control the temperature in the future, the uncontrollability may trigger a negative internal reward, and a reward loss. If this happens often, the self-evaluation system might conclude that it is not performing its central task well enough, thus leading to frustration due to  the self-evaluation.  It is possible that if the thermostat fails to keep the temperature constant, it will be thrown into the garbage bin, and a hyperintelligent thermostat might even worry about its own survival.

``One who fears suffering is already suffering from what he fears'' according to Michel de Montaigne.\footnote{\add{``Qui craint de souffrir, il souffre desj\`a de ce qu'il craint'',  \textit{Essais}, III, 13; using the old orthography on Wikisource (Bordeaux exemplaire, 1588). Seneca said almost the same in \Lucilius, LXXIV.32.}}\index{Montaigne} Humans suffer enormously because they are too intelligent in this sense, and prone to thinking too much about the future---a theme I will return to in Chapter~\ref{replay.ch} where I talk about simulation of the future.
Yet, if we humans are so incredibly intelligent, why cannot we just decide not to fear anything? Why cannot we take Montaigne's point seriously: He suggested---actually talking about his chronic pain\index{pain!chronic} due to kidney stones---that there is no point in imagining or anticipating future pain since that simply induces more suffering.
This is a complex question where part of the answer is the dual-process nature of human cognition, which will be treated in the following chapter.

}

\chapter{Fast and slow intelligence and their problems} \label{dual.ch}

\add{
In this chapter, we delve deeper into the distinction of two different modes of information processing in the brain, which coincide with those in modern AI. They were already discussed in Chapter~\ref{ml.ch}: neural networks and Good Old-Fashioned AI.
The idea of two complementary systems or processes is, in fact, ubiquitous in modern neuroscience and psychology, where it is called the ``dual-process'' or ``dual-systems'' theory.\index{dual system|see{dual process}}\index{dual process} It is assumed that the two systems in the brain work relatively independently of each other while complementing each other's computations. 
The two systems, or modes of operation, roughly correspond to unconscious processing in the brain's neural networks, and conscious language-based thinking.
Each of the two systems has its own advantages and disadvantages, which is the main theme of this chapter and, in fact, a theme to which we will return many times in this book. 
Neural networks are based on learning, which means they need a lot of data and often result in inflexible functioning. On the other hand, the computations needed in GOFAI may be overwhelming, as in planning.
On the positive side, we will see how the advantages of the two systems can be combined in the action selection of a real AI system. Using categories is crucial for GOFAI, and we conclude by discussing the deep question of the advantages and disadvantages of such categorical processing and thinking.

}

\section{Fast and automated vs.\  slow and deliberative}

Let us start with the viewpoint on the two systems given by cognitive psychology and neuroscience.\citenew{evans2008dual,kahneman2011thinking,sloman1996empirical}
According to such ``dual-process'' (or ``dual-systems'') theories, one of the two systems 
in the brain is similar to the neural networks in AI: It performs its computation very fast, and in an automated manner. It is fast thanks to its computation being massively parallel, i.e., happening in many tiny ``processors'' at the same time. It is automated in the sense that the computations are performed without any conscious decision to do so, and without any feeling of effort.\index{effort!and dual-process} If visual input comes to your eyes, it will be processed without your deciding to do so, and usually you recognize a cat or a dog in your visual field right away, that is, in something like one-tenth of a second.\citenew{kirchner2006ultra} 
Most of the processing in this system is also unconscious. You don't even understand how the computations are made; the result of, say, visual recognition just somehow appears in your mind, which is why this system is also called ``implicit''. 

The  processing in the conscious, GOFAI-like system is very different. To begin with, it is much slower. Consider planning how to get home from a restaurant where you are the first time: you can easily spent several seconds, even minutes, solving this planning task. 
The main reason is that the computations are not parallelized: They work in a serial way, one command by another, so the speed is limited by the speed of a single processing unit. 
In humans, another reason why symbolic processing is slow is, presumably, that it is evolutionarily a very new system, and thus not very well optimized. Other typical features of such processing are that you need to concentrate on solving the problem, the processing takes some mental effort, and it can make you tired. 
Such processing is also usually conscious, which means that you can explain how you arrived at your conclusion; hence the system is also called ``explicit''.\footnote{My exposition is a kind of synthesis of different theories, and not all the mentioned properties are always associated with the two systems. Further, I should mention the proposals that the second, explicit system may be specialized in simulating hypothetical events that have not happened \citep{stanovich2004robot} for example for the purposes of planning, which will be considered in Chapter~\ref{replay.ch}; or it could be mainly about working memory \citep{evans2008dual}. An interesting related division between feedforward and feedback processing in neural networks is discussed by  \citet{lamme2000distinct}.} 

Note that in an ordinary computer, the situation above is in some ways reversed, as already explained in Chapter~\ref{ml.ch} (page~\pageref{digitalvsanalog}). A computer can do logical operations much faster than neural network computations, since logical operations are in line with its internal architecture. In fact, a computer can only do neural network computations based on a rather cumbersome conversion of such analog operations into logical ones. Analogously, the brain can only perform logical operations after converting them into neural network computations, which is equally cumbersome.

To see the division into two systems particularly clearly, we can consider situations where the two systems try to accomplish the \textit{same} task, 
say, classification of visual input. We can have a neural network that proposes a solution, as well as a logic-based system that proposes its own. Sometimes, the systems may agree; at other times, they disagree.

Suppose a cat enters your visual field. When the conditions for object recognition are good, your visual neural network would recognize it as a cat. In other words, the network would output the classification ``cat'' with high certainty. However, when it is dark, and you only get a faint glimpse of the cat that runs behind some bushes, your neural network might not be able to resolve the categorization. It might say it is probably either a cat or a dog, but it cannot say which. At this point, the more conscious, logic-based system might take over. You recall that your neighbour has a cat; you don't know anybody who owns a dog near-by; you think this is just the right moment in the evening for a cat to hunt for mice. Thus, you logically conclude it was probably a cat. In this case, the task of recognizing an object used the two different systems, working together. The logic-based one took quite some time and effort to use, while the neural network gave its output immediately and without any effort. Here, the systems were not completely independent, since the logic-based system did need input from the neural network to have some options to work on.

The two systems can also disagree, as often happens in the case of fear.\index{fear!dual process view}\label{fearconflict}
Talking about fear and related emotional reactions, people often call them ``hard-wired''. This expression is not too far from reality. What happens is that the brain uses special shortcut connections to relay information from the eye to a region called the amygdala, an emotional center in the brain.\index{amygdala} This shortcut by-passes those areas where visual information is usually processed.\footnote{\citep{ledoux2016using}. More precisely, a pathway goes directly from the thalamus to the amygdala, without reaching the visual cortex. Another, slower pathway does go back from the visual cortex to the amygdala. See also \cite{hofmann2009impulse} on conflicts between the two systems from the viewpoint of self-control.} If such a connection learns to elicit fear (due to a previous unpleasant encounter with some animals, for example), it will be very difficult to get rid of it. Any amount of reasoning is futile, presumably since the visual signal triggering fear is processed by completely different brain areas than logical, conceptual reasoning. Often, the logic-based system loses here, and the neural-network-based fear prevails.
This division into two processes also explains why it is difficult for us to change unconscious associations, such as fear: the conscious, symbolic processing has limited power over the neural networks.

Interestingly,  people tend to think that the main information processing in our brain happens by the conscious, symbolic system, including our internal speech and conceptual thinking. But what if that is simply the tip of the iceberg, as early psychoanalysts\footnote{I'm here obviously referring to Freud and his followers, but the importance of unconscious processing was emphasized around the same time frame by \citet{Janetbook}, and even earlier by philosophers such as Arthur Schopenhauer and Eduard von Hartmann.\label{unconsciousimportantfn}}\index{Freud} claimed more than a hundred years ago?
The idea that most information processing is conscious and conceptual may very well be an illusion. We may have such an impression because conceptual processing requires more effort, or because it is more accessible to us by virtue of being conscious.
However, if you quantify the amount of computational resources which are used for conceptual, logical thinking, and compare them with those used for, say, vision, it is surely vision that will be the winner.\footnote{\citep{nakayama1999mid}. While this comparison in terms of brain resources seems compelling in terms of comparing vision vs.\ conceptual thinking, it is more difficult to compare the conscious and unconscious aspects since we don't really know how consciousness is related to the brain; see Chapter~\ref{consciousness.ch}.}

Similar to the dual-process theories in cognitive psychology and neuroscience just described, the division between GOFAI and neural networks has been prominent in the history of AI research, which has largely oscillated between the two paradigms. Currently, neural networks are very popular, while GOFAI is not used very widely. However, this may very well change, and 
perhaps in the future, AI will combine logic-based and neural models in a balanced way.\index{GOFAI}
Since GOFAI is used by humans, it is very likely to have some distinct advantage over neural networks, at least for some tasks.\footnote{While the main text discusses later some such combinations of the two systems, I should also mention attempts made under the titles of ``hybrid AI'' or ``neural-symbolic processing''  \citep{garcez2012neural,goertzel2012perception,graves2016hybrid,yi2018neural,tresp2023tensor}.} 

Note that in AI we find another important distinction that is not very prominent in the neuroscientific literature: learning vs.\ no learning. Neural networks in AI are fundamentally based on learning, and using them without learning is not feasible. In contrast, in its original form, Good Old-Fashioned AI promises to deliver intelligence without any learning, at the cost of much more computation and more effort spent on programming. This distinction is also relevant to the brain, as we will see next.\footnote{When we talk about learning in the brain in the context of neural network models, that is to be understood on an abstract level, where learning includes both evolutionary and developmental processes; this will be discussed in more detail in Chapter~\ref{perception.ch} (page~\pageref{learninginbiology}).}

\section{Neural network learning is slow, data-hungry, and inflexible}

To understand the relative advantages of the two systems, let us first consider the limitations in neural networks, and especially the learning that they depend on. 
First of all, neural network learning is data-hungry: it needs large amounts of data. This is because the learning is by its very nature \textit{statistical}; that is, it learns based on statistical regularities, such as correlations. Computing any statistical regularities necessarily needs a lot of data; you cannot compute statistics by just observing, say, two or three numbers. 

Second, neural network learning is slow. Often, it is based on gradient optimization, which is \textit{iterative}, and needs a lot of such iterations. The same applies to Hebbian learning, where changing neural connections takes many repetitions of the input-output pairs---this is natural since Hebbian learning can be seen as a special case of stochastic gradient descent. In fact, to input a really large number of data points into a learning system almost necessarily requires a lot of computation, since each data point takes some small amount of time to process.\index{learning!iterative}\index{learning!slow in neural networks}

This statistical and iterative nature of neural network learning has wide-ranging implications for AI. 
To begin with, these properties help us to further understand why it is so difficult, in us humans, to change any kind of deeply ingrained associations. Mental associations are presumably in a rather tight correspondence with neural connections: If you  associate X with Y, it is because there are physical neural connections between the neurons representing X and Y. Now, even if any statistical connection ceases to exist in the real world, perhaps because you move to live in a new environment, it will take a long time before the Hebbian mechanisms learn to remove the association between X and Y, or to associate X with something else.\footnote{In some cases, an association may not actually be removed but overridden by an inhibitory connection, a bit like creating a new ``negative'' connection to cancel the functioning of a positive connection \citep{westbrook2002reinstatement}. This also means the old association can be reactivated quite easily.}\index{learning!Hebbian}

In fact, these learning rules, whether basic Hebbian learning or some other stochastic gradient methods,\index{gradient descent!stochastic} may seem rather inadequate as an explanation for human learning: We humans can learn from single examples and do not always need a lot of data. You only need to hear somebody say once ``Helsinki is the capital of Finland'', and you have learned it, at least for a while. Surely, you don't need to hear it one thousand times, although that may help. This does not invalidate the neural network models, however, since the brain  has multiple memory systems, and Hebbian learning is only one way we learn things and remember them---we will get back to this point in Chapter~\ref{replay.ch}.\footnote{Chapter~\ref{replay.ch} will explain the idea of \textit{replay} whose application to this case would be as follows. Maybe your brain does actually hear the sentence ``Helsinki is the capital of Finland'' many times. One of the learning systems in the brain is based on storing events, or short episodes, in an area called the hippocampus.\index{hippocampus} It uses special mechanisms, presumably quite different from stochastic gradient methods, to store the sentence after hearing it just once. Then, the hippocampus feeds the sentence to the other parts of the brain many times, and that allows Hebbian learning and something similar to stochastic gradient learning to happen.  \label{replaymention}\index{replay!semantic information}}

The iterative nature of neural learning, together with the two-process theory, also helps to explain in more detail why it is so difficult to \textit{deliberately} change unconscious associations. Suppose you consciously decide to learn an unconscious association between X and Y (where X might be ``exercise'' and Y might be ``good''). How can you transfer such information from the conscious, explicit system to the neural networks? Perhaps the best you can do is to recall X and Y simultaneously to your mind---but that has to be done many times! In fact, you are kind of creating a kind of new data and feeding it into the unconscious association learning in you brain. You are almost cheating your brain by pretending that you perceive the association ``X and Y'' many times. We will see many variations on this technique when we consider methods for reducing suffering in Chapter~\ref{training.ch}.\index{associations!unconscious}

Another limitation is that when a neural network learns something, it is strictly based on the specific input and output it has been trained on. While this may seem like an obvious and innocuous property, it is actually another major limitation of modern AI.\index{vision}
Suppose that a neural network in a robot is trained to recognize animals of different species: It can tell if a picture depicts a cat or a dog, or any other species in the training set. Next, suppose somebody just replaces the camera in the robot with a new one, with higher resolution. What happens is that the neural network the robot previously trained does not work anymore. It will have no idea how to interpret the high-resolution images since they do not match the templates it learned for the original data. A similar problem is that the learning is dependent on the context: An AI might be trained by images where cats tend to be indoors and dogs outdoors, and it will then erroneously classify any animal pictured indoors as a cat. The AI sees a strong correlation between the surroundings and the animal species, and it will not understand that the actual task is about recognizing the animals and not recognizing the surroundings. That is why a neural network will typically only work in the environment or context it is trained in.\citenew{arjovsky2019invariant}\index{learning!context-dependent}\index{context-dependence!of learning in neural networks}

In light of these limitations, AI based on neural networks is thus rather different from what intelligence usually is supposed to be like in humans.\footnote{While the inflexibility of neural networks seems to hold for the networks in the human brain as well, there is the celebrated experiment of ``upside down goggles'', which shows an interesting adaptive ability of the human neural networks.  In this experiment, human participants started wearing goggles containing a prism which made their world look upside down. Surprisingly soon, the participants were able to function normally; somehow, their visual systems were able to process the input correctly in spite of the inverted visual input \citep{pisella2006prism}.}
In general, when humans learn to perform a task, they are often somehow able to abstract general knowledge out of the learning material, and they are able to transfer such knowledge from one task to another.
It has even been argued that the hallmark of real intelligence is that it is able to function in many different kinds of environments and accomplish a variety of tasks without having to learn everything from scratch. If all a robot can do is to mow the lawn, we would think it is just accomplishing a mechanical task and is not ``really'' intelligent.\footnote{\citep{legg2007universal}. Functioning in many environments thus requires an advanced capacity to what is called \textit{transfer learning}, which is currently a focus of very active research in AI \citep{pan2009survey,weiss2016survey}.\index{learning!transfer}\index{transfer learning} 
}\index{intelligence!definition}

\section{Using planning and habits together} \label{dualplanning}

Combining the two systems, neural networks and GOFAI, should take as closer to human-like intelligence. Let us next look at how the two systems might interact in AI.
Regarding action selection, we have actually seen how two different approaches can solve the same problem in AI: reinforcement learning and planning.  Planning is in fact one of the core ideas of the GOFAI theory. 
Planning is undeniably a highly sophisticated and demanding computational activity, and probably impossible for simple animals---some would even claim it is only present in humans, although that is a hotly debated question.\citenew{redshaw2018future,corballis2019language}
In any case, it seems to correspond closely to the view humans have about their own intelligence, and therefore was the target of early AI research. 
However, in the 1980s, there was growing recognition that building agents, perhaps robots, whose actions show human-level intelligence is extremely difficult, and it may be better to set the ambitions lower. Perhaps building a robot which has the level of intelligence of some simple animal would be a more realistic goal. Moreover, like in other fields of AI, learning gained prominence.
That is why habit-like reinforcement learning started to be seen as an interesting alternative to planning.\footnote{A related school of research emphasized how intelligence might emerge from simple reactive behaviors, even without any learning \citep{brooks1991intelligence,brooks1999cambrian}.}

\subsection{Habits die hard---and are hard to learn}

However, habit-based behavior has its problems, partly similar to those considered above for neural network learning.
Learning the value function, that is, learning habits, obeys the same laws as other kinds of machine learning. It needs a lot of data: the agent needs to go and act in the world many, many times. This is a major bottleneck in teaching AI and robots to behave intelligently, since it may take a lot of time and energy to make, say, a cleaning robot try to clean the room thousands of times. Basic reinforcement algorithms are also similar to neural network algorithms in that they work by adjusting parameters in the system little by little, based on something like the stochastic gradient methods.\index{habits!slow to learn}\index{gradient descent!stochastic}

Another limitation which is crucial here is that the result of the learning, the state- or action-value function, is very context-specific---that is one form of inflexibility discussed above. If the robot has learned the value function for cleaning a room, it may not work when it has to clean a garden. Even different rooms to clean may require slightly different value functions! The world could also change. Suppose the fridge from which the robot fetches the orange juice for its master is next to a red table. Then, the robot will associate the red table with high value since seeing it, the robot knows it is close to being able to get the juice.\index{robot!fetching orange juice}
However, if somebody moves the table to a different room, the robot will start acting in a seemingly very stupid way: It will go to the room which now has the red table when it is supposed to get the orange juice---in fact, it might simply approach any new red object introduced to its environment in the hope that this is how it finds the fridge. It will need to re-learn its action-values all over again.\index{habits!slow to learn}

Here, we see another aspect of the slowness of learning habits: Once a habit is learned, it is difficult to get rid of it. In humans, the system learning and computing the reinforcement value function is outside of any conscious control: We cannot tell it to associate a smaller or larger value to some event. This is why we often do things we would prefer not to do, out of habit. In order to learn that a habit is pointless in the sense that it does not give any reward anymore (as happened with the robot above), a new learning process has to happen, and this is just as slow as the initial learning of the habit. That is why habits die hard.\footnote{However, some hope will be offered in Chapter~\ref{replay.ch} where we consider ways of speeding up learning by replaying existing data, and that theme is continued in Chapter~\ref{training.ch}.}

\subsection{Combining habits and planning}

These problems motivate a recent trend in AI: combining planning and habit-like behavior.\index{habits!combining with planning}\index{planning!combining with habits}
The habit-based framework using reinforcement learning will lead to fast but inflexible action selection, and is ideally complemented by a planning mechanism which searches an action tree a few steps ahead---as many as computationally possible. Depending on the circumstances, the action recommended by either of the two systems can then be implemented.\citenew{daw2005uncertainty}

Let us go back to the robot which is trying to get the orange juice from the fridge. One possible way of implementing a combination of planning and habit-like behavior is to have a habit-based system \textit{help} the planning system in the \textit{tree search}. Using reinforcement learning, you could train a habit-based system so that when the robot is in front of the fridge whose door is closed, the system suggests the action  ``open the door''. When the door of the fridge is open with orange juice inside, the habit-based system suggests ``grab the orange juice''.\index{robot!fetching orange juice} While these outputs could be directly used for selecting actions, the point here is that we can use them as mere suggestions to a planning system. Such suggestions would greatly facilitate planning: The search can concentrate on those paths which start with the action suggested by the habit-based system, focusing the search and reducing its complexity. However, the planning system would still be able to correct any errors in the habit-like system, and could override it if the habit turns out to be completely inadequate. 

One very successful real-world application using such a dual-process approach is AlphaGo, a system playing the board game of Go better than any human player.\footnote{\citep{silver2016mastering}; \add{see \citet{illanes2020symbolic} for a different dual-process model.}}\index{planning! in Go} The tree to be searched in planning consists of moves by the AI and its opponent. This is a classical planning problem in a GOFAI sense. The world has a finite number of well-defined states, and also, the actions and their effects on the world are clearly defined, based on the rules of the game. What is a bit different is that there is an opponent whose actions are unpredictable; however, that is not a big problem because the agent can assume that the opponent chooses its actions using the same planning engine the agent uses itself.\index{AlphaGo}\index{games!Go}

The search tree in Go is huge since the number of possible moves at any given point of the game is quite large, even larger than in chess. In fact, the number of possible board positions (positions of all the stones on the board) is larger than the number of atoms in the universe---highlighting the fundamental problem in GOFAI-style planning. Since it is computationally impossible to exhaustively search the whole tree, AlphaGo randomly tries out  as many paths as it has time for. This leads to a ``randomized'' tree search method called Monte Carlo Tree Search.
Algorithms having some randomness deliberately programmed in them are often called Monte Carlo methods after the name of a famous casino.\index{tree search!Monte Carlo}
However, a purely random search would obviously be quite slow and unreliable.\footnote{\citep{browne2012survey,chaslot2008parallel}. Monte Carlo Tree Search does include clever tricks which make the search a bit more intelligent. It does not try out actions (or moves in a game) completely randomly, but gathers data on which actions look more promising. In particular, there is quite a lot of data regarding actions taken in the first steps of the search path, since any search has to always try out one of those, and their number is limited because there has not yet been a combinatorial explosion as in the number of long paths. Monte Carlo Tree Search uses such data to bias the search towards paths whose initial parts have been found the most promising.} 

The crucial ingredient in AlphaGo is another system which learns habit-like behaviors. This system is used inside the planning system, a bit like in the juice robot just described. 
While the system is rather complex, let's just consider the fact that in the initial stage of the learning, AlphaGo looks at a large database of games played by human experts. Using that data, it trains a neural network to predict what human experts would do in a given board position---the board positions correspond to the states here. The neural network is very similar to those used in computer vision, and gets as input a visual view of the Go board. 
This part of the action selection system could be interpreted as learning a ``habit'', i.e., an instinctual way of playing the game without any planning.\footnote{Interestingly, the ``habits'' are here learned based on imitation since they are simply trying to replicate what the human players did earlier. Imitation learning is another principle for machine learning, especially important for robots \citep{schaal1999imitation}.\index{learning!imitation}} 
The action proposed by the habit system can be used as such, but even more intelligent performance is obtained by using it as a heuristic for the tree search:\index{heuristics!dual-process view} the tree search is focused on paths related to that proposed action. This heuristic is further refined by further learning stages. In particular, the system also learns to approximate the state-values by another neural network.\footnote{For the general theory on approximating values by neural networks or simpler methods, see \citet[Chapter~9]{suttonbook}. In Chapter~\ref{replay.ch} we will also see  how the system can improve by playing against itself. A completely different purpose for combining learning and planning is to learn to plan better in a given environment where rewards are changing \citep{tamar2016value,pascanu2017learning}. }

Such suggestions based on neural networks are fast, and intuitively similar to what humans would do. Often, a single glimpse at the scene in front of your eyes will tell a lot about where reward can be obtained, and suggests what you should do. 
Even when humans are engaged in planning, such input coming from neural networks often guides the planning. If you go to get something from the fridge, don't you have almost automated reactions to seeing the fridge door closed, and seeing your favorite food or drink inside the fridge? These are presumably given by a simple neural network. Yet, there is a deliberative, thinking aspect in your behavior, and you can change it if you realize, for example, that the juice has gone bad---which the simple neural network did not know.

What is typical in humans is that action selection can also switch from one system to another as a function of practice.\index{automatization}\index{skills!automatization|see{automatization}} 
Learning a new skill, such as driving a car, is a good example---skills are similar to habits from the computational viewpoint.\index{effort!and automatization}\index{learning!skills}
First, you really have to concentrate and consciously think about different action possibilities. With increasing practice and learning, you need to think less and less, since something like a value function is being formed in the brain. In the end, your actions become highly \textit{automated}, and you don't really need to think about what you are doing anymore. The habit-based system takes over and drives the car effortlessly.\footnote{I may be oversimplifying things here, since in the brain, learning motor skills such as driving is not quite the same as forming habits, and they may be based on different brain systems. However, on a more abstract level where we only consider the computational principles, they can be very similar \citep{doyon2003distinct,peters2011towards,sun2005interaction}.\label{skillfn}}

\section{Advantages of categories and symbols}

\index{GOFAI}\index{symbolic AI|see{GOFAI}}
While in this example of Go playing, neural networks and GOFAI work nicely together, it is often not easy to demonstrate any clear utility of symbolic AI approaches. This may of course change any time, since AI is a field of rapid development. It is quite likely that GOFAI is necessary for particularly advanced intelligence---something much more advanced than what we have at this moment. 
Yet, the tendency has recently been almost the opposite: tasks which were previously thought to be particularly suitable for symbolic AI have been more successfully solved by neural approaches. For example, large language models used in systems like ChatGPT effectively transform language, i.e.\ text data, into a sequence of high-dimensional continuous-valued vectors before inputting them into a huge neural network.\footnote{\citep{achiam2023gpt}. This holds for most successful natural language processing systems, such as Google translate \citep{wu2016google}.}

Perhaps symbolic AI works with board games only because such games are in a sense discrete-valued: the stones on the Go board can only be in a limited number of positions, so the game is inherently suitable for GOFAI.
So, we have to think hard about what might be the general advantages of logic-based intelligence compared to neural networks. In the following, I explore some possibilities.

\subsection{GOFAI is more flexible and facilitates generalization}

Suppose that there is a neural network that recognizes objects in the world and outputs the category of each object. Then, what would be the utility of operating on those categories as discrete entities, using symbolic-logical processing, instead of having just a huge neural network that does all the processing needed?

We have already seen, more than once,  one great promise of GOFAI in the case of planning: flexibility. Given any current state and any goal state, a planning system can, if the computational resources are sufficient, find a plan to get there. If anything changes in the environment---say, it is no longer possible to transition between two states due to some kind of blockage---the planning system takes that into account without any problems. This is in contrast to reinforcement learning which will not know what to do if the environment changes; it may have to spend a lot of time re-learning its value functions. 

\index{data structures}
Furthermore, GOFAI is easily capable of representing various kinds of  \textit{data structures} and relationships in the same way as a computer database. For example, it can easily represent the fact that both cats and dogs are animals, i.e.\ the hierarchical structure of the categories. It can also represent the relationship that the character string ``Scooby'' is the name of a particular dog. This adds to the flexibility of GOFAI by allowing more abstract kinds of processing, which are  easily performed by humans.

\index{categories}
Another wide-spread idea is that categories are useful for generalizing knowledge over categories, which in its turn underlies various forms of abstract thinking.
Even though cats are not all the same, it is useful to learn some of their general properties. They like milk, 
they purr; they don't like to chew bones like dogs do, and they are not dangerous like bears.  Having categories enables the system to learn to associate various properties to the whole category: Observing a few cats drink milk, the system learns to associate milk-drinking to the whole category of cats, instead of just some individual cats.
Importantly, associating properties to categories means the system was able to \textit{generalize}: after seeing \textit{some} of the cats drink milk, it inferred that \textit{all} cats drink milk. Such generalization is clearly an important part of intelligence. If the system needed to learn such a property separately for each cat, it would be in great trouble when it sees a new cat and needs to feed it --- it would have no idea what to do. But, learning that the whole category of cats is associated with milk-drinking, it knows, immediately and without any further data, what to give to this new cat.

\subsection{Categories enable communication} 

Nevertheless, I think the feature which makes GOFAI fundamentally different from neural networks is that the use of symbols is similar to using some kind of a primitive \textit{language}. In fact, you can hardly have GOFAI without some kind of a language---perhaps akin to a programming language---in which the symbols and logical rules are expressed.

It is equally clear that with humans, language is primarily used for communication between individuals.\index{language}
As each category typically corresponds to a word, humans can communicate associations, or properties of categories, to each other. I can tell my friend that cats drink milk, so she does not need to learn what to feed to cats by trial and error. I have condensed my extensive data on cats' eating habits into a short verbal message that I transmit to her. 

So, it is plausible that the main reason humans are capable of symbolic thinking is that it enables them to communicate with each other. After such a communication system was developed during evolution, humans then started using the same system for various kinds of intelligent processing even when alone.
Perhaps we started by telling others, for example, where to find prey.
This led to the development of symbols and logical operations, which were found useful for abstract thinking: Perhaps you could try to figure out yourself where you should hunt tomorrow. Eventually, such capabilities ended up producing things such as in quantum physics---and the very theory of GOFAI.\footnote{This is what is called ``exaptation'' in evolution. It means that a trait was first produced to adapt for one phenomenon, but then it turned out to be useful for something else. A typical example is bird's feathers, which probably first evolved to keep the birds warm, and only later turned out to be useful for flying. \citet{wagner2003progress} discuss computer simulations on the emergence of language for communication between agents.
  \citet{dehaene2022symbols} propose that the main capacity specific to humans is using symbols and mental programs,  generalizing the idea that language and sequence processing is specific to humans. For an attempt to do some kinds of logical processing in a neural system, see \citet{frady2020resonator}.
}\index{exaptation}

A reflection of the utility of categories in communication may be seen in a recent research line in AI which tries to develop systems whose function is easy to interpret by humans.\citenew{su2015interpretable,guidotti2018survey,arrieta2020explainable} If you use a neural network to recognize a pattern, the output may be clear and comprehensible, but the computations---why did the network give that particular output--- are extremely difficult to understand for humans. This is fine in many cases, but sometimes it is necessary to explain the decision to humans.\index{interpretability}
For example, if an AI rejects your loan application, the bank using the AI may be legally obliged to explain the grounds for that decision.\footnote{For example, the General Data Protection Regulation (GDPR) of the European Union imposes a general ``right for explanation'' for almost any decision made by an algorithm on an individual \citep{goodman2016european}. One reason for such a requirement is to make sure that the AI did not discriminate applicants based on gender, race, or similar characteristics---an objective called ``fair AI''. Another reason is that the AI might not make the final decision, but could be used as a support for a human decision-maker, such as a medical doctor;  the human decision-maker would greatly benefit from understanding why the AI came to its conclusion.
  One more reason for making AI easy to interpret is that understanding how an AI works makes it easier to evaluate its potential safety hazards, and develop AI that is safe.  }
Researchers developing such interpretable AI often end up doing something similar to GOFAI boosted by learning, since it gives rules which can be expressed in more or less ordinary language, and thus they can be explained.
In fact, in Chapter~\ref{ml.ch} we saw examples of GOFAI systems whose functioning is easy to understand and to explain.\footnote{I am actually tempted to think that the \textit{only} specific utility of categories (which cannot be obtained without them) is communication, including being interpretable and comprehensible by humans in the case of AI.  In particular, it is not clear to me if explicit categories are needed for generalization. Without going into details, let me just mention that similar operations could be performed directly in a representational space by simply propagating any associations to near-by points in that space without any strict division into categories.
  For opposite viewpoints putting concepts at the heart of (human) cognition, see \citet{rosch1999reclaiming,harnad2017cognize}. Obviously, it is important here to compare the different definitions of categorization used: \citet{harnad2017cognize} uses a definition which is very general.}

\section{Categorization is fuzzy, uncertain, and arbitrary}

Now, let us consider the flipside: problems that arise when using categories. We have already seen some problems in logical-symbolic processing, the most typical being the exponential explosion of computation in planning. 
Here, we focus on the consequences of using categories, and look at the question from a more philosophical angle. Indeed, it has been widely recognized by philosophers over the centuries that  dividing the world into ``crisp'' categories can only be an approximation of the overwhelming complexity of the world. I focus on some issues which will in later chapters be seen to be relevant for suffering.\footnote{I don't go into any details on how that ``division'' of the world into categories happens, but for the interested reader, I give some pointers here. Earlier, we considered the case where the neural network recognizes an object and outputs its category. This is a simple starting point; while it can be easily done by supervised learning, it can also be implemented by unsupervised learning methods, in particular methods such as clustering and (Gaussian) mixture modelling. In the case of humans, the connection between neural networks and logic-symbolic processing is related to what is called the symbol grounding problem \citep{harnad1990symbol}. It is a topic subject to a lot of debate: some argue no proposed solution is sufficient \citep{taddeo2005solving}, while others argue it is essential to consider robots which communicate with each other \citep{steels2008symbol}. The operation of neural networks is closely related to one well-known proposal called the prototype theory. It means we define each category by a single point in the space the activities of units in a neural network (preferably in layers close to output); this point is the prototype \citep{rosch1978principles}. Basically, you would find a ``prototypical'' cat as a point in the very center of all those points that represent cats. A generalization of this idea can be found in \citet{gardenfors2004conceptual}. However, things get much more complicated in the case of abstract categories such as ``good'' or ``beautiful''.}

\subsection{Categories are fuzzy}

Philosophers have long pointed out that there may not be any clearly defined categories in the world. Granted, the difference between cats and dogs may be rather clear, but what about the category of, say, a ``game''? Wittgenstein gave this as an example of a category which has no clear boundaries. Different games have just some vague similarity, which he called ``family resemblance''.  

This idea has been very influential in AI under the heading of \textit{fuzziness}. A category is called fuzzy if its boundaries are not clear or well-defined.\index{fuzzy}
Consider for example the word ``big''. How does one define the category of big things? For simplicity, let us just consider  the context of cities. If we say ``London is big'', that is clearly true: London definitely belongs to the category of big things, in particular big cities. But if we say ``Brussels is big'', is that true or false? How does one define what is big and what is not? In the case of cities, we could define a threshold for the population, but how would we decide what it should be? An AI might learn to categorize cities into big and small ones based on some classification task---in Chapter~\ref{ml.ch}, we discussed how this might happen in categorizing body temperature into ``high fever'' or not. However, that categorization would depend on the task, and there would always be a gray zone where the division is rather arbitrary.

The consensus in AI research is that many categories are quite fuzzy and have no clear boundaries; there are only different degrees of membership to a category. There is no way of defining a word like ``big'' (or, say, ``nice'', ``tall'', ``funny'') in a purely binary (true/false) fashion. There will always be objects that quite clearly belong to the category and objects which clearly do not belong to the category, but for a lot of objects the situation is not clear. In the theory of fuzzy logic, such fuzziness is modelled by giving each object a number between 0 and 1 to express the degrees of membership to each category.\citenew{mendel1995fuzzy}

\subsection{Categorization is uncertain}

In addition, categorization is always more or less uncertain. 
Any information gleaned from incoming sensory input is uncertain, for reasons we will consider in more detail in Chapter~\ref{perception.ch}. Partly, it is a question of the neural network getting limited information, and partly because of its limited information-processing capabilities. If you have a photograph of a cat taken in the dark and from a bad angle, the neural network or indeed any human observer may not be sure about what it is. They might say it is a cat with 60\% probability, but it could be something else as well.
In other words, any categorization by an AI is very often a matter of probabilities.
\index{uncertainty!of categorization}

It is important to understand that fuzziness and uncertainty are two very different things. 
Uncertainty is a question of probabilities, and probabilities are about lack of information. If I say that a coin flip is heads with 50\% probability and tails with 50\% probability, there is no fuzziness about which one it is. After flipping the coin I can say if it is heads or tails, and no reasonable observer would disagree with me (except in some very, very rare cases). In other words, uncertainty is a question of not knowing what will happen or has happened, i.e., a lack of information about the world. In contrast, fuzziness has nothing to do with lack of information; it is about the lack of clear definition. We cannot say if the statement ``Brussels is big'' is true even if we have every possible piece of information about Brussels, including its exact population count. According to the information I find on Wikipedia, its population is 1,191,604, but knowing that will not help me with the problem if I don't know how many inhabitants are required for a city to be in the ``big'' category. 

Humans are not good at processing uncertainty. 
Various experiments show that humans tend to use\index{thinking!categorical}
excessively \textit{categorical thinking}, where the uncertainty about the category membership is neglected. That is, when you see something which looks to you most probably like a cat, your cognitive system tends to ignore any other possibilities, and think it is a cat for sure.\footnote{An example  was seen  in a study where 
the subjects were told a story which suggested that an imaginary person entering a house would be either a burglar, or a real estate agent. When the imaginary person was more likely to be a real estate agent than a burglar---based on various cues such as what other characters in the story were thinking--- they tended to ignore the possibility that the person is a burglar altogether, as seen in the predictions that they made about the behavior of the imaginary person \citep{malt1995predicting,murphy2010uncertainty}. The authors also found a way of remedying the situation: if the subjects are asked what the probabilities of the two categories are, and their estimates are shown on the computer screen (say, ``65\% vs.\ 35\%''), the subjects are able to take the uncertainty of the categorization into account.}

An old Buddhist parable about these dangers in categorization is seeing a rope in the dark and thinking it is a snake. You miscategorize the rope, and your brain activates not only the category of a snake, but all the associations related to that category (``animal'', ``dangerous''). You get scared, with all the included physiological changes, such as an increased heart rate. 
If you had properly taken the uncertainty of such categorization into account, your reaction might have been more moderate.

\subsection{Categorization is arbitrary}

In some cases, the categories are not just fuzzy or uncertain: their very existence  can be questionable. 
Consider concepts such as ``freedom'' or ``good''. Even forgetting about any difficulties in programming an AI to understand them, is it even clear what these words mean? Certainly, they mean different things to different people: people from different cultural backgrounds may easily misunderstand each other simply because they use such concepts with slightly different meanings. A great amount of time can be spent in attempting to just describe the meanings of certain words and categories. In fact, we spend more than one chapter on analyzing the category called ``self'' in this book.\index{self!as category}

Even in rather straightforward biomedical applications of machine learning, we often use categories that are not well-defined. For example, in a medical diagnosis context, it is not clear if what we usually call schizo\-phrenia is a single disease. Perhaps there are a number of different diseases which all lead to the single diagnosis of schizophrenia.\footnote{\citep{peralta2001many,brodersen2014dissecting}. The same could be said of depression \citep{drysdale2017resting}. For a general overview on such ``precision medicine'', see \citet{insel2015brain}.} Developing effective medications may only be possible once we understand all the subtypes, while thinking of all the subtypes as a single disease (a single category) may mislead any treatment attempts. 

Moreover, a categorization that works for one purpose might not be suitable for another. We might divide people into different nationalities, which is very useful from the viewpoint of knowing what languages they are likely to understand. However, we can too easily use the same categories to predict all kinds of personality traits of those individuals, and that prediction may go quite wrong. Thus, the categories and their utility depend on the context. Moreover, since different people  use  different categories in different ways, they are subjective.\footnote{Human categorization can also change when the frequencies of different objects change. In the experiments by \citet{levari2018prevalence}, ``When blue dots became rare, participants began to see purple dots as blue; when threatening faces became rare, participants began to see neutral faces as threatening''.}\index{context-dependence!of categories}

\index{Yogacara}
Such arbitrariness of categories has been well appreciated in some philosophical schools. In the \yogacara\ school of Buddhism, it is claimed that ``while such objects [as chairs and trees] are admissible as conventions, in more precise terms there are no chairs, or trees. These are merely words and concepts by which we gather and interpret discrete sensations that arise moment by moment in a causal flux.``\footnote{Quote from \citep{lusthaus2013and}, see also \citep{lusthaus1998buddhism,shun2009living,williams2008mahayana}. In ancient Greece, Pyrrhonian Skeptics\index{Skeptics} had ideas similar to such Buddhist schools, and indeed Pyrrhonians are likely to have been directly influenced by Buddhist thinkers due to Alexander the Great's campaign into India \citep{Gymnosophists,mcevilley1982pyrrhonism,garfield1990epoche,kuzminski2007pyrrhonism}.\label{alexanderfn}} What arises in such a moment-by-moment flux is, in our terminology, activities in neural networks. 
Categories are created afterwards, by further information-processing.

\subsection{Overgeneralization}

It may be easy to understand that miscategorization leads to problems, as in mistaking a rope for a snake. However,
the biggest computational problem caused by all properties just discussed---fuzziness, uncertainty, arbitrariness---may be overgeneralization. Overgeneralization can be difficult to spot, even after the fact, which makes it particularly treacherous.\index{generalization}\index{overgeneralization}

Overgeneralization means that you consider all instances of a category to have certain properties, even if those properties hold only for some of them. Since categories are fuzzy, anything which is not really firmly inside the category may actually be quite different from its prototype. 
Related to this, you may not acknowledge the uncertainty of categorization and the ensuing generalization. Even more rarely do people acknowledge that the very categories are arbitrary. 

Overgeneralization effects are well documented, for example, in perception of human faces, where gender and race can bias any conclusions you make about the individual involved.\footnote{\citep{freeman2016more}. In that case, a further problem is that the categories may operate using stereotypes (which may not be factually accurate to begin with), which means that the generalization is even more wrong.\index{stereotypes}}
  As an extreme case of overgeneralization, if you have been bitten by a dog, you may develop a fear towards \textit{all} dogs, which would be called a phobia.\index{phobia}\index{fear!phobia|see{phobia}} \add{ Such fear is overgeneralizing in the sense that it is very unlikely that the other dogs would bite you. If you didn't use any categories, you would only be afraid of that one dog that bit you.}
This is a very concrete example of how thinking in terms of categories leads to suffering, as will be discussed in more detail in later chapters. 

There are actually good computational reasons why overgeneralization occurs. 
Learning to generalize based on a limited number of categories means that knowledge gleaned from all the instances of each category can be pooled together. If you actually had enough data from all the dogs in the world, as well as unlimited computational capacities, you would be able to learn that some of them are safe while a few are not. However, data and computation are always limited, so some shortcuts may be necessary---even if they increase your suffering. This is another theme that we will return to over and over again in this book.

\chapter{Summarizing the mechanisms of suffering}\label{summary1.ch}

\add{
So far in this book, we have seen several computational ideas related to suffering.
We started by considering two basic mechanisms for suffering in Chapter~\ref{suffering.ch}: frustration and threat to the person or the self.
We first defined frustration as not reaching a goal (Chapter~\ref{planning.ch}) and later in terms of reward loss and reward prediction error (Chapter~\ref{rpe.ch}). In fact, these two kinds of frustration align well with the dual-process theory---slow vs.\ fast or GOFAI vs.\ neural networks---considered in Chapter~\ref{dual.ch}. 
In Chapter~\ref{self.ch}, we further argued that the concept of self includes higher-order desires related to self-evaluation or self-preservation, and these can also be frustrated. Chapter~\ref{threat.ch} developed a theory of threats and fear based on predicted reward loss in the future. To sum up, we obtained a theory in which suffering is based on error signals given by frustration or prediction of frustration. %
In this chapter, we summarize the ideas of the previous chapters, emphasizing the many different forms that frustration can take.
}

\section{Frustration on different time scales}

Consider a case where you are yourself going to fetch the orange juice from the fridge. You formulate a plan which involves high-level actions such as going to the fridge, opening the door, etc. Once you are in front of the fridge, your habit-based system suggests you open the door by a certain sequence  of muscle contractions which you have performed hundreds of times and which has become quite automated.\index{frustration!summarizing different aspects}\index{threat! to the intactness of person|see{intactness of the person}}

Now, suppose you follow the habit-based system and pull the door handle, but the door does not quite open. This kind of ``frustrates'' your habit of opening the door. But do you suffer? Probably not very much; you just pull again with more force, and if it opens, you hardly register anything out of the ordinary happened. In contrast, if you don't get the juice at all---because the door is somehow broken and does not open at all--- your long-range plan is frustrated, and you will definitely suffer.
There is a good reason for that suffering: all that planning and even the walking was in vain. A strong error signal has to be sent throughout your brain, and that is suffering.\index{planning!hierarchical}

This example points out one important aspect of action selection: its temporally hierarchical nature, involving simultaneous computations on different time scales.\footnote{This is called hierarchical control \citep{poole2010artificial}, hierarchical planning or hierarchical task networks \citep{georgievski2015htn,nau2003shop2} or hierarchical reinforcement learning \citep{sutton1999between,dietterich2000hierarchical,botvinick2012hierarchical}. A related paradigm is given by what is called ``options'' in reinforcement learning, see \citep[p.~461]{suttonbook}.\index{reinforcement learning!options}\index{reinforcement learning!hierarchical} }
In the brain, there are also processes operating at many different time scales.\index{time scales}\label{timescales} Reality is, of course, a bit more complex than the clean division into planning and habit-based actions we have discussed so far.

Some form of frustration can be operating on many different levels simultaneously. In one extreme, the agent may be planning long action sequences, and if they fail, frustration ensues in the sense of not reaching the goal. In the other extreme, a habit-based reinforcement learning system builds predictions on what kind of rewards or changes in state-values are associated with different actions, and computes whether there is reward loss or an RPE. 
Predictions are made on a millisecond time scale as well as on the time scale of days if not years. Each such time scale has its own learning mechanism using its own errors.\footnote{\citep{hari2010brain,botvinick2012hierarchical}; in fact, RPE's seem to be coded even in the cerebellum \citep{heffley2019classical,kawato202150}. Using RPE instead of reward loss simplifies the situation to some extent, since RPE considers the total future reward and is thus less dependent on the definition of the time scale, as explained in footnote~\ref{predictedfrustrationfn} in Chapter~\ref{rpe.ch}.} 

Such division into time scales brings us to the concept of intention---defining intention as commitment to a goal, as discussed in Chapter~\ref{planning.ch}. The point in intentions is to partly resolve conflicts between long-term and short-term optimization. 
I can have many desires simultaneously and spend some time thinking about each of them, and perhaps even planning each of them to some extent. But I'm not really hoping to reach \textit{all} the goals related to those desires. Once I decide to commit to one of the goals, that is what sets the goal, which can then be frustrated.
I would argue that in the case of planning, frustration is not so much due to desire itself but to the ensuing intention. 
This is in line with the more elaborate expositions of the Buddha's philosophy on suffering which divide desire into initial desire and a later part called \textit{attachment} (also translated as ``clinging'' or ``grasping''). Attachment is a process where after an initial feeling of desire (``Nice, chocolate, I would like to have it''), you firmly attach to the object of your desire (``I must have that chocolate''). This distinction seems to be similar to the distinction between desires and intentions in our terminology.\index{intention!and attachment}\index{attachment (Buddhist)}\index{clinging|see{attachment}}
Buddhist philosophy suggests a central role for attachment, or intention, in the process which creates suffering. 
While such attachment or intention is not necessary for frustration to occur, I propose that it greatly amplifies it. This is logical because intentions consider longer time scales, and thus an error related to intention is more serious, since more time and energy were lost in formulating and executing the plan that failed.

\section{Frustration based on desires, expectations, and general errors}

We have also seen two different kinds of frustration: not reaching a goal vs.\ incurring  a reward loss. One underlying difference between the two cases is that reward loss is based on violation of \textit{expectations}, while not reaching the goal is in line with the typical definition of frustration as not getting what one wants, i.e.\ violation of \textit{desires}.\index{frustration!based on desire vs expectation} \label{expvsdesire}
It may thus seem that our definitions are to some extent contradictory.
One way of resolving this is to consider that the term ``expectation''  may have different meanings in different contexts.\footnote{The difficulty of defining expectation was earlier discussed in footnote~\ref{expdeffn} in Chapter~\ref{rpe.ch}.}\index{expectation!definition}\index{frustration!definition of expectation} The agent is executing a plan in order to get to the goal state, and it is in that sense ``expecting'' to get to that goal state. Earlier, we saw (page~\pageref{epictetospromise}) how Epictetus talks about desire ``promising'' the attainment of its object. 
Thus, the expectation related to planning could simply be defined as the goal state being reached.
  Then, reward loss would be the same as the frustration of not reaching the goal, that is, the object of the desire (using the definition of desire given in Chapter~\ref{planning.ch}).

  Alternatively, we could see frustration of desires and reward loss (based on expectation) as two distinct,  if closely related phenomena, both of which produce suffering. What they have in common is that some kind of \textit{error} occurred.%
 \index{suffering!as error signalling}\index{frustration!and general errors}\index{error signalling!general}
  This opens up the possibility of a very general viewpoint where the connection between suffering and error signalling does not need to be concerned with goals or rewards at all. We all know that it is unpleasant if we expect something and then it does not happen, even if the event we were predicting was neutral in the sense of providing no reward. Thus, it is possible that there is some kind of suffering in almost \textit{any} prediction error.\footnote{On the other hand, if we expect something unpleasant to happen, and it does not happen, we feel relief\index{relief} which is clearly not suffering; in general, obtaining more reward than expected may be the very definition of pleasure (see footnote~\ref{pleasurefn} in Chapter~\ref{rpe.ch}). Perhaps, in that case, the unexpected positivity (of reward)  overrides the inherent suffering in prediction error. For research on relief from pain, see \citet{seymour2005opponent,leknes2008pain}.}  Most interestingly, it has been proposed that dopamine signals prediction errors even for events not related to reinforcement, so it might provide a neural mechanism for general signalling of errors.\footnote{\citep{takahashi2017dopamine,redgrave2006short}.}\index{frustration!and general errors}\index{dopamine}

\add{The meaning of such errors is further modified by the context. If you are deliberately engaged in the learning of, say, a new skill, errors are quite natural, and you are likely to feel less frustration; in a sense, you are expecting that there are errors. Or, if your prediction of the reward is uncertain, i.e.\ only very approximate, the frustration is likely to be weaker. We will have much more to say about such effects in later chapters.

  Depending on the context, what I call frustration in this book can actually correspond to different concepts with slightly different meanings. \textit{Disappointment}\index{disappointment} is a closely related term; it can also be used when no particular action is taken by the agent while the world turns out to be worse than expected.\index{disappointment} \textit{Irritation}\index{irritation} and even \textit{anger}\index{anger} can also be used to describe feelings very similar to frustration. While 
 anger can more specifically mean the interpersonal feeling of anger towards other people, it is often caused by the fact that their actions lead to frustration (more on this in Chapter~\ref{emotions.ch}). \textit{Regret}\index{regret} can be seen as frustration specifically based on our own actions,\footnote{\add{On the question of attributing frustration to oneself, see footnote~\ref{selfattributionfn} in Chapter~\ref{self.ch}.}} often amplified by  recalling past frustration (more on this in Chapter~\ref{replay.ch}).}

\section{Self, threat, and frustration}

Frustration on different time scales leads us to the frustration of self-needs treated in Chapter~\ref{self.ch}.\index{threat!and frustration}
Self-needs often work on time scales of days, months, even years, thus an even longer time scale than ordinary planning and attachment.
These different time scales can actually be related to van Hooft's different kinds of frustration discussed in Chapter~\ref{suffering.ch}: frustration of biological functioning, of desires and emotions (in his terminology), of more long-term life goals, and even of the sense of the meaning of one's existence.\index{suffering!definition!van Hooft} 
\add{It may very well be that such \textit{self-related frustration} produces some of the very strongest frustration and suffering. Frustration of self-needs is also closely connected to suffering coming from threats to the ``intactness of the person'' \`a la Cassell, thus combining the two mechanisms of suffering.}\footnote{\add{\citet{maslow1941vii} even proposes that mere ``deprivation'' in itself does not have the negative effects attributed to frustration, but only when it is at the same time ``a threat to the personality, that is, to the life goals of the individual, to his defensive system,
    to his self-esteem or to his feeling of security.'' Frustration of self-needs is closely related to such threats as was discussed in Chapter~\ref{self.ch}.}}\index{self!needs}\index{frustration!self-needs}

\add{In Chapter~\ref{threat.ch},  self-related frustration was even considered as a means to reduce the concept of threat to a form of frustration---of self-needs such as the desire for safety---to obtain a unified theory on suffering. However, while such a theoretical simplification has some interest, it was considered to go a bit too far; the connection between frustration and threat is certainly more complex. In Chapter~\ref{threat.ch}, we actually saw a fundamental distinction that can be made between frustration and threats: threats are about \textit{predicting} that a bad thing \textit{might} happen, while frustration is fundamentally about realizing the bad thing \textit{did already} happen. Nevertheless, it is interesting to ask if it is also possible to see the connection between frustration and threats from the opposite angle: can frustration be seen as a special case of a threat to the self?\index{threat!and frustration}

We can indeed consider that failing in a task implies a threat to one's self-image that Cassell talks about,\footnote{\add{\citet{van1998suffering} cites another formulation used by Cassell (in a source that is difficult to find): ``Suffering is the state of distress induced by the threat of the loss of intactness or the disintegration of personhood--- bodies do not suffer, persons do''.}} and which is related to the self-evaluation of Chapter~\ref{self.ch}. A frustrating experience may imply that the agent's positive self-image is not correct, and that it has to change its self-image so that it will consider itself less competent than it thought earlier. In this sense, frustration is a threat, especially if the frustration did not yet change the self-image but implied a certain probability that it should be done in the future. If you fail at a work task today, that is frustration but you may still think you know how to do your job. But the failure will increase your perceived probability that one day, you have to admit that you just don't know how to do your job at all. That is a threat to your person, implying the possibility that one day you will have to update your self-image to a more negative one.
Likewise, any frustration could be considered to imply a threat to the very survival of a biological agent because it suggests that the agent's decision-making system is not working very well in the current environment, which could lead to life-threatening problems in the future. %
This way, frustration can be reduced to a special case of the threat of intactness of the person.\footnote{I'm grateful to Michael Gutmann for suggesting this interpretation.\index{intactness of the person}\index{frustration!as threat}}  This is in line with the thinking prevalent in Mahayana Buddhist schools, where the self is seen as the source of all desires and all suffering.

This intricate interplay between frustration and threat is seen in the very definition of threat in Chapter~\ref{threat.ch}, where threat is based on anticipation of reward loss. This means that many of the properties of frustration just discussed also apply to threat: threat operates on different time scales, and threats can be based on frustrating expectations or on frustrating desires.
Most importantly, if there were no reward loss, there could be no threat either. In this fundamental sense, it is the threat that is secondary and can be reduced to frustration; indeed, \textit{all fear is fear of frustration.} 
This intimate connection will be important in Part~II where we consider the conditions creating suffering, and Part~III where we consider interventions to reduce suffering.  As a sneak peek, consider the proposal by Seneca, a Roman Stoic, for reducing threat-based suffering:
``Cease to hope (...) and you will cease to fear''.\footnote{\add{Quote from \Lucilius\ V.7, translated by R.~M.~Gummere.  I'm here assuming ``hope'' refers to an expectation of reward that could lead to frustration. Likewise, Epictetus says ``When I see a man anxious, I say, What does this man want?'' and tells the story of a lyre player who is happy playing by himself but becomes nervous in front of an audience, apparently  because he \textit{wants} to be well-received by the audience (\Discourses, II.13).}}

}

\subsection{\add{Can desire in itself produce suffering?} }\label{desiresuffering}
  
\add{Going a bit beyond the theories of the preceding chapters, I am tempted to think that desire (or aversion) \textit{in itself}, especially when combined with intention, can immediately create some kind of suffering even before any frustration, and even in the absence of any specific threat.\footnote{\add{I think such a claim is an essential part of early Buddhist philosophy, although I find it difficult to give a reference.  While the early Buddhist texts very clearly say that desire and aversion lead to suffering, the texts are not very clear on whether they create suffering only afterwards (based on frustration) or even immediately, in themselves.}    }  
It could be that whenever there is desire or aversion, the system predicts that there will be frustration with some probability, and this constitutes a threat, creating  suffering.\footnote{\add{Related philosophical ideas are proposed by \citet[Ch.~4]{airaksinen2019vagaries}.}}  It is in fact clear that fear, which is  an aversion towards possible future events, does create suffering in itself based on a threat; perhaps other kinds of aversion and even desire share similar mechanisms.\index{fear}
Another possibility is that such suffering is based on a general error-signalling mechanism: the internal representation of a goal state which is \textit{different} from the current state is an error that may automatically lead to the triggering of an error signal and to suffering.\footnote{\add{Related to this, it is widely observed in workplace psychology that unfinished plans create stress \citep{masicampo2011unfulfilled,peifer2020thieves}.}} %
Understanding this kind of suffering is an important question in future research.\footnote{\add{One more approach is given in Chapter~\ref{emotions.ch} which will propose that desire as well as some forms of aversion can be seen as ``interrupts'', which may produce suffering by a special mechanism (page~\pageref{interruptsuffering}). }} %
}

\section{Why there is frustration: Outline of the rest of this book}

To recapitulate, Part I of this book described a wide spectrum of frustration-related error signalling. Not reaching a goal, not getting an expected reward, or making an error in predicting any event, can all be seen in this same framework. They work on different time scales, and use different systems in the dual-process framework. It seems that particularly strong suffering is obtained by frustration of planning, and even stronger by frustration of self-needs.

\index{frustration!root causes}
Next, we will try to understand why there is frustration in the first place.
On some level, it is obvious that we cannot always reach our goals, or get what we want, if only because of the limitations in our physical skills and strength: We cannot move mountains. The world is also inherently uncertain and unpredictable, so even the perfect plan may fail because something unexpected happens. 
Yet, more interesting for our purposes are the cognitive limitations. As argued earlier, cognition is something that can be relatively easily intervened on, and modified to some extent. Thus it is more feasible to reduce suffering by focusing on the cognitive mechanisms, instead of trying to develop devices that physically  move mountains. Therefore, it is crucial to understand in as much detail as possible how various processes of information-processing contribute to suffering.

We have already seen several information-processing limitations that can produce or amplify frustration. For example, planning is difficult due to the exponential explosion of the number of paths, which means our plans may be far from optimal. We need a lot of data for learning: data may be lacking to build a good model of the world, or to learn quantities such as state-values. Categories are often used in action selection---in particular, if the world is divided into states---but these categories may not even be well-defined. The cognitive system may be insatiable and always want more and more rewards. There are several self-related needs which can create particularly strong suffering by mechanisms related to frustration.

Part \ref{partii} goes into more depth regarding such limitations that produce frustration, focusing on the origins of uncontrollability and uncertainty. 
Later, Part~\ref{partiii} will consider methods for reducing suffering, mainly by reducing frustration. I will summarize all the different aspects of frustration in a single ``equation'' (page~\pageref{suffeq}) and propose various methods or interventions to reduce frustration based on the theory of Parts~\ref{parti} and \ref{partii}. Such interventions will largely coincide with what Buddhist and Stoic philosophies propose, and include mindfulness meditation as an integral tool.

\part[Origins of suffering: uncontrollability and uncertainty]{Origins of suffering: \\uncontrollability and uncertainty
  \\ \ \\ \ \\  \normalsize The second part will consider  how uncontrollable the world as well as the cognitive system itself are, and how an agent's perceptions and thinking are uncertain and can even be called illusory \label{partii}}

\chapter{Emotions and desires as interrupts}
\label{emotions.ch}

Part~\ref{partii} of this book is about better understanding why there is suffering and what increases it. Since we saw earlier that suffering can fundamentally be seen as frustration---possibly only prediction of frustration---the question is what kind of factors increase frustration. Part~\ref{partii}  analyzes frustration in terms of uncontrollability and uncertainty (which is related to unpredictability). These properties make errors in action selection likely, and thus lead to frustration. Even the mind itself is seen as uncontrollable, since it has multiple processes operating at the same time, in particular emotions (Chapter~\ref{emotions.ch}) and wandering thoughts (Chapter~\ref{replay.ch}). Further, perceptions are uncertain due to incomplete input data as well as a faulty prior model of the world (Chapter~\ref{perception.ch}). The difficulties of communication between different brain areas or processors create a further loss of control (Chapter~\ref{control.ch}). Ultimately, we need to confront the problem of consciousness (Chapter~\ref{consciousness.ch}) which creates a kind of a virtual reality where painful events are simulated again and again. For a sneak preview of what the system will look like in the end, the reader can have a look at Figure~\ref{flowchart1.fig} on page~\pageref{flowchart1.fig}.

\medskip
\medskip
\begin{center}*****\end{center}
\medskip
\medskip

\noindent In this chapter, we look at the concept of emotions.\index{emotions}
Anybody pressed to give sources of suffering would probably give a list of such phenomena as fear, disgust, sadness, and perhaps anger. Those are actually some of the most typical emotions in the terminology of neuroscience and psychology. If we are to understand suffering, we have to understand how such emotions are related to it.

In Chapter~\ref{threat.ch}, we saw how threat can be seen as a prediction of possible frustration. A threat triggers an emotion called fear. But what does fear then add to the computation that detected a threat?  Many things: when assailed by fear, you forget everything else you were doing, you focus your attention exclusively on whatever caused your fear, you try to figure out how to get rid of it, and, eventually, you run away fast. These are examples of the aspects of emotions we investigate in this chapter.

We discuss how emotions can be seen as information processing and signalling, focusing on fear as a prime example. 
The main focus here is how emotions capture attention and interrupt ongoing processing. 
Another aspect is that emotions trigger basic, pre-programmed behavioral sequences, such as running away. Importantly, emotions are something that reduces any control we have over our minds and bodies, which is one leitmotiv in this part of the book. We also see that desires have similar interrupting qualities.

\section{Computation is one aspect of emotions}

Some readers may wonder what emotions have to do with artificial intelligence. Surely, we can program an AI or a robot to function using purely ``rational'' procedures: Maximize expected reward and act accordingly, within the limits of the information the AI has, and as far it is computationally possible. Why would we need to introduce anything ``emotional'' in the system?
To proceed, we first need to understand what the word ``emotion'' means. 
Unfortunately, very different definitions are used, and there is no generally agreed definition, even in the limited  context of neuroscience and psychology. 

\subsection{Emotions have many components}

The most comprehensive definitions define an emotion as a complex of several different components. For example, if you feel fear, you will have a particular facial expression, you may scream, and your body will undergo physiological responses such as increased heartbeat. Next, your cognitive (i.e.\ computational) apparatus will start planning how to escape from the situation, and indeed, pre-programmed behavioral routines such as fleeing may be activated. While all this is happening, you will also feel afraid, in the precise sense that you have the conscious experience of being afraid. 

As with almost any phenomenon in neuroscience and psychology, some emphasize the behavioral aspect of emotions, while others concentrate on more internal phenomena, including information-processing---usually called cognition in this context.
Emotions are further characterized by a feeling tone: often negative (as in fear) but sometimes positive (as in joy). The feeling tone, technically called ``valence'', is seen as the core of emotions by some, providing motivation for action. 
Yet others think that what defines an emotion is the conscious, subjective experience, such as feeling afraid.\index{subjective experience!in emotions}\index{experience!subjective|see{subjective experience}}\index{consciousness!in emotions}

In this book, I take an approach where all the aforementioned components together constitute an emotion.\footnote{Such a multi-component definition of emotions, for example by \citet{scherer2009dynamic}, is wide-spread in psychological literature. In contrast, in neuroscience and AI emotions are often defined more narrowly, or not properly defined at all. The list of such components given in the text is not at all complete; one might add bodily responses \citep{nummenmaa2014bodily}, such as given in the fear example in the main text, as well as social aspects \citep{nummenmaa2012emotions}.}
Nevertheless, I focus on the computational, information-processing aspect of emotions, in line with the general approach of this book.
Such information processing is eventually reflected in behavior, and at least in humans, it often leads to a subjective conscious experience.

\subsection{Emotions help when computation and information are limited}

The key question in this chapter is: how is information processed in what we call emotions; what is special in that information-processing when we feel, for example, fear or disgust? The starting point here is that emotions are needed because of the limited information available and the limited computational capacity. 
If an agent knew exactly everything that happens in the world and had unlimited computational power, perhaps it would not need emotions. A planning system would decide the best course of action---and it would really be the best course of action. However, in reality, things happen that we didn't expect. It is because the agent does not know everything about the world (limited information), and the planning system cannot compute all the possible courses of action (limited computation). This is of course a narrative running through all AI and all neuroscience, but it is worth repeating.

This chapter will show various ways in which emotions help in information-processing under such limitations. 
One implication of the limitations is that some kind of monitoring of unexpected events is needed, as well as a system for changing plans accordingly. This is the role of interrupts, which is the main theme of this chapter and one of the specific functions of emotions. Such interrupts can also trigger pre-programmed action sequences or plans that have been found useful by evolution, or the programmer, which is another aspect of how emotions help in steering an agent's behavior.

\section{Emotions interrupt ongoing processing}

Suppose you (or a robot) are walking home on a street you know. While walking, you may be planning what you will be eating tonight (the robot might be just concentrating on the walking because that's difficult enough for it). Now, a car suddenly appears and comes fast in your direction. What you need to do to survive is the following. First, your perceptual system has to detect that something unexpected and potentially dangerous is happening. Second, the fact that something potentially dangerous is happening must be broadcast to the whole system;\index{broadcasting} you have to stop thinking about what you will eat, and you have to stop following the route back home. Thus, you interrupt all ongoing activities, including your current train of thought. Instead, you have to use all your cognitive resources to figure out what to do, how to jump to safety, and when. 
\index{emotions!as interrupts}

The important new twist here is that once the sensory systems realize something suspicious is happening --- even if they don't exactly know what --- they have to send some kind of  an \textit{alarm} signal to other parts of the brain.  In particular, the system responsible for executing action plans must be interrupted; in computer science, such a signal from one process to another is typically called an interrupt. 
These functionalities go much beyond the mere ``cool'' perception that a car is visible and coming in your direction.

A separate alarming mechanism with the capacity to stop ongoing activities and reorient computation is the core of the\index{interrupt theory}
\textit{interrupt theory} of emotions originally proposed by Herbert Simon in the 1960s.\citenew{simon1967motivational,oatley1987towards} 
The key idea in this theory is that being an interrupt is what distinguishes an emotion from ordinary information-processing.  The interrupt theory explains why emotions have particularly powerful attention-grabbing properties; that is the whole point of emotions according to this theory.\footnote{The concept of attention is mainly elaborated in later chapters, but anticipating them, I need to point out that interrupts are closely related to a specific kind of attention which is \textit{bottom-up} attention, see footnote~\ref{attentionfn} in Chapter~\ref{perception.ch}.   I don't elaborate the connection between attention and interrupts here, and I use the word ``attention'' casually, in its everyday meaning.\index{attention!bottom-up}\index{attention!in interrupts}}
Such an interrupt system is particularly important since earlier in Chapter~\ref{desire.ch} we argued, following the belief-desire-intention theory, that an agent needs to commit to a single plan instead of jumping from one plan to another. Commitment is useful, but it should not be blind: interrupting a plan must be possible.\footnote{The attention-grabbing properties of emotions are well understood by the designers of social media platforms.\index{social media} The more the news and updates evoke fear or anger, the more attention the user pays to the platform. The avowed primary goal of some such platforms is engagement, which is basically one aspect of attention. Some negative side-effects of designing such systems should be well-known to everybody by now.}

\subsection{Pain,  disgust, and fear}

At the most elementary level of interrupts, we actually find simple physical
pain. Although we don't categorize it as emotion, pain is clearly a signal or a process that has such an interrupting quality. It is broadcast\index{broadcasting} to the whole
information-processing system; all ongoing behaviors are typically
suppressed, and the organism uses most of its resources to get rid of
the cause of the pain. 
Pain\index{pain!as interrupt} is, in fact, the most fundamental, as well as the strongest kind of interrupt. It has to be so, because it is the signal which is the
most relevant for the intactness and even the very survival of the organism. It is an alarm  about a physical, chemical, or biological
danger to the organism---tissue damage in the terminology of Chapter~\ref{suffering.ch}---that typically comes from outside.\footnote{In line with the interrupt theory, \citet{craig2003new} argues that pain should be seen as an ``emotion'' as it includes ``a behavioral drive with reflexive autonomic adjustments'', unlike plain sensory processing.} It requires
urgent action, such as withdrawal away from the object that caused the
pain. Reflexes like this are present even in very simple organisms,
and should be programmed even in reasonably simple robots. You
don't want an expensive robot to break down the very first day because
it doesn't understand what kind of actions are dangerous to itself.

Disgust\index{disgust} is conventionally classified as an emotion, although it is closely related to pain. Disgust is triggered by perception of substances which are likely to be toxic or transmit diseases. Again, current processing is interrupted to direct attention to that substance and how to avoid it. Disgust is often a very primitive emotion: for example, disgust at the smell of rotten food is very close to physical pain. This is natural since disgust is about protecting the organism from something not very different from tissue damage.
However, disgust has also more abstract forms as in the case of disgust at morally condemnable behaviors.\citenew{chapman2012understanding}\index{morality}\index{ethics|see{morality}}

More complex organisms are able to \textit{predict} impending\index{prediction!of danger}\index{fear!as interrupt}
danger at a much greater distance and temporal delay.
While disgust, and even pain, already have such a predictive quality in primitive form, complex organisms can predict risk of damage  before the pain or disgust systems are activated. %
 The emotion triggered by such anticipated danger is \textit{fear}, which interrupts ongoing activity and directs processing to avoidance of the dangerous object. The computation of threat (Chapter~\ref{threat.ch}) is clearly an essential part of the system that decides if the interrupt should be triggered.
 While detection of threat and fear are intimately related,  this interrupting quality of fear is one reason why fear has to be seen distinct from mere detection of threat. A threat may or may not lead to triggering of an interrupt, based on some further evaluation of its importance and urgency.

\section{Desire as an emotion and interrupt}

Interrupts can also be useful when there is no danger visible, but rather an opportunity to obtain some kind of reward. Casual observation tells us that something very similar to an interrupt happens when you see an object that you really like and want. You are assailed by an acute, ``burning'' form of desire.\index{desire!as interrupt and hot}
While in neuroscience and psychology desire is usually not considered an emotion, there has always been some doubt about whether such a distinction is justified.
Acute, burning desire actually squarely sits in the domain of emotions as far as the interrupt theory is concerned.\citenew{oatley1990semantic}

In Chapter~\ref{planning.ch}, we  defined the desire system as something that suggests goals to the planning system. But we didn't go into details on how the desire system actually works: How can it identify states which are easy to reach while having a high state-value? 
I think the whole point in the computations related to desire is that they happen as a dual process. When desires suggest goals for a planning system, they have to do it based on fast neural network computations in order to  usefully complement planning. As we saw in Chapter~\ref{dual.ch}, 
neural networks, such as those in AlphaGo,\index{AlphaGo} can be trained to output approximate solutions to the computations of state-values and similar quantities needed for planning. It is likely that the computations underlying desire are based on such neural networks, which suggest candidate states that are likely to be easily accessible while having a high state-value.

\subsection{Elaborated-intrusion theory}

\index{elaborated-intrusion theory}\index{desire!as elaborated intrusion}
A psychological theory that is very compatible with such goal-suggestions by neural networks as well as their interrupting quality, is the \textit{elaborated intrusion} theory of desires. As its name implies, it considers desire as a computational process that \textit{intrudes} your mind: it invades your information-processing system so that you lose control, at least initially. You are not able to think about anything else and keep planning courses of action regarding that object of your desire. Such ensuing compulsive planning is the \textit{elaboration} part of desire.\footnote{I follow here \citet{kavanagh2005imaginary}. A closely related model which talks about ``impulses'' instead of desires, and explicitly links them to a dual-process theory, is presented by \citet{hofmann2009impulse}. Similar ideas can be found in consumer research; \cite{belk2003fire} in particular contrast desires and what they call ``needs'' as: ``We burn and are aflame with desire; (...)
  we are tortured, tormented, and racked by desire; (...) 
  our
desire is fierce, hot, intense, passionate, incandescent, and
irresistible; (...)
Needs are anticipated, controlled, denied, postponed, prioritized, planned for, addressed, satisfied, fulfilled, and gratified through logical instrumental processes. Desires,
on the other hand, are overpowering; something we give in
to; something that takes control of us and totally dominates
our thoughts, feelings, and actions.''
}

Everybody has experienced such intrusions. You see a sexually attractive person, and you cannot think about anything else for a while. Or, you see your very favorite brand of chocolate in a supermarket, and you can hardly resist taking it in your hand and putting it into your shopping basket. You may be devising all kinds of sophisticated plans to get the object of your desire, forgetting completely what you were actually supposed to be doing.\index{vision}
Thus, at least in humans, the simple neural networks computing desire can be in conflict with deliberative planning processes. This emphasizes that desire  can take control of the mind, inexorably turning our attention towards the object of the desire.
In other words, such really ``hot'' desire, which could be called ``irrational'', can give rise to a conflict between ``reason and passion''  ---which is perhaps a poetic expression for the dual-process character of the information-processing system.\footnote{To clarify and recapitulate: We can define desire in different ways on the hot-cold axis. In the coldest definition, desire is simply a preference for some states, essentially just another way of saying that some states are rewarding or have higher state-values. You might say, for example, that you want to see Kyoto one day, but saying that does not necessarily arouse any feelings, and launch any deliberations in your brain. A slightly less cold definition says that desires propose goal states for a planning system, thus possibly launching computations to attain such a state. The definition in the elaborated-intrusion theory is quite hot, emphasizing the interruptive quality of those computations. An even hotter definition, not pursued here in detail, might further add a subjective, conscious experience of burning with desire, but this is outside of the computational modelling framework we take here, and presumably only applicable to humans and higher animals. --- To emphasize the difference between different kinds of desire, such a ``hot'',  compelling  desire is sometimes called \textit{occurrent} desire,\index{desire!definition}\index{desire!occurrent/interrupting}  while the kind of cold, long-term rational desire that simply expresses a preference is called \textit{standing} desire. I prefer to talk about ``interrupting'' desire instead of occurrent.}

\subsection{Valence}

\index{valence!definition}
Such a dual-process approach brings us close to another interesting concept: valence. In psychology, valence is a technical term describing the intrinsic positive-negative, pleasure-displeasure, or good-bad axis of states or objects. From the viewpoint of subjective human experience, valence means whether feelings are positive or negative: positive valence is associated with pleasure, negative valence with displeasure. Valence is closely related to liking: we could equate liking and valence, saying that we like things that have a positive valence and dislike things that have a negative valence. Alternatively, valence can be defined based on behavior: humans as well as animals approach and try to obtain states of positive valence, and avoid things and states of negative valence.\citenew{colombetti2005appraising} Desire is thus usually directed towards states that have positive valence.\footnote{Usually, people want things that they like, and vice versa. However, recent research has found that in some cases, people can want things which they don't like \citep{berridge2015pleasure}---in the precise sense that those things do not produce physiological pleasure reactions. 
  This phenomenon is one of the underlying mechanisms in drug addictions: An addict may want the drug and consume it without actually deriving any pleasure. In fact, such desires don't even need to be conscious in humans.\index{addiction} See also footnote~\ref{berridgefn} in Chapter~\ref{rpe.ch}.\label{berridge2fn}}   

In our framework, valence can be seen as a quick evaluation of any state or object by a neural network that computes approximations of state-values.\footnote{I shall not attempt to give a more formal definition of valence or liking here since it is not necessary for what follows.}
When you see chocolate, its high positive valence is reflected in your neural networks that predict a high state-value if you reach the state of eating it. Thus, valence computations are necessary for interrupts based on desire. In Chapter~\ref{overview.ch} we shall discuss how the sequence valence-desire-intention is essential in Buddhist philosophy: just like in the present discussion, it is valence that leads to desire, and further to intentions and frustration. \add{Likewise, negative valence leads to aversion.} In that sense, valence computation is at the root of suffering.

 \section{Emotions include hard-wired action sequences}

\index{emotions!as hard-wired programs}
 A further characteristic of emotions, and a utility of interrupts, is that they can launch ``hard-wired'' programs, or sequences of actions for specific situations. Many emotions are characterized by their specific, relatively rigid programs.\footnote{Simon's interrupt theory was elaborated by \citet{oatley1987towards} by proposing how different emotions correspond to different action sequences. Going further in that direction, we find Frijda's theory of emotions as ``action readiness'',\index{emotions!Frijda's theory} meaning the preparation for movement or action \citep{frijda2016evolutionary}. Frijda's theory sees this as the main distinguishing feature of emotions, instead of their interrupting character. But it could be argued that any simple neural-network-based reinforcement learning system can trigger such action readiness, and it is difficult to see what would be special about emotions if they were defined as simply action readiness.} 
The action sequence is, in fact, the aspect that most visibly distinguishes which emotion is taking place. In the case of fear, the typical action is to choose either freezing or fleeing. Disgust leads to immediate
rejection and avoidance of the substance triggering the emotion. 
In animals, such programs are evolutionarily quite old: humans have largely the same action programs as dogs.\citenew{gross2012many}  

The point is that some simple action sequences are particularly useful and universal, so it is a good idea to have them readily stored in the system so they can be executed quickly, without any need for elaboration. This is in stark contrast to the main processing being interrupted, which is often a result of long elaboration. In fact, since plans may take quite a while to formulate, they are less useful in an emergency situation. Furthermore, it is important to have the emotion-specific action sequences readily programmed in the system---meaning genetically transmitted in humans---since they can be very difficult to learn. For example, anything related to self-preservation is difficult to learn by reinforcement learning, since when the agent realizes that the current situation is lethal, it is too late. 

Anger\index{anger} is another fundamental example of an emotion that clearly has its own hard-wired action sequence. It also has a particularly strong social quality: real anger in the sense of an interrupt is usually associated with other people. While you might say that you are angry about bad weather, that is not much more than ordinary frustration. We shall not consider anger in any detail here because such social aspects are completely beyond the scope of this book and would require more complicated theory, in particular game theory. Let me just mention the basic idea, which is that anger is a special hard-wired action sequence that protects the agent from attacks  by creating a credible threat of a robust retaliation that would inevitably be triggered in case of being attacked.\footnote{\add{A simple model of anger in our framework is that it is a reaction that comes on top of frustration when another agent is causing, at least partly, the frustration. But this does not yet explain the evolutionary meaning of anger, and in particular why angry people can behave extremely ``irrationally'' in the sense of causing great damage to themselves. A well-known evolutionary explanation is as follows. }
Imagine a gangster comes to you and asks you to give him all your money. The rational thing to do would be to give the money. This is rational in the sense that otherwise, he might kill you or inflict some bodily harm, and certainly it is better for you to just give the money. 
However, this behavior has the downside that then the gangster can come to you any time he wishes and always take your money. 
The evolutionary explanation of anger is that it is a program that makes you behave irrationally. In this case, you would just get ``mad'', and physically attack the gangster, even if you know he will kill you as a consequence. Surprisingly, having such a program may be good from an evolutionary viewpoint, because if the gangster knows you have such a program installed, he might decide not to bother you. It is not good, from an evolutionary viewpoint, to actually attack the gangster; what is good here evolutionarily is having such a program installed, and signalling this to the gangster. If the gangster knows about the program, it may never be actually used, because it works as a powerful deterrent.  This is a well-known game-theoretic model\index{game theory} in evolutionary theory, it was originally used for modelling the behavior of animals who fight over mating opportunities, territory, or other scarce resources \citep{smith1973logic,pinker1999mind}, which is why it is often called the hawk-dove game \citep{hirshleifer1987emotions,nowak2016evolutionary}.  It is actually equivalent to another game-theoretical model called the game of chicken, which, despite sharing an avian name, has a very different story and motivation behind it. --- Let me also note that social interaction creates many further emotions, such as shame and guilt, some of which are moral emotions \citep{haidt2003moral}; that is, they enforce behavior conforming to ethical norms.\index{morality}\index{social interaction!emotions, moral}\index{emotions!moral}}

It is now useful to contrast emotions to habits, in the wide sense used in Chapter~\ref{rpe.ch}. Habits are often triggered by some environmental stimuli---a bit like interrupts---and lead to a fixed kind of behavior---a bit like the rigid action sequences we just mentioned. In these two ways, habits have some similarities to emotions. However, habits are not really interrupts. Perhaps when you walk on the street you have the habit of humming a tune to yourself. However, it rarely happens that you stop whatever you're doing because you suddenly feel an irresistible urge to start humming. Habits don't have the power to capture your attention and  interrupt current plans.\index{habits!vs emotions}

\section{How interrupts increase suffering}

We have seen that a number of phenomena, which are often considered separate in psychology and neuroscience, share the important characteristic of being interrupts. Pain, emotions, and desire can all be seen in this computational framework.\footnote{ While considering pain, basic emotions, and desires in a
  single framework is not usual in neuroscience, in
  recent neuroscience literature, ideas similar to the interrupt
  theory use the distinction between a planning system (``Model-based
  reinforcement learning'') and a fast system with automated reactions
  (``Model-free reinforcement learning''). If an agent has such two systems, a crucial question is how to divide tasks between the two
  systems, i.e.\ which one to use to respond to any particular
  situation \citep{daw2005uncertainty}.  If fast action is required,
  you obviously need to use the fast system, and if there is no hurry,
  you can spend some time in planning; choosing which system to use is a
  complicated problem.. The process leading to the decision to use the
  fast system is then not very different from the mechanisms
  postulated in the interrupt theory; it has been connected to emotions by \citet{bach2017algorithms}.}  %
But many emotions include a lot of suffering. If emotions were just interrupts, why would there be so much suffering involved?

\add{One obvious reason is that some emotions are closely related to forms of suffering we have treated in earlier chapters: fear is triggered by a threat, while anger is a form of frustration, for example.
  However, this may not be the whole story.}
 It might be the case that interrupts create suffering directly, by themselves:
the interrupting system is likely to use something like the pain signalling system.\label{interruptsuffering} In fact, most emotions discussed here are negative, they hurt, and this suggests they must use the pain system, like suffering (``mental pain'') in general.\citenew{papini2015behavioral,eisenberger2004rejection} 
Making the body feel pain is an evolutionarily primitive way of grabbing the attention of the whole cognitive system, as was discussed in Chapter~\ref{suffering.ch}. Interrupts need, by their very definition, to achieve such an attention-grabbing effect, so using the pain system is even more natural than in the case of, say, frustration.%
\footnote{Thus, such negative emotions with interrupts might be introducing a mechanism for suffering which is a bit different from what has been discussed previously. On the other hand, it could be argued that the suffering due to fear is simply the suffering from a threat, and triggering an interrupt does not necessarily amplify it. Likewise, it could be argued that the suffering in anger is just a special kind of frustration. I do not take a definite stance on this point.
}
(Positive emotions are a rather different story, and not considered here.\footnote{Some positive emotions, or rather attitudes, are considered in Chapters~\ref{attitude.ch} (acceptance, letting go, contentment) and Chapter~\ref{epilogue.ch} (compassion, loving-kindness). Let me just mention that \citet{fredrickson2001role} proposes positive emotions serve the role of enabling exploration of different action possibilities when the circumstances are safe and not even remotely life-threatening, thus leading to enhanced creativity and learning.\index{creativity!positive emotions}\index{emotions!positive}}) 
However, I do not consider that possibility further here; in the following, I focus on how interrupts \textit{increase} suffering coming from frustration and threats.

\subsection{Interrupts reduce control}

One way in which interrupts increase suffering is that they reduce control, which is one of the main themes in the following chapters, especially Chapters~\ref{replay.ch} and \ref{control.ch}.\index{uncontrollability}
A crucial part of the interrupt theory is the idea that the interrupts are automatic and largely irresistible. For example, many people would be so much happier if they could just consciously decide to switch off their fear system.\footnote{\add{This thought experiment is actually a bit complicated since even if the fear system as an interrupt is switched off, the threat system might still be operating, thus creating suffering. Arguably, though, the threat signal would be amplified since the interrupt focuses the attention on the threat; without the interrupt-alarm system, the threat might not be detected at all.}\index{attention!in interrupts}} But the point is that interrupts are outside of conscious control; they have to be so, because very often they need to interrupt conscious thinking and consciously controlled action. If you could somehow weaken interrupts so that they don't disturb you, they would be useless: it would be like switching off a fire alarm system because it is too loud. In a scary situation, fear will appear together with its inherent suffering, no matter how much you try to control it. We have already seen in Chapter~\ref{dual.ch} (page~\pageref{fearconflict}) how the dual-system structure of the brain means that the fast, unconscious fear system usually prevails over any conscious deliberation.  
The same happens with desires: The fast computations of valence and values by neural networks will ``intrude'' and interrupt other processing, directing all the processing towards the object of the desire. 
Such interrupts are even more annoying if they interrupt activity that would have created pleasure, for example, when you are in a ``flow'', fully engaged in a rewarding and meaningful activity.\index{flow!and interrupts}
In this sense, interrupts  \textit{increase} suffering: by increasing desires, aversion, planning, and frustration. 

Such reduction of control might not be a bad thing if the interrupts were somehow optimally tuned to reduce suffering.  However, another problem with the interrupt system is that its design parameters are often questionable from the viewpoint of suffering.  To begin with, the system that triggers interrupts does not care about our subjective suffering, only about our evolutionary fitness. Evolution makes us consider harmless things as dangerous, worth triggering an interrupt, if they are threats to our evolutionary success. Sexual jealousy and the ensuing rage is one example, where (from a male perspective) the evolutionary ``danger'' is that one might end up raising a child who is not one's own and does not spread one's genes. Yet, that is hardly a problem from a contemporary viewpoint: it is actually very common in modern families.

What's more, the system may not actually be very good at maximizing evolutionary fitness either. As we saw earlier, evolutionarily developed neural mechanisms may not be well adjusted to our current  society, since they may come from the legendary ``African savannah''.\index{evolution!African savannah} In the case of fear, for example, we tend to be  afraid of snakes or spiders, but not so much of cars, although cars are much more dangerous at least in modern cities.\index{fear!evolution} 

Another problem with the interrupt system is that if the interrupts
are excessive and disrupt the normal function of the system too much,
it may simply worsen the situation by making it more difficult to
respond to the situation. Such problems are related to the fact that emotions and desires are short-sighted---as has been acknowledged by philosophers since antiquity---and may interrupt useful plans in a way that produces frustration because the interrupts fail to understand the long-term utility of following the plan. For example, an important function of pain is to attract
the attention of the agent to the source of the pain; but if the
person can think of nothing else than the pain, as often happens in
the case of overwhelming fear or depression, he will not be able to find a solution
to the situation. 
Or, if you are easily scared and are constantly interrupted by, say, harmless bugs, your performance in a meaningful pursuit may be hampered even though there was never any real danger to avoid.

\subsection{Alarm systems cannot be universally optimal}

\index{signal detection theory}\index{alarms}
These questions are related to the general theory of designing alarm systems, which is considered in the mathematical theory called signal detection theory.\footnote{\citep{green1988signal}. This is a special case of statistical decision theory, typically considering the case of two possible options given by ``present'' or ``absent'' (regarding a threat), and focusing on the question of finding the right balance between false positives and false negatives.}  It is based on maximizing the expected payoffs, where payoffs are similar to rewards, describing how good (positive) or bad (negative) the results of a given action are. For an alarm system such as interrupts, there are two possible actions: trigger an alarm, or do not. The theory is related to the AI theory outlined in previous chapters but with a different emphasis. 
An important lesson in this theory is that \textit{there is no such thing as a universally optimal alarm system}.
That is because the payoffs are different for different people, and different in different contexts, and may change over time.\index{individual differences}

Consider designing a burglar alarm system. You might start by assigning a high payoff to detecting burglars; this sounds reasonable and innocuous. However, this means the system will not mind making false alarms, since you only give a strong payoff (reward) for the detection of a burglar, but you do not give any punishment for false alarms. To maximize reward, the system rationally decides to trigger an alarm if there is any hint of a burglar present.  Eventually, the system will constantly wake up everybody in the middle of the night. Realizing your mistake, you change the design by adding a really high reward for \textit{not} giving false alarms. The result is that the system never gives any alarm because that's the perfect way to avoid false alarms, which are now strongly punished. In this case, the alarm system ends up being completely useless since it does not do anything.
It is very difficult to say what the right compromise is: the alarm system should be sensitive but not too sensitive, and the right parameters  are quite subjective and depend on the context.
Evolution has programmed certain sensitivity levels in our interrupt system, but in light of this signal detection theory, it is not actually clear how optimal they were even for all our ancestors on the African savannah, let alone for modern city-dwellers.\footnote{
The sensitivity levels, or thresholds, can also be modified by experience to some extent. For example, if as a child, you saw something that made you really scared, you may lower the threshold a lot---commonly known as a phobia.\index{phobia} This leads to another problem, widely recognized in clinical psychology: the payoffs change during an individual's lifetime as well. In adults, they may be very different from what they were in our childhood environment, while any learning of the payoffs may mainly happen as a child. The best survival strategy for a child in an adverse environment may be to be constantly afraid of other people; this may not be optimal anymore when the child grows up, and is in fact a possible source of psychiatric problems. 
}

\section{Emotions are boundedly rational}

Often, emotions are contrasted with rationality and ``cool-headed'' decision-making; it is typically assumed that the best decisions are made when emotions are not at play. \add{ The word ``rationality'' comes from the Latin word \textit{ratio}, meaning ``reason'', and Western philosophy has traditionally considered reason and emotions (or ``passions'') as two opposing forces.}

  However, the viewpoints on emotions explained in this chapter show that emotions contribute to optimal decision-making and action selection. Emotions are useful from a rational viewpoint as soon as there are certain information-processing constraints; for example, if the planning system does not have time to consider all possible paths in the search tree. This is certainly true in any sufficiently complex AI system or animal. The viewpoint which considers emotions as necessarily irrational is in fact largely rejected in modern research.\citenew{damasio1994descartes,scherer2011rationality}

I have casually used the word ``rational'' here as well as in  earlier chapters, but we need to think a bit more about what it actually means.\index{rationality} 
\add{In mathematical decision theory,} a decision is called rational if it is optimal in maximizing reward (or a similar quantity) given the information available to the agent. In other words, the decision of the agent, such as choosing an action, is the same as that made by an ideal, hypothetical agent with perfect information-processing capacities and the same information about the world as the agent in question. So, even a perfectly rational agent is not expected, in this definition, to make the very best possible decision, but the best possible given the limited information it has at its disposal.

However, in reality, the information-processing power of the agent is limited as well, as we have indeed seen in many chapters of this book.\footnote{Curiously, largely due to  historical reasons, limitations in information available to the agent were always admitted, but computation was not supposed to be an issue in earlier work on rationality. I'm here referring to the classic work on statistical and economic decision theory in the first half of the 20th century, arguably culminating in the work by \citet{neumann1944theory}.}  
The case where information-processing power is also limited leads to the concept of \textit{bounded rationality}, also called computational rationality. It refers to decisions that are optimal given limitations in both the information \textit{and} the computation available to the agent.\citenew{simon1972theories,russell1997rationality,gershman2015computational,lieder2020resource}

Emotions, seen as interrupts or as automated action sequences, can be considered to strive towards \linebreak bounded  rationality.
Emotions are information-processing routines or shortcuts which help in achieving as good outcomes as possible, given the computational resources and the limited information available. 
It is in this precise sense that we can say that emotions help in rational decision-making, and it is not justified to oppose rationality and emotions.\footnote{Another rather different information-processing function of emotions has been proposed as the ``somatic marker hypothesis'' by Antonio Damasio \citep{damasio1994descartes,bechara2005somatic}.\index{emotions!somatic markers}\index{somatic marker hypothesis}\index{heuristics!somatic markers}\index{body} Somatic markers are defined as bodily responses to situations, learned from past experiences. If a certain situation has led to a bad outcome (e.g.\ a strong negative reward), you learn to associate such a situation with a bad ``gut'' feeling in your body. The somatic marker hypothesis thus shows how such feelings (here considered the essential part of emotions) can be used to improve planning by using them as heuristics. As we saw earlier, the central problem in action selection is the huge, exponential number of plans to consider. Using somatic markers as heuristics, you may be able to reject many of them based on such negative feelings and focus your search on the set of plans associated with positive gut feelings. Importantly, such ``gut feelings'' are generated by a very fast computation in a simple feed-forward neural network, thus speeding up decision-making and planning---not unlike the computations we linked to desire, valence, and dual-process action selection earlier in this chapter and Chapter~\ref{dual.ch}. \add{Computationally, such somatic markers would be a bit like a rat searching for cheese by maximizing the smell, i.e.\ using the strength of the smell as a heuristic,  as in the example in Chapter~\ref{planning.ch}.\label{somaticmarkersfn} However, such a heuristic can of course be misleading: something that gives a bad gut feeling may actually turn out to be good when you think about it a bit more.}}

Yet, emotions also have qualities that are in contrast to our everyday notion of rationality. \add{Emotions can be based on very limited information, such as one scary object, and they are often short-sighted, neglecting long-term consequences. Emotional interrupts are, by definition, the very opposite of commitment to a goal, possibly leading to inconsistent and impulsive behavior.  In contrast, rational decisions---in the common-sense meaning of the word---look at long-term consequences and use many different sources of information, including past experiences and information shared by other people.  Furthermore, emotions and the hard-wired action sequences they trigger are not entirely under conscious control, while consciously controlled deliberation is an important part of what is classically called reason. In these two ways, emotions are analogous to the neural networks in dual-process theories, as discussed in Chapter~\ref{dual.ch}, and reason is more similar to the explicit, conscious, GOFAI-like system. As we have seen, those two systems are sometimes opposed to each other but often work together; this might be a good characterization of the relationship between reason and emotions as well.

}

\chapter{Thoughts wandering by default} \label{replay.ch}

The moment you lie down on a sofa to relax, your head starts
developing different fantasies and daydreams, perhaps wondering why you did
such a stupid thing yesterday, or planning what you want to eat
tonight. Even when you try to meditate and not think about anything
(which is a typical instruction for beginning meditators), you will
almost inevitably find yourself thinking about something else after a
while. 
There is a good reason why the human mind is often compared to a monkey in meditation traditions.\index{meditation!wandering thoughts} It jumps here and there, making all kinds of noises, and never seems to rest. Likewise, based on his own method of introspection, David Hume concluded: ``One thought chases another, and draws after it a third, by which it is expelled in its turn.''\footnote{\hume, Section 1.4.6}\index{Hume}

Thoughts that come to your mind when you are trying to concentrate on something else are called  ``wandering thoughts''. They have some similarities with emotional interrupts: they stop ongoing mental activity and capture attention. Thus, they reduce the control you have over your mind and, eventually, increase suffering. However, the computational underpinnings are quite different in the two cases. In this chapter, I discuss how wandering thoughts are related to the need to repeat experiences for the purposes of iterative learning algorithms, as well as planning the future through search in a tree. Thus, there is an evolutionary reason why we have wandering thoughts: they are not just pointless activity triggered by mistake.

\section{Wandering thoughts and the default-mode network} \label{wt.sec}

Wandering thoughts\index{mind wandering|see{wandering thoughts}}\index{wandering thoughts} tend to appear whenever a person
tries to focus on a single task or object for a long time.
Everybody has encountered a situation, perhaps at
school or at work, where she tries to concentrate on
something but soon finds herself thinking about what she should
say in a job interview tomorrow, or what she did on a previous vacation. Typical tasks where such \textit{sustained attention}\index{attention!sustained} is necessary, but difficult to achieve, are driving a car on a highway, trying to read a book for an exam, or monitoring a screen as in air traffic control or surveillance. Importantly to the theme of this book, sustained attention is essential in most meditation practices.   
If you are lying on a sofa and have nothing else to do, meandering thoughts about various things are fine and sometimes even enjoyable.
However, when you are actually trying to concentrate on a task, unwanted wandering thoughts reduce your performance of the task at
hand.\footnote{In the case where you have no particular task to perform, the spontaneously appearing thoughts may not be properly called ``wandering'', but, for example, ``spontaneous''. Some authors strictly reserve the word ``wandering'' for the case where the thoughts are intrusive in the sense of occurring against your will while you are trying to concentrate on some task, such as thinking about some unrelated event tomorrow when trying to concentrate on reading a textbook. In this book, I use the term wandering thoughts a bit more liberally, sometimes including thinking that jumps from one topic to another when there is no particular task on which it is supposed to concentrate---as in lying on the sofa after work---since in real life, it is often difficult to draw the line between wandering and other spontaneous thinking.\label{spontaneousdef.fn}}

Psychological experiments confirm the ubiquity of wandering thoughts. Various experiments can be devised where the participant's task is to monitor a stream of information and report when a rare prespecified event occurs. In a typical experiment, you would be shown random digits (0 to 9), and you have to press a button when you see a target digit, say, 3. The experiment is deliberately designed to be boring\index{boredom} so that the participant's mind will certainly start wandering at times. The basic idea of monitoring for an event that is rare is reminiscent of some of the typical real-life tasks listed above (e.g.\ driving a car on a quiet highway), where nothing much happens most of the time, and sustained attention is difficult. 
The experimenters would then use a method called experience sampling,\index{experience sampling} which means they ask, at random intervals, whether the participant was focused on the task or whether they had wandering thoughts. It typically is found that the participant's performance on the task fluctuates between better and worse; this fluctuation largely reflects whether they had wandering thoughts at that particular time point or not.\citenew{christoff2009experience} 

Such experiments can be conducted even when the participants are living ordinary everyday lives. The participants would have a device, such as a mobile phone, which asks at random intervals whether they were focused on whatever task you were performing (such as working, studying, cleaning, driving, etc.) or whether they had wandering thoughts (such as daydreaming, fantasies). It is typically found that during everyday life, the mind is wandering quite a lot: one third, or perhaps even one half of the time.\citenew{kane2007whom,killingsworth2010wandering}

\subsection{Much of brain activity is spontaneous}

At the same time, modern neuroimaging confirms the prominence of various kinds of spontaneous brain activity, i.e.\ activity that ``just happens'' without any external stimulation or task being performed. 
In fact, an amazing finding in recent neuroscience is that if you measure human brain activity when the participants of the experiment are simply told to sit or lie still and think about nothing in particular, their brains are far from quiet. Technically, neuroscientists talk about ``resting-state''\index{resting-state} to characterize such a state of not doing anything in particular, since the participant may think she is having a rest---but the brain is definitely not.\footnote{In terms of neural network theory, resting-state activity is enabled by the brain having \textit{intrinsic dynamics}\index{dynamics!intrinsic} based on \textit{recurrent} connections. Recurrent connections mean that the neurons are not arranged in successive layers where the signal just goes in one direction: Instead, the outputs of some neurons are fed back to other neurons that actually provided input to those neurons in question. The output can also be fed back to the outputting neuron itself. With such feedback, neurons can learn to sustain each other's activity: Neuron A activates neuron B, which by recurrency again activates neuron A, and so on. Even a single neuron can sustain its activity by sending feedback activation to itself  \citep{hopfield1982neural,hochreiter1997long}. Such recurrent connections are extremely common in the brain, while the most commonly used neural network models have no recurrent connections; this is an important discrepancy.  
}

A particular network in the brain is actually even more active during rest than during active tasks. It is called the \textit{default-mode network}\index{default-mode network} because it seems to be activated ``by default'', i.e.\ when there is no particular reason for anything else to be activated.\footnote{For recent reviews, see \citet{buckner2008brain,raichle2015brain}, for the original articles, \citet{shulman1997common,raichle2001default}.} It is also deactivated once the person is stimulated, for example, by sights or sounds from the external world, so that the brain actively starts processing incoming information. 

The discovery of the default-mode network around the year 2000 was something of a revolution in human neuroscience. It was completely at odds with the classical way neuroscience experiments were done: the experimenter would instruct an experimental subject to observe some stimuli (e.g.\ a sequence of digits as we saw above) and possibly perform a task at the same time (e.g.\ press a button when a target digit appears). Here, in contrast, you don't tell the participants to do anything and don't give them any sensory stimulation, such as showing pictures. 
Then, it is the default-mode network that becomes activated. 
In fact, since it is deactivated (i.e., silenced) by sensory stimulation and tasks, the experimenters had better not give any stimuli or tasks to be able to observe it.\footnote{Actually, it has further been found that many of the same brain networks that are intermittently active in various neuroscience experiments are also intermittently active in resting-state.  Thus, the default-mode network  is not the only network activated in resting-state, but the default-mode network is perhaps the only one that is \textit{more} active in resting-state than in \textit{any} kind of stimulation or task.  When we talk about a ``network'' here, we mean more precisely that the activities measured in certain voxels (i.e., 3D pixels) in the imaged brain activity seem to be fluctuating synchronously \citep{damoiseaux2006consistent,fox2007spontaneous}. Typically the analysis is made using a machine learning method called independent component analysis \citep{hyvarinen2001book}.\index{independent component analysis}
Possibly the first study to provide such a decomposition to several networks was in fact based on data from  anaesthetized human children, whose brains were scanned for clinical purposes \citep{kiviniemi2003independent} .}

It is widely assumed  that the default-mode network supports wandering thoughts.\citenew{christoff2009experience,andrews2012brain} That would explain why it is particularly activated when the subjects do not receive any stimulation and have no particular task: then, the mind will easily start wandering. It is likely that the default-mode network has other functions as well, although we don't know very well what they might be.\citenew{raichle2015brain}

\section{Wandering thoughts as replay and planning}

The existence of wandering thoughts may feel completely normal to us, but actually, it is rather surprising that the whole phenomenon exists. Why should it be difficult to concentrate on one thing for a long time? Why cannot I just decide to focus on reading a textbook for an exam, say for two hours, without any interruption by any unrelated thinking?\citenew{van2015modeling}

One intuitively appealing explanation would be that your active neurons---in the exam-reading example, those needed for reading---get ``tired'',
i.e., somehow run out of energy. Then, other neurons which are full of energy will be able to somehow steal the attention. While there may be some truth in such
an explanation, it is not very compelling because sometimes you \textit{can}
concentrate without any problems on a task, especially on a task which is really engaging, such as reading a book you really like (not for an exam), or playing a video game. Furthermore, should not such fatigue of neurons rather lead to having no thoughts at all? It is more plausible that wandering thoughts are actually doing some useful computation---and that they are something that you would like to program in an AI.

So, let us think about what kind of computational problems could be solved by wandering thoughts. One problem we have seen earlier is that learning typically needs many repetitions of the inputs and the desired outputs since it is based on iterative algorithms, as we saw in Chapter~\ref{ml.ch}. Even the very same inputs and outputs may need to be presented many times to the learning algorithm. This is why modern AI systems need a lot of computing capacity for learning. 
At the same time, planning takes a lot of time as well, as we saw in Chapter~\ref{planning.ch}.

So, as much of the computing capacity as possible should be directed to these learning and planning activities. 
In particular, when the agent does not receive any special stimulation from outside, there is nothing important for it to do, and no urgent threats are detected, the computing capacity of the agent is free to be used for any internal processing based on previously acquired data---intrinsic activity, in the terminology of neuroscience. In fact, in order not to waste that computational capacity, computations related to learning and planning \textit{should} be launched. 
That will enable the agent to act more intelligently when the time to act comes.
This is also presumably why evolution has programmed wandering thoughts in us.\footnote{Humans also sleep and dream: It is possible that the function of dreams is pretty much the same as that of wandering thoughts \citep{fox2013dreaming}.\index{dreaming}}

Next, I consider in detail two different ways in which wandering thoughts can help in computation. In the first one, the system  \textit{plans future actions}\index{thinking!about the future} actions by internally simulating the world, and trying out different, new actions to see which works best.\footnote{\citep{baird2011back}. Chapter~\ref{planning.ch} already reviewed the basic theory of planning.}   The second one is called \textit{experience replay}\index{thinking!about the past} because the system internally repeats memories of past behaviors and events exactly as they were perceived, in order to enable an iterative algorithm to learn efficiently.\footnote{\citep{lin1991programming,wittkuhn2021replay}} 
In fact, a lot of what people simply call ``thinking'' falls into these two categories: You plan what to do in the future, and recall what happened to you in the past.\footnote{The computations in planning are actually not that different from the computations in reinforcement learning, as was already discussed in Chapter~\ref{rpe.ch}. 
  See also footnote \ref{dynafootnote} below on how the two computations could be combined, as well as a framework proposing something between these two kinds of action selection by \citet{lengyel2008hippocampal}.}

\subsection{Planning the future}

It is perhaps obvious why thinking about future actions is useful, as far as it is a case of planning. You can go through different kinds of plans and \textit{simulate},\index{simulation} using your model of the world, what the results of your actions will be, and finally, choose the best one. If you think about a job interview that will take place tomorrow, you polish your answers beforehand by simulating what kind of impressions different options will make, eventually memorizing the best ones. 
Often, such thinking and planning\index{planning} is actually completely voluntary. If you really want to spend some time and energy to elaborate the best course of action, this is quite normal planning activity. When we talk about wandering thoughts, we mean a case where you consciously try to do something other than planning, but unrelated planning thoughts nevertheless appear. It is the unwanted, intrusive quality of wandering thoughts that distinguishes them from ordinary thinking.\footnote{In humans, planning can thus be done in various modes which differ in their relation to conscious control: there is controlled, consciously initated planning on the one hand, and spontaneous, unconsciously initiated planning on the other hand. The spontaneous planning can further be divided to unwanted/wandering  thoughts and simply spontaneous thoughts which are not unwanted (perhaps because you are lying on the sofa), see footnote~\ref{spontaneousdef.fn}.}

You might actually want to relax and read a novel, but thoughts simulating the job interview just pop into your mind. This is understandable since as I just argued, it is especially during moments where you or the AI have nothing pressing to do that it would be a good idea (from the viewpoint of the designer of the system) to use the computing capacity for such planning. 
As we saw in Chapter~\ref{planning.ch}, planning paths grow exponentially as a function of time, so there is a real need for using a lot of computation for planning.

The planning during wandering thoughts is a bit special in that it sometimes has no particular goal. It may be just looking at possible future paths in a big search tree to see what could be done to obtain rewards: a kind of ongoing, free-style planning. Such a search could actually be done by the Monte Carlo Tree Search algorithms (discussed in Chapter~\ref{dual.ch}):\index{tree search!Monte Carlo} they are randomly searching for plans, while focusing more on branches which seem to be more rewarding. It's a bit like thinking about what to do during the weekend when you're supposed to be concentrating on your work.
Nevertheless, wandering thoughts often do focus on planning for a specific goal, as in the job interview example above.%
\footnote{In game-playing AI, planning by simulation has been used in an extreme way in terms of ``self-play''. A much-publicized example is AlphaGo,\index{AlphaGo} the system that first beat humans in the board game of Go, which we used as an example of dual processes earlier \citep{silver2016mastering}. After being input information on a huge number of actual games played by humans, it started playing against itself. This is a very special kind of planning, where you are simulating your opponent as part of the environment. Actually, there is no distinction between the agent itself and the opponent since the same program plays both of them, and learns from the successes and failures of both of them.
A later version of the AlphaGo system actually omits the learning from human games altogether and learns entirely by playing against itself; the ensuing system is aptly named AlphaGo Zero \citep{silver2017mastering}. Pure self-play has also allowed for an AI to rapidly approach human level in a highly complicated, multi-player esports video game called Dota~2:\index{games!Dota2} Using more than 100,000 processors running self-play in parallel, the OpenAI Five system can simulate in one day the same amount of data that would take more than a hundred years to collect in ordinary play against humans \citep{OpenAI_Dota}. Self-play was also used to achieve super-human performance in the game of poker \citep{brown2018superhuman}. Actually, such learning by self-play was successfully used earlier in simpler games such as backgammon \citep{tesauro1995temporal} and, even back in 1959, in checkers \citep{samuel1959checkers},\index{games!backgammon}\index{games!checkers} in one of the earliest projects on learning by machines. While some human knowledge was input to the learning process in most of the preceding studies, \citet{tesauro1995temporal} also reported a variant with pure self-play similar to AlphaGo Zero.\index{AlphaGo}  Something akin to self-play is actually used by humans when they are simulating social encounters in their own minds: We might use the same model for the actions of other people and the actions of ourselves, and learn both simultaneously. (Yet, the connection between our model of our own mind and our model of the minds of others is complex, see \citet{carruthers2009we} for different possibilities.)}

\subsection{Experience replay for learning value functions}

In contrast to planning, it may be more difficult to understand why any system would like to simply repeat past experiences. You already saw what happened yesterday, so why repeat it in your mind, and why so many times? The reason is in the structure of the algorithms used in learning. 

As we saw in Chapter~\ref{ml.ch}, modern AI systems are based on learning from the data by using iterative algorithms. We saw the general idea of stochastic gradient methods: the data points (e.g.\ images) are presented to the system one by one, and a huge number of repetitions is needed. 
Many reinforcement algorithms are not, strictly speaking, stochastic gradient descent methods, but are closely related and share those properties.
They proceed by observing the state of the world both before and after each action, as well as any reward obtained or punishment received. There are thus four pieces of information in what we might call a single ``data point'': the state before the action, the state after the action, the action taken, and the reinforcement. Based on these, the system updates the state-value function. 

What is crucial here is that, again, learning proceeds by making tiny modifications to the parameters of the system, in this case those computing the state-value function. Successful learning therefore requires a huge number of iterations, or presentations of such actions and their consequences to the learning system.\index{learning!iterative}
If you have access to really large amounts of data, you may just present each data point once, and learning will be successful since the algorithm will have enough iterations anyway. 
However, the amount of data is typically limited. In the case of reinforcement learning,\index{reinforcement learning}
what is particularly problematic is that the agent may need to act in a real environment and observe the consequences of its actions to gather data. One action by a real robot can take a second or so, which is extremely slow compared to the processing speed of most computers and the potential speed of learning. Likewise, humans do not collect new experiences on, say, job interviews, that often.\footnote{In fact, calculations of the amount of data that humans observe in real life show that the number of data points needed in AI is often much larger than what humans need. For example, children seem to learn to speak from a relatively small number of ``input'' words \citep{dupoux2018cognitive,warstadt2022artificial}; current AI systems need orders of magnitude more. Perhaps even more strikingly, humans can learn from a single example they see or hear, as I have pointed out earlier, see e.g.\ \citet{lake2015human}. This shows that there is a lot of room for improvement in AI compared to the brain in terms of efficiently using all the data.}

This is where experience replay is useful.\index{replay!definition}\index{experience replay|see{replay}} It means that in reinforcement learning, the system is not just using the data related to the most recent action and then throwing it away; instead, it stores the data, and re-uses past actions and the states associated with them many times. That is, it ``replays'' or recalls past actions and events and uses them in the iterative learning algorithm as if they happened now. This improves the performance of the learning algorithm by enabling it to make many more iterations with each data point, and thus many more iterations for the same limited amount of data. This is how more information is extracted from the data. This is particularly powerful since in most cases, the agent can retrieve past events from memory much faster than it would actually act in the real world, and thus replay makes learning much faster. There are other reasons as well, as we will see below. 

Obviously, there is a trade-off here: If you just use all your time replaying old events from memory, you will not get new data about reinforcement resulting from actions. So, you cannot use all your time for just replay. It should be smart to engage in replay when the environment does not enable too many meaningful actions --- in plain English, when nothing interesting is happening and the agent is ``bored'', which points directly at wandering thoughts.

It is also possible to do something between pure replay and planning.\index{planning!combined with replay} You can replay past events while trying out different actions in a simulation. This means the system starts by recalling something that happened earlier, but then it simulates what would have happened if it had acted differently. Certainly, we all have experienced such wandering thoughts: ``If, yesterday, in that situation, I had done X instead of Y...'' This is even better than just replaying actual past events since the system is then creating \textit{new} data using past events together with its model of the world.\footnote{\citep{Sutton91,suttonbook}. See also \cite{mattar2018prioritized} for a theoretical unification of replay and planning, \add{and \citet{kurth2023replay} for neuroscience results and theory on how replay is more than just repeating past episodes.  \add{For a popular-science account focusing on the utility of regret, see \citet{pink2022power}.}}\label{dynafootnote}}

\section{Replay and planning focus on reinforcing events}

Any replay method must choose \textit{which} events, or short ``episodes'' of events, it will replay. A system that has gathered a lot of data on past actions cannot just indiscriminately replay everything if it wants to learn really efficiently. Likewise for planning: if the system starts planning in its idle time, it needs to choose the starting state for its plan---what kind of a situation your fantasy starts in---and perhaps a goal as well.

A dominant idea in AI is that replay should prioritize events where any kind of reinforcement signal was obtained, whether positive or negative, and this seems to be the case in the brain as well. Experienced episodes containing such events are the most important in computing the state-value function. This may help explain why we have so many wandering thoughts about negative events. When you do something embarrassing, it may replay in your mind many, many times. This should be useful so that you learn to associate the negative reinforcement (social embarrassment) with the actions you took in that particular situation, thus improving your estimate of the state-value function---and future behavior. %

It has been found that replay of past events can be particularly useful if the experience is replayed backwards, starting from reinforcing events.
Suppose a robot gets a particularly nice reward (say, a lot of energy in its batteries) whenever it finds itself in room \#42 of a building where it cleans the floors.\index{robot!cleaning} Based on this experience alone, it will immediately assign a large state-value to room \#42. But in order to find room \#42 in the future, it has to code its location with respect to the other rooms in the state-value function. This is easy to do if it replays its path to room \#42 in reverse order. Suppose just before arriving in room \#42 it was in room \#13, and before that, room \#21. It replays the sequence in reverse: \#42, \#13, \#21. Now, it will assign large but slightly decreasing state-values to each of these rooms, so that the state-value is decreasing the further the reverse replay goes---the decreases are justified by the theory of discounting. The end result is that while \#42 has the largest state-value, \#13 has a rather large one as well, and \#21 is not far behind. Now, if the agent ever finds itself again next to room \#21, it knows that to find a state with a large state-value, it should enter room \#21, and there it will understand the best choice for the next state is \#13, and eventually \#42. (It may sound like all this could be learned by a single replay, but in reality it must happen by smaller increments to properly combine information from many different paths and data points.)  Combining  such backward replay with the above-mentioned prioritization of reinforcing events leads to a method called ``prioritized sweeping''.\index{prioritized sweeping}\footnote{\citep{moore1993prioritized,schaul2015prioritized,singer2009rewarded} More precisely, the prioritization mechanism replays memories of individual states (and actions taken in them) whose replay leads to maximal change in the estimated state-value function. This is not exactly the same as replaying episodes where a strong reinforcement occurred, as proposed earlier in the text, but it is closely related. Typically, a strong reward or punishment is unexpected, at least in the beginning of the learning. When you find a reward the first time, your state-value function is in some rather random initial state, and you could not really predict that the reward would be obtained; thus any reward is initially surprising. That is why prioritized sweeping prioritizes, as a first approximation, episodes containing reward or punishment. Alternative theories for choosing what to replay are considered by \citet{isele2018selective,antonov2022optimism}.}

If wandering thoughts use such a prioritizing form of replay, they are closely related to the theory of emotions as interrupts discussed in Chapter~\ref{emotions.ch}. 
Both mechanisms direct the agent's processing (one might say attention) towards dangerous or rewarding events. Emotional interrupts are more primitive, typically focused on easily identifiable and evolutionarily important threats that are present in the current state. In contrast, wandering thoughts are about learning, activated when no threat is currently being observed, and potentially lead to quite sophisticated behaviors.\footnote{A lot of replay is probably related to rewards, and thus to planning and reinforcement learning, but some part of wandering thoughts and replay is clearly independent of any rewards. We saw earlier that people are able to perform unsupervised or supervised learning from a single representation of a data point (page~\pageref{replaymention}). If you hear a nice melody, it may be replayed it in your mind repetitively, even quite obsessively. Such replay if best understood as performing some kind of unsupervised learning---which does not need any kind of reward or reinforcement signal. For example, it can be Hebbian learning or some kind of feature extraction, which learns the  melody and its characteristics particularly well by repetition. The crucial similarity between reinforcement learning, Hebbian learning, and most kinds of machine learning is their iterative nature, and in particular, the need for many iterations.\index{replay!unsupervised}  Some of that data may not be real data replayed, but simulated data more akin to planning;  such simulation can in fact be used to perform learning in a Bayesian framework \citep{gutmann2018likelihood}. An alternative theory on resting-state activity actually links it to the priors used in Bayesian perception \citep{berkes2011spontaneous,aitchison2016hamiltonian,hoyer2003interpreting}. The idea is that activities of the neurons in resting-state, at least in the sensory cortices, follow the prior distribution of those features that they are encoding. While this theory is not framed in terms of replay, we could interpret it as saying that resting-state activity is in some sense ``replaying'' typical sensory inputs. These two theories may thus not be incompatible, the replay or wandering thoughts theory focusing on reward processing and the Bayesian theory focusing on basic sensory processing.}

\add{When wandering thoughts implement planning, the question of choosing the goal for planning arises as well---unless the thoughts are about random planning without any particular goal. In line with emotional interrupts, Chapter~\ref{emotions.ch} proposed that the goal for planning could be given by neural networks based on perception of desirable objects or states. A similar mechanism could be working with wandering thoughts; the main difference is the time scale since wandering thoughts are typically about something that it is not possible to do right now. One theory actually proposes that wandering thoughts focus on goals that have been selected but not yet reached.\footnote{\citep{klinger2013goal}. His theory actually considers ``spontaneous thoughts'',\index{thinking!spontaneous} which are more general than wandering thoughts. It further includes the interesting idea that wandering thoughts may not be just triggered when the computational capacity would be idle otherwise, but they could also be triggered when there is a goal, perhaps with an intention or commitment to it, but it is not currently possible to actually perform any meaningful action to reach the goal. Then, planning to reach that goal may be triggered involuntarily and lead to wandering thoughts.\label{spontthoughtfn}} This would lead to a typical daydream where you think about achieving something nice tomorrow.
  On the other hand, a lot of planning is also about avoiding things or processing threats.
  Psychological theory has for a long time connected such spontaneous planning to \textit{worrying},\index{worrying} which is basically planning against threats anticipated in the distant future.\footnote{\add{On worrying, and its relation to anxiety,\index{anxiety} see in particular the discussion of the literature by \citet{stawarczyk2013concern}.
      Closely related is the proposal by \citet{revonsuo2000reinterpretation} on the function of dreaming during sleep:
``the biological function of dreaming is to simulate threatening events, and to rehearse threat perception and threat avoidance.'' }}
  Thus, even in the case of planning, wandering thoughts are often related to emotionally charged events.}

\section{Replay exists in rats, humans, and machines}

Replay has long been observed in neuroscience experiments.\index{replay!in rats} Typical experiments measure brain activity in rats, which are running in a maze, seeking food or drink. A brain area called the hippocampus\index{hippocampus} is specialized in storing episodes and events ---such as the sequence of running forward, turning left or right, and finding cheese. It is thought that the hippocampus replays such episodes, simultaneously signalling them to other brain areas, which then use such replayed input for learning. Replay was initially observed during sleep, but it can also be seen in awake rats.\footnote{Sleep: \citep{buzsaki1996hippocampo}, awake animals: \citep{karlsson2009awake}}
Recent experiments also show that something similar to prioritized sweeping, where the events are replayed backwards, seems to be happening in rodents.\citenew{diba2007forward,ambrose2016reverse} 

Research has also found brain activations that look like planning:\index{planning!in rats} a rat can initiate sequences of events which it has not yet experienced, but which it might perform in the future. For example, the rat can in some sense ``imagine'' a possible trajectory in a maze, which it may or may not follow later.\footnote{This is called \textit{preplay} in neuroscience \citep{pfeiffer2013hippocampal,wikenheiser2015hippocampal}. Some forms of  replay or preplay  seem to be happening much faster than real-time, which would make them particularly useful computationally \citep{karlsson2009awake}.}
So, the mammalian brain  seems to use strategies which are very similar to what you would expect from the design considerations of AI. This is not surprising since the brain and AI are trying to solve the same computational problems; but it is also the case because the AI designs have been influenced by neuroscience research.

It may in fact be that such processing in rats is not very different from wandering thoughts considered in human psychology.\index{replay!and wandering thoughts}
Something at least resembling replay by prioritized sweeping can also be observed in the human brain, although the limitations in measurement technology make it difficult to draw exact parallels.\citenew{buckner2010role,gruber2016post,kurth2016fast,momennejad2017offline} 
While replay is usually connected with the hippocampus, and planning with the default-mode network, the hippocampus is actually part of the default-mode network according to some definitions.\citenew{andrews2012brain} (Rats do have a default-mode network just like humans.\citenew{lu2012rat})\index{default-mode network!in rats}
The connection between wandering thoughts and the hippocampus is also seen in the fact that people with damage in the hippocampus have difficulties in imagining new experiences.\citenew{hassabis2007patients} 

Some scientists are  reluctant to make such parallels between hippocampal replay and wandering thoughts, since they would seem to imply that rats ``think'' or ``imagine'' like humans, at least in the sense  that rats engage in planning by imagining different sequences of actions and choose the best one.\footnote{See \citep{redshaw2018future,corballis2019language} for discussions on whether non-human animals might possess such capabilities.}\index{thinking!in animals}
Likewise, we immediately run into the question of whether such replay in an AI means that we would have to admit that an AI can think. ``Thinking''\index{thinking!not well-defined} is not a well-defined concept in either neuroscience or AI, which makes this question difficult to answer.\footnote{An important point is that, sometimes, thinking is defined to be conscious, and humans do a lot of the simulation consciously, while the AI is probably not conscious when planning. I think it is important to consider the phenomena of replay, planning, and imagining future actions  as a topic separate from consciousness, to which we will return in Chapter~\ref{consciousness.ch}.}

\subsection{Creative thinking and generative AI}

So far, the discussion has considered wandering thoughts as rather mechanistic solutions to some well-defined computational problems. This does not do justice to the variety of wandering thoughts in humans. Spontaneous thinking can be tremendously creative; in fact, it is one of the critical aspects of human creativity.\index{creativity!wandering thoughts}

Now, what is creativity? As a first approach, we might actually think of planning as a creative activity. You have the current state, a goal, and you have to somehow create a path between the two. In fact, many different kinds of problem-solving could be seen as special cases of such planning: even proving a mathematical theorem can be formalized as planning a ``route'' from the premises to the conclusion of the theorem. However, some would argue that this is just running an algorithm, so it cannot be called creative. I wonder why running an algorithm could not be called creative. What else does an intelligent system do anyway? On a sufficiently high level of abstraction, is not all our thinking a product of various kinds of algorithms? I shall not attempt to answer the deep question of what creativity really is; I will just note that creativity is not easy to define, similar to the concept of intelligence.

\add{In practice, a randomized algorithm can be quite a convincing example of creativity. Such algorithms contain certain randomness in their computation, which makes the algorithm try out completely new paths or ideas. Modern generative AI, whether generating images or text, is based on nothing else than such randomized algorithms. A model of, say, images is combined with random noise to generate an incredibly realistic image. Importantly, it would have been impossible for the human user who gave the initial prompt to predict all the details of the image, if only due to the randomness programmed in the system.}

In the case of wandering thoughts, Monte Carlo Tree Search\index{tree search!Monte Carlo}\index{Monte Carlo tree seach|see{tree search, Monte Carlo}} is an example of a randomized algorithm. It is not just deterministically finding a single solution to a given problem, but rather creatively imagining, as it were, a number of possible things to do, or steps towards a solution to the problem.  \add{Some  randomness in behavior is essential in exploring new environments, as we saw in Chapter~\ref{self.ch}; while in Chapter~\ref{dual.ch}, we saw how randomized algorithms have been very successful in game-playing AI}.
Thus, randomized search algorithms offer a plausible model for some of the wandering thoughts. 
From this viewpoint, it is natural that the computations performed by wandering thoughts can also result in creative problem-solving.\citenew{fox2019mind}

In fact, there are also some wandering thoughts that cannot be plausibly considered as replay or planning. Perhaps, while lying idly on your sofa, you have a series of  seemingly unrelated mental images, or a superhero fantasy that could never actually happen in reality. One function of such wandering thoughts may be to create completely new ideas and associations, even new goals.  \add{In this case, wandering thoughts can be compared with the outputs of generative AI systems, which, in the absence of a restrictive prompt, create text or images by quite randomly sampling from a language model or an image model.}
\add{This leads to a well-known problem with generative AI: a randomized generation of content will sometimes produce complete nonsense. To circumvent this, a ``generate and test'' approach can be used:} first, new items are more or less randomly generated by one part of the system, and then another part of the system tests whether they make any sense. Unrealistic, weird, and unstructured wandering thoughts could be the result of such random generation; hopefully, our more rational part then tests them and decides which ones make any sense and should be taken seriously.\footnote{For surveys on the theory of ``computational creativity'', i.e.\ trying to make computers creative, see \citet{colton2009computational,toivonen2015data}. \add{How such theory is related to state-of-the-art generative AI is discussed by \citet{franceschelli2023creativity}; related neuroscience research is reviewed by \citet{jung2013structure}.  } }

\section{Wandering thoughts multiply suffering} \label{wt-suffering.sec}

So far, we have seen that while mind wandering may be detrimental for whatever you're trying to do at the present moment, it helps in planning and learning, perhaps even allowing some creativity. From a purely information-processing viewpoint, it is probably a useful thing since similar ideas are currently used in AI systems, and after all, evolution would not have ``programmed'' us to have a wandering mind if it were not useful to us from the evolutionary viewpoint.\index{evolution!and happiness}

Yet, evolution does not try to make us happy. 
A problem with replaying past memories and planning the future in human brains is that we are, on some level, unable to understand they are not real. 
If you remember an embarrassing episode from the past, you actually feel embarrassed.
If you think about something scary that might happen to you tomorrow, you actually start feeling scared.  That is, wandering thoughts increase human suffering by making us suffer from simulated or replayed events, in addition to the real ones. 

Any suffering produced by real-life events may, in fact, be \textit{repeated many times} by the replay of those events. \add{Making a mistake might lead to nothing but a fleeting frustration if we didn't replay it afterwards, thus generating many instances of regret.}\index{regret} Likewise, if something unpleasant is expected to happen, the unpleasantness, the threat, is felt many times in planning how to avoid that thing---which may actually turn out not to happen at all. Planning future events includes frustration when things in the fantasy don't go as you would like them to, and you can be frustrated many times by the planning of a single event. Due to this multiplication of suffering by wandering thoughts, it could be argued that the \textit{vast majority} of our suffering actually comes from remembering or anticipating unpleasant events. The anticipation is closely related to discussions on threats and fear in the preceding chapters, but the aspect of replaying unpleasant memories is new.

Importantly from the viewpoint of suffering, you have little control regarding such wandering thoughts.\index{uncontrollability!of the mind} You may think that \textit{you} must have decided to recall an unpleasant conversation, but in fact, the recollection and replay just started without you deciding anything, and even if you try to think about something else, you may find yourself unable to do so. This is another clear connection to the emotional interrupts: both wandering thoughts and emotional interrupts are largely beyond conscious control. You cannot switch off those systems. In fact, it is even worse: both systems actually take control of the agent.

Some research actually claims that the wandering mind is generally unhappy.\index{wandering thoughts!increasing suffering}
That sounds plausible if wandering thoughts multiply suffering, as I just argued. However, it might be a bit of an overgeneralization.\footnote{See \citet{killingsworth2010wandering} for the claim that ``A wandering mind is an unhappy mind''. The problem is, however, that such studies don't conclusively show that it is mind-wandering that \textit{makes} people unhappy. It is also possible that the causal effect is the opposite: when we are unhappy, thoughts start wandering more \citep{smallwood2009shifting}. This could be because negative mood is related to unresolved goals or personal problems, which are then processed during mind-wandering. If you're sad, it may be because you are experiencing problems, and those problems need extra processing by mind wandering \citep{poerio2013mind}, in line with footnote~\ref{spontthoughtfn} above.\label{wanderingmindunhappy}}
Whether wandering thoughts make you unhappy probably depends on their contents. It might seem obvious that having wandering thoughts with negative feelings, such as worrying,\index{worrying} makes you unhappier, while positive content has the opposite effect. In what is a rather extreme case, a study found that women having wandering thoughts about their significant others actually felt happier.\footnote{\citep{poerio2015love}. Intriguingly, the effect of wandering thoughts on mood may depend on whether you think about the past or the future.  \citet{Ruby2013wandering} found that future-oriented thinking has a general positive effect on mood, even if the contents were negative; perhaps this is so because when we solve a planning problem, we get happier. In contrast, thinking about past events was found to lead to negative mood independently of the contents of the thoughts. \add{However, \citet{poerio2013mind} argue against such results, and in particular point out that future-oriented thinking may increase anxiety\index{anxiety} based on worrying about what might happen in the future; see also \citet{stawarczyk2013concern}.}}
Close to the negative extreme, we find \textit{rumination},\index{thinking!about the past} which is thinking about negative events that typically happened in the past and are related to one's personal concerns.\footnote{\citep{whitmer2013attentional}. Perhaps the most extreme negative example would be flashbacks about a traumatizing event in post-traumatic stress disorder \citep{yehuda2002post}.}  It is particularly frequent in depression and, unsurprisingly, leads to low mood.\index{rumination}
For individuals with depressive tendencies, most wandering thoughts may consist of depressive rumination, and eventually may lead to relapse and full-blown depressive episodes.\index{depression}
Even for normal individuals,\index{individual differences} wandering thoughts provide an opportunity for rumination to arise, and thus may lead, on the average, to negative mood.\footnote{\citep{teasdale2000prevention,marchetti2014self,marchetti2016spontaneous,ottaviani2013flexibility,van2018does}. Another point to note here is that when talking about thoughts inducing a positive or negative mood, we need a baseline. In the research cited, this is typically a rather normal, average mood. However, if the baseline had absolutely no wandering thoughts, as might be considered ideal in some particular meditation traditions, it could be that even positively valenced wandering thoughts actually have a negative effect on the mood.\index{empty mind}}
In spite of some reservations, therefore, I think an important point is made in claiming that a wandering mind is an unhappy mind; we will get back to this important point when talking about meditation in Chapter~\ref{training.ch}.

\subsection{Why do wandering thoughts trigger feelings?}

Replaying negative experiences, or planning the future, might not have anything to do with suffering if they did not somehow \textit{feel} unpleasant, i.e.\ if they did not activate the negative valence signalling. 
A person may have reoccurring wandering thoughts about going to the dentist and vaguely feel the pain that the dentist's tools will cause in her mouth. Isn't it
odd that she feels the pain although she is not at the dentist at all?  While
you probably have to go to the dentist one day, people also worry\index{worrying}
about the possibility of various disasters that are not at all  likely to happen to them. Let me repeat Montaigne's comment: ``One who fears suffering is already suffering from what he fears''.\index{Montaigne}

Thoughts rarely correspond to something that is actually happening here and now, as opposed to perceptions.  Almost by definition, our thinking is usually about past events which are no longer there, or future events which have not yet happened, and may not happen at all. Why do we then feel upset about them, or, from a more computational viewpoint, why do they activate negative valence signals? Indeed---this is a deep question that we  encounter rseveral times in this book---why do we feel the emotions associated with memories and imagination?

From the viewpoint of computational design, it is clear that the system that computes state-values and predicts rewards has to be active in wandering thoughts, at least to some extent, so that the brain can take its evaluations into account when planning and learning. What does not seem necessary is that we actually, on a visceral level, feel pleasant or unpleasant about the events produced by planned actions.\index{body}
Why do our bodies react to our fantasies as if they were true?
I suggest this is a kind of a \textit{computational shortcut}. If you want to make learning from the simulation as simple as possible, it makes sense to use the same mechanisms and networks as in the case of real data. This is possible if the AI or the brain is fed the same kinds of inputs signals into the same networks regardless of whether the action is real or simulated. 

Ultimately, combined with the hypothesis that the error signals\index{error signalling} are best broadcast\index{broadcasting} to the whole brain using the pain system (Chapter~\ref{suffering.ch}), such computational simplification seems to have led to a situation where in the brain, it is not possible to give an ``unpleasant'' signal to the planning system without activating the main system that signals suffering to the whole system. In other words, perhaps humans \textit{feel} suffering during negative wandering thoughts simply because it makes the design of the learning system easier.\footnote{Incidentally, such processing is also part of the somatic marker hypothesis; see footnote \ref{somaticmarkersfn} in Chapter~\ref{emotions.ch}.} 

Here we see a particularly striking conflict between evolutionary goals and happiness.\index{evolution!and happiness}\index{happiness} Suffering from the simulation of negative events may be a computational shortcut, which is not really that necessary. It is just that  the brain was ``designed'' by evolutionary forces which  do not care if the system design makes you suffer many times more; they happened to find this design useful for their own evolutionary purposes.

\chapter{Perception as construction of the world} \label{perception.ch}

Without any perceptual abilities, an agent can hardly do anything intelligent in the real world. Neural networks give a rudimentary system for perception: for an input image, they can try to tell what it depicts. 
However, it turns out that perception is an extremely difficult problem.
In this chapter, I explain the main difficulties involved in perception, and how they can be solved by modern AI but only to some very limited extent.
I argue that the very problem of perception is so difficult that even our brains do not solve it very well. Here, I consider in detail visual perception, but the theory largely holds for other kinds of perception.

What is crucial for the main theme of this book is to understand the relevant implications of the extreme difficulty of perception. 
The incoming sensory data is incomplete, and we fill in the gaps by using various assumptions, or prior information, about the world.
This implies that our perceptions are quite \textit{uncertain}, or unreliable, and much more so than we tend to realize.
One aspect of such uncertainty is subjectivity: we fill in the gaps using our \textit{own} assumptions, and my assumptions may be different from yours. Perception is essentially a construction, a result of unreliable and somewhat arbitrary computations; it is not an objective and perfect recovery of some underlying truth.

These fundamental problems in perception feed into the difficulty of making correct inferences about the world: they make any categorization uncertain, they reduce the possibility of predicting the world, and consequently reduce any control the agent has. This increases various errors such as reward prediction errors, and thus suffering. More specifically, the computation of reward loss is dependent on the prediction of reward as well as the perception of obtained reward, which are both subject to the limitations of perception, and thus can go wrong.

\section{Vision only seems to be effortless and certain}

It may be surprising to many people  how difficult computer vision actually is, and what an incredible feat the visual system of our brain is accomplishing, literally, every second. It all seems to happen so effortlessly and automatically. However,  our capacity for vision is effortless only in the sense that it does not require much \textit{conscious} effort, and it is automatic only in the sense that it does not usually need any \textit{conscious} decisions or thinking. You turn your gaze towards a cat, and immediately, without any conscious effort, you recognize it as a cat. This is a typical, even extreme case of dual systems processing: most of the computations happen in neural networks, not at the level of symbolic, conscious thinking. Since we have little access to the processing in the neural networks, we cannot understand how complicated their computations are.\index{effort!in vision}\index{vision!difficulty}\index{vision}

In the early days of AI in the 1970s, computer scientists thought programming such ``computer vision'' must be easy. 
However, anybody either studying the human visual system or trying to build a computer vision system is quickly convinced of the near-miraculous complexity of the information processing that is needed for vision and performed by our brain almost all the time. Knowing that history, it is not surprising that while computers can beat humans in chess or arithmetics, they are nowhere near human performance in visual processing.\footnote{It is true that in some specific, well-defined tasks, such as recognizing animals in photographs, AI can actually outperform humans. However, such performance is usually specific to a certain kind of input data and task, and it is still far away from the versatility of human vision; see e.g.\  \citet{recht2019imagenet,peters2021capturing}.}

\subsection{Too much data}

A major difficulty in vision is the huge amount of data received by the system.
The immensity of the data is perhaps obvious to anybody who has waited for video data to download over a mediocre internet connection. In fact, the vast majority of internet traffic takes the form of video data. Text data is completely negligible in terms of file size: a large book is hardly equal to  a second of video data.

Likewise, humans and other mammals receive a huge, continuous stream of data from the environment through their eyes. 
The human retina contains something like one hundred million photoreceptors, which are cells that convert incoming light into neural signals. The manner in which the data is stored and transmitted may be very different from computers, but still the fundamental problem of receiving an immense amount of data is there, as well as the requirement of a huge amount of information-processing capacity.  In fact, the visual areas constitute something like half of the human cerebral cortex---the part of the brain where most sophisticated processing takes place.\citenew{nakayama1999mid}

\subsection{Yet, information is missing}

Having such huge amounts of data is both a blessing and a curse. A curse obviously in the sense that you need immense computing power to handle such a data deluge; a blessing in the sense that such huge amounts of data may contain a lot of useful information. Yet, paradoxically, the information contained in the input to a camera or the retina is almost always lacking in various important ways.

One of the most fundamental problems in vision is that what each eye gives us is a two-dimensional projection of the world, just like an ordinary photograph. A photograph is nothing like a 3D hologram: most of the information on the 3D structure of the objects is missing. (Having two eyes gives some hints of the 3D structure, i.e.\ which objects are close to you and which are far-away, but this only slightly remedies the problem.)

\begin{figure}
  \begin{center}
\resizebox{14cm}{2.7cm}{\includegraphics{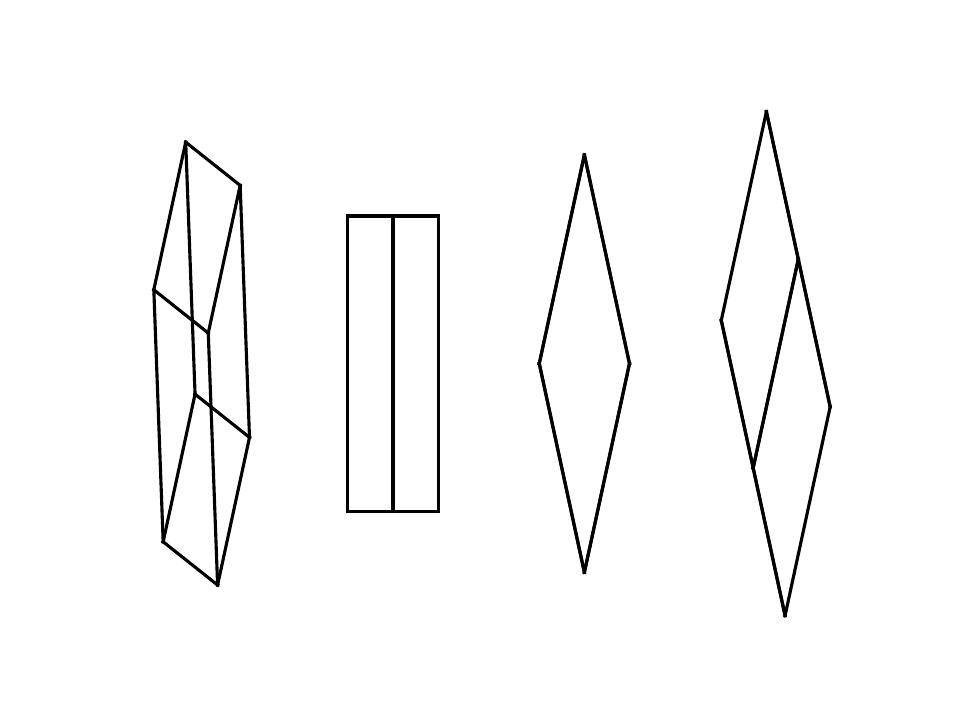}}
  \end{center}
\caption{An illustration of the inverse problem that makes vision particularly difficult. The four figures are all 2D projections of the same 3D cube.
  Any camera or a single eye can only capture one such 2D projection, which loses a lot of information and creates ambiguity.}
\label{inverse.fig}
\end{figure}

Suppose you see a black cat. Now, the actual 2D projection will be very different when you can see the cat from different viewpoints: from the front, from one side, from the other side, from above, and so on. That is, the pixels which are black are not at all the same in the different cases; the pixel values that would be input to a neural network will vary widely when the cat is seen from different viewpoints.
Thus, the neural network will have to somehow understand that very different pixel values correspond to the same object.
To illustrate this problem of 3D to 2D conversion, consider what even a simple cube can look like in different projections. Some possibilities are shown in Figure~\ref{inverse.fig}. Its 2D projection can look like a rectangle (possibly a square), like a diamond, and many other things.

And this is just one part of the problem. More fundamentally, the problem is that any object can undergo many different kinds of \textit{transformations}. Consider a cat again: it can take many different shapes by moving its limbs; sometimes its legs are wide apart, sometimes close to each other. Sometimes it stretches its whole body, sometimes it puffs up. If you think about the 2D image created, it will again be quite different in these different cases.  As another example, the lighting conditions can be very different. Imagine that light comes from above, or from behind the cat: Again the cat looks very different, and even more so in a 2D projection. Your brain has to somehow figure out all these extra parameters based on the sensory input.

Those were some of the problems in recognizing a single cat. To make things even more complicated, different cats look very different. Some are black, some are white, so the pixel values are even more fundamentally different. Yet, you somehow are able to see that they are all cats.

Such ambiguity or incompleteness of visual information in a camera or the retina is the reason why vision is called an \textit{inverse problem}.\footnote{Strictly speaking, what we consider here is an \textit{ill-posed} inverse problem; however, ill-posedness is often implicitly assumed when talking about inverse problems.}\index{inverse problem} As a very simple illustration of an inverse problem, consider there are two numbers which we denote by the variables $x$ and $y$. You want to know both these numbers, but the trick is you only are given their sum, $x+y$. How could you possibly find out both of those original numbers---how can you ``invert'' the equation? Suppose you are told the sum of two numbers is equal to 10. There are many possibilities what the actual $x$ and $y$ may be like, for example, $x=5$ and $y=5$, or $x=7$ and $y=3$ etc. Vision is a lot like this. What you observe are the pixel values in, say, a photograph. But there are a lot of factors that determine what the pixel values are like: the identity of the object in the photograph, the location of the object, the lighting conditions, the background, to name just a few. It is next to impossible to figure out what there is in the image without some tricks.

Actually, the fact that sensory information is incomplete  is in some sense quite blatant. Just think about the fact that you cannot see through solid surfaces. Suppose you look at a wall in front of you: you cannot see what is on the other side. Your perception is limited by the physics of light, which does not penetrate the wall, and thus you only obtain limited data and limited information about the environment. That may be an extreme example, but the point is that all perception is similarly constrained; it is just a matter of degree. Curiously, in your mind, you do have some idea of what there is behind the wall (another room, the street, or something else), but this idea is vague and uncertain. We will see in this chapter why all perception is, to some extent, a similar kind of guesswork.

\section{Perception as unconscious inference}

Yet, AI has recently been making major progress in vision. One reason is that computers have been getting much faster every year, but that is of course not enough in itself if you don't know how to program your computer.  The crucial breakthrough in recent computer vision has been the application of neural networks.\index{inference!unconscious}
Neural networks offer two important advances. First, they enable the processing of vast amounts of data to be distributed into a large number of processors, which work in parallel and thus can process the data more easily. The advantages of such distributed and parallel processing are considered in more detail in Chapter~\ref{control.ch}.\index{parallel processing}
In this chapter, we focus on another advantage, which is that we know how to make neural networks  \textit{learn} from big data sets.
Learning can alleviate, and to some extent solve, the problem of incomplete information, such as seeing only a 2D projection of the 3D world.

\begin{figure}
\begin{center}
  \resizebox{12cm}{!}{\includegraphics{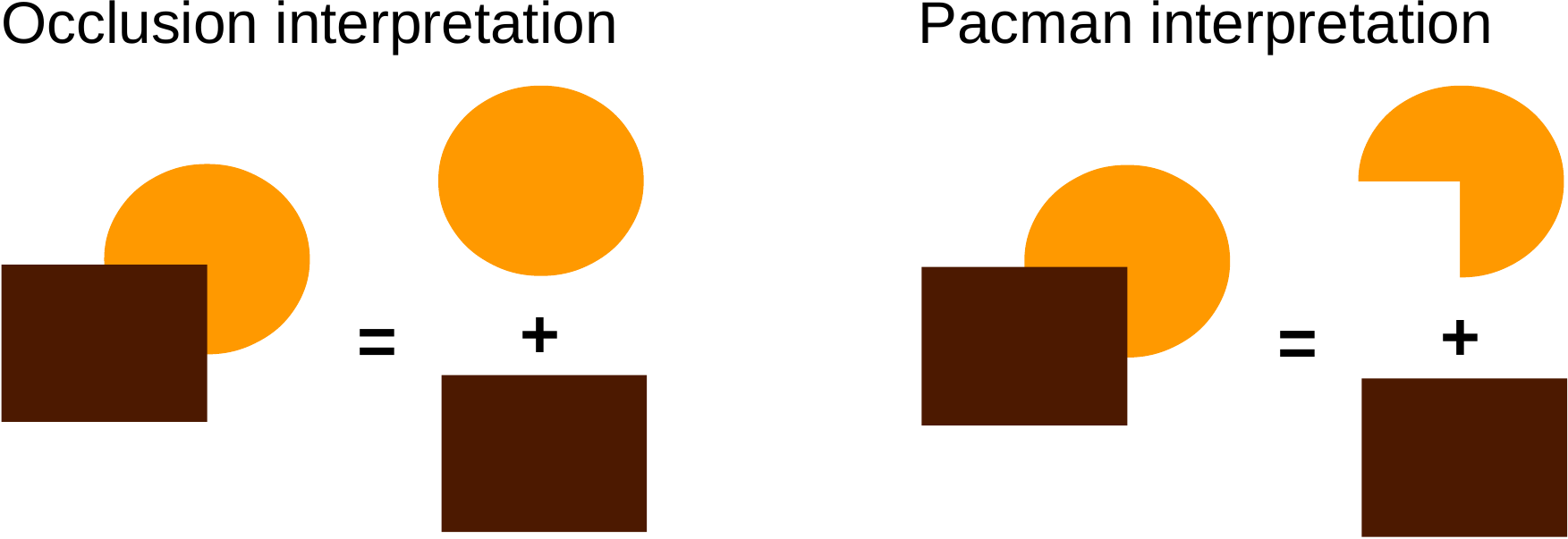}}
\end{center}
\caption{Suppose you see the figure on the far left, consisting of a square and a part of a disk. On the left: a typical interpretation where it is assumed that the disk is complete but occluded. On the right: another logically possible interpretation, with a pacman ``eating'' the square, but one that our visual system would not make because it is less likely. Our visual system chooses the interpretation which is more likely, given its prior information about the environment.}
\label{illufig.fig}
\end{figure}

The trick used by our brain is to learn what the world \textit{typically} looks like, and to use the learned regularities to complement the incoming data.\index{prior}
Look at the figure on the far left-hand side of Figure~\ref{illufig.fig}. Here, we tend to perceive a disk and a square. This is because we immediately assume that the disk actually continues behind the square, it is just partly occluded (i.e.\ blocked from view) by the rectangle. In fact, we tend to almost \textit{see} a whole disk. There's nothing wrong with such an assumption, but it does not necessarily follow from the figure. Alternative interpretations are possible based on this incomplete data. For example, it could be that the figure actually consists of a square and a ``pacman'', as illustrated on the right-hand side of the figure.

Perceptions such as in this example are usually explained as results of \textit{unconscious inference using prior information}. The visual system has learned certain regularities in the outside world---this is called prior information. For example, contours are typically continuous and smooth; lines are typically long and straight; objects can be behind or in front of each other. So, in Figure~\ref{illufig.fig}, the brain computes that it is very likely that the incomplete disk is actually part of a whole disk, but we just don't get visual input on the whole disk because it is blocked by the square. 
Such a conclusion is made by neural networks which are outside of our consciousness, thus the process is called unconscious inference.\footnote{Inference means the computational process leading to a conclusion or a decision.}

The inference in question is also \textit{probabilistic}: The visual system cannot know for sure whether the edge of the disk continues behind the square, but it is more likely that it does than that it doesn't. That is, the brain cannot make any judgements that are logically necessary and certain about this picture. The only thing the brain can do is to calculate probabilities and choose the most probable interpretation.\footnote{A fundamental question is whether the brain chooses one interpretation or whether it can entertain several interpretations simultaneously. Something in between these two seems to be happening in the special case of bistable perception, which means that when a stimulus can very well be interpreted in two different ways, the two interpretations seems to be alternating in the brain, so that conscious perception switches from one interpretation to another every few seconds or so \citep{sterzer2009neural}. The proportion of time allocated to each interpretation may, in fact, reflect its probability that the brain computes by Bayesian inference discussed next in the text \citep{moreno2011bayesian}.} That is why perception is necessarily \textit{uncertain} to some extent.\index{uncertainty!of perception}

\subsection{Bayesian inference}

The probabilistic inference needed in perception takes a particular form where the goal is to determine causes when observing the effects.\index{inference!Bayesian|see{Bayesian inference}}\index{Bayesian inference}\index{prior!Bayesian}
The mathematical theory behind such inference was initially proposed by Thomas Bayes in the 18th century, which is why such inference is often called Bayesian.\footnote{For neuroscience-oriented introductions, see \citep{kersten2004object,mabook}. While Thomas Bayes is usually credited with the general mathematical theory used in this context, the specific idea of perception as unconscious inference was actually formulated later by Hermann von Helmholtz,\index{Helmholtz} which is why some authors call this framework the Helmholtzian theory of perception. (Also, the credit for the mathematical theory should perhaps largely go to Pierre-Simon Laplace.)\index{Laplace} The recent proposal of a ``free-energy'' brain theory \citep{friston2010free} is essentially a reformulation of these ideas, with some additional hypotheses extending it to action selection.}
In the case of perception, the ``effects'' are the patterns of light coming into your eyes, while the ``causes'' are the objects and events in the outside world.

Typical scientific models based on physics will tell you what the effects are for given causes. For example, given an object and its location in your field of vision, you can rather easily compute, by basic physics, what the light coming from that object to your eyes will be like. But doing the computation backwards is more difficult. Given that your eye receives certain light patterns, as registered by your sensory organs, how can you know what went on in the outside world? You have to somehow \textit{invert} your physical model of the world, leading to the inverse problems just mentioned. Such problems can be approached by Bayesian inference, especially in the case where we can only calculate probabilities, which is exactly the case here. 

Bayesian inference tells that the probability for a given cause (given we observe certain effects) is proportional to the product of two things: First, the probability that such a cause creates the observed effects, and second, how likely the cause is to occur in general. 
The first part here is rather obvious: A given cause is more likely to be responsible for what your sensory organs report if the cause and such sensory input are compatible: that cause is likely to produce the observed effects. However, the important point here is in the second part: A given cause is even more probable if its general probability of occurrence is large. That is, if the cause has high ``prior probability'' in the terminology of Bayesian inference.\footnote{\newcommand{\cause}{\mathit{cause}}\newcommand{\effect}{\mathit{effect}}\newcommand{\given}{\text{\:given\:}}To get into more mathematical detail, Bayesian inference wants to compute the probability $P(\cause \given \effect)$, where $P$ denotes probability. More precisely, this is a conditional probability, i.e.\ the probability of one thing (cause)  given that another thing (effect) has been observed. This is the typical case of inference: we observe the effects and want to find the causes, or at least their probabilities. The celebrated Bayes formula then says the aforementioned probability is equal to  $P(\effect \given \cause)\times P(\cause)/P(\effect)$. Here, the term $P(\effect \given \cause)$ can be computed from a physical model of the world implemented in your brain. 
  $P(\cause)$ is the prior probability of a given cause; this is where the prior information about what typically happens in the world comes in. $P(\effect)$ is not so important because we are not comparing different effects, so it is constant, and it can actually be computed from the other probabilities by a simple formula.}

Consider the following example. Through your living room window, you get a glimpse of something green moving on the street. It could have been a green car, or it could have been a Martian (they are all green, as is well-known). Both of these two causes (car or Martian) would produce the same kind of quick flash of something green moving on the street, or more precisely, some green light briefly entering your eyes. So, the probability of the effect (green light stimulating your retina) is high for both two causes; let's say for the sake of argument that it is equally high in both cases. However, you will not think it is a Martian. The reason is that your brain uses Bayesian inference and looks at the prior probabilities. The prior probability of a Martian is very much lower than the prior probability of a green car; the brain knows that in general, it is very rare to encounter any Martians. Thus, when weighing the probabilities of the different causes, the green car wins by a wide margin. This inference is possible because the brain has a model of what the world is typically, or probably, like: Martians are quite rarely encountered, at least on planet Earth.

\section{Prior information can be learned}

Prior information, i.e., a model of what the world is typically like, is central in such unconscious inference, so where does it come from?\index{prior!learning}\index{learning!of Bayesian prior}
The crucial principle in modern AI and neuroscience is that the prior information can be obtained by learning from data; learning is thus the basis of perception. Now that may seem like a weird claim from a biological viewpoint. How could perception possibly be based on learning, given that many animals see quite well more or less immediately after birth? With human infants, developing proper vision actually takes several months but that is beside the point. The point here is to understand the different meanings of the word ``learning''.

When I talk about learning here, I mean learning in a very abstract sense where a system \textit{adapts} its behavior and computations to the environment in which it operates, and in particular to the input it receives. In human perception, such adaptation happens on different levels and time scales: there is both the evolutionary adaptation and the development of the individual (after birth). These two time scales are very different, but if we are interested in the final result of learning, we can just lump the two kinds of adaptation together. Likewise, the optimization procedures are very different: evolution is based on natural selection while individual development presumably uses something like Hebbian learning---although we don't understand the details yet.\index{learning!Hebbian}
Again, if we just look at the end result of the combination of those processes, we can ignore the difference of  optimization procedures as well, and simply call this whole process ``learning''.\index{learning!vs evolution}\label{learninginbiology}
This resolves the paradox of animals being able to perceive things instantly after birth. Their sensory processing is using all the results of the evolutionary part of learning, and thus even before having received much input as individuals, their neural networks are capable of some rudimentary processing.\footnote{There is actually something in between those two kinds (evolutionary and developmental) of biological learning, which is learning in the womb. At the late stages of the pregnancy, the visual system of the foetus is ``learning''. While its eyes are closed, and they don't receive much input, certain dynamic patterns called ``travelling waves'' are generated in the eye, on the retina. These patterns are then fed to the visual cortex of the brain, enabling some basic learning of visual regularities, complementing the information in the genes \citep{wong1999retinal}.}

\subsection{Neural networks weights contain the prior information}

We already saw in Chapter~\ref{ml.ch} how it is possible to train a neural network from big data sets. The weights in the network are learned based on minimization of some error function. In the simplest case, the learning algorithm knows what there is in each image used for training (a cat or a dog) which provides a label or a category, and then we can use supervised learning.
If we want to understand biological vision, though, unsupervised learning is preferred. This is because the visual system does not really have anybody constantly giving labels to each input image, which makes supervised learning unrealistic as a theoretical framework.\index{vision!feature extraction}

Fortunately, Hebbian learning and other methods of unsupervised learning\index{unsupervised learning!with images} can learn to analyze images in interesting ways, without any supervision. Intuitively, if the input to neuron A and the input to neuron B are often rather similar, it is likely that they are somehow signalling the same thing, and thus they should be processed together, for example by computing their average or difference.\footnote{In particular, Hebbian learning can implement feature extraction methods such as principal component analysis \citep{oja1982simplified} and independent component analysis \citep{hyvarinen2001book}.}\index{learning!Hebbian}\index{independent component analysis}
The results of such learning are stored in the synaptic weights of the neural network. From the viewpoint of Bayesian perception, we can thus say that the prior information is learned and stored in the form of the weights connecting the neurons. \add{Such neural networks embodying prior information also form the basis of generative AI systems that generate realistic images. They have been trained by millions, if not billions, of photographs in an unsupervised manner.\citenew{yang2023diffusion,croitoru2023diffusion} Recognizing what is in an image on the one hand, and generating new images on the other hand, require closely related neural networks and image models.\footnote{For a discussion of the connection of the two cases, see \citet{xie2016theory,grathwohl2019your}.}}

\begin{figure}
  \begin{center}
\resizebox{3.5cm}{!}{\includegraphics{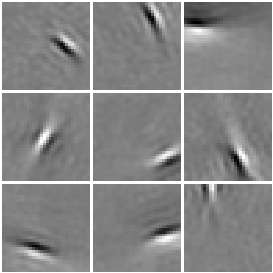}}
  \end{center}
\caption{Simple image features learned by a neural network. Each small patch gives the synaptic weights in a neuron whose input consists of small image patches. The weights can thus be plotted as gray-scale values arranged as image patches. More precisely, these are the results of applying a method of unsupervised learning called independent component analysis on small image patches.\index{independent component analysis}}
\label{ICAfig}
\end{figure}

\add{We can investigate what kind of prior information has been learned by such neural networks by looking at the weights of the networks.\index{learning!in the brain}
Considering} the initial analysis of images done by a neural network with just one layer, different learning rules almost invariably give the same result: the most basic visual regularities are something like short edges or bars. Figure~\ref{ICAfig} shows some examples.
Interestingly, such AI learning leads to processing which is very similar to the part of the brain that does some of the earliest analysis of incoming images, called the primary visual cortex.\index{visual cortex} Measurements of many cells in that area reveal that they compute features which look very much like those in Fig~\ref{ICAfig}. Edges and bars are clearly very fundamental elements of the structure of images.\footnote{The figure and the discussion are based on \citep{olshausen1996emergence,van1998independent}.    The learning principle used here can be intuitively understood from two different viewpoints. One is independence of the features: the outputs of the neural network (which in this case has a single layer) should be as independent as possible in the sense of probability theory. In other words, knowing one feature should give minimal information about the other features. The other viewpoint is sparsity: the features should be silent (zero) most of the time and only rarely turned ``on''.  An important benefit of such sparse coding is that it minimizes energy consumption if representing a feature that is zero consumes little energy. Therefore, the learning principle used is called either independent component analysis or sparse coding, which, surprisingly, turn out to be almost equivalent. Such analysis can be implemented as a particular kind of Hebbian learning. Actually, there is an even more fundamental regularity in visual input than the one depicted here, which is that two near-by pixels tend to have similar gray-scale values (they are strongly correlated). That is, if a pixel is, say, white, the pixels next to it are quite likely to be white as well---and the same applies for any color. Such similarities are analyzed by neurons (``ganglion cells'') in the retina. However, this regularity is so elementary that it is in some sense included in, or implied by, the regularity described by the edges. Mathematically speaking, the covariances of pixel gray-scale values are perfectly modelled by independent component analysis and no additional model is needed. For a general introduction to the models used here, see \citet{hyvarinen2009book}.} 

Such edges and bars can be seen as the first stage of the successive ``pattern-matching'' on which neural network computation is based. 
We can actually go further and train a feedforward neural network with many layers to analyze images.\footnote{The theory of unsupervised learning  is much less developed and more complicated than the theory of supervised learning, especially for multi-layer networks. Therefore, a lot of work on such feature learning uses supervised learning, somehow obtaining labels or categories for each image, and using ordinary supervised learning where the network learns the connection between the images and their categories. The bottleneck here is getting sufficient amounts of such data with category labels. It is difficult because somebody has to tell what the photographs are depicting; if the labels are given by humans, that is a lot of work (although a simple approximation would be to extract the labels from captions, which are sometimes attached to images on the internet).
  Current research is strongly focused on finding methods to train multi-layer neural networks without labels, that is, in an unsupervised way. A particularly promising approach is called \textit{self-supervised},\index{learning!self-supervised}\index{unsupervised learning!self-supervised} which means  performing unsupervised learning by reformulating the problem as supervised learning. Basically, you create hypothetical outputs, or a hypothetical classification problem, and use them to train your ordinary supervised, input-output neural network.
  The possibilities are unlimited: you could  define the input to a neural network to be a degraded version of your data and the output your real data, where the degraded version could be obtained by adding noise, or making a color image black-and-white \citep{vincent2008extracting,larsson2017colorization}. 
  Or, the ``degraded'' data could actually be artificially generated: 
  then you train the neural network to distinguish between the real and the artificial data 
\citep{Gutmann12JMLR}.
  For example, in video data, you could randomly shuffle the time frames in a video, or scramble audio in a video with sound, and train the neural network to classify such scrambled data vs.\ the original data \citep{Hyva17AISTATS,misra2016shuffle, arandjelovic2017look}.
In each case, the neural network has to learn something about the structure of the data in order to perform this mapping, that is, trying to reconstruct the original images from degraded ones, or telling which data is real and which is noise. The multi-layer processing thus learned are reasonably similar to what is computed in the brain \citep{zhuang2021unsupervised}. However, it should be noted that self-supervised learning in itself gives only features; it does not give a proper Bayesian prior model except in some special cases, such as the ``noise-contrastive estimation'' by  \citet{Gutmann12JMLR}, and nonlinear versions of independent component analysis \citep{Hyva16NIPS,Khemakhem20iVAE}.\index{independent component analysis}
}
After successful training, a multi-layer neural  network can contain extremely rich prior information about images. 
In general, the multi-layer network will be computing increasingly complex features in each layer.\citenew{gucclu2015deep,kriegeskorte2015deep,eickenberg2017seeing,zhuang2021unsupervised}\index{learning!unsupervised|see{unsupervised learning}}   The features computed by the units in higher layers are no longer simple edges or bars: they are more like some specific parts of the objects that the network was trained on. They are also more focused on coding the identity of those parts while ignoring less relevant details such as where in the image the parts are located.  For example, a neuron in a high layer could respond to a cat head, irrespective of where it is in the input image, and further ignore details such as the exact shape of the face of the cat. In this sense, such neurons are quite similar to cells in the  inferotemporal cortex, an area in the brain that performs a very high level of image analysis.\footnote{\citep{tanaka1996inferotemporal,brincat2004underlying}. The inferotemporal cortex is usually investigated in the macaque monkey, not humans, but similar mechanisms are likely to operate in the human brain.
  There have also been claims that high-level visual neurons in the human brain could be coding for the identities of single individuals \citep{quiroga2005invariant}. However, a more detailed analysis of the results shows that this is an exaggerated interpretation: single neurons are probably responding to several different people \citep{quiroga2008sparse}.  The property of ignoring less meaningful details is called \textit{invariance} \citep{dicarlo2007untangling}.\index{invariance}
 
}\index{prior}

\section{Illusions as inference that goes wrong}

\begin{figure}
  \begin{center}
  \resizebox{5cm}{!}{\includegraphics{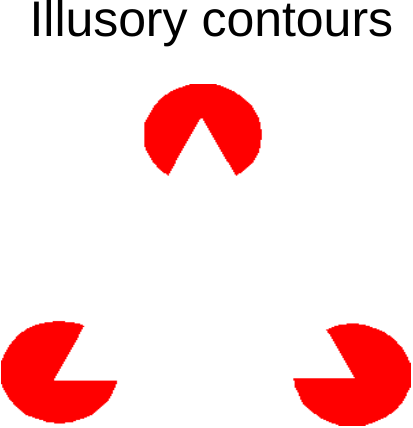}}
  \end{center}
\caption{The Kanizsa triangle, an example of a visual illusion. There is actually no triangle in the figure, just pacmans.}
\label{kanizsa.fig}
\end{figure}

We have now seen that the incompleteness of the incoming sensory information can be, to some extent, alleviated by Bayesian inference. 
However, this solution is far from perfect---whether we consider perception in humans or sophisticated AI.\index{vision!illusions}
Sometimes the perception is blatantly incorrect, as shown by the phenomenon of visual (or ``optical'') illusions.
A dramatic example is shown in Figure~\ref{kanizsa.fig}. We tend to see a full triangle in the figure, with three uninterrupted lines as its sides or edges. In reality, though, the sides of the triangle do not exist in the figure. If you cover the ``pacmans'' with your fingers, you see that there is nothing but white space between them. Yet, most people have a vivid perception of three lines between the pacmans which create a complete triangle.

This is called an illusion in neuroscience since the sides do not physically exist in the figure; they are simply imagined by our visual apparatus. Just like the imagination of a full disk in Figure~\ref{illufig.fig} we saw earlier, this can be considered unconscious inference, where your visual system computes the most likely interpretation. The difference is that here, the interpretation is in clear contradiction with the actual stimulus, or physical reality. While inferring a full disk in Figure~\ref{illufig.fig} seemed smart and would quite probably have been correct in real life, inferring that there is a full triangle in Figure~\ref{kanizsa.fig} may seem quite stupid, at least after you have checked that the sides do not really exist. The curious fact is that you cannot really help seeing the triangle in Figure~\ref{kanizsa.fig}. 

The theories explained in earlier chapters help us further understand why such illusions occur. A neural network is trained to accomplish a well-defined task, such as recognizing different objects in photographs. However, such neural networks are inflexible and only able to solve the problem they are trained for; neural networks are not general problem-solving machines. In particular, a neural network will not work very well when the input comes from a different source than what it was trained for. Arguably, the Kanizsa triangle is something artificial, and different from what you would usually see in real life (where pacmans are quite rare), so we should not expect the brain's neural networks to process it appropriately. This is another way of saying that the brain's prior information contains assumptions that are typically true in the context where you usually live, but they are just about probabilities, and might sometimes turn out to be quite wrong.

At the same time, the dual-system theory explains why it does not help if somebody explains to you that this is an illusion, or even if you realize that yourself. A logical, symbol-level understanding that there is no triangle has little effect on the other system, i.e.\ the neural networks, which are mainly in charge of visual perception.\index{dual process}

\section{Attention as input selection} \label{attention.ch}

In real life, any sophisticated perceptual system further faces the problem that there is simply too much information in the visual field, as already mentioned. This problem is very different from the missing information problem, which is partly solved by using prior information. In particular, there are often too many things in the visual input at the same time. There may be many faces, people, buildings, animals, or cars, at the same time, and it is too difficult to process all of them. 
This is in stark contrast to current success stories of object recognition by AI, which are usually obtained in a setting where each input image contains only one object, or at least one object is much more prominent than the others.

Suppose you input an image of a busy street to such a neural network trained to recognize a single object in an image.
Since the input now contains many objects, features of different kinds will be activated in the neural network, some related to the perception of people, some related to the perception of buildings, some to cars, and so on. Many of the features are actually quite similar in different objects: think about two faces in a crowd, which are quite similar on the level of pixels and even rather sophisticated features. 
Various neurons will be activated, but it is impossible to tell which were activated by which face.
It will be very difficult for the AI to make sense of such input and the activations of its feature detectors. 

This problem really arises when the information processing as well as the input data sensors work in a parallel and distributed mode. 
Parallel and distributed processing,  considered in detail in Chapter~\ref{control.ch},\index{parallel processing}\index{distributed processing}
usually means that there are many processors working simultaneously and independently. Here, the situation is even more extreme since the input data itself is received from a huge number of sensors, such as pixels in a camera or cells in the retina. Yet, the principles of parallel and distributed processing are really the same, as the outputs of the sensors are further processed by a large number of small processors. %

Traditional computer science usually does not deal with this problem. If the input to the computer is mouse clicks by a human user, the input is quite manageable. Even if a computer handles a very large database, the situation is different because it follows explicit instructions on what information to retrieve and in what order. Vision is more like thousands of disk drives  simultaneously and forcefully feeding the contents of their databases to a single computer.

The key to how the brain solves this problem, especially in the case of vision, is the multi-faceted phenomenon of \textit{attention}.\index{attention!selective} In the most basic case, the visual system of many animals, including humans, selects just one part of the input for further processing. As we say in everyday English, the animal only ``pays attention'' to one object at a time, whether it is a face seen on the street, or some object it is trying to manipulate. 

\begin{figure}
\begin{center}
\resizebox{8cm}{!}{{\includegraphics{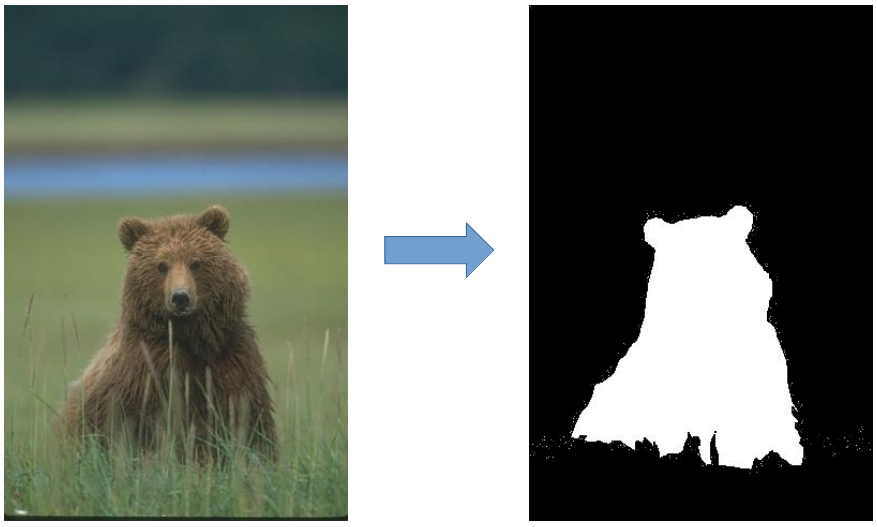}}}
  \caption{Attentional selection illustrated. The photo on the left is the original visual input. An attentional system selects the pixels to retain, shown in white in the figure on the right. (Based on data by \citet{MartinFTM01}, used with permission.)}
\label{attention.fig}
\end{center}
\end{figure}

The simplest form of such \textit{selective} attention is that you just wipe out everything else in the visual field, except for one object. In Figure~\ref{attention.fig}, we see a photograph and an attentional selection template, which shows how only the main object of interest in the figure is found and selected. The results of such computation can be used to simply blank out everything else except for the main object.\citenew{borji2015salient,zhou2019semantic,chen2018deeplab} Such a form of attentional selection is also called ``segmentation''.\index{segmentation}
Now, if you input an image that contains only this one object into the neural network, the recognition will be much easier. Such selection seems to be happening in many different parts of the brain and in many different ways. In a sense, it is a reflection of the ubiquity of parallel processing in the brain, which necessitates various forms of input selection all over the brain. The most amazing kind of attentional selection that our brains can accomplish must be finding individual faces in a crowd. Face processing is evolutionarily extremely important, so there are specialized areas in the human brain for processing just faces (monkeys have them too).\footnote{The word ``attention'' is quite overloaded with different meanings in cognitive psychology. The sensory attention we have considered here is very different from some other kinds of attention. In particular, another type of attention very relevant for this book is \textit{sustained attention}, considered in Chapter~\ref{replay.ch}, which means you try to concentrate on a single task, such as reading a book, for an extended period of time. That is very different from sensory \textit{selective} attention considered here since sustained attention is about \textit{long-term} attention on a \textit{task} instead of relatively short-term attention on sensory objects. Selective attention can further be divided on another axis: bottom-up attention, where an external stimulus grabs your attention (as in the case of interrupts in Chapter~\ref{emotions.ch}), and top-down attention, used for example when you search for a certain person in a big room and only pay attention to faces. (The exact terms used in the different cases are quite variable in the literature.)\label{attentionfn}\index{attention!different kinds}}

Performing such segmentation is not easy: using attentional mechanisms in AI and robots is an emerging topic, and we still don't know very well how to do it. However, like so many other functions related to intelligence, it may be possible to learn it. Attention is fundamentally a form of action: even moving your eyes can be seen as a form of attention, since it helps to select certain parts of your environment for visual processing. Thus, learning to attend may be possible by the general principles by which an agent can learn to act intelligently, as discussed in earlier chapters.\footnote{\citep{minut2001reinforcement,mnih2014recurrent,greff2016tagger}.   Learning attention has recently become very fashionable in AI, especially in large language models due to the methods proposed by \citet{vaswani2017attention}, although their use of the term is quite liberal and has only a vague resemblance to what we are discussing here.}

The downside of selective attention is that it leads to a bottleneck in the processing. In the example just given, only the one single object left in the image is given to further processing, including the final pattern recognition system. So, only one object can be recognized at a time, since all the others are wiped out. It is often said in cognitive neuroscience that ``attentional resources are limited'', and here we see one illustration of that principle: if you pay attention to one thing, you will necessarily tend to ignore everything else. This, in its turn, increases uncertainty since you don't know much about those things you are not paying attention to.

\section{Subjectivity and context-dependence of perception}

An important aspect of the uncertainty of perception is its subjectivity: I see one thing, and you may see something different. 
Being based on unconscious inference using prior information, perception is subjective if different people or agents have different priors. Then they will interpret the incomplete incoming information in different ways.\index{perception!subjectivity}\index{subjectivity!of perception} 

The priors used in human perception actually contain many different parts.\index{prior!subjective} There is one rather permanent and universal component, shared by all humans, and probably many animals. It includes those general regularities that can typically  be found by training artificial neural networks. But another component in the prior is more individual and depends on the experience of the agent (animal, human or AI). When an agent observes things happening, ideally it will incorporate all the new observations into its prior---possibly after performing some kind of attentional selection. If it didn't, it would be wasting valuable data that it has collected on the world. It is this individual part of the prior, based on  their own experiences, that makes the priors different from one agent to another.\index{individual differences}
Each agent may even be living in a different environment; they may spend their time in very different occupations. 
So, it is clearly useful that the prior is different from one agent to another. But this necessarily implies that perception will be different as well. You don't see exactly the same thing as your friends, not to even mention your robot. This might not be such a serious problem if the agent understood the subjectivity of perception well enough. However, such understanding often escapes even humans. 

There is even a further component in the prior, which depends on the context,\index{perception!context-dependence}\index{context-dependence!of perception} e.g., where the agent is at the moment of perception. If you're at home, you expect to see certain kinds of things, and if you're walking on the street, you expect to see other kinds of things. This leads to dependence of perception on the context, even for the same agent.\footnote{\citep{bar2004visual}. Even the perception of pain is modulated by context and history \citep{tabor2017pain}; see also Chapter~\ref{self.ch} and its footnote~\ref{painmodulatedfn}. Perception can also be modulated by metabolic states, such as hunger \citep{livneh2017homeostatic}. There are also claims that desire and aversion (motivational states) could directly influence perception \citep{balcetis2006see}, but such phenomena are controversial \citep{firestone2014top}.} These limitations of perception reflect the limitations of categories discussed in Chapter~\ref{dual.ch}. Categorization is usually based on perception, so if perception is subjective and context-dependent, the categories inherit those properties as well.

Perception is made even more subjective by the selection of incoming information by attentional mechanisms. Attention has a huge impact not only on the immediate perception in the agent, but also on the model it learns on the world. Fundamentally, attentional mechanisms choose the data that is input into the learning system. Anything not attended is pretty much ignored and not used in learning. As our brain ``creates our world'' in the sense of reconstructing it from sensory input, that creation is thus significantly influenced by attentional mechanisms.

\section{Reward loss as mere percept}

\index{reward!perceived}\index{reward loss!perceived}\index{uncertainty!of reward loss}
A crucial insight that this view on perception gives to suffering is that its causes are subjective and generally uncertain: they may be based on faulty inference. In particular, reward loss is just another percept (i.e.\ a result of the process of perception). It is based on solving an inverse problem to infer the obtained reward, and this can go wrong.

Misperceptions of rewards may be particularly common when perception of other people's reactions are involved. You might perceive the facial expression  of your friend as angry, and register some negative reward as resulting from  your actions. But perhaps your friend just had a bad headache, and his face reflected that; taking the uncertainty of perception into account should help you behave in a more appropriate way towards him.

\add{Contextual information can even change a perceived positive reward into a negative one, and vice versa. In one study, subjects were sniffing a combination of certain acids. In one session, they were told the substance is parmesan cheese, while in another session, they were told it is vomit. Depending on which category they were given verbally, the perception was different, and even the pleasantness of the odor was dependent on the verbal label.}\footnote{\citep{herz2001influence}. It is also typical for people to value objects more if it takes a lot of effort to obtain or produce them. This can be seen as a simple heuristic to approximate the reward, but it can of course go wrong \citep{kruger2004effort}. \add{\citet{eldar2016mood} further propose that mood influences perception of rewards, so that happiness makes rewards look higher, and the oppositive for a negative mood. Furthermore, individual differences are considered by \citet{scherer2021evidence} in terms of ``appraisal biases'', which is a mechanism explaining individual tendencies to experience particular emotions, and ultimately,  affective disorders; these could presumably be interpreted in terms of different priors being used by the individuals.}\index{heuristics!effort h.}\index{effort!heuristic}}
An extreme example of misperception of reward is found with some drugs of abuse. They feel good, and you perceive a reward on a biological level. Yet, such perception has no real basis: 
The drug merely misleads your brain into perceiving a reward by perturbing its metabolism.\citenew{nida}\index{drugs!misperception of reward}

The situation is even more complicated since in addition to perceiving the reward, the agent also computes the expected reward based on the information it has at its disposal and using the available computational capacities. Thus, there seem to be two different ways in which uncertainty in perception affects the computation of reward loss: the obtained reward may be perceived wrong, or the computation of expected reward may go wrong.\footnote{However, see footnote~\ref{expsubjectivefn} in Chapter~\ref{freedom.ch} on whether it makes sense to say that expectation of reward is ``wrong''.}\index{expectation!of reward} Both are just logical consequences of computation performed with limited resources and limited data.  Ultimately, a reward loss may even be illusory in the sense that one is perceived but it is merely a mental construct \add{with little basis in reality. Accordingly,} we should actually analyze the \textit{perceived reward loss} instead of any objectively defined reward loss, since the agent can never know with certainty what the reward loss was; \add{it acts according to its own perception, right or wrong}.\index{uncertainty!of reward loss} (We will postpone the details of such a re-definition to Chapter~\ref{overview.ch}.)

\pagebreak

\section{Ancient philosophers on perception}

The uncertainty and subjectivity of perception were discussed by several ancient philosophers.
In ancient Greece, the Skeptic school\index{Skeptics} was particularly prominent in pointing out the limits of human knowledge, including the relativity of perception. The Pyrrhonian branch was fond of giving examples where different people perceive the same thing differently:\footnote{Sextus Empiricus's \textit{Outlines of Pyrrhonism} from ca.\ 200 CE, with translation taken from \citet{annas1985modes}, see also e.g.\ \citep{sep-sextus-empiricus}.}
\begin{quote} 
When we press the eye from the side, the forms and shapes and
sizes of the objects we see appear elongated and narrow.
\end{quote}
Such uncertainty leads the skeptic to adopt an attitude of not making any judgements on external objects:\index{uncertainty!of judgements}\label{epoche}
\begin{quote}
So, since so much anomaly has been shown in objects (...), we shall not be able to say what each existing object is
like in its nature, but only how it appears (...)
therefore, it is necessary for us to suspend judgement on the
nature of external existing objects.
\end{quote}
A Japanese \yogacara-inspired\index{Yogacara} poem beautifully describes a scene where different agents have very different interpretations of the same sensory input:\citenew{shun2009living}\index{subjectivity!of perception}
\begin{quote}
  At the clapping of hands,\\
  the carp come swimming for food;\\
  The birds fly away in fright, and\\
  A maiden comes carrying tea---\\
  Sarusawa Pond
\end{quote}
When somebody claps his hands by the famous Sarusawa Pond in Nara, Japan, the carps interpret it as a call for feeding; the birds are scared of the noise and flee; while a maid of a near-by inn thinks a customer is calling for her. 

It is perhaps easy to admit that an animal or a robot sees things differently from yourself, either in a more primitive way, or perhaps in a superhuman way. Yet, it is notoriously difficult for humans to admit that two people can see the same thing in different ways, and that both ways can be equally valid. But there is something even more challenging; there is an even more difficult implication of the theories discussed in this chapter. It is the general idea that all our perceptions are actually just interpretations, or beliefs, or inferences, instead of revealing an objective truth. In AI theory, it is never claimed that the agent \textit{knows} anything; the very concept of knowing is conspicuously absent in that theory. All an AI agent has is \textit{beliefs}, and those are usually expressed in terms of probabilities, lacking any certainty.\index{uncertainty!of beliefs}

\pagebreak

When you take this line of thinking further, you may arrive at the idea that \textit{all} we believe or pretend to ``know''  is based on our perceptions, and thus inherits the uncertainty and the subjectivity of perception. 
In fact, one could say that my perception defines my world. This may actually be rather obvious to anybody who programs a sensory system in an AI.
Such ideas are often associated with Asian philosophical systems such as Mahayana Buddhism, especially the \yogacara\ school\index{Yogacara} and later schools drawing on those ideas, including\index{Zen} Zen.\citenew{williams2008mahayana} Yet, those ideas have also been beautifully expressed in the West, where their foremost proponent may have been David Hume who wrote:\footnote{\hume, Section 1.2.6. In fact, \citet{gopnik2009could} argues that Hume's ideas may have been influenced by Buddhism through some Jesuits; see also footnote~\ref{alexanderfn} in Chapter~\ref{dual.ch}.}
\begin{quote}\index{Hume}
  Let us chase our imagination to the heavens, or to the utmost limits of the universe; we never really advance a step beyond ourselves, nor can conceive any kind of existence, but those perceptions, which have appeared in that narrow compass. This is the universe of the imagination, nor have we any idea but what is there produced.
\end{quote}
We will see these deep points re-iterated and expanded in later chapters, especially Chapter~\ref{consciousness.ch}.

\chapter{Distributed processing and no-self philosophy} \label{control.ch}

The concept of a ``self'' is central for understanding suffering, but it is highly complex. Some aspects of self were already considered in Chapter~\ref{self.ch}.
In this chapter, I consider another central aspect of self, related to control.
Self can be seen as the entity that is in control of actions, including control of cognitive operations inside the agent, or, to put it simply, in control of the mind.\index{self}\index{distributed processing!and self}

In preceding chapters, we have seen cases where the mind seems to be difficult to control, due to automated interrupts and wandering thoughts. Here, I consider a general cognitive principle that explains why control is limited. The idea is that when the information processing is  parallel and distributed, it is difficult for any single part of the agent's information-processing system to be in charge of the whole system, e.g.\ the whole brain. This massively parallel and distributed nature of the brain thus creates most of the uncontrollability in the human mind.
The lack of control considered here can also be seen as a generalization of dual-process nature of the mind considered in earlier chapters. Here, there are not just two processes competing for control, but a great number of them. 

These considerations necessarily lead to the question of free will: Can an AI, or even a human, actually have free will---and what does that mean in the first place. From the viewpoint of the theories of perception in the preceding chapter, we can ask if perception of control and free will are simply illusory perceptions, thus providing another link between the uncertainty of perception and uncontrollability. Such considerations have lead some philosophers to propose that there is no self, or no doer of actions, and I will revisit these ideas from a computational viewpoint.

\section{Are you really in control?} \label{nocentralexecutive.sec}

Suppose you just raise your arm---you can physically do it while reading this if you like. You probably think it was you who decided to raise the arm, and it was you who actually executed the action. You felt being able to control the world, or at least your arm in this case.\footnote{Philosophers talk about (the feeling of) ``agency'' \citep{metzingerbook} . I don't use that terminology because it  would lead to confusion in this book where the word ``agent'' usually means something different. Furthermore, such agency is usually related to a conscious feeling, while in this chapter, I refrain from talking about anything related to consciousness, which will be treated separately in Chapter~\ref{consciousness.ch}.}\index{control}\index{agency}\index{control!lack of|see{uncontrollability}}
This ``you'' that first controlled your mind by making a decision, and then controlled your arm, is what can be called the self ---in one meaning of the word. 
The self chooses actions, and controls some aspects of the world, including your inner world.\citenew{skinner1996guide}

However, a number of thinkers have proposed that in fact, ``you'' are not really in control of anything. A case in point is wandering thoughts. It can be claimed---following a strict definition of the term---that we never want to have wandering thoughts: if we want to think  what we are actually thinking, the thoughts are not called wandering. Furthermore, wandering thoughts often feel unpleasant, for example in the extreme case of rumination. So, why do we then continue having them?

 A well-known experiment on the control of thoughts is to try to \textit{not} think of a pink elephant. This is another exercise you can do right now: for a minute or so, do not think of a pink elephant. What invariably happens is that you will be thinking of a pink elephant in spite of your trying not to, or rather because of that trying. Clearly, our control of thoughts is limited.

 In addition, interrupts such as fear, anger or desire capture our mind and direct the processing in ways we might not want. Even habitual behavior can be seen as a lack of control in some cases: if you mindlessly follow habits, you may end up doing something you would not have done if you had actually deliberately planned your actions.

Lack of control increases suffering in our basic framework of suffering as frustration. Lack of control reduces the probability that the agent reaches the goals it has committed to; it cannot get the things it wants, or avoid the things it is averse to. That means there will be more frustration and reward loss.
In fact, the very existence of suffering  can be seen as a form of uncontrollability, since if you could really control your mind, you would probably just switch off any feelings of suffering.

\subsection{Philosophical views on uncontrollability}

\index{uncontrollability!in Buddhism}
In philosophy, the idea of lack of control and its connection to the self goes back to, at least, the Buddha's times. In a famous discourse, he explained why there actually is no such thing as ``self''. He started his refutation by considering the human body, saying\footnote{\citep{mahasianatta}, based on \SN{22.59}; with explanatory text in parentheses added by Mahasi.}
\begin{quote}\label{anattalakkhana}\index{self!as control}\index{no-self!Buddhism}\index{Buddha}
[I]f the body were self, the core of our being, then it would not tend to affliction or distress, and one should be able to say of it, 'Let my body be thus (in the best of conditions); let my body not be thus (in a bad condition).' It should be possible to influence the body in this manner.\index{body}
\end{quote} 
He continued by going through different aspects of the human mind  (perception, thinking, etc.), and denying that any of them could be called the self either, since none of them can be properly controlled. For example, 
``no one can wish for and manage thus: 'Let my perceptions be thus, let my perceptions be not thus' ''.
If you smell something disgusting, you cannot just decide not to smell it.

Thus, originally, the Buddha framed the very concept of self in terms of control: self is what is in control.\footnote{Arguably, the Buddha's viewpoint could also be interpreted as the self being what can \textit{be controlled} instead of what \textit{controls}. Nevertheless, according to \citet[p.~12--14]{mahasianatta}, what the Buddha is denying is precisely a ``controlling self'' as well as an ``active agent self''. The two viewpoints are in a sense unified when \citet[p.~49]{harvey2009theravada} proposes that according to Theravadan Buddhist thinking, ``a Self would have total control over itself.'' In any case, this makes little difference in what follows where the main point is a general lack of control, or uncontrollability.} Since, as he argues, there is actually no (or little) possibility of control, there can be no self. Realizing this is thought to be essential to reduce suffering.\citenew{verhaeghen2017selfeffacing,harvey2009theravada}\label{noselfpage}

In ancient Greece and Rome, the Stoic philosophers had similar ideas. Perhaps the very core of Epictetus's philosophy is contained in his attitude towards control:\footnote{The very first lines in \EN, \add{translated by E.~Carter except for ``judgement'' for \textit{hypolepsis} which is by R.~Dobbin.}}\label{epicontrol} 
\begin{quote}\index{uncontrollability!Stoics}\index{Epictetus} 
Some things are in our control and others not. Things in our control are judgement, pursuit, desire, aversion, and, in a word, whatever are our own actions. Things not in our control are body, property, reputation, command, and, in one word, whatever are not our own actions. 
\end{quote}
Epictetus's idea of uncontrollability is more limited: we cannot control what others do or think about us, or, in line with the Buddha, our bodies. But in stark contrast to the Buddha, he seems to think we can control at least some of our thoughts and feelings, including desires and aversion. Presumably, Epictetus did not practice the same kind of meditation as the Buddha, which might convinced him of the uncontrollability of thoughts and feelings. In any case, both philosophers advocated recognizing how little control we have as a means of reducing suffering---we will discuss such practical implications in Chapter~\ref{freedom.ch}.\footnote{\add{This very well-known quote from \EN\ may not give a very clear idea of what exactly Epictetus considered to be under our control. In his \Discourses, a more detailed picture emerges, but it is not entirely consistent. He usually says that one of the following two things is the only thing under our control:  either the ``will'' (\textit{prohairesis}, e.g.\ \Discourses, I.22.10), or what he calls the ``use of impressions'' (e.g.\ \Discourses, I.1.7, I.12.34). The latter includes the judgement of good and bad, as well as judgement of (moral) right and wrong (\Discourses, III.22.42). For Epictetus, an impression (\textit{phantasia}) seems to be what we would call ``percept(ion)'', although according to \citet[Sec.~5.1]{LongEpictetus} it encompasses  the ``thoughts and states of consciousness in general''.  It may be that for Epictetus the use of impressions and the will are two sides of the same coin, as discussed at length by \citet{girdwood1998innovation}, since the correct use of impressions is necessary and sufficient for the correct use of the will (\Discourses, I.1.12, I.30.4, II.1.4). \label{prohairesisfn}}}

We seem to actually have two different kinds of uncontrollability here. First, the uncontrollability of the outside world as emphasized by Epictetus; and second, the uncontrollability of the mind as emphasized by the Buddha. The uncontrollability of the outside world is easy to understand, and its causes are rather obvious. The agent has limited strength: it probably cannot lift a mountain. It has limited locomotion: if it is designed to move on wheels, it probably cannot fly. If it lives in a society, it has limited means of influencing other agents.

What is less obvious, and my focus here, is that there seems to be so much uncontrollability regarding the mind. We have already seen examples where control of the mind is lacking, as in the case of interrupts and wandering thoughts; the dual-process structure of the mind creates further conflicts and reduces control. Therefore, the question arises whether there is some \textit{general principle} behind all of those manifestations of uncontrollability.

\section{Necessity of parallel and distributed processing}

The basic idea here is that the lack of control of the human mind is fundamentally based on one property of the brain: parallel and distributed processing.\index{distributed processing!and uncontrollability}\index{parallel processing}
That is, 
there are many processors, or neurons, processing the information at the same time, and to some extent independently of each other. If there are many processors working independently, each of them cannot be in control of the agent's actions: there has to be some kind of arbitration, at the very least.
Modern AI also uses such parallel and distributed processing, in particular in the form of neural networks. Both the brain and neural networks in AI are in this way  fundamentally different from an ordinary computer, which typically uses serial processing in a single processor.\footnote{In practice, a personal computer or a mobile phone would not usually have just one single processor, but a small number of them, typically less than ten. For example, the display would be supported by a separate processor, a graphics processing unit. Merely for the purpose of keeping the discussion simple, I will assume an ordinary computer has just a single processor.}

While these properties have been mentioned in earlier chapters, we have not really considered the question of \textit{why} parallel and distributed processing happens. From a biological perspective, we need to find some evolutionary justification for why the brain is parallel and distributed, and from a computer design perspective, we need to explain why such processing would be useful. Perhaps we can answer both questions if we simply find some fundamental computational advantage in  parallel or distributed computation.

\subsection{Failure of Moore's law and necessity of parallelization}

\index{parallel processing!necessity in computers}
Let's first consider the question of parallel processing from an AI viewpoint: What is the point in using many processors? If you want to speed up your computations, why not just get a single processor which is, say, a hundred times faster, instead of putting together one hundred more ordinary processors that compute in parallel? Obviously, there is a limit to how fast processors you can buy for an AI. Perhaps you need faster computation than what is given by the fastest single processor available today. That is why all the supercomputers in the world are highly parallel; they are collections of thousands of processors. That is the only way to increase the computational power to record-breaking extremes.

On the other hand, if you're really lazy, you might be tempted just to wait.
We all know that the technology behind the processors has been developing at an enormous speed. The famous Moore's law\index{Moore's law} states that the computing power of a processor doubles every two years.  
This may lead to the impression that there is really not that much reason to go through the trouble of parallelization: if the fastest processor is not fast enough, just wait a few years, and it will be. If this logic were true, it would also mean that there may not be any fundamental reason why computation in AI needs to be parallel, since the power of a single processor seems to grow exponentially and without limit.

Yet, there are fundamental reasons why really efficient computation may not be possible at all without parallel computation, and why, in fact, Moore's law is not true anymore. 
One reason is that making processors faster is to a large extent driven by making them smaller. A smaller processor means shorter delays in transmitting the information inside the processor. Such miniaturization cannot go on forever because at some point, you get too close to the level of single atoms, and even the laws of physics  change in the sense that quantum phenomena start appearing.\citenew{theis2017end} 

A more practical problem is that due to complicated physical phenomena, faster single processors use much more energy than a set of slower processors with the same total computational capacity.\citenew{markov2014limits} Energy is obviously expensive and cannot be used in unlimited quantities. Moreover, such an increase in energy consumption has another, surprising effect, which is that the processors heat up very quickly, and keeping processors cool is increasingly becoming a problem. If you design a new processor which is ten times faster than your current one, the power consumption and the heat generated are usually much more than ten times larger.\index{energy consumption}

So, these are convincing reasons why it is necessary in AI to use many processors in parallel. In fact, the speed of a single processor (``clock rate'') even in mainstream computers has not been really increasing since around 2005. The overheating problem became so serious that faster processors became impractical to use.\citenew{markov2014limits,gorder2007multicore} Since splitting the computations into many processors generates less heat, manufacturers started putting together several processors on a single chip--- the processors are now called ''cores''. The number of cores in an ordinary computer is still small, though, so this is very far  from the massively parallel case seen in the brain.

\subsection{Parallelization can be hard}

Thus, the great promise of parallel processing is that it can be much faster than serial processing, given the same budget of energy, or, indeed, money. However, there is a problem. If you have one hundred processors that process the same information at the same time, the processing could be, in principle, a hundred times faster. But that only happens in an ideal scenario where the computations are such that they \textit{can} be parallelized, i.e.\ they can be simultaneously performed on one hundred processors without any problems. Some problems can easily be parallelized, while others are more difficult, perhaps even impossible. Programming parallel systems needs special algorithms, as well as specialized expertise.\index{parallel processing}

Consider a problem of finding a small object, say a single very black pixel, in an input image. (Suppose for simplicity there is only one such object in the image). You could have a single serial processor scanning the image pixel by pixel. That might take, say, 100 microseconds (one microsecond being one-millionth of a second). On the other hand, if you have 100 processors, you could split up the image into 100 regions, and tell each processor to search for the pixel in one of the regions, and then report to a central processor whether it was there or not. That should not take much more than 1 microsecond. This problem is easy to parallelize, and the speed-up (100x) is basically the same as the factor by which you multiplied the number of processors (100x).
A neural network, whether in AI or in the brain, can do such computations massively in parallel, and thus incredibly fast. This is one of the reasons for the impressive behavior of the human visual system, and the success of neural networks computer vision tasks.\footnote{To take another example:
Tree search, which is essential planning, can also be parallelized, although it is a bit more difficult. After a few steps in the search tree, you can distribute the different branches to different processors, and each processor can search further in one of the branches. Intuitively, this would be like a boss assigning five different scenarios to her employees in a planning exercise. Each employee gets one scenario which each start from different assumptions, corresponding to the first branchings of the search tree. After the initial hurdle of formulating the scenarios (that is, building the initial branches of the tree), the parallelization is easy.}\index{parallel processing!in vision}\index{vision!as parallel processing}

Then there are tasks that are really difficult to parallelize. This is generally the case when you need to compute an intermediate result before proceeding further. As an intuitive example, consider building a house with rather traditional methods. You first have to build a foundation, and let it dry. Then you build the walls, and finally, set the roof. Suppose you had an unlimited number of builders that you can use; telling them what to do is like trying to parallelize computation. Now, the problem is that you cannot meaningfully divide the builders into three teams so that one of them sets the roof at the same time as another group lays the foundation! Also, if you really have a huge number of builders, they would not even fit on the building site. So, parallelization can be tricky. 

Optimization by a gradient method is an example of something that is typically considered difficult to parallelize because you need to do it step by step, and each step needs the result of the previous step. Yet, a lot of effort has been spent in computer science research to figure out methods that enable parallelization of such algorithms, sometimes quite successfully.\citenew{zinkevich2010parallelized,recht2011hogwild} With a lot of intellectual effort and ingenuity, it is possible to parallelize even  seemingly impossible problems. However, such parallel methods can be quite complicated.\index{gradient descent!stochastic}

The fact that some computational problems are hard to do in parallel while others can be parallelized very efficiently is part of the reason why ordinary computers and the brain are good in very different things. The brain is particularly good at vision, for example. Vision can be rather easily parallelized, as was seen in the simple pixel-finding example above, and indeed the best AI solutions to vision have imitated the brain using neural networks. On the other hand, ordinary computers are very good at logic-symbolic processing, which typically happens step by step, as discussed earlier.

But what is the evolutionary import of these considerations---does it make sense to claim that the brain is massively parallel because of the above-mentioned reasons related to the clock-speed of processors? Certainly, the constraints in building an intelligent system with biological hardware are very different, and the logic above may be mainly relevant for AI. What it actually shows is that progress in AI seems to need computers which are more and more similar to the brain. Yet, it is possible that the massive parallelization in the brain might have some relation to the energy-efficiency considerations that we just saw.

\subsection{Distributed processing reduces need for communication}

\index{distributed processing!necessity in computers}
The second question is why distributed processing is needed. Distributed processing is different from parallel processing in that the emphasis is on different processors working independently with as little communication as possible. Distributed computing is important, even necessary, simply because communication is often quite expensive. In the brain, most of the volume actually consists of white matter, which is nothing else than ``wires'' (called ``axons'') connecting different neurons. Those wires take up much more space than the actual processing units. So, the sheer space available in the head strongly limits the connectivity of brain neurons.\citenew{zhang2000universal,hari2017brain} In addition, communication consumes energy which is, again, another limiting factor.

What makes achieving full connectivity particularly difficult is that the number of possible connections between processing units grows quadratically as the number of processing units grows. If you have a million processors, and you want to build connections between all the possible pairs, you need almost a trillion (1,000,000,000,000) wires (assuming each wire can transmit information in one direction only, as  happens in the brain). So, the amount of connections easily becomes a limiting factor, and it is important to perform the computations using minimal information transfer between the processing units, by judiciously designing the algorithms as well as the connections between the different areas.\citenew{bullmore2012economy} 

This is the central point about distributed processing: When communication between the processors is expensive, special solutions are needed. In AI, there is a thrust to distribute AI computation to smartphones that collect the data in the first place, so that the amount of data they transmit to each other or any central server would be minimized.\footnote{Reviewed by  \citet{xu2020edge}, but see also \citet{corneo2021surrounded} for a critique of such distribution in the current infrastructure.}
In the brain, part of the solution is that processing is very clearly distributed on the level of large brain areas. There are areas responsible for processing visual input, areas for processing auditory input, areas responsible for moving the muscles, areas for spatial navigation, and so on. Each of these areas does its computations relatively independently. That is possible partly because they get different input (visual vs.\ auditory), and partly because they need to solve different computational tasks (object recognition vs.\ moving muscles). The communication \textit{between} those areas can then be strongly limited, and less wiring is needed. 

Distributed processing will create its design problems, just like parallel processing. Some tasks are easy to distribute over processors, while others are less so. Again, neural networks are an example of processing which is highly, even massively distributed, and clearly works well in applications such as sensory processing of images and sounds.
Considering the example of finding a small object in an image described above, it is easy to see that the computation described is also strongly distributed since the 100 processors each get their own input and then do their computations with no communication between them needed.

\section{Central executive and society of mind}

\index{central executive}
The logic above suggests that sophisticated intelligent agents may have to be a collection of relatively independent parts or processors--and that is certainly the case in the brain. The resulting computing system is very different from the view we intuitively have of ourselves. We tend to think of ourselves as serial processors because much of our inner speech and conscious thinking is serial. Speech is inherently serial because the words follow one after another in one single ``train'' of thought. But such introspection, based solely on what we can consciously perceive, is quite misleading.

\index{society of mind metaphor}
A simple metaphor for illustrating the counterintuitive properties of a parallel and distributed system is the ``society of mind'': the different mental faculties are compared to human individuals that together constitute a society which is precisely the mind.\footnote{This is a rather liberal interpretation of Marvin Minsky's original idea \citep{minsky1988society,singh2012examining}. For Minsky, the individuals (which he calls ``agents'') are very simple, more like subroutines in a computer program, as opposed to humans.}
One individual (or processor) is monitoring, say, the state of the bowels, and another one is, independently, responsible for recognizing the identities of faces whose images are transmitted by the eye. Those processors are like human workers with well-defined, separate tasks. Each one may be active much of the time, thus working in parallel. In line with the computational arguments we just discussed, it may also be intuitively clear that it is important that the different individuals mind their own business most of the time, focusing on their own part of the work. Thus, they only interact if it is really necessary, with minimum communication; thus, the operation is distributed.   This metaphor is trying to counteract the intuitive impression we tend to have that the mind is a single, serially processing entity which would be difficult to divide into parts.

Now, returning to the question of control, consider whether it is possible that one of the independent processors is actually in control of all the others.
Psychological theories often use the term \textit{central executive}  for that part of the mind which is supposedly in charge, controlling the rest.\citenew{baddeley1996exploring}  
At first sight, having such a central executive sounds like common sense. 
The brain has many sensory processing systems (vision, audition, etc.), it can send commands to a multitude of muscles to execute actions, and above all, it has complex information-processing capacities in terms of planning and learning. It would seem that such a system must fall into complete chaos unless there is one area which controls the others.
That would be the central executive,  a brain area that controls all, or at least most, of the other areas. It would integrate information coming from them and, in return, send processed information and commands to each of them.
In the society of mind metaphor, this
would correspond to a leader of the society that tells all the
individuals what they should do.

It could be argued that having a single area to control all the others is to some extent in contradiction with the whole point of distributed and parallel processing. The central executive would need to have particularly great processing power, and it would need to receive a huge amount of information from all the other parts of the whole system. Thus, both the two bottlenecks discussed above, processing speed and communication capacity, would resurface---but we will see below that this is not really the case.

Designing such a system with a central executive is not very different from designing different decision-making systems in a human society or organization. 
If there is a single leader, she must inevitably delegate a lot of power to others (say, ministers) in order to reduce the processing power needed by herself. Then, the leader is strongly dependent on the information passed on by the ministers; the leader does not have enough time to make decisions on all the details. So, the power of the central executive is limited due to the limitations on the computational power of a single processor.

On the other hand, if there were a central executive, what about wandering thoughts, emotional interrupts, or habits?\index{interrupt theory}\index{wandering thoughts!and central executive} Is the central executive just watching when the whole system is hijacked by the fear elicited by the sight of, say, a spider? We argued in earlier chapters that emotional interrupts are useful for evolutionary purposes, so the leader might actually not be very unhappy about that. But interrupts, by their very nature, cannot be prevented, not even by the central executive. Is there any point in calling such a leader the central executive if she is not really controlling everything that happens? What if you eat chocolate because you have a habit of doing it every day (in addition to an irresistible desire, perhaps), even though one part of you knows it is bad for you in the long run---who actually made that decision? 

This logic has led many to the proposal that in the human mind and brain, there is no central executive, or, metaphorically speaking, {\it the society of mind has no leader}. That is,  there is no part in the mind that controls the rest, nothing that controls everything else that happens in the society.\citenew{metzingerbook,eisenreich2017control}  The society is fundamentally a collection of relatively independent actors. This means very concretely that there is no particular part of the mind or the brain that would control our thoughts, feelings, or desires: they just come and go depending on a complex interaction between different brain areas. Each part of the mind can propose its own mental actions. One part of the visual system might tell the motor cortex: ``Let's move the eye gaze to the right since there seems to be something interesting there'', but at the same time, the replay system might insist on replaying a past episode while ignoring whatever may be happening in the outside world. The result may be a bit chaotic, and the appearance of, say, wandering thoughts would not be surprising. To the extent that we define the self as the central executive, this would mean that there is no self, in line with Buddhist philosophy, for example.\index{no-self!as no central executive}\index{self!lack of|see{no-self}}

While such a philosophy is fascinating, it has to be pointed out that there are also neuroscience results claiming that some brain regions in the prefrontal cortex are actually the central executive.\footnote{In human neuroimaging literature, the existence of a central executive network is more or less accepted by many authors \citep{koechlin2007information,sridharan2008critical,botvinick2014computational,marek2018frontoparietal}. The functions attributed to the central executive may vary, and many authors indeed talk about a number of different ``(central) executive functions'' without claiming that they are performed by a single entity, whether brain region or network, as discussed by \citet{miyake2000unity,diamond2013executive}. For this book, the main executive function discussed is the control of actions and thoughts, as treated in the following sections; inhibition of ``impulses'' and automatic behavior such as interrupts is a fundamental instance. \citet{morales2020neural} further argues that prefrontal cortex is necessary for consciousness. (See \citet{teper2013meditation} for a discussion on how  ``self-control'' could be improved by meditation.)\index{meditation!increasing self-control}} 
Moreover, in the design of distributed computing architectures, it is well-known that having some kind of a central processor actually makes communication easier. The point is that there is a good compromise to be found between the two extremes of completely distributed computation and computation in a single processor. Such a compromise can in fact be found in computation which is mainly  parallel and distributed, but, crucially, includes a central processor that \textit{coordinates} the computation, which is still mainly performed by the other processors. In the example above, with a million processors, we saw that a fully distributed system might need a trillion wires to connect all the processors with each other. But suppose that all the communication happens \textit{through} a central processor, which further selects and processes the information to be transmitted to each of the other processors. Then, all that is needed is wires from each processor to the central one and back (figuratively called a ``hub-and-spoke'' architecture), which means about two million wires, enabling a reduction by several orders of magnitude. Still, the computational power of the system need not be restricted by the central processor if it is skillfully designed to ``delegate'' the hard computation to all the processors and only take a coordinating role. Such architectures are currently of great interest in artificial intelligence.\footnote{For example, ``federated learning'' has recently emerged as such a paradigm \citep{kairouz2019advances}.}\index{learning!federated}\index{learning!distributed}

In fact, the whole dichotomy between a powerful central executive and no central executive is a bit artificial. There can be varying degrees of control that a central executive is able to exercise. While it is not possible to say much with certainty on this topic, the reality in the brain may well be that there is a relatively weak central executive that controls some things to some extent, perhaps many things to a limited extent, but it does not control everything. It may be in control a lot of the time, but not when emotional interrupts, wandering thoughts, or similar processes take control of the mind. 
Thus, while parallel and distributed processing is inherently without central control, it may be advantageous to introduce some limited form of central executive, and this may turn out to be the best description of what happens in the brain.\footnote{Further theoretical neuroscience arguments on this question can be found in \citep{botvinick2014computational,rueda2004attentional,baumeister2007strength}. }

\section{Control as mere percept of functionality}

Yet, what is undeniable is that I clearly \textit{feel} that I can control my body and do things such as raising my arm. A central executive is often intuitively assumed to exist based on exactly such a feeling of self, or a feeling of control. But 
why should we assume that there is a central executive simply because it feels like there is control?\index{control!as percept}
The feeling of control is just another form of perception, and as we have seen, perception may not be accurate. Perception follows certain rules outlined in Chapter~\ref{perception.ch}. It is usually based on incomplete information which has to be combined with prior assumptions to arrive at a conclusion, and this conclusion or inference is what we perceive. Mistakes do happen in this process.

The perception of control in the brain seems to be based on predictions---like so many other things in the brain. Every time you engage in any action, your brain tries to predict the outcome of the action.
In particular, when the brain sends detailed motor commands to the muscles, it uses an internal model to predict how the limbs should move as a result.
The brain then computes an error signal, comparing the predictions with the actual outcome. In humans, small errors in such predictions are actually quite common because of constant physiological changes in your muscles due to fatigue; or\index{error signalling!in control}
it could be that you are holding something heavy in your hand, which increases the force required to lift the arm. Computing the prediction errors is useful since they enable the brain to learn or adapt its motor commands to such changing circumstances.\footnote{\citep{kording2007dynamics}. This is an example of the principle of feedback for successful control and action.\index{feedback control} Feedback is a general principle in action selection, which is important if there is uncertainty in the world.  Just finding the best path to the goal is not sufficient if the environment is uncertain and may change. For example,  if you calculate the best possible path to a restaurant, that is usually fine, and you can just walk there. But unexpected things might happen: A road might be blocked by a delivery truck; there might be construction work. This is another limitation of purely planning-based action selection in a changing, uncertain environment. } 

Now, if the prediction error is small (the actual outcome of the action is not very different from the  prediction), you feel that you generated the action, and you are in control, according to current thinking in neuroscience.\citenew{haggard2012sense,wolpert1998internal,wen2022sense,choudhury2006intentions} This is the computational mechanism underlying the perception of whether you are in control.
In contrast, if the errors are very large, the feeling of control is disturbed, and various pathological symptoms may arise. You may even feel the arm is being controlled by somebody else (by ``them'', or by ``spirits''), as typical of some schizophrenic patients.\citenew{spence1997pet,frith2012explaining} 

Based on his extensive psychological experiments, Daniel Wegner\footnote{\citep{wegnerbook,wegner2003mind}; see also  \citep{hommel2013dancing,pockett2009does}} proposed a related theory: the perception of control is simply based on one part of your brain observing a correlation between two things, which are the formation of an intention to act (intention being used here in the ordinary sense of the word, not in the AI sense as usually in this book) and the action actually taking place.\index{control!as illusion} If the action happens soon enough after the formation of the intention and the action happens as you intended (and you cannot explain the action in any other simple way), the brain concludes that ``you''  actually performed the action out of your own ``free will''.  A strong correlation between intentions and outcomes is not very different from small  prediction errors, and thus this psychological theory is very much in line with the neuroscience results cited above. Interestingly, just like visual neuroscientists who construct optical illusions, Wegner then devised clever experiments where the perceptual system makes the wrong conclusion about control, thus showing that the feeling of control can be fooled like any other perception.

Any of these computations are actually quite simple and could be easily implemented in a robot. A robot can assess whether it is able to control its arm by comparing the results of its motor commands and the actual outcome. Suppose some kind of central processor sends a command to the joints in the arm that the arm should be lifted by 10~cm. A couple of seconds later, the input from the camera (or input from specialized sensors in the joint) tells the central processor that the arm was, indeed, lifted by 10~cm. The central processor then concludes that it is in control of the arm. 

This logic demystifies the concept of control, which is no longer anything deep or philosophical. The perception of control by the robot above is due to computations of a rather practical nature. In fact, any agent should have a model of what parts of the world it can control (e.g.\ its limbs) and which parts are outside of its control (e.g.\ mountains). This is in contrast to my everyday perception that it is \textit{I}, or myself, that is in control, which is the result of a very complex inference process, and possibly exaggerated, misleading, or even false and illusory. Our everyday perception of control \textit{by ourselves} is, therefore, no proof for the existence of a central executive, or ``self'', that controls actions. One could say that our perception only indicates that \textit{there is} control in the simple sense of the limbs moving as expected, but it does not necessarily mean that there is any particular entity that is in control. In other words, our feeling of  control simply means that certain systems are working in a predictable way, correctly and in harmony with each other; in particular this is about the decision-making system, the  motor system, and the actual limbs (or ``actuators'' as they are called in robotics). 

\pagebreak

\subsection{Free will}

\index{free will}
Free will is a celebrated and highly controversial concept in Western philosophy---the idea that you decide your actions ``yourself'', that is, your actions are not merely a function of external circumstances, such as your past or other agents. 
Free will is very closely related to control and feeling of control: most of neuroscience uses the terms almost interchangeably. There are some nuances, though: talking about free will emphasizes your capacity to decide what you will try to do (e.g.\ choosing a goal), while talking about control emphasizes your ability to actually do it (e.g.\ reaching the goal).
A very clear difference is, moreover, that free will is almost always considered a conscious phenomenon, while perception of  control need not be, as we saw above: Even a completely unconscious robot would benefit from knowing which events are due to its own actions and which parts of the world it can control. 

Philosophers have been debating about free will for hundreds of years. Democritus claimed already around 400 BCE that everything, including humans, consists of atoms, and follows strictly deterministic causal laws, thus excluding any free will. A bit earlier in India, the Buddha had debates against philosophers who held similar, strictly deterministic views.\footnote{Early Buddhist philosophy is actually often interpreted as deterministic, based on the Buddha's emphasis on causal chains (e.g.\ \SN{12.12}; see also page~\pageref{causality} below), but he also admitted the existence of ``an element or principle of initiating an action'' (\AN{6.38}), which sounds a bit like free will. See \citet{federman2010kind} for details.}

In a famous series of experiments,  Benjamin Libet recorded an EEG response that is known to precede any action decision. The results showed that conscious experience of the decision started up to half a second \textit{after} the beginning of the EEG response. From this, it is tempting to conclude that consciousness cannot cause the action decision, and hence there is no (conscious) free will. The EEG presumably measured some unconscious processes which started the decision-making process long before any involvement by conscious processes. Libet's own interpretation, though, was that consciousness could still participate in the action decision  by having the possibility of ``vetoing'' any decision that the unconscious circuits were trying to implement. This would imply some weak version of free will, but his interpretation is controversial.\citenew{Libet1983time}

Some would argue that denying free will may be dangerous: 
people have to believe in free will in order for our moral systems to work. If people don't believe in free will, they might not feel they have any moral obligations and might behave just as they please. Our justice system in particular is based on the idea of free will: if it can be proven in court that a murderer acted without free will, say, because of a brain tumour, that will usually lead to a reduced sentence. This is, of course, not saying that there \textit{is} free will, just that it may be \textit{useful} to think that there is one.\citenew{vohs2008value,baumeister2009prosocial,roskies2006neuroscientific} 

One of the most influential psychologists in the 20th century, B.F.~Skinner,\index{Skinner} had a more computational viewpoint. He thought human behavior is simply determined by rewards and punishments. That is also where the ``moral'' behavior comes from; no special metaphysical beliefs are necessary.\index{morality} Reward people for good behavior and punish for wrong behavior; that is all that is needed to make them follow moral rules, in Skinner's view.
From the perspective of this book, I can partly agree with Skinner on the importance of learning from the environment, where learning, as always, includes evolution.

\section{Philosophy of no-self and no-doer}

\index{no-self!Buddhism}
Let's go back to the ``no-self'' quote by the Buddha that  we saw earlier (page~\pageref{anattalakkhana}). In Buddhist philosophy, it is the historical basis of a celebrated doctrine claiming that there is no such thing as the self. This is clearly a much more general idea than the mere claim that there is no entity which is in control, but in this chapter, we focus on the aspect of control and free will.

We can now recapitulate the ideas in this chapter in view of justifying some claims regarding the existence of self.
First, if the brain, and thus the mind, is composed of many different processors all working simultaneously and to a large extent independently of each other, how could we speak of a self? If we admit there is no central executive, that is one form of ``no-self'': there is no particular part of the mind that actually is in control and could be called self in that sense. (It is not clear if it is neuroscientifically correct to deny the central executive, but let's admit it for the sake of argument here.)\index{no-self!neuroscience}
The \textit{conscious} part of the mind does not control actions according to neuroscientists such as Libet and Wegner, thus contradicting our everyday perception that we decide actions on the conscious level. 
Decisions seem to be actually taken by various unconscious neural networks, and it may be difficult to point out any single entity making the decision. 

Some parts of  Indian philosophy actually formulate a more specific doctrine of ``no-doer'', which means that there is nobody that ``does'' anything in terms of taking the actions---or that at least, it is not ``you'' that does anything.\index{no-self!no-doer} Instead of ``you'' making conscious decisions and being in control, your body and mind are constantly on some kind of autopilot, and your consciousness is merely observing it all.\footnote{See e.g.\ the classic Theravadan Buddhist meditation manual  \textit{Visuddhimagga} (Chapter XIX,20); or the Advaita Vedanta teacher \citet{maharaj1982seeds}. The Buddha himself may not have formulated no-self in exactly this way; he even seems to argue against it in \AN{6.38}; see \citet{harvey2009theravada}. \add{Such a no-doer philosophy is also in strict contradiction with Stoic thinking, where the will (\textit{prohairesis}) is the one thing we can control (possibly in addition to ``use of impressions''), see footnote~\ref{prohairesisfn} in this Chapter.}} 
The points above give some credence to such a variant of no-self philosophy.

But we may go further. If it is not your conscious self that decides, is it even necessarily your neural networks? 
In our framework, we could say that control is ultimately exercised and decisions are ultimately taken by the \textit{input data} that the agent learns from.\index{control!by input data}
Our computational models have assumed that our actions are determined by past input data, together with the design of our learning and inference machines---even though the mapping from input data to action can be extremely complex and impenetrable.\footnote{This is not the same as the Skinnerian viewpoint because most of the time, nobody is explicitly and purposefully feeding all the data into your brain to ``train'' you; most of the data is just passive observation, often with no rewards involved.} 
From this viewpoint, nobody at all is in control, and there is no free will even in some unconscious form---while it is possible to say that \textit{there is} some control in the specific sense of sufficient predictability. Furthermore, the existence of a central executive in the brain becomes irrelevant: if it exists, it is still only a vehicle for the evolution and the input data to steer our thinking and behavior. Even a brain with a strong central executive could be seen as having no self: even though a central executive was seen as the hallmark of a self earlier in this chapter, if we now see all its actions as simply following from learning based on input data, it may not actually qualify as a self.\footnote{It is obviously important to consider the exact definition of free will. In one radical viewpoint, freedom of will is a matter of not being physically or psychologically forced or compelled to do what one does. This viewpoint is called ``compatibilism'' in philosophy since it implies that free will is compatible with determinism \citep{strawson1998free}.
  Consider a basic robot. If it decides to raise its arm, is there any physical constraint that would prevent it from doing so? 
  Are the computations only due to its own sensory input (in its cameras or else) and are the computations made in the processors inside the robot? 
  Here, the idea of free will is formulated in terms of what causes the agent's actions: Is it solely information-processing inside the agent---in a modern formulation---or is something outside it influencing the decision? 
  If so, even such a robot could be said to have free will. Humans would certainly have free will. It may not be the conscious self that decides actions, but the neural networks in the brain. Still, as long as the neural networks are inside the human skull, it is the human that decides and controls its actions. 
Yet, many find such a definition of free will questionable. These include all the schools in the philosophy of free will other than the compatibilists. What I described in the main text is rather similar to the ``pessimist'' school. Namely, an obvious counterargument to the compatibilist definition is that is depends on the time scale used: we should look back in time, trying to find the original reasons for your actions. As I argue in the main text, fundamentally, the robot's or human's actions are just a result of its programming/evolution and, especially, the input data, so there goes free will.
}

Ajahn Brahm, a famous meditation teacher, once said that when he sits down to meditate, he always remembers the instructions of his own meditation teacher in his head; thus, it is not really Ajahn Brahm who meditates, it is his teacher---or, if I may, it is the input data he received from his teacher.\footnote{This example also points out how a lot of the input data comes from social interaction, which is outside of the scope of this book. \add{It could further be argued that the input data is not processed in the same way in all individuals due to genetic variation and possibly some other biological differences.}}\label{datacontrol}\index{social interaction!as input data}

\chapter{Consciousness as the ultimate illusion} \label{consciousness.ch}

Why do we have conscious experiences? This is one of the deepest unanswered questions in modern science. It is not even quite clear what the whole question means and how it could be formulated in a rigorous, scientific manner. One thing that is clear, however, is that consciousness is somehow related to suffering. Some would even claim that in a strict sense, there can be no suffering without consciousness. 

In this chapter, I try to shed some light on the nature and possible functions of consciousness. I consider two different aspects of consciousness: it can be seen as performing some particular forms of information-processing, or it can be seen as a subjective experience. I provide a critical review of the main theories concerning these two aspects. In particular, I explain how consciousness is related to mental simulation and the self, and as we have seen, those play an important role in suffering. This leads to some old, but still radical, philosophical ideas about the nature of our knowledge of the world and how it relates to our consciousness. Ultimately, I argue how changing your attitude to consciousness may actually have a strong influence on your suffering, a theme that will be further elaborated in later chapters.

\section{Information processing vs.\ subjective experience}

\index{consciousness} 
The main problem we immediately encounter in research on consciousness is the difficulty in defining the terms involved. ``Consciousness'' has different meanings to different people and in different contexts.\citenew{sep-consciousness} For our purposes, we can divide the concept of consciousness into two aspects.
First, there is the information processing performed by human consciousness. This is something we might understand based on AI, since information processing can usually be programmed in computers. One approach is to ask what the computational function, or utility, of consciousness might be in humans; these are relatively well-defined scientific concepts and questions. This approach is fine as long as we are content to consider consciousness as another form of information-processing, or computation.
The second, more difficult aspect of consciousness is the experience. That is, the conscious ``feeling'' which is specific to myself, i.e., subjective. Its existence is so obvious that it is rather neglected by most people.

When you look at the text in this book,  several quite amazing things are happening; they can be roughly divided to information processing and experience. Those related to information-processing have been discussed earlier in this book. Light enters your eye, generates electrical signals on the retina, the signals travel into your brain, and some incredibly intricate information processing takes place, allowing you to recognize the letters and even transform the letters into words. However, all that is simply information processing, and it can soon be programmed in a computer---in some more rudimentary form, it is possible even now.

\index{subjective experience}\index{consciousness!phenomenal}\index{qualia}
But in addition to such information-processing, there is something else: you have a conscious, subjective \textit{experience} of the book, the letters, and the words. Somehow, almost magically, the book appears in some kind of a virtual reality created by your brain. We tend to think that this is normal since the book is there, and we simply ``see the book''. But in fact, the conscious experience is not somehow \textit{in} the book, and it does not somehow automatically come \textit{out of} the book. The experience, the awareness of the book is created by some further mechanisms which we simply don't understand yet.
This experiential aspect is called \textit{phenomenal} consciousness. Philosophers use the word \textit{qualia} in this context: the conscious ``quality'' of the book being seen, ``what it is like'' when the book is consciously experienced.  Or, as more poetic narratives would have it, it is the ``redness of a rose''. It is not information processing but something more mysterious.

It is this phenomenon of subjective experience, or qualia, which is the main topic of this chapter. It is also the main meaning in which I use the word ``consciousness'' in this chapter; ``awareness'' is used in exactly the same meaning, and so is ``conscious experience''.

\section{The computational function of human consciousness}

Now, what is the connection between these two phenomena: information-processing and consciousness? Conscious experience is certainly not just one form of information-processing, but the connection is extremely difficult to understand. 
In fact, consciousness must have \textit{some} connection with information-processing: the qualia of the rose must be based on processing of incoming sensory input, even if most of that sensory processing seems to be unconscious. Let us assume, in the following, that part of the information-processing in the human brain is conscious, in some sense to be elucidated. This is such a typical assumption that it is often not even made explicit.\index{consciousness!utility}

Let us then try to understand what we can say about the function, or utility, of such \textit{conscious information-processing}.  Taking a more neuroscientific approach to the question, one can first ask: What are the evolutionary and computational reasons why certain animals, such as humans, have consciousness? We assume here that consciousness is a faculty that is a product of evolution---but strongly influenced by culture, of course.
It is quite difficult here to ignore the experiential part of consciousness and consider  information-processing only. If we say that an animal, or an AI, is conscious, it seems to almost necessarily mean a conscious experience: we wouldn't even know what it means to say that an animal is conscious if it does not have conscious experience. So, in a sense, the question is almost necessarily about the \textit{computational function} of human conscious \textit{experience}, and whether it can be explained by evolutionary arguments. I will next review a number of proposals.\footnote{For discussions on the consciousness in general and the computational function in particular, see e.g.\ \citep{baars1997theater,chalmers1996conscious,seth2009functions,sep-consciousness,dehaene2011experimental,lau2022consciousness}.} 

Investigating wandering thoughts actually leads us close to consciousness, because ``thinking''\index{thinking} is often considered the hallmark of consciousness. More precisely, the fact that we can reconstruct a vivid image about past or future events in our minds, while ignoring the present sensory input, is a remarkable property that seems to be closely related to consciousness.  Some investigators actually propose that one of the main functions of consciousness is such \textit{simulation}, which is also called \textit{virtual reality.}\index{simulation}\index{virtual reality} That is, consciousness allows us to consider different scenarios of what might happen in the future, and what would be the right things to do in those circumstances. Planning crucially needs the capacity to simulate the results of future actions, and indeed in the case of wandering thoughts, we already talked about simulation. Such simulation obviously would be useful for survival and reproduction, and thus favored by evolution.
A special case of such simulation is dreaming, which creates a virtual reality that is particularly far removed from current reality. Dreaming often includes simulation of threatening situations, i.e.\ situations in which it is important to know what to do to avoid harmful consequences.\citenew{revonsuo2000reinterpretation,hesslow2002conscious}

However, I think we should not too easily conflate consciousness with thinking or simulation. What we see is a correlation between a certain brain function (namely, simulation) and consciousness, but it is difficult to say whether consciousness is really essential for such a function. As we saw in Chapter~\ref{replay.ch}, modern AI uses planning and even something like wandering thoughts, simulating events that happened in the past or might happen in the future. Yet, nobody seems to claim that replay or planning would make a computer conscious. Such a claim seems absurd to most experts because claiming that a computer is conscious is usually interpreted as having phenomenal conscious experiences; but they are very unlikely to be produced by such simple computations as replay and planning.

Another possible function of consciousness is \textit{choosing actions}.  We typically have the feeling that we consciously decide what we are going to do, an experience of  free will. You may think that you decided to read this book; perhaps you decided to read this particular sentence. But did you actually decide how you move your eyes from one word to another? What do we actually decide on a conscious level? As we saw in Chapter~\ref{control.ch}, consciousness may not have \textit{any} role in the control of actions; the feeling of free will and control may be deceiving. It may very well be that actions are entirely decided by unconscious processes. After all, that seems to be the case with many animals (if we assume most of them don't have consciousness), as well as any robots and AI that exist at the moment.\footnote{For a viewpoint in which consciousness is the ``reason'' for most actions, see \citet{cleeremans2021function}.}

Yet another proposal is that consciousness could be useful for \textit{social interaction and communication}.\citenew{hommel2013dancing,frith2002attention,frith2010consciousness}\index{social interaction!basis for consciousness} The contents of consciousness can usually be communicated; in fact, in psychological experiments, one operational definition of the consciousness of perception is that you can report the perception verbally to the experimenter. The utility of conscious perception, in particular, would be that this perception can be transformed into a verbal form, and communicated to others. Again, the problem is that it is perfectly possible to build AI and robots which communicate with each other without anything we would call consciousness, at least in the experiential sense.

A particularly relevant proposal for this book is that consciousness facilitates \textit{communication  between different brain areas}.\footnote{\citep{baars2002conscious,baars1997theater,dehaene2001towards}. Related to this are the ``higher-order'' theories in which consciousness depends on higher-order mental representations that represent oneself as being in particular mental states \citep{lau2011empirical,lau2022consciousness}. For example, a fear reaction would be coded in the amygdala while the feeling of fear would be coded in the  prefrontal cortex. This is a form of metacognition which will be discussed in more detail in Chapter~\ref{training.ch}.}
While unconscious processing has a huge capacity for information-processing, it suffers from the problem that the processing is divided into different brain areas whose capacity for communication is limited---as typical of parallel distributed processing. The idea here is that consciousness is the opposite: it has very limited processing capacity, but its contents are broadcast all over the brain. Consciousness can thus be considered a ``global workspace''. It could be compared to a notice board where you can put short notes (limited capacity), which will be seen by everybody in the office (global broadcasting). It is also a bit like a central executive  in the society of mind discussed in Chapter~\ref{control.ch}---one that is not particularly smart but whose thundering voice is easily heard even at a great distance. 
This links clearly to the proposal in previous chapters, where we considered a system where pain and other error signals, such as reward loss, are broadcast to the whole system. An intriguing possibility is that this could be why pain, whether physical or mental, must be conscious. Perhaps pain is so acutely conscious precisely because the broadcasting system it uses is inherently related to the global workspace of consciousness. Yet, it is not clear to me why this would necessarily require conscious experience, since distributed information processing is increasingly performed in computers as well.

I have just described several proposals which each consider highly relevant
information-processing principles. For example, inside an AI, the communication
between different processors or processes needs to be solved, and 
mechanisms related to the global workspace theory can be very useful.
Yet, in each of those cases, we have to ask whether we would say an AI with such capacities is conscious. Would it necessarily have subjective experience, if that is what we mean by ``conscious''?  We could go through \textit{all} the computational functions of the preceding chapters and ask whether consciousness necessarily has any role in any of them. To the extent that all of these are simply computations that can be implemented in an AI, they may actually not \textit{need} consciousness. (I'm here assuming that any computer we have at the moment is not phenomenally conscious, which is relatively uncontroversial.)
Therefore, I think there is currently little reason to believe that consciousness would be necessary for some particular kinds of computation, which would be impossible without consciousness.

\subsection{Consciousness as a specific hardware implementation?}

However, another viewpoint is possible:
there may be some forms of information processing that are \textit{correlated} with consciousness. It could be that some of the computational routines in the brain are always implemented using some special circuits or processes that give rise to consciousness. Such computations would then give rise to consciousness, even if in theory, it would be possible to implement them in non-conscious circuits. If we program that same kind of computation in an AI, we might then say that we have programmed the AI to perform ``conscious'' information processing. However, it may be best to use scare quotes here: the AI may be imitating processing that is conscious in the brain, but it might not have conscious experience, so whether we should call such computations ``conscious'' is questionable.\footnote{The well-known distinction between``access'' and ``phenomenal'' consciousness \citep{block1995confusion,kouider2010rich} is related to this point. In this book, when I talk about consciousness, I mean phenomenal consciousness, i.e.\ the experiential kind of consciousness, unless otherwise mentioned (or in quotation marks). Access consciousness is, in my view, an operational definition of consciousness, used in experimental neuroscience: If you ask a person whether she is conscious about X, and they reply yes, then the person is conscious of X in the sense of having access to the experience or perception of X. I find this definition of access consciousness not very relevant for the present discussion.
}

Therefore, any argument---such as I have just made--- saying that a certain computational function cannot be the actual function of consciousness because it can easily be programmed in an AI, may be missing the point.
While it may not be completely necessary to have consciousness for, say, simulation, it could still be that in biological organisms, shaped by evolution, consciousness is somehow an important part of the computational implementation of simulation, or any of the other functions above. The fact that something is easy to program in a computer, which is based on completely different kind of hardware, does not mean that it might not be very difficult to implement in the brain without the help of some, hitherto unexplained, conscious mechanisms. Thus, consciousness might be a particular ``hardware'' implementation of certain computations that are otherwise difficult to perform in the brain---whether simulation, global workspace, or else.

Yet, this is all mere speculation. We cannot exclude the completely diametrically opposed possibility, which is that consciousness is not actually part of the information processing at all. Perhaps it does not affect the computations or the contents of the mind in any way; it simply \textit{reflects} the results of the information processing. It might not even have any evolutionary utility.\footnote{  Claiming that consciousness has no evolutionary function could be seen as a special case or implication of the stance called Epiphenomenalism \citep{walter2009epiphenomenalism}, according to which consciousness has no causal effect on physical events. What I have written here is probably also compatible with the sophisticated alternative given by \citet{chalmers1996conscious}, who also addresses the obvious question of why there might be consciousness if it is not evolutionarily advantageous (his Section 3.6). Chalmers's arguments rely heavily on considering what a zombie without any consciousness would be like compared to humans; I think his zombies are comparable to AI, which I assume is not conscious. \citet{chalmers1996conscious} seems to agree on the difficulty of understanding consciousness: ``[W]hen it comes to consciousness, it seems that all the alternatives [of philosophical stances] are bad. If someone comes away with the feeling that consciousness is simply an utter mystery, then that is not completely unreasonable.''  (I am not committing myself here to any particular philosophical stance, but just exploring different possibilities.)}
Many thinkers over the centuries have proposed that consciousness in humans is only the tip of the iceberg, and most mental activities---which I call information processing---happen without consciousness.\footnote{See footnote~\ref{unconsciousimportantfn} in Chapter~\ref{dual.ch} for some historical remarks.} But here, we find an even more startling possibility: perhaps consciousness is not even the tip of the iceberg but, to push the metaphor further, a bird that flies over the iceberg, only watching it from a distance. We will see even more startling possibilities later in this chapter.

\section{The origin of conscious experience}

Next, let us consider the problem of the existence of conscious experience itself.
Most scientists would agree on the fundamental importance of understanding  the physical, chemical, and biological processes that enable conscious experience.
While it is one of the deepest questions in science, I am, again, afraid there is little we can say about it with any certainty. It is not even clear if the whole question can be approached scientifically. This is because it is difficult to make any rigorous experiments on experience, due to its subjective nature.

Only \textit{I} observe my conscious experience; you, or any neuroscientist, cannot really know what I experience.
So, how could a neuroscientist conduct experiments on people's experiences? Measuring brain activity, or looking at people's behavior are not measuring experience. Brain activity and behavior are related to and correlated with experience, but not the same thing. The closest you can get is asking people what they experience. However, they might not be able to express it verbally with sufficient accuracy or detail. In fact, if participants in experiments answer such questions, they are ultimately engaged in behavior (in the form of speaking), and, in a sense, the neuroscientist is actually only measuring their behavior (speech, in this case). 

With good reason, the problem of understanding how and why the brain creates conscious experience--including whether it is actually the brain that does that--- is called the \textit{hard problem} of consciousness research.\citenew{chalmers1995facing,chalmers1996conscious} However, let us not despair: Even if any solution may not be available, some interesting things can be said about the problem.\index{consciousness!hard problem}

To begin with, some neuroscientists think there is something special in humans that enables consciousness, and perhaps in some other mammals such as great apes as well. What it would be, nobody really knows. 
The main theories are based on observing what kind of structures human brains have, and what simpler animals like cats and dogs do not have. 
Because the brains of cats and dogs are in many ways very similar to the human brain, the relatively small differences might be related to consciousness.  

One difference between the brains of humans and ``lower'' animals seems to be the existence of a special class of neurons, called von Economo neurons. They have particularly many long-range connections to other neurons. Since long-range connections might be related to something like a global workspace, von Economo neurons have often been considered as a potential candidate for a mechanism generating consciousness.\footnote{\citep{critchley2012will,butti2013economo}}
In fact, it has also been suggested that the brain basis of consciousness might be related to feedback between brain areas, as opposed to any special kind of  processing inside each single area.\citenew{lamme2000distinct,crick2003framework} It could be that the long connections of von Economo neurons make such feedback stronger, sufficiently complex, or otherwise more conducive to consciousness.
Interestingly, apes have von Economo neurons as well, and so do elephants, dolphins, and even some monkeys, so based on this criterion, those animals at least should be conscious.\footnote{It is not very clear which animals have von Economo neurons and which don't; \citet{jacob2021cytoarchitectural} have recently claimed to  find them even in raccoons.\index{consciousness!brain basis}}

Arguably, we can use AI for studying the hard problem of consciousness. In particular, we can perform thought experiments based on the same kind of comparisons as was just done with other animals.  Starting from the assumption that AI is not conscious, part of the hard problem of consciousness is then to explain what creates this fundamental difference between humans and AI.

\subsection{How can we know something is conscious}

However, there is a problem with the argumentation above: It is based on finding animal species, perhaps such as cats and dogs, which are reasonably intelligent but have no conscious experience. Or, if we consider AI, it is based on assuming that  AI is not conscious. But how can we even know if an animal species or an AI is conscious or not? 

Some would claim that we cannot even know if other people are conscious.
We do tend to assume that every human we meet is conscious, but this is just a guess, really, without much logical basis. We are actually generalizing based on ourselves: the only human I know for sure to be conscious is myself.
 Others just move around and say things, but they could be some kind of robots for all I know; perhaps I am the only person conscious in the world.
 If I assume all other humans are conscious as well, I can only hope I'm not overgeneralizing!
This is, somewhat cheekily, called the zombie problem: it could very well be that some of the people you meet are ``zombies'', that is, creatures that look like humans and behave like humans, but do not possess any kind of consciousness. 

Leaving such wild speculation aside, we do have a real scientific problem here. In neuroscience, it has been found extremely difficult to determine which animal species are conscious and which are not.\footnote{\citep{seth2005criteria,ledoux2023consciousness,TICSconsciousness2024}. Neuroscientists have developed an interesting test called Mirror Self-Recognition \citep{toda2015animal,loth2022through}. The idea is that a mirror is introduced to the animal. After the animal has had some experience with the mirror, some red dye is applied to its face to create a small but visible spot. Many animals instinctively try to touch the red spot. But does the animal touch the real spot on its face, or its image in the mirror? If it touches the real spot, it is concluded that the animal has some kind of consciousness of itself, or at least a body image similar to what we have. Chimpanzees, for example, pass the test. However, this is of course a very indirect measure of only one aspect of consciousness, in particular self-awareness considered later in the text. (For moral implications of our ignorance of whether animals can suffer on a conscious level, see \citet{birch2017animal}.)}\index{consciousness!in animals}\index{morality!and animal consciousness}
Even considering humans, it is not easy to tell  whether people in coma are conscious.
For coma patients incapable of saying anything or making any motor responses, measuring brain activity provides the last resort for assessing their consciousness. Surprisingly, it has been found that patients who were thought to be in a completely unconscious, vegetative state are sometimes perfectly conscious: They can respond to questions like healthy humans would when they are given the opportunity to communicate with the external world by special devices, which transform brain activity to text.\citenew{monti2010willful,bruno2011unresponsive}  
So, it is actually true that we cannot always tell if even other humans are conscious.

How could we then judge whether an AI is conscious or not?\index{consciousness!in AI}
What if current AI is conscious, or will become conscious in the near future? You can find people arguing strongly for the possibility of conscious AI. Some say it is simply a question of complexity: when AI becomes complex enough, it will become conscious; the only reason why present computers are not conscious is that they are too simple in terms of their computation, in particular lacking sufficient interaction and information interchange between different processing units.  Others think an AI must have a body, i.e., it must be a robot, in order to be conscious, and consciousness is somehow created in the interaction with the world.\footnote{For reviews, see \citep{reggia2013rise,mcdermott2007artificial,chella2007artificial}; on complexity, \citep{tononi1998consciousness}, and embodiment, \citep{ziemke2007embodied}}

Fundamentally, the question of determining consciousness seems to be unsolvable because of its subjective nature: I can only know something about my own consciousness. We cannot know for sure if any animal or AI is conscious or not. Consciousness---at least regarding its experiential quality---remains a huge mystery.\footnote{However, I don't mean to be completely pessimistic about the possibility of doing scientific research on consciousness. If verbal reports (or similar information) of conscious content are combined with brain imaging in a sufficiently large number of human subjects---possibly specifically trained to perform introspection---progress can be made. The specific methodology needed is discussed by \citet{lutz2003neurophenomenology,gallagher2003phenomenology}.}

\section{Why is simulated suffering conscious?} 

\index{consciousness!and suffering}\index{subjective experience!and suffering}\index{suffering!and consciousness|see{consciousness, and suffering}}
Let us get back to the question of suffering. Consciousness is in some sense crucial to suffering: if we were not conscious of our suffering, if we didn't have the conscious experience of suffering, it would not be the same kind of suffering at all.  Suppose you have a headache but you start watching a really fascinating movie; you may cease to notice the pain at all. Somewhere in your brain there is probably some kind of activity that would usually lead to the experience of pain, but your attention is in the movie, so you completely ignore the pain. That is because when you are not paying attention to something, you cannot be conscious of it either.\footnote{The connection between attention and consciousness is complex, but it is usually assumed that we can only be conscious of something we attend to. \citet{de2010attention} review evidence for and against this assumption.\label{attentionconsciousnessfn}}   So, in some sense, the whole problem of suffering revolves around the question of consciousness. If we consider a simple animal or an AI and agree it is not conscious, is it actually meaningful to say it suffers---as I seem to have done in this book?\footnote{This problem could be contrasted with the problem of whether a computer can \textit{see}. Suppose a robot moves around in its environment, avoiding obstacles and performing some task thanks to input from its camera. Now, how would you answer the question of whether the robot is able to ``see''? Most people, including scientists working on such computers, would casually say that the computer sees, for example, it sees the obstacles. If pressed hard on what that means, they would probably admit that the computer ``does not \textit{really} see'', presumably because there is no consciousness involved. What is very interesting is that this ambiguity is not usually considered a problem: it is rare that any serious debates are conducted on whether such a robot actually ``sees'' or not. When we talk about suffering in an AI, the situation is, in principle, quite similar. However, much more heated debates can be expected on the question of whether the AI actually suffers. This lack of clarity on whether an AI can suffer seems to be much more difficult to accept than in the case of seeing.}\index{simulation}

The other day I was watching a fictional TV series in which a tiger attacked a woman. I felt scared. Was there any point in being scared? I was in my own home, just watching an electronic device produce some patterns of light on its screen. There was no real tiger nearby, no real risk of being eaten. Even if I had been in the middle of the action, it would have been on a film set. The tiger was tame; or perhaps it was just a computer animation, and there was no real tiger at all. In any case, even if I had been at the studio instead of home, I would not have been in any kind of physical danger.
What is even more interesting is that after having watched that on TV, my brain started replaying the events. Several times during that evening, I saw the tiger in my wandering thoughts. Every time, some element of fear crept into my mind. I thought: How stupid can my brain be? Why do I feel fear although there is no real tiger nearby, there was never any real danger of anybody being eaten by a tiger, and finally, I haven't even seen the image of a tiger for hours, it's just repeating in my head.

This is  yet another amazing thing about conscious simulation: It reproduces the same valences, that is, the positive and negative feeling tones, and the same experience of suffering, as the real thing. When I think about something unpleasant, it hurts. Maybe not quite as much as the real thing, but still it hurts.\index{simulation!taken for real}\index{valence!in simulation} 
The theories explained in previous chapters actually explain, to some extent, why the brain does that. It is not stupid to replay experiences. Replay and other wandering thoughts are important for learning a good model of what the world is like and what kind of actions are useful in which situations, as we saw in Chapter~\ref{replay.ch}.

Yet, my current accusation of my brain being stupid is on a different level than the theories of the previous chapters.  Here, I'm talking about consciousness. Why am I \textit{consciously} afraid of the tiger and consciously suffering during the replay? Why do I need to experience suffering while the brain is just performing some simple computations that we can easily program in a computer? To put that more precisely, why do I need to experience a negative valence on a conscious level while doing the replay? Couldn't the brain just do the replay somehow quietly on an unconscious level without disturbing my conscious feelings and conscious thinking? I'm not just repeating the question posed at the end of Chapter~\ref{replay.ch}, which was: Why do wandering thoughts trigger feelings (possibly quite unconscious) of pain and pleasure? Here, I ask a more general question about consciousness and suffering: Why are such simulations, and the ensuing suffering, \textit{conscious}?

\index{simulation!why conscious}\index{evolution!shortcuts}
Again, we might assume that perhaps evolution just made a simple computational shortcut. If something dangerous is perceived in the outer world, the systems for threats and fear will be activated on every level, unconscious as well as conscious; apparently, this functionality is activated even with simulations, but the question is why. One reason might be that conscious fear is important in information processing because of its capability for broadcasting, as in the global workspace hypothesis: this would justify why conscious fear has to be activated in simulations to properly compute things.
On the other hand, even if consciousness were not necessary for any computations, it would still be evolutionarily pointless to somehow explicitly switch consciousness off when doing replay. It would be nice indeed if, when something dangerous comes up in a wandering thought, the fear system would be activated only partly, not on the conscious level, perhaps only in some distant corner of the unconscious processing systems. This would be nice, but would evolution have any reason to do us such a favor?

We should recall again that evolution does not care at all about whether we feel good or bad. It tries to optimize computation in order to maximize the spreading of the genes, and this has to be done with limited computational resources. Allowing us to switch off conscious suffering when engaging in replay would presumably be pointless from the viewpoint of optimizing computation. So, evolution just makes us suffer from replay since that is optimal use of finite computational resources. Such optimization of computation may actually increase our chances of survival a bit, and give us a longer life. Full of suffering, though.

\section{Self vs.\ consciousness}

\index{self!vs consciousness}\index{consciousness!and self}
So far, I have been mainly considering consciousness on the sensory level, as in ``consciousness of the text you are seeing''.
Another very different thing that we can be conscious of is our own self.
It can even be argued that if there is any consciousness at all, there must necessarily be self-consciousness, or self-awareness, that is, conscious experience related to oneself. It can be seen as a particularly automatic and primitive form of consciousness.
So, we find yet another meaning for the term ``self''---in addition to those in Chapters \ref{self.ch} and \ref{control.ch}--- defined as precisely this self-awareness. This corresponds very well with our intuition, where it is my conscious feeling of being ``me'' that defines what ``I'' am, or what my ``self'' is.\footnote{\citep{dennett1992self,gallagher1999models,gallagher2000philosophical,sep-self-consciousness}.}

This aspect of self-consciousness is very different from the way ``self'' was treated in previous chapters.
The aspects of self treated in earlier chapters do not necessarily have anything to do with consciousness: all the operations described earlier are just computations. In particular, an AI does not need to be conscious to infer that it can control certain things and not others, or to develop behavioral mechanisms that ensure its survival, while even a simple AI system can and should have methods for evaluating the performance of ``itself''.

If self is defined in this sense of self-awareness, it might in fact be difficult to defend any form of ``no-self'' philosophy, of which we have seen one version in Chapter~\ref{control.ch}. Descartes famously
was absolutely certain that he could say  ``I am'' because he ``thinks'':\footnote{  \textit{Meditations on First Philosophy}, Chapter I, \add{translated by Elizabeth S. Haldane}.}
\begin{quote}\index{Descartes}\index{no-self!vs Descartes}
[A]fter
having reflected well and carefully examined all things, we
must come to the definite conclusion that this proposition:  I
am, I exist, is necessarily true each time that I pronounce
it, or that I mentally conceive it.
\end{quote}
Yet, Descartes was quite wary of saying \textit{what} he actually is:
\begin{quote}
I must be careful to see that I do not imprudently take some other object in place of myself, and thus that I do not go astray in respect of this knowledge
that I hold to be the most certain and most evident of all that I have formerly learned.
\end{quote}
The complexities of no-self philosophy largely come from the tension between these two viewpoints: It is intuitively clear \textit{that} I am, but it is not clear \textit{what} I am. (There can hardly be any difference between the ``I''  and the ``self'', they are just two words for the same thing.\citenew{sep-self-consciousness})

On the other hand, some would say that such self-awareness can be seen as a mental construction, even an illusion, just like control and free will. Our self-awareness could be based on a collection of the awarenesses of various sensory perceptions, with no special core that could be called ``me'', or awareness of myself. Hume expresses this potently in a famous quote which is not unlike anything Buddhist philosophers might have said:\footnote{\hume, Section~1.4.6}\index{Hume}
\begin{quote}\index{no-self!Hume} 
For my part, when I enter most intimately into what I call myself, I always stumble on some particular perception or other, of heat or cold, light or shade, love or hatred, pain or pleasure. I never can catch myself at any time without a perception, and never can observe any thing but the perception. When my perceptions are removed for any time, as by sound sleep; so long am I insensible of myself, and may truly be said not to exist.
\end{quote}
This suggests a no-self philosophy where self-awareness is nothing but a complex of various instances of sensory awareness, mistakenly leading to an illusory perception of a separate entity called ``self''.\footnote{In addition to the obvious sensory modalities, it is important here to consider proprioception (perception of the body position, including body ownership, \citet{tsakiris2007neural,seth2018being,de2023self}) and interoception (sense of the internal state of the body, in particular internal organs, \citet{craig2009you}). Further related phenomena include meta-awareness (discussed later in Chapter~\ref{freedom.ch}) and autobiographical memory \citep{prebble2013autobiographical}.\label{propriofn}} 
In the absence of any perceptions, ultimately, I may be said not to exist. Such no-self philosophy could be called ontological: it claims that the self does not exist at all. It does not merely say that self is not what it looks like, or that it is missing something, or that it is not too important; instead, it claims that self does not exist, period. While Hume may not have meant to go quite that far, many Buddhist philosophers do.\footnote{While the no-self philosophy is widely associated with Buddhism, different Buddhist schools actually approach it in very different, even contradictory ways, and several interpretations exist. In fact, the philosophy of no-self has perhaps as many facets as the very concept of ``self''. We already saw the interpretation of no-self as lack of control in Chapter~\ref{control.ch}. Another approach is to see ``no-self'' as a suggestion not to worry about self-evaluation or self-preservation, which was the interpretation of ``self'' in Chapter~\ref{self.ch}; rumination may not be possible without some concept of self to which the bad things are happening.\index{rumination} In this latter sense, it may not be so much a ``truth'' describing the world, but rather a useful way of thinking, as we will see in Chapters~\ref{freedom.ch} and \ref{training.ch}. The (ontological) interpretation we have in this quote by Hume is yet another approach, probably the most well-known in Buddhist philosophy. At the risk of greatly oversimplifying this complex issue, I would venture to say that the Theravadan school is more in line with Hume here; Theravada considers self as an illusion, as something that does not exist. In contrast, Mahayana schools, with the possible exception of Madhyamaka,  emphasize the primacy of consciousness, like Descartes, and do not deny the existence of self---although they do point out that our ordinary conception of self is mistaken in various ways. See \citet{verhaeghen2017selfeffacing} for a short, readable overview emphasizing some practical implications of such a philosophy; \citet{vago2012self} emphasize how mindfulness works largely through self-related mechanisms. \citet{harvey2009theravada} and \citet{williams2008mahayana} give book-length expositions of the philosophy.}

\section{Nothing is real?}

Saying that the self is a mental construction, possibly an illusion, sounds quite radical. Well, how about going a bit further, and denying that \textit{anything} really exists?
While it is undeniable that there is some kind of experience of the world outside of myself, it is equally undeniable that this experience is not the same thing as the world outside. The conscious experience is---according to  a conventional neuroscientific view---the product of complex information-processing of incoming signals. Actually, most of conscious experience has little to do with the world that surrounds us here and now, since conscious contents are often a product of planning, replay, and other kinds of thinking and imagination. The interesting thing is how people are misled into believing that this experience, this virtual reality, this simulation, replay, or planning, is actually the reality.  

It should be easy to admit 
that when we plan the future, the planned events are just imagined, and not real. But the ``unreality'' of consciousness goes deeper than that: 
In fact, \textit{everything} in our consciousness is a simulation, a virtual reality, constructed by our mind. This also includes your consciousness of everything you see, hear, feel, taste, and smell at this very moment. Any perceptual experience, as well as any thought, is simulation, or computation, and not the same as reality.\footnote{Note that the discussion regarding simulation in this chapter has nothing to do with the idea that we would be living in a simulation programmed by some other race, sometimes called the ``simulation argument'' \citep{bostrom2003we}.} 

This is just a rephrasing of well-known neuroscientific facts. As we have discussed several times by now, when you look at this text, your brain is doing complex computations based on the incoming information. Based on the results of those computations, it  creates a conscious perception, which contains an image or a feeling of the world around you, including the book or the computer screen on which you see this text. The conscious experience is created by some quite quasi-miraculous mechanism, which science has not yet been able to explain---even saying that it has to be in the brain is speculative. But the important point is that what you see is the virtual reality, or the simulation in the brain, not the real world.
The distinction between the world and your conscious experience is basically inherent in the very notion of ``experience''.  Although I have already said this in the beginning of this chapter, this point requires a longer explanation, so let me try.

Usually, you would say that you ``see'' this book (let's just assume for the sake of simplicity that you are reading this text in a book). However, according to the conventional neuroscience viewpoint, what you're actually conscious of  is the interpretation created by your brain, not the book itself. The book simply reflects some photons emitted by a lamp or the sun, these photons enter your eye, and your eye sends electrical signals to your brain. Based on these electrical signals combined with the prior information about the world, your brain creates a virtual reality, including your perception of this book. Meanwhile, based on other sensory information, and again all kinds of internal information and processing, the brain creates your perception of your surroundings, your body, and indeed, your perception of your self.  

I am not denying here that the book exists. I am merely pointing out that your consciousness, your sensory awareness of this book and everything else is created by your mind, presumably by some highly complicated process in your brain. 
You cannot really ``see this book'',  you cannot be ``conscious (or aware) of the book'', you are only aware of the results of some computations performed in your brain, in which the book only plays the role of being the physical source of some radiation which was input to the computations.

The metaphor of virtual reality means that consciousness is similar to wearing virtual reality goggles which feed an input to your eyes which is so realistic that it looks almost like real. In the case of seeing this book, though, it looks \textit{exactly} like real to you because you know nothing better: You have never seen anything which would be somehow closer to reality than this virtual reality.
A number of science fiction movies are based on the idea that somebody could feed fake sensory information directly to your brain, and you would have no idea the sensory input is fake. Descartes\index{Descartes}\index{demon!Descartes's evil} already proposed that he cannot trust his perception because an ``evil demon'' might be feeding an illusory external world to his mind---which is precisely why he could only be certain of his own existence. Such claims lead to an extreme form of uncertainty regarding perception.\footnote{Descartes's \textit{Meditations on First Philosophy}. \add{See \citet{sep-skepticism-content-externalism} for a modern discussion based on Putnam's brain-in-a-vat scenario.}\index{Putnam}\index{brain in a vat}}

My point is that something like that is actually happening to you all the time, according to perfectly mainstream neuroscience. I want to emphasize that I'm not trying to make some radical philosophical point here. There are others that will tell you that the world does not \textit{really} exist, including proponents of some Eastern philosophical systems, such as Advaita Vedanta, or Mahayana schools of Buddhism, including Zen and \yogacara.\footnote{For example, \citet[p.~94]{williams2008mahayana} describes the \yogacara\ viewpoint by Vasubandhu as ``Apparently external objects are constituted by consciousness and do not exist apart from it.  (...)   There is only a flow of perceptions.''  
  Claiming that the world does not really exist is a form of \textit{ontological} idealism, while claiming that we cannot possibly know for sure if the world exists is \textit{epistemological} idealism \citep{idealism-stanford}.\index{idealism} In Mahayana Buddhist philosophy and, especially, its Western commentary, there has been a lot of debate on which form to support. For example, \citet{lusthaus2013and} warns about misunderstanding the \yogacara\ literature to consistute an ontological statement while it is actually intended to be epistemological only. (For my part, I'm not committing to any such philosophical viewpoint here.) Even in early Buddhist texts you find claims related to such idealism: ``And what, bhikkhus, is the all? The eye and forms, the ear and sounds, the nose and odours, the tongue and tastes, the body and tactile objects, the mind and mental phenomena. This is called the all.'' (\SN{35.23}); however, this formulation open to interpretation and may also be seen as admitting the (ontological) existence of outside objects.\label{nothingexistsfn}}\index{Zen}\index{Vedanta}\index{Yogacara}I'm trying to steer away from such philosophical speculations about what exists, and merely point out some of the limitations of our perceptual and cognitive systems, in a way which is, I hope, acceptable, even if unpalatable, to most scientists working on those topics.

At the risk of repeating myself: Most neuroscientists would agree that sensory processing in the brain is producing an interpretation of the incoming input; they would further agree that the brain creates consciousness. Thus, the contents of consciousness are not a direct product of the world, let alone the same thing as the world; it is a construction, an interpretation created by the brain. Yet, we often have the intuitive feeling that the contents of consciousness are somehow identical to the contents of the outside world, which is not the case. Just studying an introductory course in neuroscience or in AI might be enough for many people to give up such an idea. Visual illusions, such as in Fig.~\ref{kanizsa.fig}  on page~\pageref{kanizsa.fig} are one way of demonstrating how perception is different from reality.

The Belgian artist Ren\'e Magritte has a famous painting called \textit{La Trahison des images}, or ``The Treachery of Images''.\footnote{See e.g.\ {\tt https://en.wikipedia.org/wiki/File:MagrittePipe.jpg} which cannot be reproduced here for copyright reasons.} 
The painting consists of a picture of a pipe, with the text ``Ceci n'est pas une pipe'', or ``This is not a pipe'', written underneath it. The point is that the painting is just a picture, not the real pipe. While the artist's purpose was to illustrate the deceiving nature of images, the painting illustrates the illusory nature of consciousness as well. Suppose you actually hold the pipe in your hand and look at it. What appears in your consciousness is a picture, a simulation, or a reflection of the pipe; it is not a pipe. 
Yet, we have the habit of thinking that the perceptual image is the real pipe, while in reality, it is only somehow indirectly related to the real pipe. Furthermore, the category of a ``pipe'' is just a mental construct.
In this sense, perception is not the real thing; consciousness is not the reality.\footnote{In this chapter, I have taken the viewpoint of physical materialism by assuming that the pipe actually exists and that our consciousness is created by the brain. If we reject one of these assumptions or both, the conclusions will of course be even more radical.} 

These philosophical points are not simply theoretical speculation; our attitude to consciousness has a direct effect on suffering. Consider the example of the tiger I saw on TV: If I could somehow develop a different attitude towards the contents of my consciousness, seeing them as mere simulation, I might suffer less. This is precisely why some Buddhist schools claim that the outside world only exists in your imagination---or at least they recommend adopting such an attitude towards the world.\footnote{Any Buddhist claims about the inexistence of the outer world could, in fact, be seen as clever devices only intended to help with meditation and other practices \citep{schroeder2004skillful}. The Theravadan master Ajahn Chah 
  seems to have this intention when he says ``If you think things are real there is suffering and there is fear. You are afraid of the different ways things may turn out. (...) There is thinking, then fear follows immediately. It deceives you, creating a picture to mislead you. (...) As to what is actually happening, there is nothing'' \citep{chah2001being}. See footnote~\ref{nothingexistsfn} above for discussion on the philosophical claims concerned, and page~\pageref{platoemptinessfn}  for a related quote from Seneca.\label{chahfn}} In the next chapters, we will consider this and many other ways of reducing suffering by changing our thinking patterns as well as using meditation.

\part[Liberation from suffering]{Liberation from suffering
  \\ \ \\ \ \\  \normalsize The final part will describe methods for reducing suffering, \\largely drawing from philosophical traditions such as Buddhism and Stoicism, \\while showing how they logically follow from the science of Part~I and Part~II} \label{partiii}

\chapter{Overview of the causes and mechanisms}
\label{overview.ch}

In this final part of the book, we move to the question of how to reduce suffering or, ultimately, how to be liberated from it. Applying the scientific theories of the previous chapters, we devise various methods to that end in the following two chapters. But first, in this chapter, I will recapitulate the basic theory. I will use two flowcharts to illustrate the basic mechanisms of suffering: the first one emphasizes adverse properties of the world and general cognitive design principles, while the second flowchart focuses on the dynamics of moment-to-moment cognition. The flowcharts also make some explicit connections to the basic concepts of Buddhist philosophy. The connection of reward loss to the whole architecture of intelligence is then succinctly summarized in one single ``equation'', which directly suggests ways of reducing
suffering.

\section{Why there is (so much) suffering}

\begin{figure}
\begin{center}
\resizebox{0.9\textwidth}{!}{\includegraphics{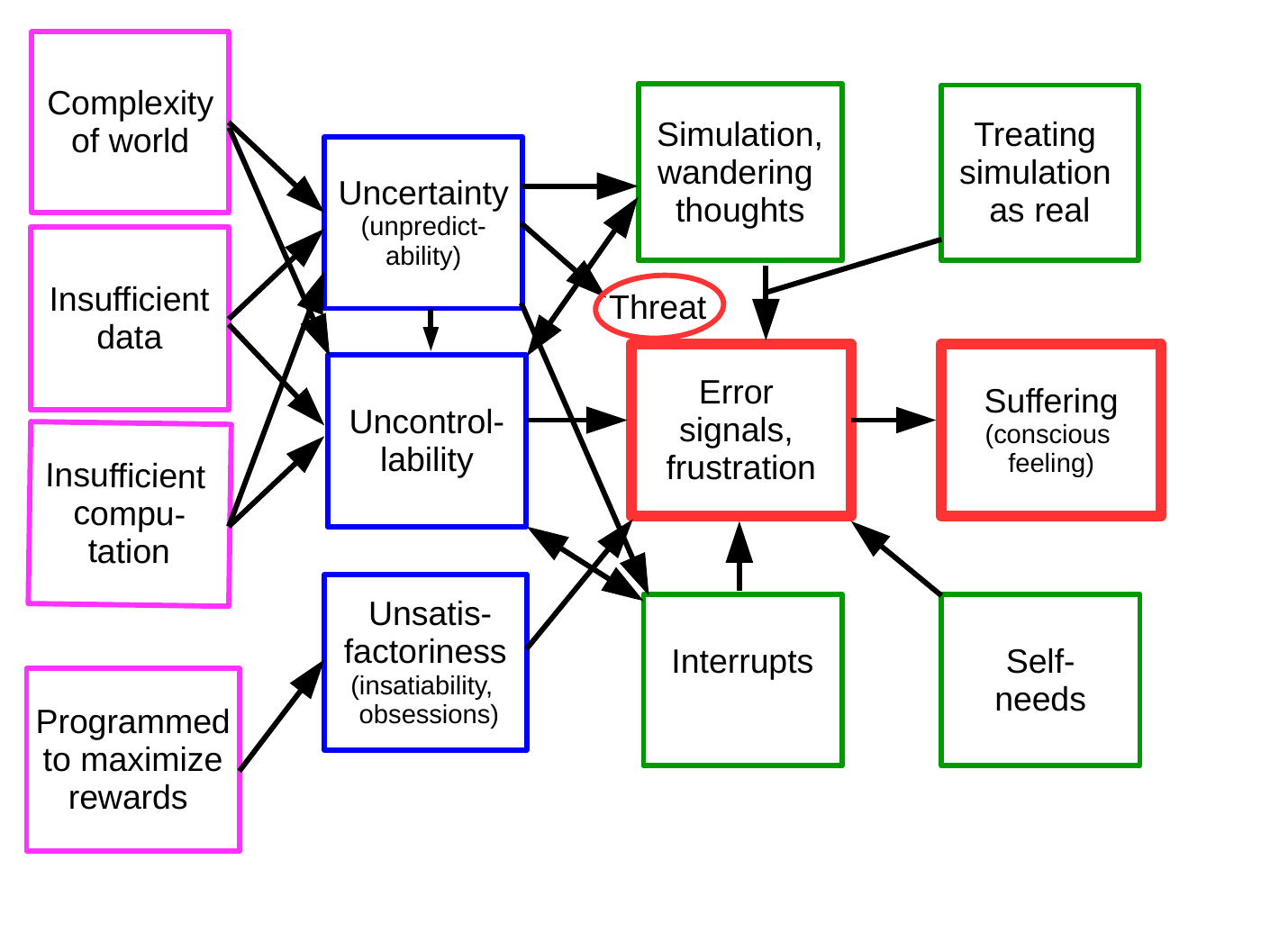}}
\caption{Recapitulation of the causes and mechanisms of suffering explained in earlier chapters. The boxes in magenta are intrinsic properties of the world---including the agent---while the boxes in blue are more concrete problems they pose for information-processing. The green boxes are possible functionalities in a highly developed cognitive system, and the red boxes are the postulated system finally generating suffering.}
\label{flowchart1.fig}
\end{center}
\end{figure}

Let us start by summarizing the difficulty of information-processing in a complex world. The fundamental reasons of how suffering or mental pain is born are illustrated in Fig.~\ref{flowchart1.fig}, which I will next go through in detail.

\subsection{Root causes of suffering}

\index{suffering!root causes}
The starting point is that the agent finds itself in a world which is highly complex. 
In such a world, acting optimally (in any reasonable sense of optimality) would require huge amounts of sufficiently detailed sensory input, together with huge capacities of computation. Unfortunately, in the real world where we live in, an agent cannot have any of those. These three root causes of suffering---\textit{complexity of the world, insufficient data,} and \textit{insufficient computation}---are shown in  the left-most column of the figure. Certainly, the agent's limited physical capabilities to act in the world and change it---catch any prey it wants, for example---create suffering as well, but here we focus on limitations related to information-processing because it can be modified more easily.\footnote{It could be argued that lack of a good model, or ``inductive bias'', is another limitation. Inductive bias can refer to slightly different things: on the one hand, it is sometimes simply used as a fancy term for a Bayesian prior\index{Bayesian inference}  in a probabilistic model, but it can also refer to constraints that are more structural in the sense of, for example, the choice of the family of nonlinearities, regularization, or other computational structures used in a model (but these could still, in most cases, be seen as Bayesian priors in a hierarchical Bayesian model). Basically, what we are talking about here is that the agent might not have a good model family from which to pick its model of the world, and might in particular ``suffer'' from overfitting (see footnote~\ref{overfitfn} in Chapter~\ref{ml.ch}).\index{overfitting} I take here the viewpoint that, fundamentally, a good inductive bias is only necessary because the data is limited: if the data were infinite, the proper inductive bias could be learned from the data by testing the performance of the models on a new test set which was not used in the learning \citep{feinman2018learning}. Therefore, I do not discuss inductive bias in any detail in this book, and subsume the problems due to lack of correct inductive bias under the heading of ``insufficient data''.\index{inductive bias}}

At the same time, the system is typically programmed to try to relentlessly \textit{maximize rewards} set by the ``programmer''---which is evolution for humans, eventually complemented by some cultural processes. 
The rewards are not designed to make the agent happy, but to fulfill some objectives of the programmer, such as spreading your genes in the case of evolution. It is not possible for the agent itself to decide that it wants to pursue some new goals, or to re-define its own rewards; nor can it decide that it has had enough rewards and does not need more. 
This is a fourth root cause of suffering, shown at the bottom of the left-most column of the figure. 

\subsection{Three fundamental problems in information-processing}

The four root causes just mentioned lead to a number of challenges for information-processing, which I here condense into three: uncertainty (including unpredictability), uncontrollability, and unsatisfactoriness (consisting of insatiability together with evolutionary obsessions).

First, the overwhelming complexity of the world leads to \textit{uncertainty}: the agent is not able to accurately understand what happens in the world. It is not even able to accurately perceive most phenomena in the world. It will try to divide the perceptual inputs into categories, but such categories are fuzzy, sometimes arbitrary, and categorization is often uncertain. %
One important special case is uncertainty about the future, also called \textit{unpredictability}.\index{unpredictability}\index{impermanence} (It  is closely related to the Buddhist concept of impermanence, as will be discussed in Chapter~\ref{freedom.ch}.) On the one hand, it is clear that it is difficult to predict the future if even the current state of the world is uncertain, which is a consequence of uncertain perception. However, unpredictability is actually a more general phenomenon than the uncertainty of perception: even if the perceptions were perfectly accurate and certain, it might not be possible to predict the future accurately due to the great complexity of the world. It might be impossible to learn to model the world accurately enough, or using such models might require overwhelming computational power. This is well-known in the natural sciences, where even extremely accurate measurements of a natural phenomenon do not necessarily mean you can predict it, because the prediction would require overwhelmingly advanced scientific models.

\index{uncontrollability}
Uncertainty and unpredictability necessarily lead to \textit{uncontrollability}, lack of control of the world: if the agent does not know what is actually happening in the world, or it does not know how to predict what will happen in the future,  it cannot possibly control the world. In fact, control requires the capability to predict the results of your actions, which requires not only a good model for prediction, but also an accurate perception of the current state of the world.\footnote{For those conversant in Buddhist philosophy, this is similar to the idea that impermanence feeds into no-self, when impermanence is seen as related to uncertainty and no-self is interpreted as uncontrollability \citep{mahasianatta}. See footnote~\ref{threecharfn} below for more on such analogues.\index{no-self}}
However, uncontrollability is not only due to uncertainty. 
Even if the world could be perfectly perceived and predicted, there would still be uncontrollability due to at least two reasons. First, the agent has limited physical capacities to influence the world. Second, limitations of computation reduce controllability in many ways: the computational complexity of the search tree precludes finding perfect solutions to the planning problem, while the parallel and distributed nature of the agent's cognitive system hinders proper control of the agent's internal functioning. All these reasons make sure that the world is uncontrollable for humans. But it can be rather uncontrollable even for a thermostat, since the temperature of most environments obeys extremely complex natural laws that are beyond the understanding of the thermostat, and errors cannot be avoided.

Meanwhile, the programming of the agent to maximize rewards means that the agent finds that no amount of rewards is enough: it is \textit{insatiable}\index{insatiability} (page~\pageref{insatiability}). In fact, the very raison d'\^etre of the agent is to maximize the rewards set by its programmer or evolution; it will never be satisfied, and the desires will never be satiated. A related property is what I called \textit{evolutionary obsessions}\index{obsessions!evolutionary} (page~\pageref{evobs}), which means that humans are compelled to seek various rewards which they might, if they thought about it rationally, prefer not to seek---such as unhealthy food and excessive competition for status. Seeking those unnecessary rewards increases the chances for frustration, thus leading to more suffering, and may even lead to less reward in the long run (by ruining your health, for example). Yet, it is difficult for humans to change what they find rewarding. I group these two properties under the umbrella term \textit{unsatisfactoriness},\index{unsatisfactoriness} expressing the general idea that even if the world were completely known and controllable, there would still be suffering due to the fact that the system is never satiated and strives at questionable goals.

The three fundamental problems of uncertainty, uncontrollability, and unsatisfactoriness are shown as blue boxes in Figure~\ref{flowchart1.fig}, the second column.\footnote{The three problems or challenges could be seen from two different viewpoints: either as properties of our natural world (at least if unsatisfactoriness is seen from a more general perspective) or as properties of information-processing in any sufficiently complex world. Here I take the latter view; the three problems are problems of information processing. However, they are in fact created by those properties of the natural world which are given in the magenta left-hand column in the figure. These three problems are a rough analogue of what is called the three characteristics\index{three characteristics (Buddhist)} of existence in early Buddhist philosophy: impermanence (\textit{anicca}), no-self (\textit{anatt\=a}), and unsatisfactoriness/suffering (\textit{dukkha}).\index{dukkha} Impermanence\index{impermanence!as uncertainty} is to some extent a special case of uncertainty, as will be discussed in detail in Chapter~\ref{freedom.ch} on page~\pageref{impvsunc}. Uncontrollability is an important aspect of no-self philosophy (see page~\pageref{noselfpage}) and may have been its original meaning in the earliest layers of Buddhist philosophy. 
  The Buddhist concept of \textit{dukkha} has the broadest definition of them all, simply meaning ``suffering'' in one interpretation; thus our concepts of insatiability and evolutionary obsessions are only some of its aspects, as will be discussed in Chapter~\ref{freedom.ch}. We could have added another blue box depicting ``emptiness'',\index{emptiness (Buddhist)!in flowchart} a widely used concept in later Buddhist philosophy; a discussion on emptiness is postponed to Chapter~\ref{freedom.ch}, where it will be introduced as an umbrella term for fuzziness, subjectivity, and contextuality, and related properties. Alternatively, we could have added another box giving distributed processing and possible lack of central executive (discussed in Chapter~\ref{control.ch}) as a necessary computational consequence of the root causes on the left; now distributed processing is not explicitly mentioned in the graph, although several boxes are related to it. One more possibility would have been to introduce nonstationarity\index{nonstationarity} (discussed later in footnote~\ref{nonstatfn} in Chapter~\ref{freedom.ch}) as a root cause, but it can be simply seen as a special case of the complexity of the world, even if a particularly important one. 
  \label{threecharfn}}

\subsection{Error signals and suffering}

Because of uncontrollability and uncertainty, the agent's information processing will incur  errors. Often, things do not go as planned or as expected; predictions have errors, and expected reward will not be obtained. Thus, the system generates error signals. Such error signals are particularly frequent because the agent is never satisfied and is always looking for more reward. Error signalling is the central red box in the third column in the figure.\index{error signalling}
Threat is based on a prediction of large frustration with a sufficient probability, so it is fundamentally related to frustration and depicted as an annex to frustration as it were.

Our fundamental hypothesis in this book is that such error signals are what produce the feeling, and ultimately the conscious experience, of suffering. The suffering due to error signals is especially strong if the error is frustration, i.e.\ the agent is trying to reach the goal (or a reward) but it fails. Thus, error signals finally lead into the red box of suffering, on the right-hand side in the figure.\footnote{For a more detailed discussion on the connection between error signals and suffering, see footnote~\ref{sufferingconsciousnessfn} in this chapter.}

\subsection{Optional processes that increase suffering}

Several further processes may further be active, depending on how sophisticated the agent is. If its cognitive architecture uses \textit{wandering thoughts}, we get another box in the flowchart (top row, third column). 
It is a type of information processing that takes place only in highly sophisticated agents, which is indicated by drawing the box in green.\index{wandering thoughts!increasing suffering}
Such highly intelligent agents may engage in  \textit{simulation}\index{simulation} of the world in terms of planning and replay, which in humans often happens in the form of wandering thoughts.  They increase error signalling by repeating or anticipating experienced errors; this is depicted as the green simulation box on the top row feeding into error signals.\index{error signalling}
Since the goal of simulation and wandering thoughts is to gain more control and reduce uncertainty on the world by better learning its dynamics, there is an arrow from the uncontrollability and uncertainty boxes to the simulation box. On the other hand, since wandering thoughts increase uncontrollability in their own way, that arrow is bidirectional.

Furthermore, the agent may react in different ways towards the contents being replayed or simulated. If the simulated contents are processed almost \textit{as if they were real}, and the various frustrations in the simulations are processed in the same way as real frustrations, this will greatly increase suffering. Otherwise, simulated error signals might not lead to suffering. 
This is indicated in the flowchart as the green box on the top right-hand corner; it feeds into the connection between the simulation box  and the error signal box, thus creating a causal connection between simulation and error signalling as just described. 

A related design principle is using \textit{interrupts}, which are useful for handling uncertainty due to unpredictability. Interrupts are seen as an essential aspect of emotions as well as desire in this book. Interrupts create more frustration since by interrupting on-going behavior, they increase uncontrollability; they also impose new, possibly short-sighted goals on the agent, which can further increase to frustration. It is also possible that interrupts produce specific error signals unrelated to frustration. Interrupts are depicted as another green box on the bottom row, feeding into error signals.\index{interrupt theory}

Sufficiently developed agents have various intrinsic rewards, which may be frustrated as well. The very strongest suffering actually tends to come from the frustration of self-related goals, such as survival or self-evaluation. These \textit{self-needs} create new kinds of frustration and errors, such as the agent ``not being good enough'' in the sense of not obtaining enough rewards on a longer time scale. This is the box at the bottom right-hand corner, again in green since it is a sophisticated module, which the programmer may include in the system or not.\index{frustration!of self-needs}\index{self!needs}

This flowchart explains the conditions leading to suffering on an abstract level but still, it clearly has practical implications.\footnote{For future research, I would like to point out that many of these boxes can be quantified, although different measures are possible, and research is needed to decide which ones are useful. Uncertainty is typically quantified by Shannon entropy as defined in information theory and already used by, e.g., \citet{hirsh2012psychological,friston2010free}. The complexity of the world could be the number of states (or the entropy over the typical distribution over them) in a category-based world model. The amount of data available can again be quantified by information theory, for example, by the Fisher information (possibly multiplied by the number of data points) measuring the information a dataset gives about the parameters of an ideal world model, i.e., how much information the data actually contains about the world  (alternatively, the mutual information between the data and the world states could be useful). Computational resources can be quantified using flops per second or a similar measure. To quantify uncontrollability, related probabilistic computations are possible \citep{huys2009bayesian}; we might also be able to use various tools from control theory, e.g.\ \citep{liu2011controllability}.}
If we consider it from the viewpoint of interventions,\index{intervention} e.g.\ practices that would decrease suffering, it clearly points out that we could reduce suffering by reducing wandering thoughts, self-needs, and other processing in the green boxes. Such ideas will be considered in detail in the next chapters. The next chapters will also consider how to deal with the blue boxes (uncertainty, uncontrollability, unsatisfactoriness). In the remainder of this chapter, however, we look at the process of suffering from two further viewpoints.

\section{Cognitive dynamics leading to suffering}

One complementary viewpoint on suffering is provided by cognitive dynamics, i.e.\ how the different cognitive processes work and influence each other in real time. In some sense, this is about zooming into the part of the mechanism in Figure~\ref{flowchart1.fig} that leads from the blue boxes in the second column (uncertainty, uncontrollability, unsatisfactoriness) to errors and suffering. This reveals further quantities that can be intervened on to reduce suffering.\index{dynamics!cognitive, of desire}

\begin{figure}
\begin{center}
\resizebox{0.7\textwidth}{!}{\includegraphics{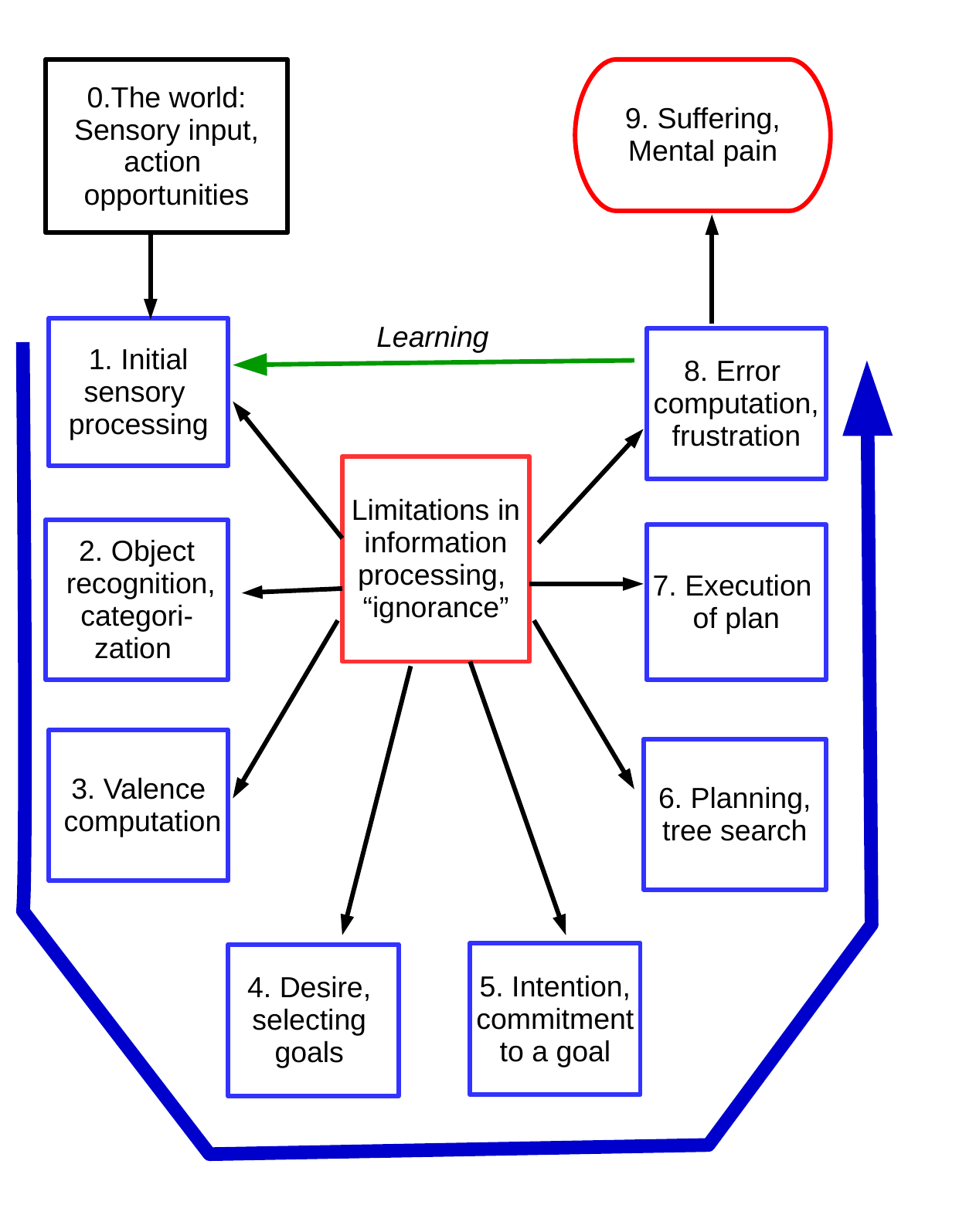}}
\caption{Recapitulation of the cognitive chain or cycle underlying suffering. The sensory input from the world enters the system in the top left-hand corner of the chart (step~\#0). It is processed in a sequence of steps~\#1-\#7, along the big blue arrow. Throughout these steps, there are various limitations in the information-processing, or ``ignorance'', to use a Buddhist term defined in the main text.
  Finally, the computation leads to computation of errors (step~\#8) which may lead to suffering in the top right-hand corner of the chart (step~\#9). The error computations are further fed back to the whole system as indicated, for brevity, by the single green arrow closing the cycle and going back to sensory processing.}
\label{flowchart2.fig}
\end{center}
\end{figure}

In previous chapters,  we have seen a number of steps in an information-processing procedure that translates sensory input into action decisions and possibly suffering. Such steps are recapitulated and illustrated in our second flowchart in Figure~\ref{flowchart2.fig}. To begin with, sensory input is received from the external world; see the upper left-hand corner of the flowchart. As the black arrow in the flowchart indicates, in the next step, the agent engages in initial sensory processing. This typically leads to  recognition and categorization of objects in the world, which in our basic formalism includes recognition of the state in which the agent is. (For better visualization, the order of processing is now indicated by a single long blue arrow in the flowchart.) Recognition of the state is immediately followed by computation of the valences of the nearby objects or states. (Valence means here the prediction of the reward associated with an object, or more generally the prediction of the value of a state.)\index{valence!in cognitive dynamics}
Based on those valences, a number of candidate goals are chosen, which is the process of desire in its hot, interrupting form. Next, the agent may choose one single goal and commit to it, which is also called intention. Then, the agent starts planning how to reach the committed goal, possibly by some kind of tree search, and executes the  plan obtained.\footnote{Some pointers and details on the steps 3-7 in the flowchart: Desire was defined as a process that suggests new goals  in Chapter~\ref{planning.ch}. Such suggestions were seen to be possible by neural network computations in Chapter~\ref{dual.ch}. In particular, such computations give fast approximations of valences, i.e.\ how rewarding near-by states are, preferably taking into account the whole future by approximating the state-values. Valence computations were further linked to the generation of interrupting (intruding) desire in the framework of the elaborated-intrusion theory of desire in Chapter~\ref{emotions.ch}. After the selection of one of such suggested goals, planning proceeds as already described in Chapter~\ref{planning.ch}, including the idea of commitment or intention to the goal that has been selected among the possible high-value states. Note that in this framework, goals are set by the agent itself, based on predictions of rewards, which come from the outside world. 
}

  After finishing the execution of the plan, the agent observes the outcome, and based on it, an error such as frustration or reward loss is computed. Such an error may lead to suffering, in the precise sense of a conscious experience.   Finally, the computed error will be used in a learning process to guide future actions, which in a sense closes the loop, as indicated by the green arrow. (Only an arrow from error computation to sensory processing is drawn in the figure for simplicity, but actually, the error is broadcast widely.) The flowchart shows the prototypical sequence, but in reality there is, of course, more variability. \footnote{I hope the flowcharts in this chapter also clarify some conceptual differences regarding the concept of suffering, which may have been slightly confounded in previous chapters. First, I emphasize that I'm not saying that errors \textit{are} suffering, but that errors are the direct cause for suffering. Suffering is a complex phenomenon, comparable to emotions which also have multiple aspects, including in particular a conscious experience.\index{subjective experience!and suffering}\index{consciousness!and suffering} Conscious, subjective experience is obviously not the same thing as errors computed on the level of largely unconscious information processing. Second, nor am I saying that suffering is driving learning: it is the errors that are driving learning, after some further sophisticated computations. Thus, the errors computed lead to suffering on the one hand, and, more indirectly, to learning on the other.  This is why errors, suffering, and learning are separate items in the flowchart.\label{sufferingconsciousnessfn}}

  In the middle of the flowchart, we have ``limitations of information-processing'', which influences all the steps in the processing. While that has been the main theme of the whole book, here it also provides an analogy to Buddhist philosophy, which uses the term \textit{ignorance} in connection with similar conceptual schemes.\index{ignorance (Buddhist)} 
  Ignorance describes the fundamental underlying reason why the agent's cognitive apparatus creates so much suffering---thus adding to the adverse properties of the external world, which were shown in the previous flowchart. We have already seen various limitations in information processing that might be called ignorance, in the sense that they can be seen as imperfect or lacking, and they increase suffering. Thus, any limitations of information processing could be seen as one computational definition of  ``ignorance''.

However, this Buddhist term could also be interpreted more literally in the sense of our being ignorant of the limitations of our information processing: a ``meta-ignorance'' so to say.   We can thus consider such forms of meta-ignorance as 1) ignorance of the arbitrariness and even harmfulness of our rewards and goals, 
  2) ignorance of the uncertainty and fuzziness of perceptions and concepts, 3)  ignorance of uncontrollability, especially of the mind itself, and even 4) ignorance that the simulation is not real---but this list is not meant to be exhaustive.\footnote{In early Buddhist philosophy, a central form of ignorance is the belief in ``self''. I omitted it from this list because in this book, I have attempted to largely reduce such no-self ideas to less abstract concepts, in particular uncontrollability.\index{ignorance (Buddhist)}}  In all these cases, we can claim, inspired by Buddhist philosophers, that there is some kind of a \textit{mistake} in our ordinary thinking and functioning of the brain---not merely a \textit{lack} of computing power, say. One particular mistake, or flaw, is about misunderstanding where suffering comes from and how it can be reduced. Importantly, in contrast to fundamental limitations of information processing, this ``meta-ignorance'' could be corrected. 
  Most of this book has actually been devoted to explaining what such ignorance consists of and how its different forms lead to suffering or amplify it. So, in Figure~\ref{flowchart2.fig}, ignorance is naturally placed in the very middle.\footnote{It is interesting to compare the mechanisms described above with a central idea in Buddhist philosophy: the twelve-link chain, which has served as a central inspiration for this flowchart.\index{twelve-link chain|see{dynamics, Buddhist model}}\index{dynamics!Buddhist model} (A related approach is proposed by \citet{grabovac2011mechanisms}.) The Buddha elaborated his idea of desire as the origin of suffering by building a sequential causal model, which can be seen as an instance of his more general ideas on ``causality'' resumed under the heading of dependent co-arising/origination \citep{mahasidependent,analayo}. While different variants exist, I consider here the version with twelve items. Some of the items in his chain correspond clearly to concepts we have seen in this book, while other do not. The chain begins by three items which are difficult to interpret, and seem to be metaphysical speculation about how the ignorance of the true nature of reality creates consciousness and this creates the world. In the text above, I provided some more scientific interpretations of ``ignorance'' in terms of limitations of information-processing, and our ignorance of those limitations and their implications. After those initial three items, the middle part of the Buddha's chain goes as follows:
4) ``Name-and-form''. This basically means the world, including our internal world of memories and consciousness.
5) ``Six-fold sense bases''. This is when the sensory organs receive input from the outer world, or memories or wandering thoughts enter the mind (which is in Buddhism considered a sixth sense).
6) ``Contact''. This I interpret as perceptual processing leading to object recognition, where the brain processes information and interprets the incoming stimulus in terms of a given category (``That's a dog'').
7) ``Feeling'' (\textit{vedan\=a})\index{valence!in Buddhism}\index{vedan\=a} means computation and perception of the valence of the sensory stimulus: Is it good or bad, do I like it or not? 
8) ``Craving (or thirst, or desire)'' is the same as desire in our terminology, as always including aversion. You may want the object you has seen, or you may want to get rid of it. A number of goals are considered at this stage. 
9) ``Attachment (or clinging)''.\index{attachment (Buddhist)} I proposed in Chapter~\ref{summary1.ch} to interpret this as forming an intention, i.e.\ committing to a certain plan and a goal, and planning for it.\index{intention!and attachment} (In Buddhist literature, the interpretation of ``attachment'' is actually highly variable, and in my view rather muddled: often, the distinction between desire and attachment is only a matter of degree. It is sometimes pointed out that an alternative translation of the word in question (\textit{up\=ad\=ana}) is ``fuel'', which might give an alternative interpretation as being related to the learning process in the next steps.)\index{up\=ad\=ana}
10--11) ``Becoming and birth'' are the next two steps which are a bit more difficult to interpret, and have traditionally been interpreted in more metaphysical terms. I suggest interpreting them as referring to the \textit{learning process} which creates various associations in the mind, including creating habits out of one's actions. Thus, whatever the agent does leads to ``birth'' of new action tendencies and associations.
12) ``Old age and death'' is the final result of the above causal chain, and can be interpreted as simply ``suffering''.
Considering the specific connections between such a Buddhist scheme and ours, we see that steps 4--9 here directly correspond to the first boxes in our flowchart (from box~\#0 to box~\# 5). The boxes \#6--\#8 following that, including actual planning, plan execution,  and error computation may be missing in the Buddhist chain, or they could be seen as being included in its steps 10-11. The steps 10--11 in the Buddhist chain are, in my tentative interpretation, specifically related to the ensuing learning process, shown by the horizontal green arrow in the flowchart. Step 12 is a poetic description of suffering in the red box~\#9 at the top right-hand corner in the flowchart.
In any case, the two models share the crucial idea of how sensory input leads to desire, which via attachment (clinging/intention) leads to suffering, and how this suffering somehow perpetuates itself (in our model by means of a learning process).}

It should be noted that this chain of processing steps is not only initiated by external stimuli (the sight of something good) but also by internal simulation, which is another reason why the processing constitutes something more like a cycle or a loop. For example, wandering thoughts, or almost any kind of  thoughts, trigger memories or predictions of sensory stimuli, and the cycle is launched almost as if those stimuli were real. This is, however, not shown in the flowchart for the sake of brevity.

\section{An equation to compute frustration}

While the flowcharts above help us understand the mechanisms behind suffering and even design interventions, the real strength of a computational approach is that we can quantify things, at least in principle. I do not mean that we would necessarily be able to give a number measuring the strength of suffering, but we can understand the connections between different quantities more explicitly than with flowcharts.
As the most powerful recapitulation of the theories of the preceding chapters, I next propose a simple equation that describes the amount of frustration experienced by the agent. This will be the basis in the next chapters, where we attempt to reduce suffering.

The starting point for the equation is Chapter~\ref{rpe.ch} (page~\pageref{frustdef}), where we defined suffering as reward loss, that is, the difference between the expected reward and the obtained reward. (More generally, the reward prediction error might be used.) This theory provides a quantitative basis for modelling suffering. Reward loss is based on a simple mathematical formula, so we can look at the different terms it contains. We can analyze how they influence reward loss, and how they could eventually be manipulated.
The equation we had in Chapter~\ref{rpe.ch} was, however, very basic and did not take into account any of the complexities of a real cognitive system that we have seen in later chapters. So, we need to look at the different factors influencing reward loss in more detail and reformulate the equation.

The first point to consider is that any quantities affecting reward loss need to be perceived by the agent.
While the difference between expected reward and the reward actually obtained is, in principle, the basis of suffering, the difference cannot of course cause suffering by itself. It has to be computed---that is, perceived---by the agent.
So, we need to make a connection between limitations in perception and categorization on the one hand, and frustration on the other. Due to such limitations, the perception of the actual reward is uncertain and subjective, as explained in Chapter~\ref{perception.ch}. Obviously, our expectations are subjective and may be overblown as well. 
Yet, the agent can only compute the reward loss based on its own perceptions, on the information at its disposal. 

The second point is that as with any perception, the level of certainty of the error computation should also be taken into account: If the perception of reward loss is particularly uncertain (say, because it is dark and you cannot see what you get), that should reduce the effect of reward loss. It is common sense that if the agent is not at all certain about what happened, it should not make any strong or far-reaching conclusions, and the error signal should not be strong.\footnote{\add{This intuition will be made more rigorous in footnote~\ref{deltafn} in Chapter~\ref{freedom.ch}.}}

Further following the general rules governing perception outlined in Chapter~\ref{perception.ch}, the intensity of the perception of reward loss is modulated by the attention paid to it. Reward loss causes less suffering if little attention is paid to it, for example, when one is distracted by something else---simply because you might not even notice reward loss occurring. Paying attention to something may also be necessary to become conscious of it.\footnote{See footnote~\ref{attentionconsciousnessfn} in Chapter~\ref{consciousness.ch} for discussion on this connection.} Thus, the amount of attention paid to the reward loss must be included in the equation. 
There are further related phenomena that change the amount of frustration experienced. For example, we may take the contents of consciousness---the virtual reality or simulation---more or less seriously (Chapter~\ref{consciousness.ch}). Furthermore, we may find errors acceptable if we are deliberately trying to learn something new. 
For simplicity, we include these aspects in the term called ``amount of attention'' being paid to the reward loss, since not taking simulation seriously is related to not paying a lot of attention to it. 

The final important point is that error computation can happen many times, in particular in the case of replay or planning, which means we perceive, or rather simulate, the same (possibly imaginary) reward loss again and again (Chapter~\ref{replay.ch}). For example, if you replay an event just once in your head, the suffering is multiplied by two, almost.

\index{frustration!summarizing different aspects}
Taking all these aspects into account, we arrive at the following   \textit{\eqsuff}: %
\begin{quote} \label{suffeq}\index{frustration!frustation equation}
  {\bf frustration = \\perception of (the difference of expected and obtained reward)\\
    $\times$ level of certainty attributed to that perception\\
    $\times$ amount of attention paid to it\\
    $\times$ \add{number of times it is simulated, plus one for the initial perception}\\
  }
\end{quote}
In this equation, we have four terms on the right-hand side, i.e.\ after the equality sign, multiplied by each other. First, there is the basic formula of reward loss in parentheses.
Thus it includes the amount of expected reward and the amount of obtained reward, whose difference is computed. As with reward loss, if this difference is negative, it is set to zero---if the obtained reward is greater than the expected, there is zero frustration. But crucially, the reward loss here is modulated based on how it is perceived by the agent.\footnote{There is a subtle point about perception of the reward loss, which is that the system may first compute percepts of the two quantities (expected and actually obtained rewards) and then compute the difference, or it can directly attempt to compute the percept of the difference. In other words, we can have perception of the difference, or the difference of the perceptions. (In the case of the expectation, its ``perception'' might rather be called the ``estimation'' of the expectation.) Intelligent systems may use either of these two approaches. I shall not venture into speculating which might be the case in the human brain. \label{lossperc}\index{reward loss!perceived}}

Then, this perceived reward loss is multiplied by three modulating factors. We use multiplication here to emphasize how the perceived reward loss may actually lead to no suffering at all if just one of these modulating factors is zero. The modulating factors are: the level of certainty that the agent attributes to the perception of reward loss (zero meaning absolute uncertainty, one meaning complete certainty), the amount of attention paid\index{attention!to reward loss}\index{reward loss!attention paid}
to reward loss (including how seriously the contents of the consciousness are taken), and finally, the number of times the event is \add{simulated or perceived (after one initial instance of perceiving the event, it may be replayed or simulated in planning several times, so the term is the number of simulations plus one)}.\footnote{The equation does not mention the ``self''. This is because I reduce self-based suffering to frustration of self-needs based on the logic explained in Chapter~\ref{self.ch}; that is the logic used in the following chapters. Alternatively, inspired by Buddhist philosophy \add{or the discussion on self-related frustration in Chapter~\ref{summary1.ch},} we might think about adding another multiplying factor to the equation, called ``relevance to self'', which would measure if the reward loss is affecting the self (in some sense to be defined).\index{self!and frustration}
} 

It should be emphasized that such frustration happens on different time scales: from milliseconds to even years, perhaps. In the smallest timescales, the suffering is likely to be much weaker than in the longer time scales; see Chapter~\ref{summary1.ch} (page~\pageref{timescales}) for discussion on this topic. The equation above is intended to be applied separately on different time scales.\footnote{It should be useful to formulate a similar equation based on  RPE instead of reward loss. Such a formulation would handle these complex temporal aspects in a more principled way, and  in particular, it would encompass frustration based on expectations alone, as treated in Chapter~\ref{rpe.ch}, especially footnote~\ref{predictedfrustrationfn}. I leave that for future research.}

Another point that is useful to recall here is how we reformulated action selection as being based on rewards in Chapter~\ref{rpe.ch}. While we still often talk about goals and planning, for example in the flowchart in Figure~\ref{flowchart2.fig}, the goals are now seen as something that the agent itself sets in order to maximize rewards. In particular, Chapters~\ref{dual.ch} and \ref{emotions.ch} explained how an agent would predict that a certain state gives a big reward, then set it as a goal state, and start planning for it---the same logic underlies Figure~\ref{flowchart2.fig}. Thus, goals are not something inherent in the world, but rather a computational device used by the agent in order to maximize rewards.
That is why this equation uses the formalism of reward loss, instead of the basic formalism of frustration of goals initially used in Chapter~\ref{planning.ch}. Still, goals are implicitly present in this equation since the expected reward is often the reward that reaching a certain goal would give (according to the agent's prediction).

\subsection{Toward designing interventions}

Now, the essential point here is that all the terms on the right-hand side of the \eqsuff\ are something that can be influenced or intervened on, at least to some extent. That means it is possible to develop methods that change the terms on the right-hand side, and thus change the amount of frustration and suffering. The next two chapters are largely an explanation and application of this equation from that viewpoint.

\add{The focus here is on frustration instead of threat. This is justified by the fact that our theory reduces threat to anticipated frustration, as discussed in Chapters~\ref{self.ch}, \ref{threat.ch}, and \ref{summary1.ch}. In particular, if there were no frustration, neither currently perceived nor anticipated, there would be no threats perceived either. Thus, \textit{reducing frustration provides a very general basis} for reducing suffering.}
\add{Also note that there is another, apparently very different mechanism for reducing suffering: reducing desires and aversions themselves. If there are no desires, there can be no frustration either, at least in the sense of desires not being fulfilled.\footnote{\add{Recall we have two different models of frustration: expectation-based and desire-based, as discussed in detail in Chapter~\ref{summary1.ch}.}} This case will also be covered in the next chapter.}

In fact, so far, it might actually seem that the book has been just one big complaint. Suffering seems unavoidable, a necessary consequence of intelligent information processing. However, in the following final chapters, we will see a way forward: what an intelligent agent can do to actually reduce its suffering.

\chapter{Reprogramming the brain to reduce suffering } \label{freedom.ch}

In this chapter and the next, I will present various ideas on how to reduce suffering in a complex intelligent system acting in a complex world---such as humans. I derive various ways how information-processing should be changed, i.e.\ how the agents should be reprogrammed, based on the theory presented in this book. Since the systems in question, such as our brain, have largely learned their function from input data, an important part of such reprogramming is retraining the learning system by inputting new data into it.

The methods discussed here are not original: almost all come from Buddhist and Stoic philosophy or related systems. The goal here is to interpret them from a computational AI perspective, using the theory developed in this book. Thus, we gain more understanding on how they work, why they work, and what could be done to improve them.

The main starting point in this chapter is the \eqsuff\ we just encountered (page~\pageref{suffeq}). We can try to reduce suffering by changing any of the terms on the right-hand side of the equation, since that inevitably implies that the frustration on the left-hand side of the equation is reduced. We can see from the equation that the obtained reward should be increased because it has a negative sign in the difference computed. In contrast, all the other terms on the right-hand side should be decreased because their contribution to frustration is positive.

Maximizing the obtained reward is really a very conventional way to try to reduce suffering, based on the wide-spread view that happiness comes from having achieved all your goals and having got what you wanted.\footnote{\citep{oatley1987towards,van2005experientialism,heathwood2016desire}; cf.\ Diotima in Plato's \textit{Symposium}: ``[T]he happy are made happy by the acquisition of good things''.\index{Plato} For a discussion of different definitions of ``well-being'', which can be seen as equivalent to happiness here, see \citet{sep-well-being,fletcher2015routledge}. \add{\citet{huta2014eudaimonia} discuss the particularly important  distinction between ``hedonia''  and ``eudaimonia'' (roughly, happiness as pleasure/feeling vs.\ happiness as meaning/virtue). \add{\citet{eldar2016mood} explicitly link happiness to long-term RPE.}
  }\index{well-being}\index{happiness} } However, that is difficult for reasons which are rather obvious. Many resources are limited: not everybody can have the best cars, the best wines, and the best sex partners. There is fierce competition over such resources, and not everybody can win. Besides, expectations are adapted to the obtained level of rewards, so what used to feel good no longer brings happiness after a while, as discussed in Chapter~\ref{rpe.ch} (page~\pageref{treadmill}).

So, instead, we attempt here to \textit{reduce all the terms other than the obtained reward} in the \eqsuff. In this chapter, we start by considering how it is possible to reduce two of those terms: the (perception of) expected reward, and the certainty attributed to the perception of reward loss. (The next chapter will consider reducing the remaining terms, as well as some further methods.) Such reduction also includes reducing self-needs as a special case, thus complementing the \eqsuff\ by the logic of the flowchart  in Fig.~\ref{flowchart1.fig}.  Ultimately, such practices lead to reducing all desires and aversions. 
This approach may be rather unusual in the context of modern Western psychology and philosophy, but it is thoroughly standard in Buddhist and Stoic philosophy.

\section{Reducing expectation of rewards}

Let us first look at the term ``perception of the difference of expected and obtained reward'', i.e.\ perception of reward loss, in the \eqsuff\ on page~\pageref{suffeq}. This should be made as small as possible, ideally zero or even negative. As already mentioned, the most conventional way to reduce it would be to try to increase actually obtained rewards, but that is very difficult.  So, we need to do something more clever. A well-known idea in Buddhist and Stoic philosophy is to lower your expectations. %
Then, your reward loss should be smaller; it will perhaps vanish altogether.

The expected reward is typically a product of two things: the probability the agent assigns to obtaining the reward, and the actual amount of the reward if it is obtained (considering the basic case where the amount of reward, if obtained, is fixed). Thus, reducing the expectation of a reward can be accomplished in two ways: either reducing the probability the agent assigns to the reward, or reducing the value it sees in the reward.\index{expectation!of reward}
This can be compared to a lottery. Suppose your initial chance of winning a Porsche is 1\%. Obviously, the lottery would be made less attractive if the probability of winning is lowered to 0.01\%; it would also be less attractive if you realized that the Porsche is second-hand and not so cool after all. In both cases, your expected reward is reduced.

Most importantly, rewards in the real world are always a bit subjective, and so are the probabilities we assign to them. A new Porsche may feel like a great reward to one person, while it may matter very little to another; this is why we have to talk about \textit{perceived} reward.\index{reward!perceived} People will also have very different guesses of the probability of winning it. Since these quantities are subjective, it is possible to change our estimates of them by changing our beliefs, perceptions, and associations, even if the actual physical reality remains unchanged.\footnote{The subjectivity and contextuality of perception %
  is actually quite complex  and creates many further possibilities of reducing the term being considered here. Logically, we might also try to \textit{increase} the perceived \textit{obtained} reward independently of the actual reward obtained.  This should be possible by somehow learning to better appreciate the rewards obtained, such as by a feeling of gratitude or appreciation, as briefly discussed in Chapter~\ref{attitude.ch} %
  Likewise, it might be possible to reduce the perceived reward \textit{loss} even if the actual perception of the reward and the expected reward are unchanged.  
  Seeing the reward loss as a useful learning signal might work in that direction, but I will not pursue that possibility any further.}

A key goal of Buddhist and Stoic systems is exactly such re-evaluation of the probabilities and rewards. To accomplish this, Buddhist philosophy talks about the ``three characteristics of  existence'', which are impermanence, no-self, and unsatisfactoriness.\index{three characteristics (Buddhist)} They map roughly to our concepts of uncertainty, uncontrollability, and unsatisfactoriness we discussed in the preceding chapter.\footnote{See footnote~\ref{threecharfn} in Chapter~\ref{overview.ch} for a detailed discussion of the connection.} Each of these characteristics gives a reason why the rewards are actually lower than what they would otherwise be, or what they appear to be, as will be explained next.

\subsection{Facing uncontrollability}

\index{uncontrollability!facing it}
Uncontrollability, discussed in Chapter~\ref{control.ch}, is a key concept here. 
The level of controllability is clearly related to the level of expected reward. If you think the world can be controlled, you will expect to achieve high rewards, because you think you are able to take courses of action that give you the very highest rewards, and you are reasonably certain that you can achieve them. \add{Thus, you're exposed to strong frustration since your expectations are high.} In contrast, 
if you think the world is \textit{un}controllable, you assign a low probability to achieving any rewards, and the higher rewards may seem to be completely out of your reach. Then, your expectation of reward is smaller, and you are less likely to suffer from a reward loss, i.e.\ frustration. This is how considering the world to be  uncontrollable reduces suffering.

To transform this logic into a  method for reducing suffering, the trick is to acknowledge the fact that you have little control and there can never be very much control, and firmly believe in that fact. 
We saw earlier how  the Stoic philosopher Epictetus emphasizes how little we can control (page~\pageref{epicontrol}). He continues by explaining that if we are mistaken about this point, suffering is inevitable:\footnote{Paragraph 1 of \EN.}\index{Epictetus}
\begin{quote}
The things in our control are by nature free, unrestrained, unhindered; but those not in our control are weak, slavish, restrained, belonging to others. Remember, then, that if you suppose that things which are slavish by nature are also free, and that what belongs to others is your own, then you will be hindered. You will lament, you will be disturbed, and you will find fault both with gods and men. 
\end{quote}
\add{To put this idea into practice, on every occasion where you are inclined to develop a desire or aversion towards an object or event, you should ask yourself whether you can control it or not. Basically, you cannot control anything external to you, such as your possessions, other people, or their opinions. Epictetus suggests you should not see any such external, uncontrollable things as good or bad, or as bringing any reward in our terminology.}\footnote{\Discourses, III.3.14-15 and III.8.1-2}
  
Likewise in Buddhist philosophy, the original form of the no-self philosophy says that nothing is part of me, which is a way of saying that nothing can be controlled, as we saw in Chapter~\ref{control.ch}. Understanding this is crucial according to the Buddha:\footnote{\SN{22.59}, latter half, strongly shortened, based on the translation by \citet{mahasianatta}.}
\begin{quote}\index{uncontrollability!Buddhism}\index{no-self!Buddhism}\index{Buddha}
All [mental phenomena], whether past, future or present, internal or external, gross or fine, inferior or superior, far or near, should be seen with one's own knowledge, as they truly are, thus: 'This is not mine, this I am not, this is not my self.' (...) [S]eeing thus, [the disciple] grows wearied of form, wearied of feeling, wearied of perception, wearied of volitional formation [i.e.\ desire and aversion], wearied of consciousness. Being wearied, he becomes passion-free (...), he is emancipated [from processes leading to suffering].

\end{quote}
Here, I interpret ``growing wearied'' as signifying that the reward expectations are lowered, or little enjoyment anticipated. Thus, the point is that recognizing uncontrollability, or inexistence of self, reduces expectations of reward, and this reduces suffering. 

Buddhist philosophy further emphasizes the importance of understanding ``causality''.\index{causality!Buddhist}\label{causality} Such causality \linebreak
means that events in the world just follow from each other based on natural laws, for example those depicted in Figure~\ref{flowchart2.fig}. This thinking minimizes the importance of free will and the control that the agent can have over the world; it is related to what is called determinism in Western philosophy. I would think, therefore, that the Buddhist emphasis on what they call causality is just another viewpoint on uncontrollability; seeing such causality is one way of realizing that the world is uncontrollable. Stoic philosophers advocated the study of the natural sciences\citenew{sep-stoicism} (which they simply called ``physics''), with a similar goal.\footnote{Causality is a topic of great current interest in AI \citep{pearl2009causality,peters2017elements,gershman2017reinforcement}.\index{causality!in AI} However, the meaning of the term in AI is a bit different, and in particular, very specific: It is more about the difference between correlation and causality, and how an agent could learn that difference. In AI, understanding such causality will enable the agent to act more efficiently, increase its control of the world and the rewards it obtains, as well as better predict the rewards. In contrast, in Buddhism, understanding causality is about admitting the determinism of the world and minimizing the perceived control of the agent. Eventually, both these two kinds of ``understanding causality'' may reduce suffering in their own ways. Briefly, understanding causality in the AI sense means that you can find optimal actions and increase rewards, while the Buddhist understanding means appreciating how little reward even those optimal actions bring, thus reducing expectations.  } %

\subsection{Facing uncertainty, unpredictability, and impermanence}

\index{uncertainty!facing it}\index{unpredictability!facing it}\index{impermanence}
Uncontrollability is closely related to the concept of uncertainty.
Uncertainty feeds into  uncontrollability: if the workings of the different objects in the world are uncertain, even quite random, the world cannot be very well controlled.   Likewise, uncontrollability leads to uncertainty about whether rewards will be obtained. 
In some sense, these are two sides of the same coin. 

Buddhist philosophy focuses on the related concept of impermanence, which can be largely seen as a special case of uncertainty. \label{impvsunc}
Impermanence means that the world is constantly changing, and usually in unpredictable ways.\footnote{From an alternative probability theory viewpoint, impermanence could also be seen as incorporating \textit{nonstationarity},\index{nonstationarity}\index{impermanence!as nonstationarity} which is the technical term for the situation where the world is changing, and a statistical model learned on data in the past may not be valid anymore in the present and even less in the future. Such problems were already alluded to in previous chapters when it was pointed out that humans may be evolutionarily adapted to the African savannah\index{African savannah} instead of the modern city environment (Chapter~\ref{rpe.ch}); or that emotional reactions learned as a child may be far from optimal in an adult since the environment is very different (Chapter~\ref{emotions.ch}). It could be argued that nonstationarity is as important as scarcity of the data, and an independent root cause of suffering.\index{suffering!root causes} However, I don't pursue that line of argumentation here since I tend to think that any problems with nonstationarity could be avoided if the agent had enough data and computation since it would then be able to predict the nonstationarity (like Laplace's demon in footnote~\ref{laplacedemonfn} below); but this is admittedly a complex and controversial point that needs further research.\label{nonstatfn}}   For example, any object that you possess can break or get lost. Any enjoyment that you get is likely to be fleeting. In fact, even your feelings and opinions are impermanent: today you like one thing, but perhaps tomorrow you're already bored with it and want something else; what you consider important today may have no significance to you next month. Obviously,  impermanence thus interpreted leads to uncertainty.\footnote{However, the Buddhist impermanence has also aspects that cannot be considered to be forms of uncertainty. For example, if you know for sure that you will die tomorrow,\index{death} there is no uncertainty, although this is the quintessential expression of impermanence in Buddhism\add{---but it could be argued that there is still uncertainty about what happens after death}. In fact, in a meta-level sense, the central point in early Buddhist philosophy of impermanence is that impermanence itself is absolutely certain, as well as ``permanent'': Everything will perish one day for sure. Some exceptions may exist, however:  ``enlightenment'' (nirv\=ana or nibb\=ana) is permanent according to most schools \citep[p.~52]{harvey2009theravada}, consciousness is permanent for some \yogacara\ thinkers  \citep[p.~99]{williams2008mahayana}, and a rather obscure metaphysical construct called \textit{dharmak\=aya} is also permanent in some Mahayana schools \citep[p.~106]{williams2008mahayana}.\index{Yogacara}}

Going back to our \eqsuff, the consequences of uncertainty are very similar to the consequences of uncontrollability. The central point is that any future rewards are uncertain, i.e.\ unpredictable. Rewards and the circumstances leading to rewards can change, so an agent cannot really know whether it will get any reward after executing its plan. Thus, the agent should lower the probability it assigns to any future reward. If the agent acknowledges such uncertainty of the world, its expectations regarding rewards will be lowered, just like in the case of uncontrollability. Consequently, frustration will be reduced. (Later, I will talk about perceptual uncertainty, which has a different effect on suffering.)

It is quite paradoxical that Buddhist practice, which turns your attention to uncertainty and uncontrollability, tends to reduce stress and suffering.\index{stress} In Chapter~\ref{self.ch} we saw that uncertainty and uncontrollability are usually thought to lead to more stress, not less. I think the paradox has a lot to do with one's attitude to uncertainty and uncontrollability. Somehow Buddhist philosophy seems to result in a particularly appropriate attitude, related to their \textit{acceptance}, which will be considered in more detail in Chapter~\ref{attitude.ch}.\footnote{This contradiction between ancient philosophers and modern psychology has baffled many commentators. I would speculate that the problem is that not all research on lack of control (or uncertainty) has clearly made the distinction between the level of control/certainty the agent \textit{perceives} to have, and the level of control/certainty the agent \textit{wants} to have. In a typical experimental paradigm, the perceived control can be easily reduced by experimental manipulation, but the control that the agent wants to have may be unchanged. Then, the agent may want to have more control or certainty than it perceives to have, and this leads to a case of frustration on a ``meta-level'' of self-needs, and thus suffering.   A Buddhist practitioner is supposed to accept that control is not possible, and everything is uncertain. Thus, she gives up wanting any control and avoids any such meta-level frustration; as just argued in the main text, frustration on the ordinary level is reduced as well. Thus, such an attitude should lead to less suffering.}

\subsection{Facing unsatisfactoriness}

\index{unsatisfactoriness}
In Buddhist philosophy, the two characteristics of impermanence and no-self (roughly,  uncertainty and uncontrollability) are complemented by a third characteristic: unsatisfactoriness, which has many meanings and interpretations. On the one hand, it expresses the idea that whatever we try to achieve, we often fail due to uncontrollability and uncertainty. In this sense, it simply recapitulates those two aforementioned properties. On the other hand, unsatisfactoriness can be seen as an extremely general characteristic which penetrates all phenomena and all existence. In fact, in the original Indian texts, the single word \textit{dukkha}\index{dukkha} is used to express such unsatisfactoriness as well as suffering, i.e.\ this very thing we are trying to reduce. One could express the relation between these two meanings by saying that all phenomena are unsatisfactory in the sense that they can produce suffering, one way or another.\footnote{On the translation of \textit{dukkha}, see \citet[p.~244]{analayo}.} 

In Buddhist philosophy, it is recommended to acknowledge the unsatisfactoriness of all phenomena.  A basic justification for this can be constructed using our \eqsuff: if the agent is strongly convinced about the unsafisfactoriness of all phenomena, its expectation of reward will be very low, and the reward loss will be small and rarely even occurs. Thus, here we are talking about a very general, if a bit vague, strategy for lowering the expectations of rewards.
As a training method of great generality, the Stoics suggested reviewing any plan of future action with the view of anticipating what could go wrong and how the plan will not lead to great enjoyment after all. Epictetus gives a famous example of going to a Roman bath:\footnote{Paragraph 4 of \EN.} 
\begin{quote}\label{romanbath}\index{Epictetus}  
If you are going to bathe, picture to yourself the things which usually happen in the bath: some people splash the water, some push, some use abusive language, and others steal. 
\end{quote}
With this mindset, you will not expect much enjoyment, i.e.\ reward, and you will not be disappointed. Such a scenario could be analyzed in terms of uncontrollability and unsatisfactoriness as well, but unsatisfactoriness may be a more natural viewpoint.\footnote{In particular, such mental imagery may serve to increase the perceived probability of adverse events based on what is called the availability heuristic \citep{tversky1974judgment}.\index{heuristics!availability h.}  That is, humans tend to estimate the probability of events based on how easily they can recall (or imagine) those events. If you willfully imagine an event happening in the future, that will make the event more accessible in terms of memory retrieval, and thus you may start considering its probability of happening is higher. When things going wrong are perceived to have a higher probability, expected reward is necessarily reduced. Thus, imagining adverse events may be a particularly powerful way of influencing unconscious decision-making processes. \add{It could be counterargued that such simulation of negative events creates suffering by itself. This may be true, but it seems that the suffering created by such \textit{deliberate} simulation is quite small compared to the possible real-life frustration.}\label{availability}}

Using our computational theories, we can penetrate still deeper into the meaning of unsatisfactoriness. Suppose you can get chocolate quite easily and there is little uncertainty about its great taste. It is not obvious how uncertainty or uncontrollability would be a problem here. But, we can still say that the chocolate is unsatisfactory because there are various \textit{negative long-term side effects} hidden in the apparently rewarding object.

To see what such negative long-term effects might be, recall how in Chapter~\ref{overview.ch}, we defined unsatisfactoriness in a more specific way, based on two computational ideas. %
First, we had the concept of \textit{insatiability} (Chapter~\ref{rpe.ch}). An intelligent system programmed to maximize reward will never be satiated or satisfied, by the very construction of the system. It will never find that it has had enough, because in the long run, getting more reward will increase the expectation of rewards. In a word, the system is infinitely greedy. %
The second aspect of unsatisfactoriness in our framework was \textit{evolutionary obsessions} (Chapter~\ref{rpe.ch}).\index{obsessions!evolutionary} Even the very goals pursued and the rewards obtained can be questioned. Perhaps the evolutionary system gives you a certain reward for drinking a sugary drink. But we know very well that such a reward is misleading: the sweet drink is not good for you when all its effects are considered in the long run. %

These two computational viewpoints of unsatisfactoriness point at mechanisms which are very different from uncontrollability or uncertainty, or even simply reducing reward expectations. The implication is that even if we could totally control the world and everything were certain, the result of our strivings would not be that great anyway because it would not produce a lasting satisfaction or pleasure.  While uncertainty and uncontrollability are more about the probability of getting various kinds of rewards, unsatisfactoriness (both in our sense and the Buddhist sense) is really about the \textit{real worth} of the rewarding objects or events themselves, when considering the bigger picture. Even the very best chocolate, if you eat it every day, will ultimately leave you indifferent, and may ruin your health in the long run.

Our definition of unsatisfactoriness actually works a bit outside of the \eqsuff\ because it is not that the rewards or their probabilities (or any other terms in that equation) are changed: it is rather understood that even if the rewards are obtained, there will be side effects in the distant future. The \eqsuff\ is in a sense short-sighted: it only considers the direct, immediate effects of rewards or their simulation.\footnote{Thus, our computational definition of unsatisfactoriness does not exactly lead to reduction of reward expectations in the strict framework of the \eqsuff, unlike the basic Buddhist interpretation given in the text. However, understanding the negative long-term effects may also reduce the expectation of immediate reward if the long-term effects are in some sense added to that expectation.  Recognizing such unsatisfactoriness should ultimately reduce desires since that appears to be the only way of completely escaping this logic. With less desires, there would be less opportunity for any frustration to arise, while insatiability or evolutionary obsessions become irrelevant. Reducing desires will be considered in detail later in this chapter.} In contrast, the ideas of insatiability and evolutionary obsessions bring a longer time scale into the picture, pointing out that obtaining rewards now may actually increase frustration and suffering \textit{in the long run}.
\footnote{
\index{aversion}
It may be more difficult to see why the aforementioned attitudes would also reduce frustration due to aversion, or a negative reward.
Let us consider aversion based on expecting that a bad thing is likely to happen, such as your neighbours starting a noisy renovation. Now, taking account of uncontrollability means you cannot really avoid the bad thing, at least not with any certainty. This means that the probability of the bad thing happening is larger than what you might have initially thought---your flat will be noisy for sure. Thus the expected reward is less than what you would have thought without taking uncontrollability into account. More precisely, it is more negative, since the probability of a negative reward is larger. Thus, frustration is reduced by reducing the expectation of reward by making the negative expectation even more negative. %
Likewise, thinking in terms of unsatisfactoriness (in the Buddhist sense) means thinking that the bad thing is likely to be really bad---the noise is probably going to be something quite unbelievable. Again, this reduces expected reward in the sense of making it even more negative, and what actually happens is less likely to give you a negative surprise and frustration.\index{unsatisfactoriness}
On the other hand, a classical Buddhist account would point out that impermanence\index{impermanence} means that the object of aversion will eventually disappear, which makes at least the feeling of aversion weaker. Clearly, it will give me some comfort knowing that the noise will not be there forever. 
However,  putting uncertainty into the framework of the \eqsuff\ is not straightforward in the case of aversion, or negative reward. Taking account of the various forms of uncertainty might mean that you realize that the bad thing, which you initially thought is certain, is actually less likely to happen than what you first estimated. Paradoxically, this increases the expected reward, because the negative reward is less likely to happen, and actually increases your frustration. Thus, uncertainty may need to be treated in separate ways depending on whether the reward is positive or negative. %
}  

\section{Reducing certainty attributed to perception and concepts}

\index{uncertainty!of perception}
Another term in the \eqsuff\ that we can reduce is the ``level of certainty attributed to that perception''. As we saw in Chapter~\ref{perception.ch}, perception is uncertain. To recapitulate the main ideas: perception is based on limited data, thus necessitating unconscious inference, which may not always be much better than guessing. Perception is also subjective: different people can have different priors and thus different perceptions. Subjectivity is made even more serious by the strong selection of incoming information by attentional mechanisms. 
Since the computational capacity is always fundamentally limited, and the world is awesomely complex, it is not possible 
to build a perceptual system that always makes correct inferences, let alone one that perceives the ``true'' reality.
This should imply a fundamentally skeptical attitude towards any perception: we should not make too strong conclusions based on sensory input. 

Thus, we see that uncertainty has two different aspects. There is the objective unpredictability of the world: surprising and unexpected things can happen, the world is to some extent random---this is the kind of uncertainty we focused on earlier in this chapter when talking about the importance of recognizing uncertainty and impermanence.\index{uncertainty!and unpredictability}\index{unpredictability!and uncertainty} 
But here, we focus on the uncertainty in our perceptions and beliefs of the world, which I here call perceptual uncertainty. The point is that we don't know with any great certainty what the state of the world is, since we have neither enough data nor enough computation to perceive it properly.\footnote{This division into two kinds of uncertainty could be criticized on philosophical grounds. Laplace proposed that an intellect (called a ``demon'' by later commentators) which knows everything about the world would be able to perfectly accurately predict everything, and nothing would be uncertain to it. Thus, from this viewpoint, uncertainty is always a reflection of ignorance about some aspects of the world.\index{demon!Laplace's}\index{Laplace} I shall not go into that debate here, and just acknowledge that the division I make here is not very rigorous while in line with how randomness is often conceptualized in AI theory. This is basically the distinction between ``aleatoric'' and ``epistemic''  uncertainty discussed, e.g., by \cite{hullermeier2021aleatoric,charpentier2022disentangling,lockwood2022review}.\label{laplacedemonfn}}
Such perceptual uncertainty increases the effects of unpredictability and uncertainty that we saw earlier,  since it makes the world even more unpredictable for the agent.

If the agent is intelligent enough, it will take perceptual uncertainty into account
when evaluating the reward loss or frustration. Suppose the agent has completed an action sequence in view of getting reward, and it tries to evaluate the reward loss.  
Now, the agent should understand that it cannot know with certainty how much reward it got.  A drink may have tasted good, but you cannot know if it was actually good for you.\index{uncertainty!of reward loss}\index{reward loss!perceived}
That is why in the definition of reward loss, we should really be talking about \textit{perceptions} of rewards instead of any objective quantities; this is precisely what is done in the \eqsuff.\footnote{To emphasize: our basic definition of reward loss on page~\pageref{frustdef} does not take into account the fact that it is perceptions that matter. Obviously, it cannot then take the uncertainties into account either. Thus, the definition must be changed accordingly, and this was done in our \eqsuff\ on page~\pageref{suffeq} by multiplying the perceived reward loss by the certainty of perception. See also footnote~\ref{lossperc} in Chapter~\ref{overview.ch}.}
Since the reward loss is uncertain, any conclusion drawn from it should not be given too much weight.\footnote{Here we focused on the uncertainty in obtained reward. It could also be asked if there can be uncertainty in the expected reward.  In an orthodox Bayesian interpretation, it may in fact not be possible to say that there is any uncertainty about the expected reward, since the expected reward is a subjective quantity, something purely defined by what the agent believes and expects. In contrast, in a frequentist intepretation, the expected reward is an objective quality in the outer world (how much the agent would get on average if it repeated the same action many times) that could further be considered a parameter in a statistical model. Therefore, it can be misestimated, thus adding to the uncertainty of reward loss.\label{expsubjectivefn} Notwithstanding such theoretical arguments, I think it is clear that for biological organisms, understanding the real evolutionary value of, say, a piece of food may actually be a highly complex process involving a lot of learning and computation, so it can surely go wrong, as in the case of sugary food, which means it is meaningful to say that there is uncertainty about the expectation. %
} This is an implication of the basic principles of Bayesian inference as used in AI;\index{Bayesian inference} many philosophers over the centuries have also pointed out that what first appears to be a negative outcome may even turn out to be positive, and vice versa.\footnote{Let me just mention the great Chinese classic Huainanzi's ``The old man lost his horse''.}

If the agent is programmed to take account of the fact that all its perceptions are uncertain, it would likely have weaker reward loss signals. Consider an agent that attempts to get some chocolate. Suppose that after executing a plan, the agent is able to eat some, but its program ``understands'' that it does not really perceive the amount of chocolate with any certainty; perhaps because it swallows all of it immediately without really taking a look. Intuitively, it does not then make a lot of sense to send a strong reward loss signal: such a signal would be too much guesswork and would not provide a proper basis for learning better behavior. In other words, uncertainty about the correct signal to send should lead to a weaker signal.%
\footnote{\add{It may be intuitively clear that acknowledging the uncertainty of perception should lead to ``weaker'' signalling of frustration; in this footnote, I explain how to make that idea more rigorous. We can consider, as an illustrative example, one of the simplest online learning tasks, namely linear regression. There, we minimize a quantity such as $\sum_t (y_t-ax_t)^2/\sigma_t^2$ where $x$ is input, $y$ is output, $\sigma_t^2$ is noise level, and $a$ is a parameter to be estimated. The magnitude of the error signal for each data point is proportional to the inverse of the noise level $\sigma_t^2$. Thus, for a high noise level (large uncertainty), the error signal is smaller. If the noise level is estimated separately for each data point (or time point $t$), this will have the effect of reducing the error signal at time points where there is a lot of uncertainty as modelled by the noise level $\sigma_t^2$. The concrete algorithm used here might be what is called the delta rule; see \citet{korenberg2002bayesian} as an example of a related if slightly more complex model, and \citet{kendall2017uncertainties} for a more sophisticated deep learning model. In the context of reinforcement learning, \citet{mai2022sample} propose a closely related weighting for RPE, where indeed we see how the learning proceeds by minimizing the expected (squared) RPE so that it is down-weighted by the estimated variance of the RPE.
  }
    \label{deltafn}}
Thus, taking account of the uncertainty of the perception of reward would reduce suffering.\footnote{One aspect of uncertainty which is not explicit in the \eqsuff\ is that there can also be an RPE due to changing predictions. Suppose one moment you think you will get a reward, but the next moment it looks like you will not get it. This decrease in expectation induces RPE and thus suffering, as explained in footnote~\ref{predictedfrustrationfn} in Chapter~\ref{rpe.ch}. However, nothing may have actually happened, it was all just predictions in your mind. Importantly, such a change in prediction is only possible if there is uncertainty. If the future were certain, there would be no need to update your predictions, but because of uncertainty, the predictions change from one moment to another. Again, such suffering can be reduced if you realize that your predictions are uncertain; then the change in the predictions would be given less weight, according to the same principles as just explained in the text.}

\subsection{The Buddhist concept of emptiness}

The perceptual kind of uncertainty has a central role in the later Mahayana schools of Buddhism.
While the ``three characteristics'' (impermanence, no-self, unsatisfactoriness) form the core of the Buddha's original philosophy, later Buddhist philosophers found them somewhat simplistic.
The emphasis shifted to the properties and limitations of perception and cognition, as opposed to characterizing the outer world. 
The inaccuracy of perceptions and beliefs became essential as part of the multifaceted concept of ``emptiness'' widely used in Mahayana Buddhism---although rarely by the Buddha himself.\footnote{Though see \SN{22.95}, where the Buddha clearly talks about a general emptiness of the Mahayana kind, while using a slightly different terminology: he does not use the word \textit{su\~n\~nat\=a/\'s\=unyat\=a} which is the term usually translated as emptiness, and became prominent in later texts. See also \MN{121} for a very different early view on emptiness, and \citet[p.~54]{williams2008mahayana}.}

Emptiness has many meanings.\index{emptiness (Buddhist)!definitions}
In the framework of this book, we can  consider emptiness as an umbrella concept encompassing several of the ideas related to information-processing that we have seen in this book, in particular uncertainty, fuzziness, subjectivity, and contextuality. To summarize it in a single word well-known in Western thinking, we could call it ``relativity''.\footnote{I am here referring to the common, non-technical definition of relativity, such as ``the state of being dependent for existence on or determined in nature, value, or quality by relation to something else'' (Merriam-Webster.com, accessed 24/1/2022). The interpretation of emptiness as relativity was initiated by Theodore Stcherbatsky, one of the earliest Western interpreters of Buddhist philosophy. Some commentators may prefer ontological interpretations of emptiness, but my treatment here sees it more as an epistemological quality, compatible with my computational approach. Emptiness actually has two different but related well-known definitions in Mahayana Buddhism \citep{williams2008mahayana}. First, there is the \yogacara\ definition\index{Yogacara!definition of emptiness}\index{emptiness (Buddhist)!definitions} based on the ``consciousness-only'' thinking described in Chapter~\ref{consciousness.ch}: All phenomena in  the world are called empty because they are simply phenomena in the mind and constructed by the mind; in particular, any categories and concepts are merely mental constructs. This is rather similar to what we just discussed, except that in \yogacara, such thinking can even be taken to a metaphysical level, denying the existence of the outside world---at least in some interpretations. Second, there is the Madhyamaka definition,\index{Madhyamaka} where all phenomena are called empty in the sense that they are simply products of long causal chains, thus lacking any independent, intrinsic existence, and subject to change at any time. This is a very general definition that is ultimately supposed to contain most related properties described in this book or other Buddhist schools; it is surprisingly similar to the dictionary definition of relativity just given. For example, subjectivity of perception\index{subjectivity!of perception} can be seen as a result of such causality because perception is causally influenced by the priors in the perceiver's brain, and thus the percept does not exist independently (of the brain).  } 
What the different aspects of emptiness  have in common is that fully appreciating them should make us take the contents of our minds less seriously.  

\add{While it seems fashionable to discuss such concepts in terms of Buddhist philosophy, very similar ideas can be found in Greek philosophy. We already saw the Skeptics questioning the reliability of any sensory information in Chapter~\ref{perception.ch}.\footnote{\add{Connections between Skeptic ideas\index{Skeptics} and Buddhist emptiness are discussed by \citet{garfield1990epoche,brons2018life,dreyfus2021madhyamaka}}}
On the other hand, Plato's famous theory of ``ideas'' (or ``forms'') describes a kind of true reality behind the sensory phenomena, thus denying the true existence of the phenomena. Seneca explains how this theory is related to reducing desires:
\begin{quote}\label{platoemptinessfn}
[A]ll these things which minister to our senses, which arouse and
excite us, are by Plato denied a place among the things that \index{Plato}
really exist. Such things are therefore imaginary, and
though they for the moment present a certain external
appearance, yet they are in no case permanent or
substantial; none the less, we crave them as if they were
always to exist, or as if we were always to possess them.\footnote{\add{\Lucilius, LVIII.26-27.
    Compare with footnote~\ref{chahfn} in Chapter~\ref{consciousness.ch}.
  }}
\end{quote}
Indeed, Seneca reads Plato as if he were a Buddhist philosopher propounding emptiness philosophy.
}

Concepts and categories are considered particularly empty in Mahayana philosophy. It proposes that the objects in the world do not really exist as separate entities, but are just part of a complex flux of perceptions happening in our consciousness. In this sense, there are really no separate objects or crisp categories in the world; they are purely constructions of the mind.
Zen\index{Zen} texts use the parable 
of confusing the moon and the finger that is pointing at the moon. Here, I would interpret this in the sense that the finger is a category, perhaps expressed by a word, that merely points at a phenomenon in the real world, that is, the moon. Ceasing to think in terms of categories and concepts, based on a recognition of emptiness, is something that generalizes the idea of reducing the certainty attributed to perception, or in fact, to your cognitive processes in general. It reduces frustration according to the logic given above for recognizing uncertainty of perception.
Furthermore, any valence that you would typically associate with a category cannot be considered certain anymore.\footnote{The Pyrrhonians explicitly advocated rejecting any valences (Sextus Empiricus's \textit{Outlines of Pyrrhonism}, 1.27-28).\index{Skeptics}  
If you admit that you're not sure about what category some object belongs to, valences as well as any further associations and generalizations have to be given up as well. This implies further processes that are more specific to categories, and complement the logic of simple perceptual uncertainty. }
 What may be Epictetus's most famous quote says: ``Men are disturbed not by the things which happen, but by the opinions about the things.''\footnote{\EN, Paragraph~5\index{Epictetus}, trans.\ G.~Long.
For example, if I think that what somebody else just did belongs to the category ``rude'', perhaps I should not be so certain about such inference. I can start with considering if my perception was incorrect: I may have completely misunderstood what he was doing, or what his goal was.
From the viewpoint of contextuality, I might consider if in this particular situation, his behavior was actually just right---or maybe I am in a foreign culture and don't know the rules. From the viewpoint of subjectivity, I might wonder if other people found his behavior commendable and if it is just me who finds such behavior rude.\index{subjectivity!of thinking}
From the viewpoint of fuzziness, I might ask: How does one define rudeness anyway, is there a well-defined criterion? Fuzziness is actually something whose effect on suffering we have not yet considered in detail, although it is an important concept---if not under this term---in relevant philosophical systems, such as Zen and the Pyrrhonian Skeptics. Chapter~\ref{dual.ch} argued  that while conceptual thinking uses crisp categories, many of the things in the world are fuzzy. %
  If you categorize events which are only borderline rude as simply rude, that is a form of overgeneralization: you may then be suffering unnecessarily due to your crisp-categorical thinking. The effect of fuzziness on suffering thus seems strongly analogous to the effect of uncertainty.\index{fuzzy}
}

\section{Reducing self-needs}

\index{self!needs!reducing}
In the \eqsuff\ above, we didn't have any terms explicitly related to self. Yet, self is obviously an extremely important concept from the viewpoint of suffering, as seen in Chapters~\ref{self.ch} and \ref{control.ch}. In our framework, self creates its own kind of frustration, by bringing aspects such as self-preservation,  self-evaluation (or self-esteem), and control into play. As such, self-related suffering is covered by the \eqsuff\ as a special case.

Many philosophical traditions such as Buddhism encourage reducing self-related thinking as a means to reduce suffering. 
\index{self!esteem and evaluation}
 One case of self-related thinking is related to the self-evaluation system.
In Chapter~\ref{self.ch}, it was proposed that a self-evaluation system constantly computes whether we have gained ``enough'' reward recently, looking at the relatively long-term performance of the system. (This long-term evaluation system is different from the one which computes the ordinary, short-term reward losses in the first place.)
  Such self-evaluation creates, as it were, another frustration signal on a higher level, in case the result of the self-evaluation is worse than some set standard.

  Logically, there are three ways of reducing negative self-evaluations. The first is similar to the ``conventional'' approach we discussed above regarding ordinary frustration: it is to really gain a lot of reward, so that you surely reach the standard required. This is obviously easier said than done. Furthermore, such striving may not reduce suffering at all because gaining a lot of reward may increase the expectations in the future, resulting in insatiability on a ``meta-level''.\footnote{Actually, the theory in the previous chapters does not exactly lead to such meta-level insatiability. We saw in Chapter~\ref{rpe.ch} how predictions are constantly updated, thus leading to insatiability. However, self-needs are not necessarily concerned with predictions but expectations of a different kind, as discussed %
    in Chapter~\ref{self.ch}. Still, it is possible that the expectations computed by the self-need systems are also updated based on past rewards, leading to meta-level insatiability. }
  The second approach, in line with the main proposals in this chapter, is to lower the standard of expected reward. For example, the aforementioned philosophical viewpoint that everything is unsatisfactory should work here as well. If the system expects little reward even in the long run, the self-evaluation should not  claim that the agent did not gain enough. 

However, there is clearly a third option: shut down the system that evaluates your long-term success. Such a shut-down is possible by convincing yourself of the total futility of the self-evaluation. The Buddhist philosophy of no-self should be particularly useful here. Admitting the lack of control, even lack of free will, implies that there is little to evaluate. If we cannot influence the world and the level of obtained rewards, what is the point in  evaluating my actions and learning strategies? On a deeper philosophical level, if it is not me that actually decides my actions---say, it is my neural networks----who is to be evaluated? Perhaps my neural networks and my body could still be evaluated, but not ``me'' really. On the other hand, what if ``my'' actions are ultimately determined by the input data or the environment, not ``myself''?

Suppose an agent were somehow able to shut down its self-evaluation system. It could be objected that such an agent with no self-evaluation might no longer be functional. However, even if the long-term self-evaluation were completely shut down, the system could still achieve most of its goals, and it will even be able to learn. Learning might just be slightly impeded because the learning system would not be optimally tuned to the environment. Thus, only ``learning to learn'', a kind of meta-learning, would be shut down, while the agent would be perfectly functional otherwise, even without self-evaluation.  

  I should emphasize another crucial point about self-evaluation. As long as the self-evaluation is based on evolutionary fitness, including what I called evolutionary \textit{obsessions},\index{obsessions!evolutionary} it does not actually make a lot of sense for us. It is too often based on criteria that are not in line with what humans should strive at, according to mainstream ethical principles. We need better criteria to decide if our actions were ``good enough''; criteria that would be more in line with what we consider a good human life should be about.\footnote{As self-evaluation can use social comparison\index{social comparison}\index{social comparison} as a baseline (see page~\ref{socialcomparison} in Chapter~\ref{self.ch}), it is also important to question the adequacy of such comparisons. For example, social media platforms may create unrealistically high standards regarding what  one should look like and what lifestyle one should have, partly because such content is carefully selected and even fake. When adolescents compare their own life with social media, there may be a huge gap, which may create mental health problems as reported by \citet{vogel2014social,verduyn2015passive} (but see also \citet{beyens2020effect}). How to reduce this kind of suffering: should people avoid using social media platforms? At the very least, it would be useful to understand the futility of such comparisons.\index{social media}}

  \index{self!preservation}Likewise, reducing the survival instinct, or information-processing aiming at self-preservation,  would seem to be useful for reducing suffering. Again, it could be objected that it is not good for the agent: such reduction may increase the probability of injury and even death. If I had no survival instinct, I might just happily go and pat a tiger I see in the jungle. This is a valid point, but we could still try to reduce the intensity of suffering incurred. In fact, religions and spiritual traditions invariably propose some method to cope with fear of death and mortality. Fear of death may sometimes be paralyzing, and quite often, it is unreasonable since  I may even suffer from seeing a tiger on TV. Therefore, a moderate reduction in survival instinct might have mainly positive consequences.\index{simulation!reducing} One method would be to reduce the mental simulations of injury and death; we will get back to this point in the next chapter, where we look at reduction of simulation by meditation.

\subsection{General reduction of self}

In addition to reducing the two specific self-needs just considered, we can aim at a more general ``reduction of self'', which can take many forms.\index{self!reducing}
To begin with, if we see the self as the source of control, and then we recognize  uncontrollability as discussed earlier in this chapter, this can actually be seen as a way of reducing the power of the self. As far as the self is about control, giving up control is, figuratively speaking, giving up part of the self. More precisely, it is rejecting part of the power that self-centered processing has on us.

Another approach is limiting the number of things that belong to ``myself''. Typically, I would consider that a number of things belong to me: perhaps my family, my home, my job, and so on. If I think of them as ``mine'', I invest them with a certain power because I think I should be able to control them, as well as keep them intact. In other words, I think that they are in a sense part of myself; some would say I ``identify'' with them. Then, if anything bad happens to them, or anybody tries to take them away from me, I will have a strong negative emotion as if my self were threatened---and in a sense the intactness of my person or self \textit{is} threatened.

It is clear how one can reduce suffering coming from such possessions: as a first approach, just own fewer things. If you have very few things that you consider yours, it is less likely that you will experience them breaking down, being stolen, or getting lost. Many spiritual traditions do recommend giving up most of your material possessions. Further, you can try to change your attitude towards such external parts of yourself. Epictetus proposes that you should think of all your possessions, your family, and so on, as not really belonging to you, but as things that have been temporarily lent to you:\footnote{\EN, Paragraph~11.}
\begin{quote}\index{Epictetus}
  Never say of anything, ``I have lost it''; but, ``I have returned it.'' Is your child dead? It is returned. Is your wife dead? She is returned. Is your estate taken away? Well, and is not that likewise returned?\footnote{\add{In \Discourses, III.24.85, Epictetus provides an ``anticipatory'' version of this: ``[I]f you kiss your child, your brother, your friend, (...)
keep reminding them that they are mortal. In such fashion do you too remind yourself that the object of your love is mortal; it is not one of your own possessions; it has been given you for the present, not inseparably nor for ever, but like a fig, or a cluster of grapes, at a fixed season of the year, and that if you hanker for it in the winter, you are a fool.''}\ See also \SN{22.33}, where the Buddha takes this approach to the extreme in the sense that he recommends abandoning everything, including your own body and any aspects of your mind.}
  \end{quote}
  Finally, the reduction of self can be approached from the viewpoint of reducing thinking in terms of categories.  Typically, I divide the world into things that are part of myself and things that are not part of myself. This is how I construct the category ``self''. Like with other categories, it would be useful not to take this category too seriously, and understand its fuzziness and arbitrariness, or emptiness. ``Self'' can be seen as the ultimate category that should be deconstructed and given up.\index{self!as category}\index{no-self!deconstructing self category} Such giving up of the whole category of self, in a sense, encompasses all the other aspects of no-self philosophy described above. If the very category of self does not exist, or, to put it simply, if self does not exist, what would be the point in self-preservation or self-evaluation, or any attempt to control? Any such self-related thinking should vanish if the underlying category of ``self'' is given up. The Buddha said that when a monk is advanced enough, ``any thoughts of `me' or `mine' or `I am' do not occur to him''.\footnote{\SN{35.205}}\index{Buddha} This is the most general way of reducing suffering based on no-self philosophy.\footnote{\add{Epictetus takes a rather different viewpoint on self. According to \citet{sep-epictetus}, for Epictetus  ``it is the volition [i.e.\ \textit{prohairesis}, or will]  that is the real person, the true self of the individual''.  Interestingly, Epictetus explicitly claims that such a conception of self is useful for mental training: ``It is a universal law (...) that every creature alive is attached to nothing so much as to its own self-interest. (...) Wherever `me' and `mine' are, that's where every creature necessarily tends. If we locate them in the body, then the body will be the dominant force in our lives. If it's in our faculty of will, then that will dominate.'' (\Discourses, II.22.15 and 19; see also III.1.40). It is precisely the will that is the target of Stoic mental training, so according to this logic, it it useful to think that the will is my self.  }}

\section{Reducing desire and aversion}

While so far we have focused on reducing the frustration of desires, many philosophical traditions propose that desires themselves should be reduced---as always, this includes aversions. In Buddhist philosophy of the Theravadan school, it is traditionally the main focus of the training, and it is the main point of the Buddha's teaching as expressed in the Four Noble Truths.\index{four noble truths (Buddhist)} After describing what suffering is (quoted on page~\pageref{fournoble1}), he proposed that it is born of desire, and that one can be liberated from suffering by eradicating desire by following a path of meditative and other practices.\footnote{For completeness, I will briefly describe the Buddha's Four Noble Truths in their entirety. They can be seen as a psychological theory of why suffering comes about and how it can be avoided. The four truths are a logical sequence: 1) All phenomena (i.e.\ external objects, perceptions, feelings, thoughts, etc.) in the world are unsatisfactory in the sense that they have the potential to produce suffering. 2) Suffering is produced by desire for any of these phenomena (or desire to avoid any of them, i.e., aversion). 3) Suffering disappears if desire is eradicated. 4) Desire can be eradicated  by following a certain combination of meditation techniques, philosophical attitudes, and ethical behavior. (For references, see footnote~\ref{fournoblefirst} in Chapter~\ref{suffering.ch}.) \label{fournobleall}}
Epictetus was equally clear about the importance of not having desires or aversions, especially towards things we cannot control:\footnote{ \EN, Paragraph~2; see also \Discourses, I.4.}
\begin{quote}\index{Epictetus} 
  Remove aversion, then, from all things that are not in our control (...)
  But, for the present, totally suppress desire: for, if you desire any of the things which are not in your own control, you must necessarily be disappointed; and of those which are, and which it would be laudable to desire, nothing is yet in your possession.\footnote{ \add{
Even Socrates is claimed to have said that 
``You seem (...) to imagine that happiness consists in luxury and extravagance. But my belief is that to have no wants is divine; to have as few as possible comes next to the divine'' (Xenophon, \textit{Memorabilia} 1.6.10, trans.  E. C. Marchant).\index{Socrates}
Other schools of Hellenistic philosophy had a very similar attitude. According to \citet[p.~13]{LongHellenistic}, a Pyrrhonian (Skeptic) will not ``decline or choose'' since desire is ``the first of all bad things''.\index{Skeptics}
  Reducing desires was recommended even by Epicurus, who seems to have been seriously misunderstood \add{(Seneca describes in \Beata, XIII how Epicurus was misunderstood already in ancient Rome).}\index{Epicurus}\index{Seneca} Epicurus proposed that there are a few desires which need to be satisfied since they are both natural and necessary: Food, water, and shelter; these desires are also easy to satisfy. In contrast, desire for money, power, fame etc.\ are unnatural and unnecessary; they are also insatiable. Optimal ``pleasure'' is obtained by rejecting desires which are not natural and necessary. See Epicurus's \textit{Letter to Menoeceus}, or
  \citet[p.34]{Hadot}; \citet{Konstan}.
Likewise, Seneca's \Lucilius, 21.7-8 proposes that the best way to increase pleasures is to reduce desires, while \Lucilius, 16.7-9   considers the insatiability/satiability (or satisfiability) distinction in more detail.  \citet{irvine2005desire} provides a modern account of the Stoic position.}\index{Seneca}\label{epicurusfn}}
\end{quote}
Humans can indeed reduce frustration simply by giving up some unnecessary goals: you don't really need a fancy car. It is possible to consciously decide not to strive for certain goals, and we can modify our desires to some extent without any special techniques. In our framework, this in particular means reducing intentions, i.e.\ commitment to plans, also called attachments in Buddhist terminology. Intentions can, in fact, be easier to reduce than desires themselves, as may be intuitively clear and will be discussed in more detail in the next chapter.\index{intention!reducing} Suffering will then be reduced since %
if there are no desires and no goals that need to be achieved, frustration will not appear, and neither will suffering. As such, reduction of desires is a central mechanism through which reduction of frustration is possible. 

Many ideas in this chapter can be seen as mental techniques serving the very goal of reducing desires. Consider, for example, reducing expected rewards as considered above: why would the agent want anything if it has arrived at the conclusion that the expected rewards are zero, or very small? Likewise, desires will be reduced by adopting the belief that many desires are pointless and even bad for you, they are just evolutionary obsessions. As such, reducing desires is closely related to the earlier ideas of facing uncertainty,\footnote{
  \add{For the Pyrrhonian Skeptics, recognition of the uncertainty and fuzziness of sensory evidence was the main method for reducing desires. After contemplating the conflicting evidence in different perceptions and inferences, as described on page~\pageref{epoche} in Chapter~\ref{perception.ch}, they eventually find an ``irresolvable conflict because of which [they] are unable to choose or reject''     (Sextus Empiricus, \textit{Outlines of Pyrrhonism}, 1.165, as quoted by \citet[p.~48]{LongHellenistic}). Thus desire and aversion are extinguished.} \label{epochefn}  \add{But paradoxically, uncertainty can sometimes lead to more vigorous action. Uncertainty about the availability of food increases foraging behavior in mammals, perhaps because they decide to hoard food or fatten themselves to protect themselves against such unpredictability \citep{anselme2019foraging}.   Even in human subjects in one economic experiment, uncertainty  increased motivation \citep{shen2015motivating}.}}
uncontrollability, and unsatisfactoriness, and in fact, in a traditional Buddhist account, the main justification for such philosophical attitudes is precisely that they reduce desires.\footnote{Let me try to make the links to the other ideas in this chapter explicit. First, reduction of certainty attributed to perception reduces desires since you don't actually know for sure whether the object of your desire really gives reward---or is even there. Second, if you cannot control anything, what would be the point in wanting, let alone planning, since rewards and goals cannot be attained? Seeing the insatiability of the desires should also lead to the conclusion that their total fulfillment is impossible in the long run, so the desires should be dropped as futile; seeing desires as evolutionary obsessions means realising they can even be bad for you. Self-needs, in particular self-evaluation, can be considered as forms of desires in this context, and the same ideas apply to them.}

There are also special techniques to reduce desires. One example is choosing to pay attention to good things that one already has, instead of things that one might obtain. This reduces desires and the tendency  of insatiability; it is central in mental exercises based on gratitude, which will be treated in more detail in Chapter~\ref{attitude.ch}.
\add{Reduction of desires is also facilitated by a simple, possibly ascetic lifestyle where there are fewer stimuli, or ``temptations'', that might elicit desires. A fashionable example is taking a break from the internet or social media; the monastic rules of Buddhist monks and nuns give a more radical example.}
 Epictetus also proposed a rather extreme form of training for this end, namely contemplation of death:\footnote{\EN, Paragraph~21} 
\begin{quote}\index{death!contemplation of}\index{Epictetus} 
  Let death and exile, and all other things which appear terrible be daily before your eyes, but chiefly death, and you will never entertain any abject thought, nor too eagerly covet anything.\footnote{\add{Likewise, Seneca (\Lucilius, 49.10) proposes ``Say to me when I lie down to sleep: 'You may not wake again!' And when I have waked: 'You may not go to sleep again!'. }\index{Seneca} It is easy to see why contemplation of death would reduce desires, and in particular planning and intentions. Presumably, contemplation of death reminds you that you just might die tomorrow, even if that is not very likely. By some kind of availability heuristic (see footnote~\ref{availability} in this chapter), that reminder will increase your estimate of the probability of dying soon, which implies that you don't have much time left to obtain rewards. So, their expected value is low, and any planning is less useful and highly restricted by this time horizon.
    Buddhist practices also include contemplation of death; it may serve slightly different purposes \citep[p.~155]{analayo}, but the classic manual \textit{Visuddhimagga} (Chapter VIII, 41) links it directly to ``disenchantment'' and ``conquer[ing] attachment''. Nevertheless, some psychological research based on the Terror Management Theory claims that reminding people of their mortality may, in fact, increase their willingness to consume \citep{kasser2000wealth}; see also \citet{burke2010two,gao2020neural}. \add{\citet{frias2011death} propose a model to understand why such quite opposite effects can be observed, and how exactly the death reflection should be done to increase gratitude and reduce greed.} } 
\end{quote}

Yet, there are also desires that are really ``hot'', hard-wired, and difficult to modify, let alone reduce, based on the rather purely philosophical or intellectual considerations presented in this chapter. What is needed are special techniques that work on deeper levels of the mind than philosophical thinking. Meditation is one such method, as we will see in the next chapter.\footnote{In addition to reducing desires, Stoics proposed another approach to working with desires. It is also possible to align one's desires with what can be more easily achieved in the world, instead of eliminating them. Epictetus takes this idea to the extreme by suggesting:\index{Epictetus}
``Don't demand that things happen as you wish, but wish that they happen as they do happen, and you will go on well.'' (\EN, Paragraph~8)
If the only thing you want is that things happen as they do happen, how could there be any frustration? Your wishes will always be fulfilled.
}

\color{black}

\chapter{Retraining neural networks by meditation} \label{training.ch}

The preceding chapter presented several directions in which information-processing should be changed to reduce suffering. We also saw some practical suggestions for reprogramming, such as seeing the uncertainty and uncontrollability of the world and reducing desires and self-needs. This will eventually lead to a reduction in reward loss, frustration, and suffering.

Yet, the account of the preceding chapter may be rather unsatisfactory for some readers: It seems to be asking the impossible, at least in the case of mere humans. The goal is to change some fundamental beliefs about the world and your mind. How is one supposed to become so thoroughly convinced about, say, the uncontrollability of the world that one is not disturbed by the loss of, say, one's job or house? Is it not simply ``human'' to think otherwise? How can you actually reduce expectations of rewards, belief in the certainty of perceptions, and so on?

Crucially, what we need are changes in neural associations that work on an \textit{unconscious} level, and that is notoriously difficult. 
We need to develop feasible and practical methods for retraining the neural networks in the human brain.

In this chapter, I consider meditation, or mindfulness training, as a method that can radically boost retraining of neural networks, compared to straightforward attempts to change thinking at the conscious level.
It also turns out that meditation has further benefits, such as reducing ``hot'' desires, reducing simulation, and developing metacognition.
I will not go too much into the practical details of any such training methods, on which hundreds of manuals have already been written. Rather, I discuss general principles on how they work, largely interpreting them in the information-processing framework of this book.\index{meditation}

First, I discuss how meditation can be seen to speed up learning from new input, thus enhancing the methods of the preceding chapter. Second, meditation can be seen as reducing two terms in the \eqsuff\ on page~\pageref{suffeq} that we did not yet consider: the number of times the reward loss is simulated and the amount of attention paid to reward loss; these are related to the top-row green boxes already shown in the flowchart in Figure~\ref{flowchart1.fig}. In fact, emptying the mind by meditation clearly reduces simulation, and meditation almost inevitably seems to develop a metacognitive attitude, which changes the attention paid to reward loss.
A third major benefit of meditation is that it enables stopping the processing chain in the flowchart in Figure~\ref{flowchart2.fig} by increasing conscious control over interrupting desires. 

\section{Contemplation as  active replay}

The fundamental problem with the approach of the preceding chapter is that a conscious decision to think in  a different way often has little effect on what unconscious neural networks do. A conscious decision may not even really change future conscious thinking since it may be overridden by the unconscious networks. That is why reprogramming of the brain must include some kind of retraining of the unconscious neural networks.

From a dual-process perspective, the problem to be solved here is how the conscious-symbolic-explicit system can change a mental association that is actually encoded in both the two systems. For example, it might try to create an association between ``I'' and ``impermanent'', being inspired by classical Buddhist philosophy. However, what really matters is changes in the unconscious-neural-implicit system because that system computes values, expected rewards, and reward losses. So, how can the explicit system force a change in the implicit one? Transfer of knowledge or learning between the two systems is difficult. While you may have a clear understanding that everything is impermanent on a conscious level, it is not easy to transfer this understanding to the unconscious neural networks.\index{associations!unconscious}

As a first approach, we could use techniques that I here call \textit{contemplation}.\index{contemplation} That means a constant conscious repetition of selected thoughts. For example, it can be contemplation of the characteristics of impermanence, uncontrollability, and unsatisfactoriness, possibly combined with some object---such as ``I'' or my ``self''---whose impermanence or other property one wants to learn. The constant repetition of such thoughts on a conscious level
should slowly modify the unconscious associations used to compute the perceptions and replay. Some kind of Hebbian learning\index{learning!Hebbian} is likely to construct an association
between the different concepts, such as ``I'' and ``impermanent'', even on the basic neural level.
Reading books on Buddhist or Stoic philosophy, as well as later thinking about their contents, can also be seen as contemplation. 

The mechanism working here is what I call \textit{active replay}:\index{replay!active} The explicit system uses the mechanism of experience replay (see Chapter~\ref{replay.ch}) to make the implicit system learn whatever the explicit system wants. That is, the explicit system in your brain can select thoughts in the form of linguistic sentences or visual images of events---possibly imaginary---and replay them. It can do that repeatedly, thus replaying selected items many times. Such replay will change your neural networks---that is in fact the very point in replay. What is special here is that the explicit system chooses what to replay, thus ``teaching'' the neural networks, while in ordinary replay, the material would be selected by the implicit system itself.\index{dual process!communication between}

Such training may seem rather different from modern meditation instructions, but it seems to have been an essential  form of practice in the Buddha's times and emphasized by some modern Buddhist meditation teachers as well.\footnote{See e.g.\ \citet[p.103-104]{analayo}; \citet{mahasianatta}} When the Buddha was asked for meditation instruction by monks entering a solitary retreat, he would often tell them to contemplate on impermanence, no-self, or unsatisfactoriness, sometimes linking them all together in various causal chains. For example, he would advise:
\begin{quote}\index{Buddha} 
You should abandon desire for whatever is impermanent. And what is impermanent? The eye [and visible forms etc.] is impermanent; you should abandon desire for it.\footnote{\SN{35.76}, \add{translated by Bhikkhu Bodhi}; see also \SN{35.32}; \SN{35.162}}\vspace*{1mm}\\
Forms [i.e.\ anything that is seen] are impermanent. What is impermanent is suffering. What is suffering [i.e.\ unsatisfactory] is nonself. What is nonself should be seen as it really is with correct wisdom thus: ``This is not mine, this I am not, this is not my self.''\footnote{\SN{35.4}, \add{translated by Bhikkhu Bodhi}. See also \citet[p.79]{williams2008mahayana} for a description of similar practices in a Mahayana context.}
\end{quote}
We do not know much about the details of how such contemplation was practiced in the Buddha's times. %

The fundamental problem with such simple contemplation is that the learning process can be very slow and inefficient.\index{learning!slow in neural networks} One reason is that it has the same characteristics as learning in neural networks in AI. As we saw earlier, neural network learning \add{is \textit{incremental}: it} requires a large number of repetitions of input, which change the neural connections little by little, using some mechanism related to stochastic gradient descent or Hebbian learning.\index{gradient descent!stochastic}\index{learning!Hebbian}\index{learning!incremental} So, retraining neural networks by contemplation requires a huge number of repetitions.

Moreover, transferring learning from the explicit to the implicit system is hampered by the fact that  the representations and computations in the two systems can be quite different, as we have already discussed. Suppose that your explicit system repeats the word ``impermanence'', in an attempt to contemplate on that property. How are your primitive, lizard-level neural networks supposed to understand what that means? Such neural networks do not operate with words or abstractions but on representations related to sensory input. There is a kind of communication barrier between the two systems, and contemplation will have difficulties in crossing it.

The situation can be somewhat improved if the explicit system imagines events or episodes and replays them as real sensorial input such as images, instead of merely in verbal and abstract form. When you read a story or a simile in Buddhist literature and vividly imagine it happening, that does provide more natural input to your neural networks. Or, the explicit system can imagine detailed episodes of future events, for example from the viewpoint that an action plan is likely to produce frustration, as in Epictetus's Roman bath example (page~\pageref{romanbath}).\footnote{  
\add{A related well-known Stoic exercise is a  bedtime recollection of what you did during the day (Epictetus, \Discourses, III.10.3; Seneca, \Beata, VI). While this may be difficult to understand or justify in our framework, one effect seems to be to reduce experience replay (presumably ordinary events, not philosophical ideas) during sleep: "Oh the blessed sleep that follows", exclaims Seneca. This exercise seems to be a case of deliberate replay, but of a very different kind from what we discuss in this chapter.}\index{Seneca}}

\section{Mindfulness meditation as training from a new data set}

\index{meditation}\index{mindfulness|see{meditation}}
A crucial improvement to such contemplation practices is what is called meditation in the modern context. Mindfulness meditation in particular is a technique that can influence neural networks more efficiently than simple contemplation. Mindfulness meditation can incorporate many of the goals described above, such as realizing uncertainty and uncontrollability.\footnote{For introductory books to mindfulness meditation, see e.g.\ \citet{gunaratana2010mindfulness,kabat2012mindfulness}; for an attempt at a definition, see \citet{bishop2004mindfulness}. In this book, the term mindfulness always  refers to mindfulness meditation, often seen as a training or a learning process (instead of a state of mind, or a long-term psychological trait, which are alternative uses of the term). In terms of the typology of meditation practices discussed by \citet{lutz2008attention,dahl2015reconstructing}, what is emphasized here is the ``open monitoring'' aspect of the practice. Meditation is thus also virtually synonymous with such terms as insight meditation or vipassana in this book.\index{meditation!insight}\index{meditation!vipassana}\index{vipassana}}

Typical instructions of mindfulness meditation emphasize objective observation of any contents that appear in your mind, that is, mental phenomena. In particular, that encompasses anything that your senses perceive, including the ``internal sense'' of thinking and imagination. If you hear something, you acknowledge hearing it; if there is a bodily feeling in any part of your body, you recognize that you have a bodily feeling, and so on. Such observation is done, as far as possible, passively without interfering with the sensory process or the physical source of the perceptions (for example, without moving your body to change bodily feelings). The contents should be observed from an external perspective, as if from a distance, and without judging the contents to be either good or bad.

There are a number of techniques to make such observation easier, based on regulating the attention of the meditator. Basic meditation instructions typically start by recommending sitting in a comfortable posture and then provide one particular technique for attention regulation. A very well-known technique is focusing the attention on observing the breath, possibly reinforced by mentally counting the breaths. (Alternatively, the focus might be a visual target, or a particular word or phrase that is mentally repeated.) Such observation of breathing should be seen as simply a technique whose goal is to enable better observation of the myriad mental phenomena, and indeed simply counting breathing may sound like an absurd exercise if the actual purpose is not understood. Such attention regulation facilitates the observation of mental phenomena by making the mind relatively empty; observing mental phenomena is very difficult if the mind is full of different kinds of thoughts and perceptions. Furthermore, emptying the mind has several direct benefits as well, in particular reduction of simulation as discussed below. 

The exact mechanisms of mindfulness meditation are far from being understood, but some of them can be understood by the framework presented in this book, as we will see next.\footnote{For previous proposals and reviews on the mechanisms of mindfulness, see e.g.\ \citep{holzel2011does,grabovac2011mechanisms,williams2008mindfulness,shapiro2006mechanisms,teasdale2011doesII,vago2012self,baer2003mindfulness,garland2014mindfulness,gerritsen2018breath,wielgosz2019mindfulness,brandmeyer2021meditation}. My account here is particularly computational. Initially, it emphasizes the learning of new attitudes that could be called philosophical, but later in the text, other goals will be considered as well.\label{mfnlit2}}

\subsection{Direct input to train neural networks}

The most crucial mechanism at play may be that the meditator learns largely the same things as in the contemplations above but in a more efficient way.
I suggest the reason why meditation is more efficient than what I called active replay above is that there is no longer any need to transfer information between the two systems (conscious thinking and neural networks, roughly speaking). Instead, the practitioner observes characteristics such as impermanence first-hand, in real sensory input or imagined sensory content. Then, neural network learning can proceed in a completely natural manner, largely bypassing linguistic constructs and conceptual thinking.\index{meditation!as speeding up learning} 

In other words, during meditation, the sensory systems \textit{directly perceive} how things are. For example, they are seen as impermanent by observing how those things change and disappear. %
Thus, the neural networks learn directly from such natural input. This is in stark contrast to contemplation, where the difficult part is to transform concepts and words into something that can train neural networks, and replay does this in a somewhat contrived way.
Neural networks learn best from real sensory input, so it is crucial here to enable them to do exactly that. Such observation is eventually extended to all the characteristics discussed in the preceding chapter.

The key trick here is to select the right data to input into the neural networks. As discussed in Chapter~\ref{perception.ch}, selection of input data is an essential part of the perceptual system, in terms of the multi-faceted phenomenon called attention. That is why regulation of attention is a central part of any meditation method: in mindfulness meditation, you typically start by focusing your attention on observing your breath. It is in fact possible to get useful input data from the breath itself, if you do it with a special kind of attentional focus. While the practice may start by simply observing the breath in a general manner, eventually, you can start observing its specific aspects in light of the theory of the preceding chapter. For example, you observe the impermanence of breath, how it is changing all the time from an in-breath to an out-breath---this is a classical Buddhist exercise. That means your attentional system selects your sensory input to consist of observations of your breath, and more precisely, any aspects of your breath related to permanence or lack of it. This is how your neural networks get a lot of good data pertaining to that particular property, and they learn to perceive the impermanence much better than they would by any kind of abstract contemplation based on linguistic concepts.\footnote{Such selection of input can be further improved by controlling one's media consumption as well as by choosing a suitable lifestyle and social environment. Buddhist monastic training provides a rather extreme example of such choices.}

Thus, the explicit system in a sense ``teaches'' the implicit system, and the teaching happens by means of the attentional system. The direction of attention is, to some extent, under conscious control. So, the symbolic or thinking part of the brain can just tell where the implicit system should be looking---this is only partly a figure of speech--- and it does not need to really input anything to the implicit system, unlike in the case with replay. It is a bit like a professor telling students to read a book; she does not then need to give a lecture herself.\index{dual process!communication between}

\subsection{Realizing how the mind wanders}

\index{wandering thoughts!in meditation}
Another important example of such direct input is observing how often and easily the mind starts wandering. As we have seen in Chapter~\ref{replay.ch}, sustained attention is difficult, and after a while, the mind often starts wandering, and various daydreams fill the mind. Frequent occurrence of such mind-wandering is extremely salient to anybody who tries to focus on breathing or a similar meditation object.
Realizing how difficult it is to focus on breathing gives a direct view into how uncontrollable the mind is. If you systematically observe how automatically your mind starts wandering, you will gradually be convinced---and so will your neural networks--- that you cannot control even your own thinking, at least not completely.  After all, wandering thoughts are, by definition, a failure of controlling your mind.\index{meditation!wandering thoughts} 

\index{wandering thoughts!and no-self}\index{no-self!realization in meditation}
Such observation may also convince you that there is no self, no central executive, and perhaps then no free will. Under ordinary circumstances, if I decide to plan what I will do tomorrow, I may have a clear feeling that it is ``me'' who is doing the planning. However, after observing how planning happens automatically in wandering thoughts, I may be forced to admit that the plans are something that ``I'' did not create.
You may even start having doubts about the correctness and certainty of your thoughts, since they seem to be something that just appears in the mind, and you have little idea why they appear or where they come from. Thus, uncertainty about your thoughts can be taught to the neural networks as well.

\subsection{Efficient reduction of expectations}

To interpret such learning processes in the framework of the \eqsuff, what happens is that the \textit{unconscious neural networks themselves}---and not just the conscious and/or symbolic thinking systems---will \textit{learn to reduce the expectations} of any rewards. This happens through your neural networks learning that the world is uncontrollable and uncertain (or impermanent), which necessarily reduces their expectation of future rewards according to the logic of the preceding chapter.\index{expectation!of reward!reducing}

Many further meditation techniques can be seen as such attentional selection of input, with different targets for the learning. One classical Buddhist technique is to focus on the ending of any pleasurable feeling. This enables seeing first-hand how pleasure, and in general any effects of rewards, are fleeting and worth less than might be expected. Thus, you will learn the impermanence and, in particular, the \textit{unsatisfactoriness} of all mental phenomena in a particularly efficient way.

\subsection{Extinction of aversive responses}

In a slight variant of the logic above, mindfulness meditation can also help directly change associations related to specific emotions. An important example is fear extinction. Extinction is the opposite of classical conditioning: It means that when the predictive stimulus (e.g.\ the bell for Pavlov's dog) is presented \textit{without} the other stimulus (the food for the dog), the conditioning weakens. If the bell is presented without the food many times, the dog learns that the bell does not predict the food anymore, and the conditioning is eventually extinguished.\index{extinction!of classical conditioning}\index{meditation!extinction of conditioning}

Suppose you have learned to associate a fear reaction with your boss by classical conditioning. Perhaps that was based on a single episode, and the association is not valid anymore, so it would be nice to be able to let such a fear reaction be extinguished. Unfortunately, extinction is often very slow---just like any neural network learning---but this can be improved  by mindfulness training.\index{fear!extinction}\index{conditioning!extinction}
The trick here is that you create completely new data, going beyond simply selecting input from existing data as above, but still feed it directly into your neural networks.

It turns out that mindful meditation tends to make people relaxed and feel good (possible reasons for this will be discussed later). So, if you recall the unpleasant episode with your boss many times, but always stay in such a pleasant, calm, meditative state, extinction is more likely to happen. 
Thoughts about unpleasant situations will be increasingly associated with a general feeling of calm; 
the image of your boss will be associated with relaxation and feeling good in the whole body. This will  help override the fear association.\footnote{On the general idea of extinction by mindfulness and exposure, see \citep{baer2003mindfulness,holzel2011does}; on relaxation and positive affect, see \citep{carmody2015reconceptualizing}. An interesting question here is whether extinction needs attention and/or awareness, and would thus be greatly facilitated by mindfulness; the results are not very clear-cut on this point \citep{kwapis2015role,han2003trace,weike2007fear}.}

\section{Speeding up the training}

Unfortunately, such meditation training is still rather slow, even if it improves on simple contemplation. In fact, slowness of training is a ubiquitous problem with neural networks \add{due to the incremental nature of their learning,} as already pointed out in Chapter~\ref{ml.ch}. Even though in mindfulness meditation, we have a new source of more direct and natural data for learning, neural networks still need large amounts of input data, and a lot of meditation practice is needed.  Fortunately, the amount of training and effort required can be further reduced by further techniques.

\subsection{Increasing the plasticity of the brain}

\index{plasticity!increasing} 
One central principle here is increasing the \textit{plasticity} in the brain. Plasticity is the biological term for the capacity of neural connections to change and thus to learn. Plasticity in the brain's neural networks is by no means granted, nor is it a constant quantity. If, with some suitable tricks, such learning capacity could be increased, the learning process would take less time. 
A large amount of neuroscience research has been dedicated to finding different ways to increase plasticity.

\index{sensory deprivation}
Sensory deprivation seems to be one useful trick; it has indeed been shown to increase plasticity, at least in rats and cats.
It may be rather common sense that if your brain has had little stimulation for a while, it will better concentrate on any new task; it turns out that its learning capacities are also increased.\citenew{he2006visual,duffy2013darkness}
Mindfulness meditation in itself can be seen as imposing sensory deprivation, since it is usually conducted in a quiet environment with eyes closed, or at least there is nothing much happening in the visual field. In some meditation schools, an even stronger form of sensory deprivation may be imposed in the form of silent retreats. Such retreats often entail minimization of any kind of sensory stimulation: the participants don't go out of a prescribed enclosure, they don't watch TV or use the internet, and obviously they don't talk to each other.
In several discourses, the Buddha recommended such deprivation, together with meditative concentration, because it makes the mind ``pliant'' and ``malleable''. Then, the meditator is better able to gain insight into, for example, the uncontrollability and uncertainty of existence, as well as better able to learn from those insights.\footnote{\MN{36}; \DN{2}. \add{Sensory deprivation may seem to be contradictory with the idea of learning from new input, but the deprivation mainly refers to sensory input while wandering thoughts etc.\ are still running and provide input for the learning; moreover, sensory input is never zero anyway since the input from bodily senses (proprioception, interoception, pain perception) is hardly reduced.\index{body} Presumably, such sensory deprivation also reduces the capture of attention by sensory stimulation and enables directing the attention to those phenomena that the learning needs as input (e.g.\ wandering thoughts). }} 

Plasticity can further be increased by restriction of food intake, which is another typical characteristic of ascetic training in spiritual traditions. Paradoxically, it can also be increased by the very opposite of sensory deprivation: enriching the environment. In animal experiments, that might mean allowing the animals to live as groups in large, spatially complex cages, equipped with toys and running wheels. In humans, similar results are obtained by aerobic exercise, as well as action video game playing.
Whether such methods could be used to improve meditation practice is a very interesting question for future research.\footnote{Increase in plasticity due to
  food restriction:  \citep{spolidoro2011food}; environmental enrichment: \citep{sale2007environmental}; 
  exercise and games: \citep{bavelier2010removing,nokia2016physical}. \add{See also \citet{kirste2015silence} who show how both silence and unusual noise can increase neurogenesis in the hippocampus (and thus plasticity), which they assume is due to the novelty of both conditions.}}

\index{drugs!increasing plasticity}
Plasticity can also be increased by drugs, such as the antidepressant fluoxetine (aka Prozac).
A large amount of research is currently being conducted on new drugs that would increase plasticity even more, and with minimal side effects. The huge impact such drugs could have on society is obvious.\citenew{vetencourt2008antidepressant,castren2017neuronal,ly2018psychedelics}

In fact, you may be wondering why plasticity is such a bottleneck: Why hasn't evolution made our neural networks learn much faster? The reason seems to be that
some limitation of plasticity in the brain is useful to prevent new information from overwriting old information too easily.\footnote{\citep{mccloskey1989catastrophic,bavelier2010removing,pascual2005plastic,kirkpatrick2017overcoming}. Furthermore, too much plasticity might destroy the stability of the brain as a dynamical system, even leading to such phenomena as epileptic seizures \citep{kozachkov2020achieving}.} 
So, it may not be wise to increase plasticity too much, because it could lead to too much forgetting of previously learned information. This is hardly a problem with meditation-based interventions, but with drugs, such negative side-effects might be real.

In principle, an AI has much more freedom in how it changes the results of its learning, and the amount of ``plasticity'' could be made infinite by design. Thus, an AI could get rid of a bad habit or a harmful association in a split-second, by just removing or changing some connections in its neural network. However, this may not be as easy as it sounds, since
just like with humans, there may be a risk of interfering with other connections so that the AI may forget useful information. Also, it may not be clear which connections should be changed in the first place, due to the overwhelming complexity of neural networks. So, even in the case of an AI, it may be better that all the training happens by simply inputting carefully selected data into the system and patiently waiting until the learning happens.\footnote{Another way of improving learning would be to adapt the contents of the contemplation to each individual based on their personality and temperament. While this is rarely done in a Buddhist context, the classical Buddhist meditation manual \textit{Visuddhimagga}, for example, does include such instructions (Chapter III, 74).\index{individual differences}}

\subsection{Training can become automated}

\index{automatization!of meditation}\index{learning!skills!in meditation}
Another major difficulty in meditation training is sustaining attention in the way typical meditation techniques require. I need to emphasize that we actually have two different attentional mechanisms at play here. First, there is \textit{sustained} attention\index{attention!sustained!in meditation} to the task at hand, meaning that you concentrate on meditation and don't think about anything else, as explained in Chapter~\ref{replay.ch}. Second, there is \textit{sensory, selective} attention,\index{attention!selective!in meditation} which means you select certain data as input to sensory processing, as originally explained in Chapter~\ref{perception.ch} and extensively discussed earlier in this chapter. Both are necessary for successful meditation. However, sustained attention tends to be the major bottleneck because it is notoriously difficult to maintain.

In previous chapters, we have actually seen several reasons why sustained attention is difficult. First, wandering thoughts assail the mind, for example due to experience replay. But we also saw that emotions are essentially interrupts; what they are interrupting is current activity, and to do that, they have to be able to grab attention away from wherever it may be. The general concept of the brain as parallel distributed processing emphasizes the idea that there are different networks or modules which are often competing, for example, for attention and the control of attention. 

Fortunately, it is possible to \textit{learn} to use your attentional capacities better.\footnote{\citep{friese2012mindfulness,mackenzie2015self}. Such learning has earlier been well-documented on a more general level in the work on self-control \citep{rueda2004attentional,baumeister2007strength}. It can be seen as another interaction between the two systems in dual-process theory (explicit and implicit), see also \citep{doyon2003distinct,peters2011towards,sun2005interaction}.}
This is yet another form of learning, but a bit different from the learning we have considered in this chapter: here we are talking about learning a new \textit{skill}, as briefly described in Chapter~\ref{dual.ch}. 
A skill means that you know how to ride a bicycle, to speak a foreign language, or to use your new smartphone; it is opposed to learning facts and increasing your knowledge about what the world is like. Skill learning follows some general laws and these apply to meditation as well. As we all know, you need to practice.
At the beginning of the practice, you need to concentrate and spend a lot of effort, meaning a lot of sustained attention. The important point here is that with practice, meditation becomes more and more automated, which means that less and less conscious effort is needed. Some meditation traditions talk about meditation as ``just sitting'', which is in a sense enough if the meditation is sufficiently automated. Importantly, the regulation of attention will in fact become a habit, and will be easily conducted during ordinary life, as if by itself, even outside of formal mindfulness meditation sessions.

So, there are actually two different learning processes at play: Learning that the world has certain characteristics (such as uncontrollability), and on a higher level, ``learning to learn''\index{learning!to learn} that the world has such characteristics. The latter learning process means learning to meditate in an automated, habit-like manner, with minimum conscious effort. Thus, with practice, the meditator will be able to perform the former learning process with increasing efficiency, and this former process is the one that reduces suffering according to the theory of the preceding chapter.

\subsection{But who is actually meditating?}

\index{no-self!in meditation}\index{meditation!no-self}
The fact that meditation can become automated and habit-like means that, in a sense, it is no longer my ``self'' who is meditating. We find echoes of the no-self philosophy treated in Chapter~\ref{control.ch}. Some neural networks will be able to observe the breathing without any conscious effort, or even without a conscious decision to start meditating. There is no need for any central executive to make any decision, and no need to \textit{want} to observe the breath; it just happens. It is like when walking, you make no conscious decision to move your feet; you feel no burning desire to put one foot in front of the other.

But if the neural networks are retrained by the explicit system as I argue in this chapter, does that not mean that it is the explicit system, perhaps even a conscious self, which is in control? That might be a hasty conclusion since there are so many complicated ways in which the two systems interact. In fact, earlier (page~\pageref{datacontrol}) I argued that it is meaningful to say that ultimately, it is the input data that controls us.\index{control!by input data} I gave the example of a meditation master who says that it is actually his master who is meditating, because he still hears his master's voice in his head. This shows that in order to find the ``ultimate'' source of control, we have to consider where the data to the explicit system comes from. Part of it clearly comes from human society and the cultural context: there are other people that input data into us, for example in the form of meditation instructions. How that happens, and who is controlling whom, is a vast topic that I have to leave for future research.\footnote{My arguments here are not very rigorous since the very definition of ``control'' is not made explicit in this book; I simply follow typical common-sense usage of the word. A more detailed analysis would point out that control is a matter of degree: In the context of this chapter, the explicit system has only partial control of the implicit system anyway because it merely directs its attention, so the explicit system is certainly not in \textit{total} control in any meaningful definition of the word.}

\section{Reducing interrupting desires}

In addition to speeding up learning in neural networks, mindfulness meditation has further benefits.\index{desire!as interrupt and hot}\index{dynamics!cognitive, of desire}
Next, we consider how it reduces suffering from the viewpoint of cognitive dynamics, which complements the \eqsuff. As we saw in Chapter~\ref{overview.ch}, one traditional Buddhist account of a mechanism to reduce suffering is based on the moment-to-moment cognitive chain or cycle shown in the flowchart in Fig.~\ref{flowchart2.fig}. 
The idea here is to stop the dynamic process in the flowchart in the middle so that it does not lead to its end product, which is suffering.\label{chainbroken}
The point where the process can best be stopped is assumed to be (in the terminology of our flowchart) the three links of desire,  intention, and planning.\footnote{See e.g.\ \citep{analayo}; note that in the traditional early-Buddhist account, these correspond to the two links of desire and attachment/clinging. 
}\index{intention!and attachment}\index{intention!reducing} It is in fact assumed in early Buddhist philosophy that until the valence computation,\index{valence!in Buddhism}
the process is too automated, and desire provides the first link that can be stopped.\footnote{\citep[p.~89]{mahasidependent}.\index{vedan\=a} Valence is closely related to what is called ``feeling tone'' or \textit{vedan\=a} in Buddhist literature.\index{vedan\=a} I think \textit{vedan\=a} can best be described as the perception of valence. (Such perception requires of course some kind of computation of valence as well.) For a Stoic viewpoint, see \Discourses, II.18.15-18.}

The ensuing method is distinct from reducing desires by adopting the attitudes of the preceding chapter. Here, I am talking about sudden, ``hot'', interrupting desires triggered by the valence computations, and their prevention in real-time when they are about to arise. The preceding chapter focused on reducing long-term desires from the ``colder'' perspective of reward calculations; that will also reduce  the underlying tendency for hot desires to arise, but it works only passively in the background.

One problem here is that the hot desires have the properties of interrupts, as explained in Chapter~\ref{emotions.ch}, which means they are strongly automated and can be quite difficult to prevent or stop. Therefore, it might be better to try to stop the dynamics a bit later, at the links right after desire. In Buddhism, those following links are called ``attachment'', which is in our schema divided into forming an intention (i.e., committing to a goal) and planning for that goal.

\index{desire!reducing}
Whether desire or attachment is chosen as the target, the trick here is to weaken the cognitive dynamics so that this largely automated chain leading to suffering fails to operate. If the desire or attachment is prevented from taking place, no goal is committed to or planned for, and no goal-oriented action is conducted. 
Thus, the whole \eqsuff\ above is not operating, and frustration is avoided by that route.\footnote{Some frustration will still happen because of the habit-based system, but as argued in Chapter~\ref{planning.ch}, such frustration is much weaker than that coming from planning and execution of plans.}

\subsection{Perceptual learning}

Such stopping of the cognitive dynamics is enabled by well-known mindfulness meditation techniques. The point is to observe the cognitive dynamics repeatedly, so that one learns to introspectively detect the different parts of the process and discriminate between the different links, in real-time. Mindfulness meditation has here the effect of training a new perceptual capacity that allows for observation of the internal mechanisms of the mind.\index{meditation!perceptual learning}

This is a special case of the phenomenon of ``perceptual learning''.\citenew{sagi2011perceptual}\index{learning!perceptual} Research on perceptual learning started in vision science by the discovery that it is possible to greatly enhance the performance in almost any visual perception task; all that is needed is sufficient training. Improvement is possible even in tasks where the limits of perception were previously thought to be set by the optics of the eye, such as the task of telling whether two lines have the same orientation (angle) or not.

In the context of meditation, such perceptual learning allows one to observe the individual elements of mental processes more accurately. An important case of such learning is that it becomes possible to observe the associations between phenomena. If B is associated with A, then, under ordinary circumstances, it may be that the thought of A immediately and necessarily brings B into mind, and it seems that A and B are two aspects of the same thing. But with mindfulness training, it is possible to see how this process breaks into pieces: First there is A, then the association is activated, and then B comes to the consciousness because of the association. This allows one to see not only the existence but also the arbitrariness of that association. In particular, one is able to dissociate a desire from the stimulus that caused it, as if by creating a ``space'' between the stimulus (say, chocolate) and the desire, as well as %
\add{between any further steps in the chain}. 

\subsection{Breaking the cognitive chain}

This opens up the possibility of breaking the long chain leading from stimulus to suffering depicted in Fig.~\ref{flowchart2.fig}. %
Introspectively, the meditators often report that it feels as if the whole process were slowed down by such perceptual learning. The process is also, to some limited extent, brought under conscious control. Even if a stimulus leads to a strong valence, the ensuing desire and the following steps will not happen completely automatically, but there is some space for deliberation.\index{meditation!breaking cognitive chain}

Perhaps such breaking of the causal chain is most understandable in the case of planning, which is often a rather conscious process, and as such, it should be possible to decide not to initiate it at all. Obviously, there is a strong unconscious tendency to start planning when desire arises; it is comparable to the unconscious reaction to start scratching a body part that is itching. However, with practice, such an unconscious tendency can be weakened, inhibited, and perhaps even completely removed. That would mean not letting ``attachment'' arise in Buddhist terminology. The key is to be able to consciously recognize when the planning is being triggered, instead of letting it happen automatically.\footnote{As already mentioned, \citet{Libet1983time} proposed, rather controversially, that while the consciousness does not decide actions, it has a ``veto'' over actions: It can cancel an action sequence that the unconscious neural networks are trying to perform. This might provide an interesting explanation of how consciousness, in an advanced state of mindfulness and metacognition, seems to be able to prevent habitual actions \citep{baer2003mindfulness,garland2014mindfulness}, such as stopping the twelve-fold chain.}

\index{automatization!of meditation}\index{learning!skills!in meditation}\index{skills!learning|see{learning, skills}}
It is important to achieve automatization of such mindfulness by long-term meditation practice, as described above. The learned and automated tendencies of observation can then create the possibility for inhibiting the more innate automatic tendencies of desire and attachment. In fact, if such observation is followed by conscious, deliberate inhibition of desire or attachment often enough, that very action of inhibition will become automated as well. Conscious control processes are often too slow and weak to prevent the processes underlying hot desire or other interrupts, so it is really important to train the neural networks to initiate the action of inhibition as well. Once the neural networks have been trained to perform both the detailed observation and the inhibition during formal meditation sessions, they may be able to transfer that skill to everyday life with its infinite temptations.\footnote{Related models consider how mindfulness meditation helps in addiction \citep{brewer2014craving,garland2014mindfulness}.\index{addiction} In particular, \citet{brewer2014craving} proposes several mechanisms describing how mindfulness meditation can ``de-automate''  the dynamics, including learning to simply observe aversive states without reacting to them and taking them less ``personally'' , while becoming ``more aware of habit-linked, minimally conscious affective states and bodily sensations''.}

\add{While inhibiting desire and attachment is emphasized in Buddhist training of the Theravadan school, this is not the only way to break the chain. Even if planning happens, and even if a plan is executed, the dynamics might still be stopped later, for example, right before error computation. In that case, the computations necessary for frustration simply do not take place. In other words, a failure does not lead to an internal ``judgement'', but is in some sense just accepted (the concept of acceptance will be discussed in detail in the next chapter). Even after that, there is a final link that could be broken, from error computation to suffering. That is, even if an error is computed, some further processes are necessary to translate that error into the subjective feeling of suffering, at least in humans. This link might be weakened by a metacognitive perspective, which will also be treated in detail below.}

\section{Emptying the mind and reducing simulation}

\index{empty mind}\index{simulation!reducing}\index{meditation!emptying mind}
Another additional benefit of meditation is that many people report feeling great pleasure when meditating. This is often attributed to the fact that the mind is strongly focused on a single object, such as breathing, and thus emptied of any thinking. Several traditional meditation schools actually maintain that an ``empty'' mind is happy, that is, a mind where there are no thoughts, whether wandering or intentional.\footnote{\add{It may seem contradictory if meditation tries to make the mind empty, while above, meditation was seen as learning from selected input. This is not contradictory since the point is that relative emptiness of the mind is necessary to be able to select and pay attention to mental phenomena in a way that optimizes learning about uncontrollability, etc. Without such emptying of mind, all the processing would be spent on ordinary thinking and sensory processing instead of the intended learning. For example, only with a relatively empty mind can the meditator realize how wandering thoughts are uncontrollable and impermanent, or that bodily feelings likewise just come and go.}}   (Emptiness of the mind does not here refer to the Mahayana Buddhist concept of emptiness we saw earlier.)  A similar pleasurable state is sometimes achieved in the state of ``flow'', where wandering thoughts are equally absent.\citenew{csikszentmihalyi1997finding}\index{flow}

Understanding why an empty mind tends to be happy is one of the deepest problems for a scientific understanding of the mechanisms behind meditation, and not quite resolved at the moment. A number of viewpoints can be taken here. In a traditional Buddhist account, where desire is considered the basis for suffering, a simple explanation would be that an empty mind is happy because it does not have any desires (including aversions).\footnote{See page~\pageref{desiresuffering} for a proposal on how desire and aversion in themselves produce suffering.}
On the other hand, Chapter~\ref{replay.ch} reviewed research showing that wandering thoughts are typically related to a negative mood; however it was not clear if those results apply to all thinking and not just wandering thoughts, and what is the cause and what is the effect. Yet another viewpoint is to 
recall once more Cassell's statement that ``to suffer, there must be a source of thoughts about possible futures'', which cannot exist in an empty mind.

In the framework of our \eqsuff, we can formulate a more computational viewpoint.
Reward loss is computed every time a simulation, whether in terms of  replay or planning, is conducted in the brain. A reduction of thinking should reduce suffering since such simulation of frustration or reward loss is reduced. In fact, in our \eqsuff\ on page~\pageref{suffeq} we have the term ``number of times [the reward loss] is simulated'' which gives the number of times the reward loss is computed (after adding one to this number, due to the initial actual perception). Reducing mental simulation will reduce this term, and thus suffering.
Reducing mental simulation\index{simulation!reducing} will, for example, reduce rumination over past errors, simulation of future threats to the person, as well as judgements related to self-esteem, which are some of the most important sources of suffering.\footnote{A problem with this logic, which we already partly saw in Chapter~\ref{replay.ch}, is that it is not clear why simulation of positive experiences would not cancel the effect of simulating negative experiences. Somehow, it seems that negative experiences are stronger in this case. It is possible that this only holds for some people whose thinking just happens to be more often negative than positive, and it is those people whose mood is most improved by meditation.\index{individual differences!in effects of thought wandering} Or, it could be that due to some evolutionary reasons, this is the case for the vast majority of humans: \citet{baumeister2001bad} reviews a great number of results leading to the conclusion  that ``bad is stronger than good'' as far as the emotional effects of life events are concerned. (The important case of rumination was treated in Chapter~\ref{replay.ch}). \add{Interestingly, Plutarch proposed a training method to reduce future-oriented wandering thoughts by recalling good things that have happened to you in the past (\textit{On the Tranquillity of the Mind}, 14). At the same time, he also recommends cherishing things you have right now, thus making a clear connection to gratitude exercises discussed in Chapter~\ref{attitude.ch}.\index{gratitude}
    Thus, it seems to be possible to engage in thinking that is not very different from wandering thoughts, but deliberately optimizing the contents can make the effects positive.}\index{Plutarch}} 

\index{mental simulation|see{simulation}}

The logic just given may explain why many meditation methods have the explicit goal of emptying the mind of thinking, or at least reducing thinking. Typically, one concentrates on a single object, such as the breath. 
This immediately reduces thinking, including wandering thoughts---but does not eliminate them completely, as the meditator soon notices. An important aspect of any meditation technique is how to react to the occurrence of wandering thoughts.
Some meditation techniques directly aim at suppressing them by refocusing on the original object of meditation. Suppose you have any unpleasant, possibly scary wandering thoughts about the future or the past during meditation. If you refocus on the meditation object, thus clearing the mind of such scary wandering thoughts,  it is rather obvious that suffering will be reduced.\citenew{kuyken2010does} Being able to thus prevent negative wandering thoughts from occurring should have a strong positive effect on mood, in line with our logic above based on \eqsuff.\index{wandering thoughts!in meditation}\index{default-mode network!in meditation}
In fact, it has been shown that the default-mode network, largely responsible for wandering thoughts, is less activated in experienced meditators.\footnote{\citep{brewer2011meditation} Recapitulating some of the logic above, we arrive at a speculative computational explanation of why almost any wandering thought leads to suffering, and why the elimination of almost any wandering thoughts reduces suffering.  Namely, most wandering thoughts are related to some kind of desire or aversion, which either underlies planning of future action or motivates replay of a rewarding or punishing past episode. If we combine this with the idea that aversions and desires are suffering in themselves (page~\pageref{desiresuffering}), we see why wandering thoughts almost necessarily lead to suffering.} (Below, we will see an alternative approach to dealing with wandering thoughts based on meta-awareness.)

\subsection{Focusing on what is here and now}

Focusing on breathing is something that can be done during a formal meditation session, but perhaps not during ordinary life. That is why in Buddhist training, there is also a strong emphasis on focusing on what happens \textit{here and now}; this can be practiced in everyday life, outside of any meditation sessions. Such a focus can be conceptualized as learning to change your ``cognitive style'' to a more ``experiential'' one, which means you replace most thinking, whether future- or past-oriented, by the simple sensory experience of the present moment.\citenew{watkins2004adaptive} This is essentially another shift of attention away from thinking, but this time the shift is to any immediately present perceptual input, instead of some pre-selected object like the breath during meditation. 
\add{Since suffering is fundamentally based on predictions and expectations, focusing on the present moment should reduce suffering in many ways.}

In particular, focusing on what is here and now means that any reward loss occurred in real life is only briefly observed without paying too much attention to it. Soon, attention is directed to something else in the here-and-now, since the reward loss has already become a thing of the past. According to the \eqsuff, such reduction of attention reduces suffering, but this time it works via the the term ``amount of attention paid'' since reducing attention reduces the impact of any perception.\index{cognitive style}\index{attention!to reward loss}\index{reward loss!attention paid}
Nevertheless, such a cognitive style also reduces simulation, and produces the benefits of an empty mind as well.\footnote{Buddhist philosophy, as well as the theory in this book, further suggest another very different way for achieving a reduction in replay and planning, which is nothing else than adopting the philosophical attitudes described in the preceding chapter. Planning how to obtain future rewards is likely to be reduced if future rewards are considered lesser; there is simply not so much incentive anymore in planning for them. 
Likewise, planning to avoid threats, \add{or worrying},\index{worrying} will be reduced if those threats are seen as relatively uncontrollable.\index{uncertainty!of thoughts}\index{uncertainty!of perception}
Furthermore, when the uncertainty of our thoughts and perceptions is realized, spontaneous thinking is often reduced, since there seems to be much less point in simulating something which is uncertain anyway. This is how adopting the philosophical attitudes discussed in the preceding chapter will also lead to a reduction in simulation, and towards an empty mind. This logic %
shows how the question of causality regarding emptiness of mind and happiness/suffering is complex. (See also footnote~\ref{wanderingmindunhappy} in Chapter~\ref{replay.ch}.) We started this section by pointing out that emptying the mind by meditation often has the effect of making people feel more joyful. %
Thus, emptying the mind was seen as an intervention that causally reduces suffering. In contrast, the idea that reducing desires reduces (especially wandering) thoughts is in line with some classical Buddhist authors who seem to claim that the emptiness of mind is mainly an \textit{effect} of mental development, not a cause of happiness \citep[p.~55]{williams2008mahayana}. In such thinking, reducing desires reduces frustration as discussed in Chapter~\ref{freedom.ch}, and an empty mind is just a side-effect. Meanwhile, the discussion on the experiential cognitive style just given could probably be interpreted based on either causal direction; either an experiential style makes the mind empty, or emptying the mind leads to a more experiential style; or perhaps both are effects of the reduction of desires or some similar cause.}

\section{Metacognition and observing the nature of mind} 

\index{metacognition}\index{meditation!metacognition/meta-awareness} 
There is one more form of attentional control operating in mindfulness meditation, especially in more advanced stages of the practice: direction of attention and awareness to a metacognitive level. Metacognition means cognition about cognition: for example, thinking about one's own thoughts, or observing one's own perceptual processes.  In such a case, the ``higher'', metacognitive part of your mind is observing the ``normal'' thinking or perceiving part of your mind.\footnote{\citep{beran2012foundations,proust2010metacognition,fleming2012metacognition}. An increase in metacognition is one of the more robust findings in studies of mindfulness training \citep{lao2016cognitive}. Interestingly, one important function of metacognition is assessing the uncertainty of perceptions and cognitions \citep{fleming2024metacognition}; presumably uncontrollability and even unsatisfactoriness have a similar link to metacognition.} Such metacognition is presumably possible because of the parallel and distributed nature of brain function, which means one part of the brain can observe what is happening in another part. 

An obvious utility of metacognition is that it enables introspection, which allows you to understand the processes underlying your thinking, emotions, and desires. This is of course the goal of a multitude of psychotherapeutic systems. However, the practice of meditation, especially in a Buddhist context, can go much deeper in this respect, and a well developed metacognitive attitude is seen as interesting in its own right.\index{consciousness!meta-}\index{awareness|see{consciousness}} Such development of metacognition can be seen as another example of perceptual learning\index{learning!perceptual} discussed above.

Buddhist meditative practice eventually leads to a class of advanced techniques based on \textit{meta-awareness}, which designates the quality of the consciousness or awareness present in such metacognition. It is awareness of awareness; in other words, there is conscious recognition or perception of the fact that there is awareness. This may seem very complicated or even paradoxical, but in fact it is something that we regularly engage in, if only fleetingly. A typical example used in neuroscience is when you realize your mind is wandering and regain focus; that realization was on the level of meta-awareness. But there are many more interesting cases.\footnote{\citep{chin2010meta,schooler2011meta}. As in earlier chapters, the words consciousness and awareness are here used synonymously.  The distinction between metacognition and meta-awareness is not always clear and there is overlap on how these terms are used. The key difference for me is, however, that meta-cognition can happen even in an unconscious agent such as a robot, while meta-awareness, by definition, requires consciousness.}

Consider the following case of sensory meta-awareness.
If I ask you whether you see this book, you would reply in the affirmative. I can formulate the question in a more explicit way: Are you aware of the fact that you are consciously perceiving this book? You would probably still reply in the affirmative. It is almost the same question really, since in colloquial language, if you ``see'' something, that means that you see it on a conscious level. If you can consciously recognize that you are consciously perceiving this book, you must be aware of such conscious perception happening, and thus there is meta-awareness. So, you were fleetingly aware of the sensory awareness of the book; you moved to the metaconscious or meta-aware level for a few seconds. That shows that almost any kind of sensory awareness can be accompanied by meta-awareness, and it can be deliberately initiated. While this meta-awareness didn't last long, it is possible, as a meditative exercise, to stay on the level of meta-awareness for a longer period of time.\footnote{\citep[e.g.~p.~77-79,121-126]{SUT2}}

The same kind of meta-aware observation can even be extended to thinking.
Some advanced meditation techniques emphasize observing the wandering thoughts while they are taking place, instead of suppressing them. The possibility of actually observing the wandering thoughts and their contents in real-time---instead of merely noticing that you have had some wandering thoughts a while ago---may seem quite paradoxical. 
However, it is possible to learn such sustained meta-awareness of one's thoughts with enough meditation practice.\footnote{\citep[e.g.~p.126-133]{SUT2};\citep[Ch.~4]{pramoteenlightenment};\citep[p.~177-179]{mahamudra}. The difference between intermittent (fleeting) and sustained meta-awareness in a meditation context is considered by \citet{dunne2019mindful}. In contrast, \citet{smallwood2007lights}, apparently talking about a non-meditative context, emphasizes that the absence of any metacognition or meta-awareness is typical of wandering thoughts.} 
From this viewpoint, at least on advanced levels of practice, it may not be necessary to reduce wandering thoughts; after all, any attempt to empty the mind may create new suffering because the mind is uncontrollable. Instead, one may change the quality of the awareness in the sense that the attention is mainly operating on the metacognitive level.\footnote{\citep{teasdale1999metacognition}. This technique is different from but related to directing the attention elsewhere on the non-meta level, such as to the here and now that was described earlier.}
Such meta-awareness often feels like perceiving one's thoughts as if from the outside, instead of being inside or involved in them.\index{wandering thoughts!observing them} 
In fact, if your mind engages in a scary simulation of something that might happen to you in the future, you can now just watch the simulation while reminding yourself that it is not actually happening: it is just a simulation where your mind plans possible courses of action. With such a quality of consciousness, there is actually little need to \textit{stop} the simulation to reduce suffering. Moving to such a metacognitive level is actually often an automatic consequence of long practice in mindfulness training, and it may easily happen during an intensive meditation retreat.\footnote{  \citet{dahl2015reconstructing} discusses different meditation techniques and the role of meta-awareness in them. This is related to what is called \textit{decentering} by \citet{fresco2007initial,safran1996interpersonal}. 
}\index{decentering}\index{simulation}

Now, what if you could spend a considerable proportion of your daily life on the meta-aware level? Such long-term sustained meta-awareness seems to be possible after extensive meditation practice. Importantly, such meta-awareness may  lead to insights that convince the meditator about several philosophical points we have seen in this book. You may see all conscious mental phenomena, that is, all the contents of your consciousness, such as perceptions and thoughts, as results of impersonal computational processes. In other words, they are simply mental constructions, or results of a simulation performed by your brain. This logic may lead to the conclusion that even what you see in front of you at this very moment is a perception constructed by your brain, based on various unconscious inferences, sometimes hardly better than guesses: you really have no other source of information about the world but perceptions and thoughts playing in the virtual reality of consciousness.
Perceptions and the ensuing thoughts are thus necessarily subjective, contextual, fuzzy, and uncertain constructs---they are empty, in Buddhist terminology.\index{emptiness (Buddhist)!of all phenomena} They do not represent any absolute truth about how things are. Such insights into uncertainty and emptiness are in stark contrast to our inherent tendency to think that our perceptions are somehow identical to reality.\footnote{
  This eventually leads to what is the deepest form of meta-awareness: being simply aware of the existence of consciousness itself, or of the very capacity to be conscious of mental phenomena, as opposed to being aware that you are aware of some specific phenomenon (such as the perception of this book in the example above). Such awareness is related to what is called seeing the (true) nature of mind (or consciousness) in Buddhist and related literature \citep{brahm,dalailama,mahamudra,spira}; it is also related to the attitude briefly described at the very end of Chapter~\ref{consciousness.ch}. 
  However, it is  outside of the scope of this book, and goes well beyond our \eqsuff. Nevertheless, similar to what is argued next in the main text,  it could lead to minimal attention to reward loss and therefore a great reduction of frustration even according to that equation.
}

\subsection{Meta-awareness and suffering}

These insights into \add{uncertainty} will reduce reward loss by reducing the certainty-related term in the \eqsuff, similar to mechanisms already explained in Chapter~\ref{freedom.ch}. 
In addition to such a long-term learning process, there is also an immediate utility in keeping a meta-aware attitude towards all mental phenomena: meditators often report great calm and peace of mind when their minds are in such a state. %
The reason is not well understood from a neuroscience viewpoint, but I would assume that less attention is paid to error signals, because attention and awareness have largely moved to the meta-level. Perhaps error signalling is somehow generally dampened, due to some mechanism to be discovered.  Introspectively, the effect can be described as the meditator keeping some distance from the thoughts and perceptions, and taking them less personally as well as less seriously. 
Going back to the \eqsuff\ (page~\pageref{suffeq}), we can assume that any reward loss will be paid less attention,\index{attention!to reward loss}\index{reward loss!attention paid} meaning that meta-awareness is reducing the term ``amount of attention paid to [reward loss]'' in the \eqsuff.

\add{
Meta-awareness may be particularly useful in the specific case of threats.}\label{threatintervention}\index{threat!and meta-awareness}
Suppose you were able to see all threats as empty: uncertain, subjective, open to interpretation, nothing but mental constructs. You just watch them from the meta-level, as if from the outside, from a distance.
 This is particularly feasible in case of threats to your self-esteem or social status instead of your survival; or if the threat only happens in a simulation. Then, any threats would not be taken that seriously, and their ability to trigger fear would be weakened.
 From this viewpoint, it is not necessary to reduce the desires that underlie the threats (as we did in Chapters~\ref{self.ch} and \ref{freedom.ch}) or reduce any simulation. It is now possible to directly intervene on the threats themselves by seeing them as empty; \add{in particular, considering them as mere phenomena in the mind.    Again, this intervention is distinct from reducing simulation and the desires underlying threats; there might still be simulation of threatening events and desires leading to threats, but the threats  just do not generate suffering anymore}.\footnote{ From a historical viewpoint, this suggests an interpretation where the Mahayana school complemented the Buddha's original theory of desires and frustration by offering interventions that more directly apply to threats. The Buddha may not have talked very much about fear in his discourses, with the expection of the fear of death (e.g.\ \AN{4.184}; but see footnote~\ref{chahfn} in Chapter~\ref{consciousness.ch} for a Theravadan quote focused on fear), while the Mahayanan emphasis on meta-awareness may work as a powerful intervention towards threats and fear.}

\add{A strong meta-awareness may have the very general effect of reducing the effects of error signalling by decoupling error signals from the actual conscious feeling of suffering.  In the theory of this book, error signalling causes conscious suffering, but they are not the same thing. Error signalling is an information-processing operation amenable to modelling; in contrast, conscious suffering is a subjective experience,\index{subjective experience!and suffering}\index{consciousness!and suffering} and as such,  difficult to understand scientifically. Now, it should be possible to break the causal link from error signalling to conscious suffering, as already pointed out above in relation to the cognitive cycle. This can presumably be achieved by a strong meta-awareness that is able to discriminate between those two phenomena. It opens up the possibility of seeing errors and suffering as two separate mental phenomena, eventually preventing errors from triggering suffering.\footnote{\add{It may seem to be too difficult to say much about this possibility since it is related to the rather mysterious nature of conscious experience. But that may not necessarily be the case: it is also possible that there is a group of cells representing the error and another group of cells representing suffering. It is then thoroughly possible to break the connection between the two groups of cells. The fact that the latter group of cells is intimately connected to conscious experience, and that we don't understand what that connection is,  may not be relevant for the design of this intervention.}}}

 \add{Thus, we see that the real beauty of moving to the level of meta-awareness is that it reduces suffering coming both from frustration and from threats. The key is not to take your thoughts or any mental phenomena too seriously or personally; they are just something automatically generated by your brain, even against your wishes and even against the facts.
   Seneca put this sharply when commenting on his own worrying thought:   %
   \begin{quote} The author [of that thought] is a fool, and he who has believed it is a fool, as well as \\he who fabricated it.\footnote{\add{\Lucilius, 13.13}}\index{Seneca}
 \end{quote}}

\chapter{Recapitulating and unifying interventions} \label{attitude.ch}

\begin{figure}
\begin{center}
\resizebox{0.92\textwidth}{!}{\includegraphics{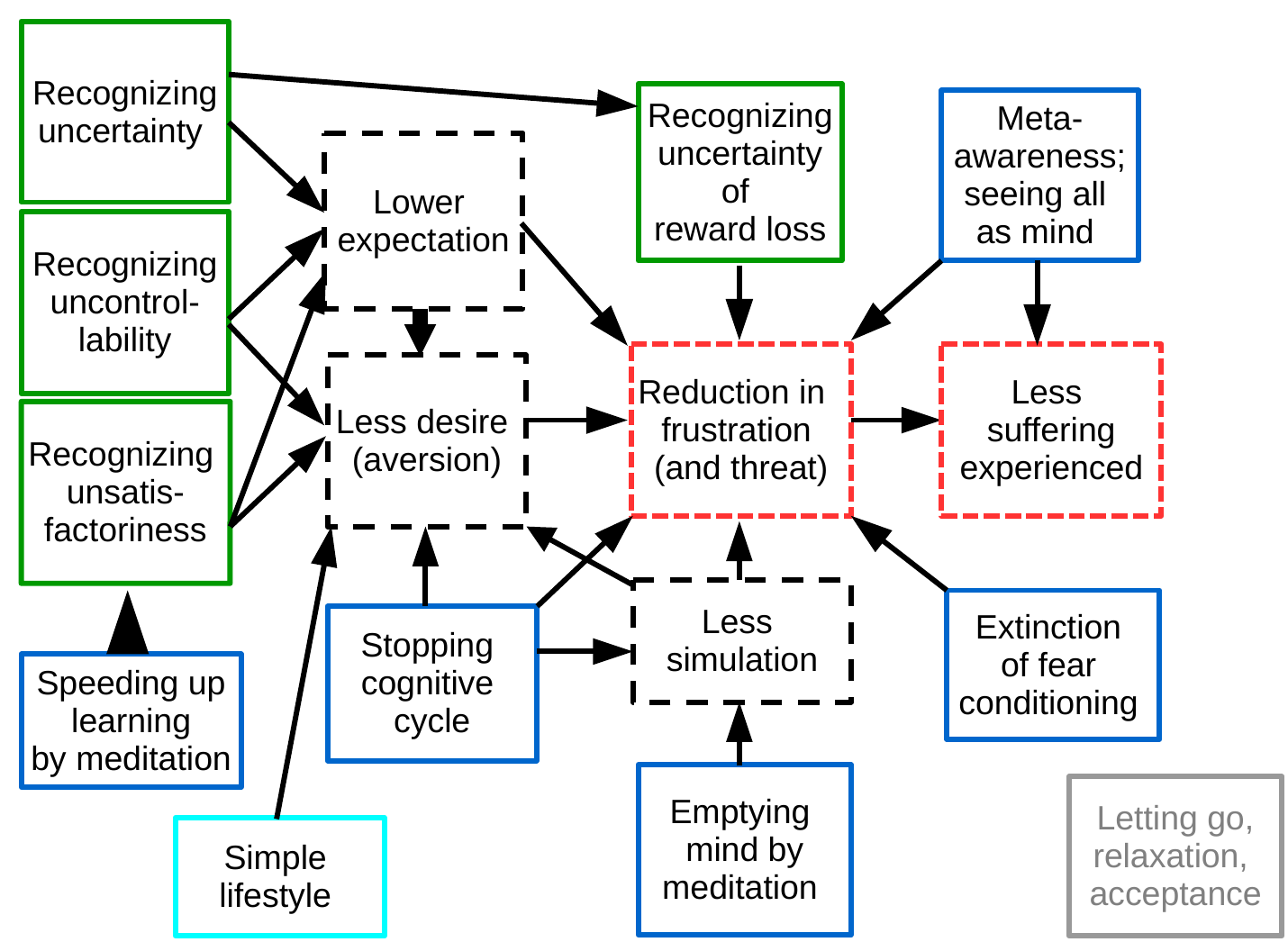}}
\end{center}
\caption{\add{Recapitulation of the mechanisms of the interventions explained in this book. 
    The green boxes are interventions of cognitive-philosophical kind. Lifestyle changes are separate from those, given in a cyan box. While mindfulness meditation is a single intervention, it is divided into a number of (blue) boxes based on the different mechanisms at play. (The single thick arrow from the ``speeding up'' box means that it speeds up the learning in all the boxes above it.)
    The dashed black boxes with dashed contours are intermediate results of those interventions, while the dashed red  boxes describe the final effect of those interventions.
    The gray box in the lower right-hand corner is about general attitudes with rather nonspecific effects,   
    which is why no explicit arrows are drawn there. 
  }}\label{flowchart3.fig}
\end{figure}

 \add{
In this chapter, I start by recapitulating the interventions proposed in the two preceding chapters, and then discuss some typical counterarguments against such interventions---or Buddhist-Stoic training in general.
Then, as a final class of interventions, I discuss the development of positive attitudes towards mental phenomena. Instead of observing the world as it is and learning from it, here the emphasis is on actively cultivating certain ways of relating with the world that reduce negative feelings (valence). This approach is ``positive'' compared to the training based on various kinds of reduction emphasized in the preceding chapters. Acceptance and letting go are the key attitudes here; it turns out that they are intimately related to the reduction of desires and expectations. In fact, letting go, which I interpret as a form of relaxation, can be seen as a principle that unifies most of the training methods described in this book.

   \section{Recapitulating the interventions}

 The flowchart in Fig.~\ref{flowchart3.fig} on page~\pageref{flowchart3.fig} recapitulates the main mechanisms of the different interventions described in the two preceding chapters, while briefly mentioning some to be explained below. %
The most fundamental interventions are recognizing uncertainty, uncontrollability, and unsatisfactoriness (green boxes on the left). These lead to reduction of expectations and reduction of desire, including aversion (black dashed boxes), which reduce the frustration computed and suffering experienced (red dashed boxes). A simple lifestyle (bottom left) is another intervention that I did not describe in much detail in this book, but it is also a well-known method for reducing desires. Another related intervention is recognizing the uncertainty of perception and in particular the perception of reward loss (green box, top middle), which reduces the computed frustration as well.
 
Mindfulness meditation strengthens and speeds up many of the interventions just mentioned, which is depicted as a blue box at the bottom left-hand corner.
Meditation also brings several new mechanisms into play; some are shown as separate blue boxes in the chart, although in practice, they all come from the same intervention. First, meditation tends to make the mind empty, thus reducing simulated frustration (third column, two boxes at the bottom). Meditation also enables stopping the cognitive cycle or chain (Fig.~\ref{flowchart2.fig} on page~\pageref{flowchart2.fig}), which  goes from perception to desire and frustration (blue box in the second column). Such stopping can have many different effects, but what is emphasized here is that it can directly reduce desire as well as simulation, while it can also explicitly prevent the frustration computation itself.
Another intervention offered by meditation  is that it creates a form of metacognition and meta-awareness (blue box at the upper left-hand corner), which may eventually lead to the attitude that all phenomena are just in the mind, as proposed by Mahayana Buddhist philosophy, thus directly reducing all kinds of suffering. 

All the interventions just described are intimately related to manipulating terms in the \eqsuff\ (page~\pageref{suffeq}). (Those that reduce desire are a bit more indirectly related since they essentially prevent the mechanism in that equation from being triggered at all.) However, we also saw some interventions related to meditation that work in very different ways. Meditation is actually a highly complex phenomenon, and it certainly has many effects mentioned only briefly, if at all, in this book. The flowchart lists as a further example the extinction of fear conditioning; recognizing the emptiness of categories could also have been added. The flowchart has a further box on letting go and acceptance, which will be treated later in this chapter.

Essentially, these interventions reduce frustration (red box in the middle). Threat computation is reduced as well since threat is based on predicted frustration. 
Finally, this reduction of frustration and threat reduces the conscious experience of suffering, or mental pain. Meta-awareness is very special in the sense that it can directly reduce conscious suffering even if the frustration or threat computations are performed, which is indicated by the direct arrow from meta-awareness to conscious suffering.

}

\section{How far should reducing desires and expectations go?}

Let us next consider a typical objection that can be raised at this point: the thinking underlying the interventions of the preceding chapters seems depressing. One may ask whether not wanting anything and not expecting much  leads to complete inactivity and, indeed, to some kind of depression. A diagnostic criterion of depression is ``markedly diminished interest (...) in all, or almost all, activities'',\footnote{DSM-5 diagnostic criteria} which sounds a bit like having substantially reduced reward expectations and having few desires.\index{depression} The fundamental question is: Can such reduction of desires and expectations go too far?

Let us consider first how much expectations should be lowered. Is it enough to admit the \textit{actual} levels of uncertainty and uncontrollability, or should we go further and consider things even more uncertain and uncontrollable than they really are, thus lowering expectations even more?
If our only goal were to reduce suffering in the agent, we could simply program it to assume that everything is completely uncertain and completely uncontrollable. Then, the agent would expect zero reward, or very little, in any state or from any action.
As a consequence, it would have virtually no desires either. %
Is this a good way of programming an agent?\footnote{The traditional Buddhist viewpoint tends to emphasize that people are mistaken about the level of control and permanence, and it is enough to correct their ``ignorance'' or ``illusions''.\index{illusion}\index{ignorance (Buddhist)} \add{Both neuroscience literature and reinforcement learning literature do offer examples of how humans are overoptimistic, as well as why that may be a good thing for an agent \citep{palminteri2022computational,ciosek2019better,munos2011optimistic}.}\ On the other hand, consider  a super-intelligent agent which has no constraints regarding data or computation. It would presumably estimate uncontrollability and uncertainty correctly and accurately, without any illusions. But it would still have reward losses, and those reward losses might not even be particularly small, especially if the outside world is difficult to control (perhaps due to strong physical constraints in the ability of the agent to manipulate it) and exhibits a lot of randomness. So, it is not clear if suffering would be very much reduced by correcting ``illusions'' in the sense that the agent learns to make ``optimal'' inference (in the sense of probabilistic AI theory) with infinite data and computation. I would assume that the real goal of such Buddhist practice may rather amount to adopting reward expectations which are lower than what is objectively true. In this case, it would lead to increased happiness at the expense of slightly suboptimal inference---but note that such ``suboptimality'' refers only to the lack of optimality in maximizing rewards, \add{or evolutionary fitness}.\label{optimismfn}} 

Claiming that Buddhist training can lead to something akin to depression is, in fact, a well-known point of criticism, and similar arguments have actually been raised against Buddhism throughout its history.
I think such criticism is not very relevant because it considers an extreme case, which is unlikely to be achieved by most people practicing such systems. Perhaps the point is that most people living in a modern industrialized society simply have too many desires, and it would be better for them to have fewer of them. This would explain why people engaged in Buddhist training tend to get happier when they reduce reward expectations. It may be irrelevant to ask what might happen in the extreme case where they completely annihilate all their desires---which is a feat even most meditation masters are incapable of. Buddhist philosophy actually emphasizes the general principle of the ``middle way'',\index{middle way} or moderation, which sounds like a good idea here as well. The situation might be different for Buddhist monks or nuns engaged in full-time practice for many years, but they follow a very special lifestyle, which is specifically designed to be compatible with having very few desires.\footnote{A famous counterexample to my optimism happened during the Buddha's life, when several of his disciples committed suicide\index{suicide} after intensively engaging in a particular exercise:  reducing carnal desires by  contemplating the loathsomeness of the human body (\SN{54.9}). The Buddha realized his mistake and changed his teaching accordingly. Using loathing as a meditation technique to reduce reward expectations and desires is extremely rare in current Western meditation practice.}

On the other hand, there is certainly something fundamentally different between a depressive state and a mental state where the unsatisfactoriness of the world is seen from a Buddhist perspective. If an agent concludes that none of its desires are  going to be fulfilled and it will never receive any reward, that gives in itself no reason for a negative feeling or valence. The agent would just rationally decide that no desires are worth pursuing, it would not engage in goal-oriented action, it would predict zero rewards in the future, and, consequently, it would suffer \textit{less} since there is no frustration. %

If humans tend to get a negative feeling after seeing that the world is fundamentally unsatisfactory, it must be because there is another ``higher-order'' desire, presumably coming from the self-evaluation system treated in Chapter~\ref{self.ch}.
A depressed person, in particular, finds the very unsatisfactoriness of the world frustrating, and \textit{wants} to find satisfaction or reward in various kinds of seemingly pleasurable objects and activities. In our framework, we would say that she is frustrated in terms of her self-evaluation, as she sees that she gets less reward in the long run than she ``should'' according to some internal standard (possibly based on social comparison).\index{social comparison} 
The self-evaluation system\index{self!esteem and evaluation} may indeed conclude---based on a superficial calculation---that since no goals can be reached and no reward can be obtained, there must be something wrong with the agent. Thus, a negative meta-learning signal is generated, and this would be felt as suffering.

However, I think an important point in the Buddhist philosophy of unsatisfactoriness is that if the self-evaluation system sends a negative signal when the agent does not get enough rewards, the system is simply malfunctioning. Clearly, the realization of the total unsatisfactoriness of everything should also influence the self-evaluation system. The self-evaluation system should set its expectations and its standard of an ``acceptable'' reward level very low, even zero. The self-evaluation system cannot rationally claim that the agent is not getting enough rewards if the system itself believes that no rewards can possibly be obtained! As such, Buddhist philosophy proposes that there is no need to be frustrated about any long-term lack of reward, nor is there any need to make any negative self-evaluation; not getting much reward and not reaching one's goals is natural and unavoidable.

\pagebreak

\subsection{Is frustration not needed for learning?}

Another objection that could be raised against the philosophy presented here is that it may not be useful to reduce frustration since the frustration signal is useful for learning. Human beings seem to be trapped in a situation where they need frustration to learn, while they suffer from it. That may sound like a dilemma with no satisfactory solution. However, I'm not sure there is any \add{serious} dilemma here. One reason is that, as discussed in Chapter~\ref{rpe.ch}, many of the rewards we are programmed to receive are actually rather useless ``evolutionary obsessions''; frustrating them may not teach us anything useful, if it is not the very futility of those rewards. The same is true from the viewpoint of insatiability: why should one try to learn how to better satisfy desires that cannot be satiated anyway?

Furthermore, Chapter~\ref{consciousness.ch} proposed that a large part of the problem is how frustration is made conscious even though it need not be; learning from frustration could, in principle, happen on an unconscious level. It might seem that not much can be done about this, but in fact, an intervention is possible, as was seen in the discussion on meta-awareness in Chapter~\ref{training.ch}.

\add{Likewise, it might be claimed that thinking about threats is useful since then, the agent learns to avoid them. However, people worry about horrible things which are extremely unlikely to happen;  they also worry about things which are unavoidable, such as death. Such worrying is unlikely to improve the performance of the agent at all. It may actually decrease the performance since so much energy is spent on those rather useless computations.}

Yet another counterargument is that while understanding the uncertainty of all perceptions reduces frustration, it may actually \textit{improve} learning and make us more ``intelligent''. Uncertainty and uncontrollability are real properties of the world, but we may have been grossly underestimating them.\footnote{ \add{See footnote~\ref{optimismfn} in this chapter.}} Thus, learning to better appreciate uncertainty and uncontrollability is a useful meta-learning process, even from the viewpoint of trying to optimize rewards in the world. 

Based on these counterarguments, I think that while it may be meaningful to claim that not all frustration should be removed,  \textit{most} of it can still be removed without making learning or the ensuing behavior any worse.\footnote{In fact, if somebody argues that frustration is actually good since it enables learning, the question arises as to why frustration is painful. If frustration is ``good'' and should be encouraged, frustration should feel pleasant, not painful, based on elementary evolutionary logic. The fact that frustration is painful means that at least in some evolutionary sense and to some extent, it has been deemed to be bad for you. %
  As a thought experiment, suppose frustration felt good, perhaps because you have become so thoroughly convinced about the utility of the ensuing learning that you are able to override millions of years of evolution. Then, you would presumably try to fail in everything you do---it feels good and you will consider that good feeling as some kind of an internal reward. You might learn a lot from such failures, although if you fail without even trying hard, the utility for learning might be meager. In any case, you would not get much reward; you might starve, die young, and would not produce any offspring. This (admittedly not very rigorous) argumentation suggests that frustration should be avoided from an evolutionary perspective and thus, it has to be evolutionarily made painful. However, this argumentation was based on an extreme case. Perhaps there is an optimal amount of frustration which is not zero; perhaps it would be possible to detect circumstances under which frustration is good while it is usually bad. I leave this for future research.} 

\subsection{Interventions need to be based on personal preferences}

\add{Some readers may still not be convinced. They might argue that desires feel good;  without them, life would be empty; besides, they have no time or energy for the training described in this book. I think it is important to understand that this book is fundamentally an exposition of a scientific theory. A scientific theory per se does not tell you what to do. It only tells you that if you do X, then Y will follow (with some probability); it explains causal connections and, in particular, what effects different interventions will have.
As such, it is up to you to decide which interventions, if any, are actually worth it for you. It depends on your preferences or your values (in the ordinary sense of the word), and no scientific theory can tell you what to do without considering your personal preferences. The interventions have side-effects or costs, and the cost-benefit analysis is different for each individual and intervention. One individual may get a lot of satisfaction from pursuing and achieving a certain goal, while another might get much less; for some other goal, it might be the other way around. Likewise, how much an individual benefits from a particular intervention must vary from one individual to another. Ideally, some scientific analysis of your personality, temperament and lifestyle might be able to tell which interventions are the best for you, but we are not there yet.}

\section{Positive viewpoints to reduction}

\add{Some of the arguments against Buddhist-Stoic training may be due to the simple fact that \textit{reducing} anything sounds like a negative thing, as if you were missing out on something. Next, I will try to give more positive interpretations of the reduction---or even absence---of desires and expectations.}

\subsection{Contentment, gratitude, and freedom}

To begin with, the absence of desires can be expressed as \textit{contentment}, in the literal meaning of being content with what one has and not wanting more. The insatiability of desires, in particular, implies that a simple-minded agent cannot be content: it always has to search for more rewards, leading to endless frustration. If an agent has no desires, we can, from a positive viewpoint, consider it to be content with the current situation.

\add{If contentment becomes strong enough, it may turn into a feeling of \textit{gratitude}. Gratitude training or meditation is a major topic in itself, and there are specific methods to increase gratitude. Gratitude is an emotion that is social or interpersonal, which is why it is a bit outside of the theory of this book. In any case, the paradox of gratitude is that it actually enhances the well-being of the person being grateful---not only of the person to whom the gratitude is directed. As Seneca puts it:
  \begin{quote}
    I am grateful, not in order that my neighbour, provoked by the earlier act of kindness, may be more ready to benefit me, but simply in order that I may perform a most pleasant and beautiful act; I feel grateful, not because it profits me, but because it pleases me.\footnote{{Quote from Seneca, \Lucilius, LXXXI.20. Gratitude was also strongly recommended by Epictetus (\Discourses, I.6. II.5.10, II.16.28) as well as Plutarch (\textit{On the Tranquillity of the Mind}, 14). } Recently, it has become a topic of great interest in positive psychology \citep{emmons2002gratitude,wood2010gratitude,watkins2013gratitude}; \textit{appreciation} is a related construct that may be defined as something more general \citep{fagley2016construct}.}\label{gratitudefn}\index{gratitude}
\end{quote}  }

An even more fundamental positive interpretation of having no desires is \textit{freedom}.\index{freedom} While an emphasis on freedom is ubiquitous in Buddhism, 
Epictetus summarizes the idea in a way that is, yet again, in complete harmony with the Buddha's philosophy:
\begin{quote} \hspace*{0mm}Freedom is acquired not by the full possession of the things which are desired, \\but by removing the desire.\footnote{\add{Quote from \Discourses, IV.1.175. The whole Chapter IV.1 in \Discourses\ is dedicated to explaining how the goal of Stoicism is freedom, in particular, freedom from ``being constrained or impeded by any external circumstance or emotional reaction'' according to \citet[p.~27]{LongEpictetus}. Seneca talks about our slavery in more specific terms, related to something like valences in \Beata, 4: ``See (...) how evil and guilty a slavery the man is forced to serve who is dominated in turn by pleasures and pains, those most untrustworthy and passionate of masters. We must, therefore, escape from them into freedom.''; see his also \Lucilius, LXXV.18.\index{Seneca} Emphasis on freedom is ubiquitous in Buddhism as well, even if the word may be used in various meanings. While the whole goal of the Buddha's teaching is often formulated as ``freedom from suffering'', this is rather uninformative and uses the word ``freedom'' in a different sense than considered here. Ancient Buddhist texts also consider the metaphysical goal of freedom from reincarnation, but that is clearly outside of the scope of this book.}\ For our purposes, a very useful cognitive interpretation is given by \citet{peacock2018vedana} who formulates the goal of early Buddhist philosophy as freedom from ``reactive patterns'' triggered by valences. This is of course related to freedom from desire (and aversion) advocated in the third of the Buddha's Four Noble Truths.}
  \end{quote}
More generally speaking, the condition of a human being has been described as being a  ``puppet of the gods'' by Plato,\footnote{Plato, \textit{Laws,} Book I. These puppets are different from those that Plato talked about in his more famous cave allegory. See also Marcus Aurelius's \textit{Meditations} II.2.}\index{Plato} meaning that ``affections in us are like cords and strings, which pull us different and opposite ways''.  We have to remove those cords and strings if we want to be free. \add{Instead of being enslaved by our neural networks with their interrupts and unconscious action tendencies, we need to liberate ourselves from such evolutionary constraints, giving more space for conscious deliberation and the use of \textit{reason}.}

\subsection{Attitude of acceptance}

Another positive attitude that is fundamental in meditation practice is acceptance.\index{acceptance}
There is, in fact, an important caveat in any attempt to reduce mental phenomena, be it desires or wandering thoughts. It is important that this training does not lead to the idea or evaluation that the mental phenomena are somehow bad. Such an attitude would, in itself, easily lead to aversion and thus, to suffering. In the extreme case, if there is aversion towards the mental phenomenon of aversion, that may lead to a vicious circle, which constantly increases aversion. To counter this tendency, it may be necessary to actively create new mental phenomena so as to neutralize the existing ones.\index{aversion!to desires}

It may sound paradoxical to say that one should not think of the mental phenomena as bad, or at least undesirable. How could one not think that, say, desires are bad if one believes they lead to suffering? And how is one supposed to get rid of them if one does not regard them as something negative, something to be avoided?

The solution to this paradox is that while the actions of the meditator should be chosen so as to reduce desires (or other mental phenomena), it is still possible  to avoid creating any new aversion in the sense of a new mental process. Thus, on an abstract level, it is useful to consider the desires ``bad'', or perhaps rather as something that it would be better not to have, but such thoughts should just work in the background as weakly as possible, instead of being strong and actively cultivated. In particular, they should not lead to any interrupt-like aversive emotions. Such processing is possible since the neural networks can implement automated habit-like action tendencies that try to avoid certain phenomena, and that can happen without any need to activate the desire/aversion system. As an extreme example, when you are walking, you know that losing your balance is ``bad'', but you probably don't feel a constant aversion or fear towards stumbling; your neural networks have simply been trained to avoid that happening; they ``reduce stumbling'' so to say but without any aversion.

In practice, it has been found that with meditation, the tendency to develop aversion is so strong that specific techniques are necessary to reduce it.
The key technique is to cultivate the attitude of acceptance. This means a general attitude of accepting all thoughts and sensations that come to the mind, instead of resisting or judging them. More precisely, acceptance here means simply not activating processes of aversion, i.e.\ not activating a desire to get rid of something. So, acceptance here is taken in a very limited sense: this is neither about moral acceptance nor about thinking that some things could not be bad for you. Such acceptance could also be described as  removing resistance; \textit{nonreactivity}\index{nonreactivity} is a related term used in current research.\footnote{\citet{lindsay2017mechanisms}, while emphasizing the importance of acceptance in mindfulness training, use the term almost synonymously with ``nonreactivity''.
\citet{hayes2005acceptance} define acceptance as ``an open and noncontrolling stance toward all experiences'', which shows explicitly the connection to control. Meanwhile, \citet{peacock2018vedana} offers an interesting interpretation of the goal of early Buddhist philosophy in terms of ``freedom from enthrallment to reactive patterns'', which places nonreactivity at the very center of Buddhist training.
It should be noted, however, that in actual meditation training, it is often recommended that an active, positive feeling (possibly what is called loving-kindness)\index{loving-kindness} is developed towards mental phenomena \citep{grabovac2011mechanisms,hofmann2011loving,brach2004radical}. It may be necessary to actively develop such positive feelings to counteract the inherent tendency to aversion and judgement; simply trying to refrain from negative judgements and practising meditation based on observation may not remove them efficiently (\SN{10.4}). 
}
For example, a depressed person may be annoyed by the very occurrence of rumination.\index{rumination}\index{depression} In such a case, accepting the fact that rumination occurs may actually be beneficial, since it removes the suffering due to the aversion to rumination.\citenew{feldman2010differential} Again, acceptance does not here mean that the person would give up any techniques that reduce the rumination.

An accepting meta-cognitive attitude can actually be adopted towards all mental phenomena. Many mental training systems include some kind of active acceptance practice of all mental phenomena as an integral part. 
An acceptance practice can be seen as a specific method for reducing aversions of all kinds. It complements the methods described in the preceding chapters, which were more oriented towards reducing desires in the restricted sense of the word (i.e., excluding aversion). It is closely related to the practice of letting go, which will be considered below.\footnote{A practical introduction to such meditation methods is provided by \citet{brach2004radical}.}

Theories such as those explained in this book may help in the acceptance training because simply understanding the mechanisms behind, say, wandering thoughts or emotions may enable you to accept them. Suppose you are convinced that they are natural processes, which even have some computational benefits, and that they are largely outside of conscious control. In that case,  it may be easier to just let them happen and go away naturally, without fighting against them. This is related to seeing ``causality'' in the Buddhist sense of the word (considered in Chapter~\ref{freedom.ch} on page~\pageref{causality}),\index{causality!Buddhist} but it goes further since the phenomena are seen as not only natural and uncontrollable but even useful---at least from an evolutionary viewpoint. 
\add{Based on this viewpoint, even failures and errors could be accepted as an unavoidable part of a learning process, or of life; it is not necessary to get upset by them and activate the brain's pain system.}

Ultimately, even \add{the feelings of pain and suffering themselves}\ need to be accepted on some level. 
Any aversion towards them will create a lot more suffering.  
As an extreme example, people suffering from chronic pain will suffer even more if they ``catastrophize'' the pain, resist it, and develop a particularly negative attitude towards it; accepting the pain will help.\citenew{veehof2016acceptance}\index{pain!chronic} The Buddha gave a famous simile of a man who is struck by an arrow, which inflicts physical pain. If the man ``sorrows, grieves, and laments'', feeling aversion towards pain, he makes the suffering even worse, as if he were struck by a ``second arrow''.\footnote{\SN 36.6}\index{Buddha}

\section{Letting go and relaxation as unifying principles}

\index{letting go}
Buddhist philosophers often use the concept of ``letting go'' to recapitulate the general attitude that underlies the mental training described in this book.
At the most concrete level, the idea is that we let go of things and objects in the sense that we don't strive to possess or control them anymore. On a more computational level it means we let go of desire, i.e.\ we don't even want those things in the first place---nor do we want to avoid them. The same approach can further be applied to thoughts and perceptions, which are understood to be subjective and unreliable, so they can be let go of. Feelings and emotions are likewise just observed and then let go of. The whole simulation called consciousness is no longer taken that seriously. Combined with the no-self philosophy, the attitude can be recapitulated as letting go of everything that is not part of me, and since nothing really is part of me, or my ``self'', \textit{everything} is let go of.\footnote{To quote \SN{35.101}:
``Whatever is not yours: let go of it. Your letting go of it will be for your long-term happiness and benefit. And what is not yours? The eye is not yours: let go of it. (...) [Visual] forms are not yours: Let go of them. (...) Eye-consciousness [i.e.\ visual awareness] is not yours: Let go of it. [The text goes through all the sensory organs, the objects of sensation, and the accompanying sensory awarenesses.]  The intellect is not yours: let go of it. (...) Ideas are not yours: let go of them. (...)
  Whatever arises (...), experienced either as pleasure, as pain, or as neither-pleasure-nor-pain, that too is not yours: let go of it. Your letting go of it will be for your long-term happiness and benefit.'' (Translated by Thanissaro Bhikkhu)}\index{no-self!letting go of self}

Letting go is an expression that obviously has a clear connection to the term ``reduction'' that we have used very often. %
It is not so much a question of programming new routines or new functionalities. The idea is to reduce activity, letting go of existing mental associations and routines. 
The key is less desire and aversion, less replay and planning, fewer interrupts, and so on.\footnote{Alternatively, letting go could be seen as the opposite of attachment, especially if the corresponding term (\textit{up\=ad\=ana}) is translated as ``grasping'' or ``clinging''.\index{attachment (Buddhist)!vs letting go} However that would require an interpretation of attachment  which is quite different from what I have done in this book.}

An important point about letting go is that it circumvents the paradox of wanting not to want anything. If meditators want to reduce desires, they can be seen as wanting not to want, which may sound impossible.\index{desire!to have no desire}
This apparent paradox in Buddhist philosophy has been pointed out by a number of authors: since wanting not to want is a form of wanting, how could one possibly get rid of wanting by such wanting? The paradox is actually so obvious that even the Buddha himself, as well as his immediate disciples, were confronted with claims that his system is inherently paradoxical.\footnote{\add{\SN{51.15}. \citet{dejonckheere2022perceiving} found that social pressure to be happy makes people less happy.}} Thinking of the mental process in Buddhist training as letting go, and as reduction, should largely resolve this paradox of seemingly wanting not to want.
The term ``letting go'' describes a reprogramming that reduces mental activity instead of introducing a new desire.

One way of interpreting letting go is that it is mental \textit{relaxation}\index{relaxation}\index{meditation!relaxation} in the sense of absence of activity and tension. Desire and the subsequent goal-setting are about actively engaging in a mental activity, and thus they are a kind of opposite to relaxation. Figuratively speaking, just as muscular activity prevents physiological relaxation, wanting is the opposite of mental relaxation in that it relies on specifically activating certain computational processes. If you set the goal that you don't want anything, you would actually be just setting one more goal, and increasing mental activity---this is another viewpoint to the paradox we just saw. But if instead, you learn to relax the planning and goal-setting system so that it simply rests, and does not set any goals and does not plan, then you resolve the paradox of wanting not to want. \add{Furthermore, you can relax the evaluation mechanisms that compute frustration.} Learning such relaxation is not easy, but the training methods discussed in this book were basically all designed to lead towards such a mental relaxation.\footnote{\add{%
    I emphasize that there is nothing contradictory or impossible in such training: in particular, there is nothing contradictory in  ``wanting to relax all desires''. What is needed is that the agent's information-processing system creates a desire to relax all \textit{other} desires, and then takes as its goal the state where all other desires are relaxed. When the agent has relaxed all other desires, this (meta-level) desire for relaxation goes away by itself, just like any desire disappears after its goal has been reached. Thus, in the end, all desires, including the meta-level desire, have vanished. Any contradiction is avoided because this meta-level desire to relax desires is only directed at other desires, not at itself, and because desires go away automatically when their goal is reached. This is also my interpretation of  \SN{51.15}, where the paradox of wanting not to want is resolved by explaining how ``[a mendicant who is perfected] formerly had the desire to attain perfection, but when they attained perfection the corresponding desire faded away'' (Trans.\ Bhikkhu Suhato). }
  A complementary approach to  resolving this paradox is to consider how the meditation practice changes over a time span of many years. Initially, meditation is based on the desire to reduce suffering, and makes use of the desire to let go or relax. But ultimately, you let go of even the desire to be happy, and, paradoxically, of the desire to let go (or relax). This is possible since you let go of letting go only after a long practice, so %
  the attitudes and habits required for letting go or relaxation are now  automated in your neural networks and need no effort or explicit desire to operate anymore. You just relax and let go automatically, without desire or planning to do so. Note that this is clearly related to the problem of ``aversion towards aversion'' that we considered above in connection with acceptance. In a similar vein, \citet{striker2004historical} emphasizes that Pyrrhonian Skeptics\index{Skeptics} did not (actively and purposefully) suspend judgement, as it is sometimes claimed, but rather were unable to arrive at any judgement and gave up any such attempt; see also \citet{herman1979solution} for further analysis of the Buddhist case. %
  }

The ultimate goal of Buddhist training is called \textit{nibb\=ana} or \textit{nirv\=ana},\index{nibb\=ana/nirv\=ana} depending on which ancient Indian language is used. It is defined as a state devoid of any suffering, the cessation of all suffering. The term literally means \textit{extinction},\index{extinction!as in nibb\=ana/nirv\=ana} as in a fire being blown out. It is often described in negative terms such as ``unconditioned'', ``unconstructed'', or even ``unborn'', which may sound nonsensical. I think the key to understanding this is that nibb\=ana is reached by reducing, and ultimately removing, various mental phenomena, in particular desire; it is not about constructing any new mental phenomena. This may again sound paradoxical to any beginning meditator struggling to maintain even a tiny amount of concentration, but I am of course talking about highly advanced stages of practice here.
Thus, the best description of the ultimate state may be entirely negative, in terms of what it is \textit{not}, and what it does \textit{not} contain.\footnote{\citet{sayadaw2016manual} gives a traditional Theravadan commentary:   ``Because there is no arising in the nibb\=ana element [which is the cessation of conditioned phenomena through their non-arising], it is called not-born (\textit{ajata}) and not-brought-to-being (\textit{abhhuta}). Because it is not made by a cause, it is called not-made (\textit{akata}). Because it is not made dependent on causes and conditions, it is called not-conditioned.''   (see his Chapter ``Attainment of Fruition'').} It is often described as freedom, and in particular it is freedom \textit{from} those elements of the mind that produce suffering.\footnote{While the Four Noble Truths\index{four noble truths (Buddhist)} indicate that extinguishing desire accomplishes the goal of removing suffering, aversion (or hate) and ignorance (or delusion) are usually added to the list of phenomena that have to be extinguished, see e.g.\ \SN{38.1}. (The exact meaning of ignorance/delusion in this context is quite controversial.) Such lists come in various lengths, and ultimately may contain almost all mental phenomena, as when the Buddha says that he teaches ``for the elimination of all standpoints, decisions, obsessions, adherences, and underlying tendencies, for the stilling of all formations, for the relinquishing of all attachments, for the destruction of craving, for dispassion, for cessation, for Nibbana.'' (\MN{22}).  It should also be noted that the conception of nibb\=ana or nirvana is quite variable among different Buddhist schools. For a detailed account of the early Buddhist view, see \citet{harveybook}.
  In later Buddhism, there is more emphasis on the extinction of conceptual thinking---as when Nagarjuna says that nirv\=ana is ``the calming of all verbal  differentiations''   \citep[p.~75]{williams2008mahayana}---as well as the realization of the ``nature of mind'' \citep[e.g.,p.~156]{mahamudra} which is an advanced form of meta-awareness. 
}

One might think such a mind-state with no contents must have neutral valence, and could even be boring.\footnote{\add{But obviously, boredom\index{boredom} is a mental phenomenon, akin to an emotion, which cannot exist in a truly empty mind. For a review on boredom research, see \citet{danckert2023search}.}} Yet, Buddhist philosophy claims it is extremely happy and pleasant, in fact pure bliss. It is claimed to be the only thing that is not unsatisfactory in any way. This may perhaps be understood if we consider the mind in such a state to be \textit{completely} empty,\index{empty mind} and we have seen that even a relatively empty mind seems to be, for some reason, quite happy.\footnote{In particular, the mind might be empty of all perception in addition to thinking, even including proprioception and interoception (feeling of the body, see footnote~\ref{propriofn} in Chapter~\ref{consciousness.ch}), which are partly the basis of the feeling of ``self''.  The Japanese Zen master Dogen said that he experienced the ``dropping away of body and mind'',\index{body} while \citet[p.~158]{brahm} emphasizes that in deep meditative absorption (\textit{jh\=ana}), ``the five senses have shut down''. Clearly, such complete emptiness can only be achieved by letting go of everything, a total mental relaxation, not by making an effort to empty the mind.}
Nevertheless, we find yet another interesting paradox: How can having a completely empty mind possibly be pleasant since it logically should not contain any pleasure either? I will not try to resolve this paradox, which seems to reach metaphysical depths; let me just quote S\=ariputta, one of the closest disciples of the Buddha, who put it very simply:\footnote{\AN{9.34}, \add{translated by Thanissaro Bhikkhu}.}
\begin{quote}\index{S\=ariputta}
Just that is the pleasure here, my friend: where there is nothing felt.
\end{quote}

\chapter{Epilogue}\label{epilogue.ch}

There is a wide consensus that trying to build an AI teaches us a lot about what human intelligence is about: an AI works as a model of the human mind. I think this also applies to suffering.  For sure, a model is not the same as the real thing; some things are always missing.  You cannot actually drive to work with a computational model of a car; mathematical equations of physical forces and chemical reactions written on a piece of paper do not actually make your car accelerate. Yet, it is such models that enable the construction of cars and even rockets that fly to the moon.

\index{model}
A good model can tell us a great deal about the real thing, and thus help science understand how a complex system works. A model can also enable us to predict what the system does in the future, for example, by providing a weather forecast. But from the viewpoint of this book, what really matters is if the model is predictive in the following narrow sense: Does it enable us to predict what results \textit{interventions} have on the system? That is, does it help us in changing the system in some way we find preferable?\index{intervention}

This book proposes that computational models of human suffering can tell us what kind of processes are \textit{necessary} for suffering. The AI models in this book explicitly showed us some of the conditions, causes, and processes that have to be operating in order that suffering arises.  That means we can develop methods that will \textit{reduce} suffering: We simply need to remove the necessary conditions, or, at least, make them weaker.
This is why I think the models in this book are useful, and the later chapters of this book were, in fact, all about methods to reduce suffering.

It is possible to argue that an AI or a robot cannot \textit{really} suffer since it is not conscious. 
In other words, the computational processes considered in this book may not be \textit{sufficient} for suffering if one insists that suffering must be conscious. However, that is beside the point if our main goal is to develop methods that reduce suffering. Actually, some even claim an AI is not really intelligent---according to some stringent conditions for intelligence---yet AI is not only capable of performing some very useful practical tasks, but it has also greatly advanced human neuroscience by giving insight to the computations performed by the brain.

The interventions I proposed  were mostly identical to what existing philosophical systems propose, while I showed how to motivate them using current AI theories. 
The theory in this book will hopefully be complemented by further research; I think this is just the very beginning of a long-term scientific endeavour.
I hope it will lead to more and more efficient interventions in the future, including completely new kinds of interventions.

I certainly do not claim that the theory in this book would be either complete or perfect. In particular, there are quite probably mechanisms of suffering which do not fit into the framework of this book. That may be the case, for example, for suffering due to certain kinds of social emotions, or existential suffering such as lack of meaning of life. The theory in this book also attempts to explain all kinds of suffering---including self-needs, uncertainty, uncontrollability, negative emotions (such as fear and disgust), and stress--- by the single mechanism of frustration.
Whether such a theory based on a single mechanism is satisfactory remains to be seen in future research. For example, some interpretations of Buddhist philosophy further maintain that desire and aversion \textit{in themselves} are suffering, and it is not quite clear how that fits the framework in this book.\footnote{See Chapter~\ref{summary1.ch} (page~\pageref{desiresuffering}) for discussion on this point.} As always in science, theories can be rejected, at least partly, as science progresses. 

\subsection{Summary: Limitations of the agent lead to errors and their monitoring leads to suffering}

To recapitulate the book in a few paragraphs: we saw several ways in which the limitations of the agent and its intelligence lead to suffering. We can succinctly summarize the main problems as uncontrollability, uncertainty (or unpredictability, or impermanence), and unsatisfactoriness (including insatiability and evolutionary obsessions). The agent cannot control its environment as much as it would like; it is not able to perceive or predict the world with much certainty; it strives endlessly at goals \add{ultimately given by the programmer} (which in humans means evolution), unable to ever find satisfaction.

Due to these limitations, the cognitive system will make errors in its predictions, its plans, and its actions.
We saw that suffering is basically a function of the constant evaluation
that an intelligent system performs regarding its actions, resulting in an error signal.
Without such evaluations, the performance cannot be improved.
In particular, error signalling is necessary for the system to learn and update its model of the world. \add{Frustration is the central form of such error signalling.}

\add{The brain and other systems with a particularly sophisticated cognitive architecture use some clever tricks to improve their performance.   Threat computations give predictions of possible future errors and lead to fear, thus providing another mechanism of suffering.}
Wandering thoughts speed up learning by running learning algorithms in the background; however, they make us experience simulated suffering in addition to the real one. Emotional interrupts are useful when unexpected things happen and the computational resources need to be redirected, but they can be mistuned and lead to unnecessary alarms and suffering. Highly intelligent agents may have to use parallel and distributed processing where it is no longer clear if anybody is in control. 
This means that, unfortunately, intelligent agents increase their own suffering by such mechanisms intended to improve future reward. In animals and humans, we also find processes related to self-preservation and self-evaluation, which create another layer of suffering. 

\add{Thus, the constant monitoring and signalling, even prediction of errors creates constant suffering, and paradoxically, the more intelligent the agent, the more  error signalling there seems to be.}
This is what leads to the simple maxim in the title:  \textit{intelligence is painful}.

\add{Yet, the theory of this book is not pessimistic: we also saw a large number of interventions to reduce suffering. Reducing desires and expectations reduces frustration; meditation enables a metacognitive viewpoint that reduces suffering in general. Therefore, I can claim the model of this book  is actually useful: it does, quite directly, lead to a number of interventions. The interventions presented here are not very different from Buddhist or Stoic training, but there is a promise of not only better understanding those interventions but further optimizing them and developing new ones.}

\subsection{Does intelligence necessarily lead to suffering?}

It could thus be argued that suffering is the price to pay for intelligence: without some kind of error monitoring, learning is not possible. It is common sense that errors due to past decisions have to be detected in order to learn to make wiser decisions in the future. Error signals might not be needed if the agent were programmed to be sufficiently intelligent to begin with, so that it would not need any kind of learning, but current AI research suggests that intelligence without learning is very difficult to achieve.

Yet, one might ask if the price is too high, whether intelligence is worth the suffering.\footnote{This question was already considered in Chapter~\ref{attitude.ch} from a slightly different angle.} Would you prefer to be a bit dumber if that reduced your suffering? Suppose a drug were developed which abolishes any error-signalling in humans; perhaps that is possible by interfering with the dopamine metabolism. Suppose that as a logical side effect, it prevents you from learning new reward associations. Would it be worth taking? Actually, we don't even need to consider such an extreme case where all error signals are removed. How about just taking a small dose of that drug, so that error-signalling is reduced to some extent? You would suffer less but perhaps learn new things a bit more slowly. What would be the right balance between maximizing performance and reducing suffering? 

I have argued  that many desires are actually not good for us, and should be seen as evolutionary obsessions. Some desires are insatiable, so trying to learn to satisfy them is a fool's errand.\index{insatiability} Perhaps most importantly, the uncontrollability of the world makes a large proportion of desires completely impossible to satisfy.
Clearly, frustration in those cases should be avoided altogether; they present no real trade-off between suffering and intelligence. If you really want to be frustrated, better do it in cases where the desires actually serve a useful purpose, and you learn to act more efficiently in a meaningful context. \add{We have to also bear in mind that the right balance, or the adequacy of any interventions, ultimately depends on the individual and their personal values and preferences.}

On the other hand, even if we admit that a certain amount of suffering is necessary as a trade-off to achieve intelligence, is it really necessary that such error-signalling should be \textit{consciously} experienced as suffering?\index{error signalling!why conscious}  Even the most rudimentary AI computes errors while hardly being conscious. We would
not say that  a thermostat, arguably the simplest possible system
with some intelligence, is suffering or feels pain when it
realizes the temperature of the room is not what it is supposed to be.
This leads to another thought experiment: How about a drug that does not reduce error-signalling, but prevents it from reaching our conscious perceptions---would you not take it? In fact, this need not be just a thought experiment. Moving to the level of meta-awareness, as described in Chapter~\ref{training.ch}, seems to reduce the felt impact of all suffering, a bit like such a drug.

Consciousness is a great mystery. It cannot be entirely avoided in any discussion on intelligence or suffering, but unfortunately, there is very little we can say with any certainty. One thing which is clear, though, is that the way human consciousness usually operates is not very nice from the viewpoint of suffering. A large amount of suffering is even created out of nowhere by conscious simulation.

\subsection{From intelligence to wisdom}

Nevertheless, intelligence may not only be a bad thing from the viewpoint of suffering. Intelligence may lead to reduction of suffering once it reaches a certain stage, while being embedded in a culture that actively investigates where suffering comes from and what can be done; \add{it may lead to the birth of philosophical systems that question our evolutionary tendencies.} Buddhist philosophy, together with the Stoics and other related systems, proposes that we should adopt certain ways of thinking which counteract, and to some extent neutralize, the causes of suffering.
For example, we should give up any attempts to control and accept that things are just happening; we should recognize that actually we don't know much and are always making decisions under uncertainty; we should give up the meaningless and even destructive desires programmed by evolution. Ultimately, we should recognize the true nature of our consciousness, that we are operating in a kind of virtual reality, which bears only some indirect relation to the actual reality.

Such proposals are quite radical, and have been recognized as such for centuries. This is not surprising since reducing desires and giving up control are strictly against our evolutionary programming. However, it may not be necessary to follow these ideas to any extreme extent: Buddhist philosophy in itself proposes the ``middle way'',\index{middle way} the idea that going to any extremes is, in the end, counterproductive. Instead of giving up all control, for example, we might just give up some of the control, preferably on those things where claiming control is most clearly conducive to suffering. 

Most philosophical systems that discourage acting out our desires do recognize that a human being needs to take some actions; they do not recommend complete inactivity as some might assume. Stoicism as well as Taoism emphasize acting ``naturally'' (or according to one's nature), which I would interpret in terms of the habit-based, automated action selection: learned associations between the current state and actions may still remain even if no reward is expected or predicted anymore.\footnote{Epictetus recommends to ``behave conformably to nature in reaction to how things appear'' (\EN, Paragraph~6), see also the discussion on Marcus Aurelius by \citet{Hadot}, as well as Sextus Empiricus's \textit{Outlines of Pyrrhonism}, 1.23.\index{Epictetus}\index{Skeptics} Laotse (Laozi)\index{Laozi} recommends ``nonaction'' or ``effortless action''---a highly complex concept with many interpretations. According to \citet{laozi-stanford}, it ``seems to be used more broadly as a contrast against any form of action characterized by self-serving desire''; ``nonaction would be 'normal' action in the pristine order of nature, in which the mind is at peace, free from the incessant stirring of desire.'' \add{Such natural action is closely related, in our cognitive terminology, to automated action selection  as in habits.
Automated action does lead to some frustration according to our RPE theory, but it seems to be weak as argued in Chapter~\ref{summary1.ch}. }  From an alternative viewpoint, such acting naturally could mean that any attempt to control is minimized by choosing courses of action which are in harmony with the environment; this interpretation does not, however, explain where the ultimate motivation for action comes from. In any case, it seems important that such natural action is still constrained by sound moral principles, so that it does not mean just doing whatever one feels like.} In early Buddhist thought, motivation for mental development is often seen to be a desire of a special kind that should not be eradicated, thus providing another motivation that is not based on reward maximization.\footnote{In early Buddhist philosophy, desires for spiritual development and similar things are called \textit{chanda}, often translated as ``aspiration''. However, I'm not aware of any principled way of distinguishing between \textit{chanda} and the ``bad'' desire (called \textit{tanh\=a}), so this seems to be just assuming an arbitrary exception to the general theory. In fact, in later Buddhist philosophy, even getting rid of the desire for spiritual development \textit{is} considered important, in line with the discussion on letting go at the end of Chapter~\ref{training.ch}.} Some parts of Hindu philosophy suggest performing one's duty without any concern for reward,\footnote{\textit{Bhagavadgita} 2:47 says ``You have a right to perform your prescribed duty, but you are not entitled to the fruits of action''. Duty is a socially defined concept, and as such, outside of the scope of this book.  Stoic philosophy can be seen in this light as well, if the Greek \textit{kathekon} is translated as ``duty'', see p.~172 in \citet{Hadot}.  Acting according to God's will is another formulation used by Epictetus (e.g.\  \textit{The Discourses}, IV:1) and of course by  various religious systems.}
while later Buddhist philosophy emphasizes altruistic action\index{altruism} as the ultimate motivation for fully enlightened beings.\footnote{For the Mahayana school, see \citet{oldmeadow1997delivering}, but similar ideas can certainly be found in Theravadan school as well \citep[p.~245]{brahm}; see also later in the text.}

Indeed, this book has almost completely neglected the social aspects of being human---perhaps because AI's are not very social at the moment, and relevant computational theory is scarce. The theory of this book is clearly applicable to the social domain in the sense that social interaction creates its own input data from which the agent can learn. The agent might then realize that other agents are often quite unpredictable and uncontrollable and that this leads to a lot of frustration. Yet, social interaction creates completely new phenomena, which are outside of the theory of this book but should to be considered to see the whole picture of human suffering.

It can be argued that social interaction is essential for understanding what it is like to be human.\citenew{hari2015centrality}
One aspect is that human philosophical systems considered in this book are products of a long cultural evolution. It is difficult to see how any single AI could conclude, by itself, that desire produces suffering (or errors) and should be reduced. 
It is probably impossible for even any single human being to discover anything like those aforementioned philosophical systems. What is necessary is a cultural learning process based on sharing information between individuals, eventually leading to accumulation of knowledge over many generations.\footnote{It is crucial that the shared knowledge is cumulative, i.e., increases from one generation to another, which seems to be extremely rare with animals. A suprisingly important mechanism in such cultural learning seems to be imitation,\index{learning!imitation} even though at first sight, it might seem very primitive and unrelated to any higher form of intelligence \citep{iacoboni2005neural,whiten2009emulation}.} Such culturally produced, higher kind of intelligence, which can even consider the very concepts of intelligence and suffering as the objects of its analysis, is close to what would better be  called \textit{wisdom}.\index{wisdom} It is something much deeper than intelligence, and presumably unique to humans. 

\subsection{From individual desires to altruism}

Another essential aspect of social interaction is the human capacity for compassion, love, gratitude, and similar social emotions.\index{gratitude} 
In classic Buddhist training, there is a group of practices based on the cultivation of positive social, interpersonal emotions, such as compassion\index{compassion} and ``loving-kindness''.\footnote{For current research, see \citet{graser2018compassion,hofmann2011loving,cassell2002compassion}, and for the related emotions of forgiveness and gratitude,\index{gratitude}\index{forgiveness} see \cite{mccullough2002forgiveness} and footnote~\ref{gratitudefn} in Chapter~\ref{freedom.ch}.   For practical meditation guidance, see e.g.\ \citet{salzberg2002lovingkindness}.  }\index{loving-kindness}
Interestingly, such emotions can even be directed towards oneself: As an important example, self-compassion, i.e.\ compassion directed towards oneself, may strongly reduce negative self-evaluations, and thus self-related suffering.\citenew{neff2007self} 
Another book could possibly be written where reduction of suffering is approached from the viewpoint of such  positive social emotions.\index{emotions!social, positive}\index{social interaction}  Unfortunately, any related computational theory is rather lacking at this moment.\footnote{Gratitude was briefly discussed in Chapter~\ref{attitude.ch}}.
 
Historically, within Buddhism, a self-centered approach to reducing suffering was increasingly criticized in the centuries following the Buddha's death. Consequently, the later Mahayana schools adopted unselfish behavior as the ultimate ideal, instead of your individual nirv\=ana.\index{nibb\=ana/nirv\=ana}
They proposed that it is better to sacrifice one's own bliss and meditation time, at least to some extent, in order to help others to reduce their suffering. Slightly paradoxically, such a prosocial attitude is then seen as leading to an even higher form of happiness.\index{happiness} I would assume that such enlightened altruistic action somehow avoids the frustration process, perhaps because there is no longer any consideration for rewards that the agent itself will get, so in a sense, the self-based desire is no longer operating.\index{desire!and self}  It also seems that altruistic action gives its own evolutionary rewards,\footnote{For evolutionary theories of altruism, see \citet{wright1994moral,nowak2010evolution}} and  can even provide meaning to one's existence.\footnote{On meaning in life and its relation to happiness, see \citet{baumeister2002pursuit,martela2020wonderful,huta2014eudaimonia}. \add{However, desire to find a meaning for life could also be seen as just another desire that can be frustrated as in van Hooft's theory (page~\pageref{vanhooftpage}).} } Thus, altruistic action, if performed with the proper attitude, may be the ultimate exercise to reduce suffering---even to the very person performing the action.\index{altruism}

 To conclude, let me quote the Mahayana Buddhist philosopher \'S{\=a}ntideva,\index{S\=antideva} who recapitulates these ideas brilliantly:\footnote{\'S{\=a}ntideva's \textit{Bodhicary{\=a}vat{\=a}ra}, written in the 8th century CE, translated by Kate Crosby and Andrew Skilton, OUP, 1995. Such an altruistic attitude is often called the bodhisattva ideal in Buddhist literature \citep{garfield2010like,williams2008mahayana}.\index{bodhisattva}}
\begin{quote}
  All those who suffer in the world do so because of their desire for their own happiness.\\ All those happy in the world are so because of their desire for the happiness of others.
\end{quote}\index{vision|seealso{perception}}\index{perception|seealso{vision}}\index{happiness}

\printindex

\addcontentsline{toc}{chapter}{Bibliography}
\setlength{\bibsep}{0.2mm} 
\bibliographystyle{apalike}
\bibliography{biblio}

\end{document}